\documentclass[12pt]{article}

\usepackage{authblk}
\usepackage{geometry}
\usepackage{subcaption}
\usepackage{graphicx}
\usepackage{epstopdf}
\usepackage{xcolor}

\title{Optimal Vector Compressed Sensing Using James Stein Shrinkage}
\author{
    Apratim Dey\thanks{Corresponding Author. Email: \texttt{apd1995@stanford.edu}} \, and \, David Donoho\thanks{Email: \texttt{donoho@stanford.edu}}
}
\affil{\textit{Department of Statistics, Stanford University}}

\usepackage{amsthm,amsmath,amsfonts,bm}

\newtheorem{theorem}{Theorem}[section]  % Theorems numbered within sections
\newtheorem{lemma}[theorem]{Lemma}      % Lemmas follow the numbering of Theorems
\newtheorem{remark}[theorem]{Remark}    % Remarks follow the numbering of Theorems

\newcommand{\BST}{\text{BST}}
\newcommand{\JS}{\text{JS}}
\newcommand{\CBST}{\text{ColorBST}}
\newcommand{\CJS}{\text{ColorJS}}
\newcommand{\cvx}{\text{cvx}}
\newcommand{\MM}{\text{MM}}
\renewcommand{\deg}{\text{deg}}
\newcommand{\arr}{\text{arr}}
\newcommand{\LF}{\text{LF}}

\newcommand{\Stein}{\text{Stein}}

\usepackage{algorithm}
\usepackage{algpseudocode} % For algorithm writing
\usepackage{xcolor}

\def\eqref#1{equation~\ref{#1}}
\def\Eqref#1{Equation~\ref{#1}}
\def\1{\bm{1}}

\def\eps{{\epsilon}}

\DeclareMathAlphabet{\mathsfit}{\encodingdefault}{\sfdefault}{m}{sl}
\SetMathAlphabet{\mathsfit}{bold}{\encodingdefault}{\sfdefault}{bx}{n}

\def\gF{{\mathcal{F}}}

\def\gN{{\mathcal{N}}}

\def\gP{{\mathcal{P}}}

\def\gR{{\mathcal{R}}}

\def\sP{{\mathbb{P}}}

\def\sR{{\mathbb{R}}}

\newcommand{\E}{\mathbb{E}}

\DeclareMathOperator*{\argmin}{arg\,min}

\DeclareMathOperator{\Tr}{Tr}

\usepackage{hyperref}
\usepackage{url}
\usepackage{graphicx}
\usepackage{placeins}
\usepackage{float}
\usepackage{natbib}

\begin{document}

\maketitle

\begin{abstract}
% We present SteinSense, a Compressed Sensing algorithm, to reconstruct undersampled sparse high dimensional vectors with almost optimal number of measurements. The algorithm is simple to implement, does not require any tuning parameter or data to train, is practically unimprovable as the vector dimension gets large, and high computational speed is achievable with modern open-source software. Fundamental to the success of the algorithm is a key component from statistical decision theory - namely, James Stein shrinkage. The algorithm enjoys firm theoretical grounding with ideas borrowed from the Approximate Message Passing style reasoning. A wide array of numerical experiments confirms that the algorithm delivers according to expectations.
The trend in modern science and technology is to take vector measurements rather than scalars, ruthlessly scaling to ever higher dimensional vectors. For about two decades now, traditional scalar Compressed Sensing has been synonymous with a Convex Optimization based procedure called Basis Pursuit. In the vector recovery case, the natural tendency is to return to a straightforward vector extension of Basis Pursuit, also based on Convex Optimization. However, Convex Optimization is provably suboptimal, particularly when $B$ is large. In this paper, we propose \textbf{SteinSense}, a lightweight iterative algorithm, which is provably optimal when $B$ is large. It does not have any tuning parameter, does not need any training data, requires zero knowledge of sparsity, is embarrassingly simple to implement, and all of this makes it easily scalable to high vector dimensions. We conduct a massive volume of both real and synthetic experiments that confirm the efficacy of SteinSense, and also provide theoretical justification based on ideas from Approximate Message Passing. Fascinatingly, we discover that SteinSense is quite robust, delivering the same quality of performance on real data, and even under substantial departures from conditions under which existing theory holds.
\end{abstract}
\section{Introduction}\label{sec:intro}

The current trend in science and technology is to collect high dimensional vectors rather than scalars; ruthlessly, inexorably scaling to ever higher dimensional vectors. Important applications include Magnetic Resonance (MR) Spectroscopy, hyperspectral imaging, RNASeq with multiplexing, etc. In MR Spectroscopy, radiologists record the concentrations of multiple chemical substances in the tissue, thereby creating a rich source of data that is much more informative than traditional MRI scans in predicting the presence of tumors. Hyperspectral images have had enormous impact in deep space exploration and geo-spatial imaging. Satellites are routinely collecting images in hundreds of spectral wavelengths (often bordering on thousand), enabling important scientific discoveries and transforming precision agriculture. For instance, CRISM hyperspectral images have provided deep insights into the geological history of Mars. Wearable devices such as smart watches regularly collect data per second for multiple health metrics. Modern methods enable RNA sequencing and gene sequencing of multiple different samples all at once. Thus, today, it is much more practical and powerful to measure vector data, and vector data offer unprecedented scientific insights and opportunities for statistical and machine learning exercises that are simply not available from scalar data.

Collecting ever higher dimensional vector data, unfortunately, poses challenges in terms of high acquisition time and costly transmission. In MR imaging, it is important to speed up acquisition to offer comfort to the patient. In hyperspectral imaging and wearable devices, the data collected need to be transmitted to earth and a central server respectively, and the transmission of such high dimensional vector data can be quite slow and resource-inefficient. Thus, it is highly desirable to take many \textit{fewer} measurements than what may be apparent, and transmit only these few measurements instead of the original high dimensional vectors. As a tradeoff, one needs to allocate resources to reconstruct the vectors, although this is much more desirable given the computational advancements and high quality software tools available today.

Formally, suppose one wants to measure a large number $N$ of vectors $X_{i\star}\in\sR^B$, where $B$ can be large as well. Define $X\in\sR^{N\times B}$ to be the matrix containing $X_{i\star}$ as its $i$'th row:
\begin{align*}
    X &= \begin{bmatrix}
        X_{1\star}^\top \\
        \cdots \\
        X_{i\star}^\top \\
        \cdots \\
        X_{N\star}^\top
    \end{bmatrix}\in \sR^{N\times B}
\end{align*}Then, instead of recording and transmitting $X$, one records and transmits instead $Y\in\sR^{n\times B}$, where
\begin{align*}
    Y &= AX
\end{align*}with $A\in\sR^{n\times N}$ being known as the \textit{measurement matrix} or \textit{sensing matrix}. Taking fewer measurements implies $n<N$, and the goal becomes reconstruction of the original $X$ given $(Y, A)$.

With this formulation, Compressed Sensing \citep{donoho2006compressed,candes2006robust} naturally enters the picture. In the scalar case, that is when $B=1$, Compressed Sensing provides the message that it is possible to have $n$ significantly smaller than $N$ and still \textit{perfectly} reconstruct $X\in\sR^{N\times 1}$ provided that $X$ is sparse. Compressed Sensing has offered major benefits in speeding up acquisition time in pediatric MRI by 6-8 times \citep{lustig2007sparse,lustig2008compressed,vasanawala2010improved}. Since the FDA approval to incorporate Compressed Sensing in MRI hardware in 2017, leading companies like Siemens and Philips have developed products such as Compressed Sensing Cardiac Cine and Compressed Sense (respectively) benefiting thousands of patients.

In the scalar case, perhaps the most popular algorithm for Compressed Sensing reconstruction is Basis Pursuit \citep{chen1994basis,chen2001atomic,donoho2006compressed,candes2006robust}, which is based on convex optimization. However, it is known that while Basis Pursuit indeed promises perfect reconstruction with $n<N$ (depending on the level of sparsity in the scalars), there is a limit to how small $n$ can be for successful reconstruction, and further improvements typically require knowledge about the distribution of the non-zero elements in $X$. Besides the fact that it can be extremely challenging to know precisely the distribution of the entries of $X$ in most real applications when $N$ is large, the gains in going beyond Basis Pursuit can be marginal.

For $B>1$, the performance of the corresponding convex optimization does improve over $B=1$, but very soon hits a wall, which can be traced mathematically. Indeed, as $B$ gets large, after a point, convex optimization starts providing only marginal benefits since it encounters a fundamental limit, and this prevents it from achieving oracle performance.

Fascinatingly, when $B$ is large, this curse of convex optimization can be broken by a certain non-convex procedure. Not only can one outperform convex optimization, one can also achieve essentially oracle performance (namely, vector reconstruction using essentially minimal possible number of measurements) \textit{without} any extra information about the vectors $X_{i\star}$.

To achieve this, we introduce \textbf{SteinSense}, an iterative algorithm employing the James Stein denoiser \citep{james1992estimation} in a suitable way. The algorithm is free of any tuning parameter, does not need any data to train, does not need knowledge of sparsity to run, enjoys essentially optimal reconstruction performance and thus is unimprovable. Further, it enjoys firm theoretical basis with completely predictable performance, as it is built on the grounding provided by Approximate Message Passing (AMP) algorithms, which have proven to be powerful theoretical tools in analyzing many problems in high dimensional statistics during the last decade. Since it attains oracle performance, any other procedure that employs any other knowledge, no matter how much, can only offer minor improvements, that too for small $B$. Crucial to the success of SteinSense are insights from statistical decision theory, enabling optimal Vector Compressed Sensing.

\noindent

\section{Related works and our contributions}\label{sec:related_works}

Compressed Sensing \citep{donoho2006compressed,candes2006robust} has emerged as a powerful paradigm to reconstruct sparse signals from undersampled measurements. Specifically, in the case $B=1$, traditional scalar compressed sensing attempts to recover sparse $X\in\sR^N$ (thus, $N$ scalars) given measurements $Y=AX$, where $A\in\sR^{n\times N}$ is the measurement matrix (with $n<N$ representing \textit{undersampling}). One of the most popular approaches towards achieving this goal has been Basis Pursuit \citep{chen1994basis,chen2001atomic,donoho2006compressed,candes2006robust} which is based on convex optimization:
\begin{align}\label{eqn:cvx_opt}
    \text{Minimize }\|X\|_1 \text{ such that }Y = AX
\end{align}
Understanding (\ref{eqn:cvx_opt}) and its variants, both with and without noise (one of the noisy variants being the LASSO \citep{tibshirani1996regression}), has been a source of intense research exploration (see for example \cite{donoho2003optimally,candes2006stable,tsaig2006extensions,candes2007sparsity,donoho2008fast,wainwright2009sharp,raskutti2010restricted} among many others, also see \cite{davenport2012introduction} for a book-length treatment). While much of the work in the early compressed sensing days focused on getting bounds on the reconstruction error under favorable circumstances, a parallel line of work explored the precise sparsity-undersampling \textit{phase transition} exhibited by \ref{eqn:cvx_opt} \citep{donoho2005neighborliness,donoho2006high,donoho2009counting,donoho2009observed,amelunxen2014living}. In particular, \citet{donoho2009observed} presented what is today popularly known as the Donoho-Tanner phase transition in the context of compressed sensing. We describe it as follows.

Let $\epsilon=k/N$ denote the fraction of nonzeros in $x$, henceforth to be termed \textit{sparsity}, where $k$ denotes the number of nonzero entries in $X$. Let $\delta=n/N$ denote the \textit{undersampling ratio}, where $n$ denotes the number of measurements. Suppose the measurement matrix $A$ is filled with iid $N(0,1)$ entries. The central message from the above mentioned phase transition literature was that given sparsity $\epsilon$, there exists an analytically tractable function $M_\cvx(\epsilon,B=1)$ such that the following holds as $N\to\infty$ (assuming $\epsilon,\delta\in(0,1)$):
\begin{center}
    If $\delta>M_\cvx(\epsilon,B=1)$, $\sP((\ref{eqn:cvx_opt}) \text{ succeeds})\to1$,\\
    If $\delta<M_\cvx(\epsilon,B=1)$, $\sP((\ref{eqn:cvx_opt}) \text{ fails})\to1$.
\end{center}In other words, a phase transition occurs at $\delta^*=M_\cvx(\epsilon,B=1)$. The formula for $M_\cvx(\epsilon,B=1)$ was originally calculated using convex geometry and polytope theory \citep{donoho2005neighborliness,donoho2006high,donoho2009counting,donoho2009observed,amelunxen2014living}. Later, via Approximate Message Passing \citep{donoho2009message,donoho2013accurate}, it was established that $M_\cvx(\epsilon,B=1)$ is the minimax risk of soft thresholding, over the class of $\epsilon-$sparse probability distributions (to be appropriately defined later). Remarkably, this shows that the phase transition is independent of the actual characteristics of $X$, and sparsity is all that matters.

The vector case, namely the case $B>1$, is often referred to as the Multiple Measurement Vector (MMV) problem in signal processing, with perhaps the earliest works traced back to \citet{cotter2005sparse,chen2006theoretical}. Over the last two decades, extensive research has been performed in the MMV problem \citep{van2010theoretical,duarte2011structured,chen2011hyperspectral,yang2011multiple,li2015off}. 
Most of these works focus on a convex optimization based natural extension of (\ref{eqn:cvx_opt}):
\begin{align}\label{eqn:cvx_opt_B}
    \text{Minimize }\|X\|_{2,1} \text{ such that }Y = AX
\end{align}where $\|X\|_{2,1}=\sum_{i=1}^N \|X_{i\star}\|_2$. 
% Note that (\ref{eqn:cvx_opt_B}) is also based on convex optimization.
It has been documented in multiple studies that as $B$ grows, the performance of (\ref{eqn:cvx_opt_B}) improves. One then encounters multiple questions. 
\begin{enumerate}
    \item Does the improvement happen indefinitely as $B$ gets larger? (The answer is NO.)
    \item If not, can we outperform convex optimization by resorting to a different algorithm as $B$ grows?
    \item Can we achieve optimal performance? Is the procedure scalable?
\end{enumerate}
% Unfortunately, prior work does not lead to answers of this level of granularity.

In this work, we establish that convex optimization, although improves with increasing $B$, \textit{does} have a fundamental limit, which is given by the limiting minimax risk of a certain denoiser. The connection is forged by a suitably defined Approximate Message Passing (AMP) - style algorithm, which we will call \textbf{SoftSense}. Further, a simple change of the denoiser leads one to formulate the \textbf{SteinSense} algorithm - the main deliverable of this work - which is able to significantly outperform convex optimization even for moderate choices of $B$, as small as $5$.

The algorithm we study share notable differences with the usual AMP algorithms presented in the literature \citep{donoho2009message}. The main reason is that although the measurement matrix $A$ \textit{looks} like it is composed of iid gaussians, the actual measurement matrix $\tilde A$ needed to recast this problem into the traditional AMP formulation becomes highly structured; in particular, it becomes a block diagonal matrix with repeated blocks, all equal to $A$. The usual Onsager divergence term, which is quite common in usual AMP algorithm formulations, no longer works for such a structured measurement matrix, and needs to be replaced by a Jacobian Onsager term.

Such a correction has been leveraged in prior works, for example in \citet{hara2022blind,hara2020grant,zhu2016performance}; however, there is an important difference with our work. Prior works usually focus on Bayes estimators and other carefully constructed denoisers to extract optimal performance. This requires one to have full knowledge of the distribution of the non-zero entries in the vectors - but who knows the exact distribution of real world signals? Estimating this distribution from samples is known to be notorious when the vector dimension $B$ gets large. One of our main messages is that when $B$ gets large, no knowledge of the non-zero distribution is needed, and no supposedly clever denoising procedure is needed. The James Stein denoiser in SteinSense adapts to the distribution of the vectors and becomes essentially unimprovable, along with scalability. More clearly, SteinSense succeeds as soon as the undersampling ratio $\delta$ exceeds the minimax risk of James Stein denoiser, to be denoted by $M_\JS(\epsilon,B)$ (as discussed in Theorem \ref{thm:js_mse}). The reason behind the success of SteinSense comes from the simple fact that $M_\JS(\epsilon,B)\to \epsilon$ as $B\to\infty$. This means that SteinSense succeeds with approximately ``sparsity" fraction of measurements, which confirms optimality, since there cannot be any other procedure that achieves perfect recovery with \textit{less than sparsity} fraction of measurements.
% \citet{hara2022blind} assumes a simple \textit{scalar} model in Equation 8 which results in the corresponding State Evolution to be scalar, but an unignorable consequence of structure in the measurement matrix causes the State Evolution to be matricial (as we clarify later). While the prior assumed in Equation 4 of \citet{hara2022blind} indeed \textit{reduces} the matricial State Evolution to a scalar State Evolution, it is important to highlight that in general, such a simple reduction is far from true. For example, if the non-zero entries in the vectors $X_{i\star}$ are positive, the scalar State Evolution would be an incorrect description of the dynamics of the corresponding AMP algorithm. In our experience, this distinction does not seem to be adequately clearly pointed out in the literature, and might lead a scientist attempting to use AMP astray. 
% The denoising procedure employed in \citet{hara2022blind} is dependent on knowing the precise distribution of the high dimensional vectors, which would be a challenging requirement to fulfill for most real world signals. 

The connection between the minimax risk of a denoiser and the phase transition of an Approximate Message Passing algorithm with that denoiser has been pointed out in \citet{donoho2013accurate} (also see \citet{oymak2012relation,amelunxen2014living} for the connection between the phase transition exhibited by convex optimization and minimaxity). However, the AMP algorithms and corresponding State Evolution presented in \citet{donoho2013accurate} would work only when one \textit{concatenates} all the vectors to form an enormously long array $X_{\arr}\in\sR^{NB}$ containing $NB$ scalars, with a huge $O(NB)\times NB$ dimensional measurement matrix $A_{\arr}$ consisting of iid gaussian entries. Such a measurement scheme is clearly impractical for $B$ even moderately large. Indeed, in realistic applications, and in the usual MMV setup, as described before, the sensible way to take measurements is to record $Y=AX\in\sR^{n\times B}$. In such a situation, the corresponding State Evolution is no longer scalar, rather matricial. This is discussed in more details in Sections \ref{sec:array_cs} and \ref{sec:theory}.

Finally, we emphasize that our work is primarily computational in nature, with the goal to demonstrate convincingly the quality of SteinSense over a wide variety of experiments. Theoretical conclusions are drawn leveraging the powerful theory of generalized Approximate Message Passing \citep{bayati2011dynamics,rangan2011generalized,javanmard2013state} to draw important insights from the behavior of State Evolution to determine and improve the phase transition of Vector Compressed Sensing. Consequently, we are able to demonstrate, both empirically and theoretically, that when $B$ is large, James Stein becomes optimal, and that there is virtually no need to go for any denoiser more advanced or a procedure more complex.

\paragraph{Remarks on Notations.} $I_B$ will denote the $B\times B$ identity matrix. $\gN_B(\mu,\Sigma)$ denotes the $B$-variate Gaussian distribution with mean vector $\mu\in\sR^B$ and covariance matrix $\Sigma\in\sR^{B\times B}$. $\E$ will denote expectation, $\sP$ will denote probability. A collection of random variables $X_n\stackrel{\text{a.s.}}{\to}X$ (in words, $X_n$ converges almost surely to $X$) if $\sP(X_n\to X)=1$. For a matrix $M$, $\|M\|_F$ denotes the Frobenius norm of $M$, which is the sum of squares of entries in $M$. The letter $\delta$ will be used multiple times in this paper, unfortunately in different contexts to maintain notational obedience with existing literature and discourse in Compressed Sensing, with the hope that there would not be substantial confusion regarding its use. Several times $\delta$ will be used to denote undersampling ratio and phase transitions. At other times, $\delta_x$ will be used to denote the degenerate distribution at $x$.
\noindent

\section{Experimental Methodology}\label{sec:LD50}

The process of discovering the phase transitions of different vector Compressed Sensing algorithms required intense computational effort. Over a long period of time, several tens of millions of embarrassingly parallel experiments were conducted, varying different axes such as algorithm, measurement pattern, number of vectors ($N$), vector dimension ($B$), distribution of the nonzero entries of the vectors, and so on, covering both synthetic and real data. Computations were performed mostly on Stanford's high performance compute cluster, Sherlock, and at times also on Google Cloud Platform (GCP) and on personal supercomputers. All the data are stored securely on Google BigQuery, ready for download and use by anyone with appropriate permissions.

The general experimental methodology to obtain the plots presented in this paper is explained as follows:
\begin{itemize}
    \item Sparsity $\epsilon$ varies in the grid $\{0.02, 0.04, \cdots, 0.98\}$. We use this grid for most experiments, and for those where the plot looks \textit{coarser}, sparsity was varied in $\{0.05, 0.1, \cdots, 0.95\}$.
    \item We now need to choose the undersampling ratios $\delta$, or equivalently the number of measurements $n=N\delta$ for our experiments. Without any idea about the potential phase transition location, one would naively vary $n\in\{1,2,\cdots, N\}$. When $N$ is in the thousands, choosing this grid for each $\epsilon$ produces a humongous number of experiments that would increase the time required to get all the experimental evidence by several orders of magnitude, given the sheer number of experiments performed. Further, the notion of a phase transition makes one expect approximately deterministic results once one is far away from its location; for larger $\delta$ we expect most experiments to result in success, and for smaller $\delta$ we expect most experiments to result in failure. Consequently, we only perform experiments with $n$ varying in integers in $\epsilon-$dependent grids around an expected phase transition point informed through smaller scale pilot experiments.
    \item For each $(\epsilon,\delta)$ pair, we choose an iid Gaussian measurement matrix $A\in\sR^{n\times N}$ filled with $N(0,1/n)$ entries. The choice of $1/n$, and not $1$, as the variance is unimportant for Convex Optimization since it only alters $Y$ multiplicatively in the constraint $AX=Y$, but is important for SoftSense and SteinSense.
    \item For each $(\epsilon,\delta)$ pair, we generate $N$ vectors $X_{1\star},\cdots, X_{N\star}\in\sR^B$ such that exactly $k=N\epsilon$ of them are non-zero, with the nonzero entries generated from a user-specified distribution. The support set $S:=\{1\leq i\leq N:X_{i\star}\neq 0\}$ is chosen as a uniformly random set of size $k$ picked without replacement from $\{1,\cdots, N\}$. The $X_{i\star}$ are stacked row-wise to form the matrix $X\in\sR^{N\times B}$. For real data experiments, the procedure of generating $X$ differs slightly, with necessary modifications clarified in Section \ref{sec:real_exp}.
    \item For each $(\epsilon,\delta)$ pair, several such $(A,X)$ pairs are generated, and for each, an algorithm was run. If the output $\hat X$ resulted in small relative error, specifically $$\dfrac{\|\hat X-X\|_F}{\|X\|_F}<0.001$$ we declare the experiment to be a \textit{Success}, recording $1$, otherwise a \textit{Failure}, recording $0$. Here $\|M\|_F$ denotes the Frobenius norm of the matrix $M$. The relative error threshold $10^{-3}$ is significantly more conservative than what the current computational literature on Compressed Sensing has used, for example \citet{donoho2013accurate} use the threshold $0.1$ which is much more relaxed than what we use. Of course, such a threshold is user-dependent at the end of the day.
    \item In the plots in this paper, we present heatmaps showing the fraction of successful reconstructions for each $(\epsilon,\delta)$ pair. We also overlay an empirical phase transition curve (details below), some analytically computed curves, and the diagonal, whenever appropriate.
\end{itemize}

\paragraph{Empirical phase transition estimation.} We use the classical median lethal dose / LD50 estimation method from clinical trials to estimate the location of the empirical phase transition for each considered experiment. Such a procedure has been widely used in Compressed Sensing, see for example \citet{donoho2009observed,amelunxen2014living,donoho2013accurate}. In short, we pick that value of $\delta$, henceforth to be denoted as $\delta_{\text{PT}}$, at which the fitted probability of getting a success (and hence a failure) is $1/2$.

For each sparsity $\epsilon$, we have data $(\delta_i, r_i)$ where $\delta_i$ represents the undersampling ratio and $r_i$ denotes a binary outcome $0/1$. We fit a logistic model 
\begin{align*}
    r_i\sim \text{Logistic}(f_{\deg}(\delta_i))
\end{align*}where $f$ is a polynomial of degree $\deg$. The coefficients of $f_\deg$ are estimated by usual polynomial logistic regression. Call the fitted polynomial $\hat f_\deg$. The phase transition $\delta_{\text{PT}}(\deg)$ is obtained as a properly chosen root of $\hat f_\deg$. For $\deg=1$, writing $f_1(\delta) = \beta_0+\beta_1\delta$, it becomes standard logistic regression. Using estimates $\hat\beta_0,\hat\beta_1$, our estimate for the empirical phase transition becomes
\begin{align*}
    \delta_{\text{PT}}(\deg=1) &= -\dfrac{\hat\beta_0}{\hat\beta_1}
\end{align*}For $\deg>1$, we define $\delta_{\text{PT}}(\deg)$ to be the root of $\hat f_{\deg}$ closest to $\delta_{\text{PT}}(1)$. In our experiments, we use $\deg=2$ or $3$ to produce a better fit than $\deg=1$.

\paragraph{App for plots.} An exorbitant amount of data has been collected through massive experimentation over a long period of time, and consequently, a huge number of plots have been generated. Experiments will continue to be performed in future, and more data will be generated and added to the existing already massive database. Plots will henceforth be updated on \href{https://vector-cs-plots-apratim.streamlit.app/}{https://vector-cs-plots-apratim.streamlit.app/}. All codes will be made available on \href{https://github.com/apd1995/Vector-Compressed-Sensing}{https://github.com/apd1995/Vector-Compressed-Sensing}.
\section{Fundamental Limit of Convex Optimization}\label{sec:cvx_opt}

\begin{figure}
    \centering
    \includegraphics[width=\linewidth]{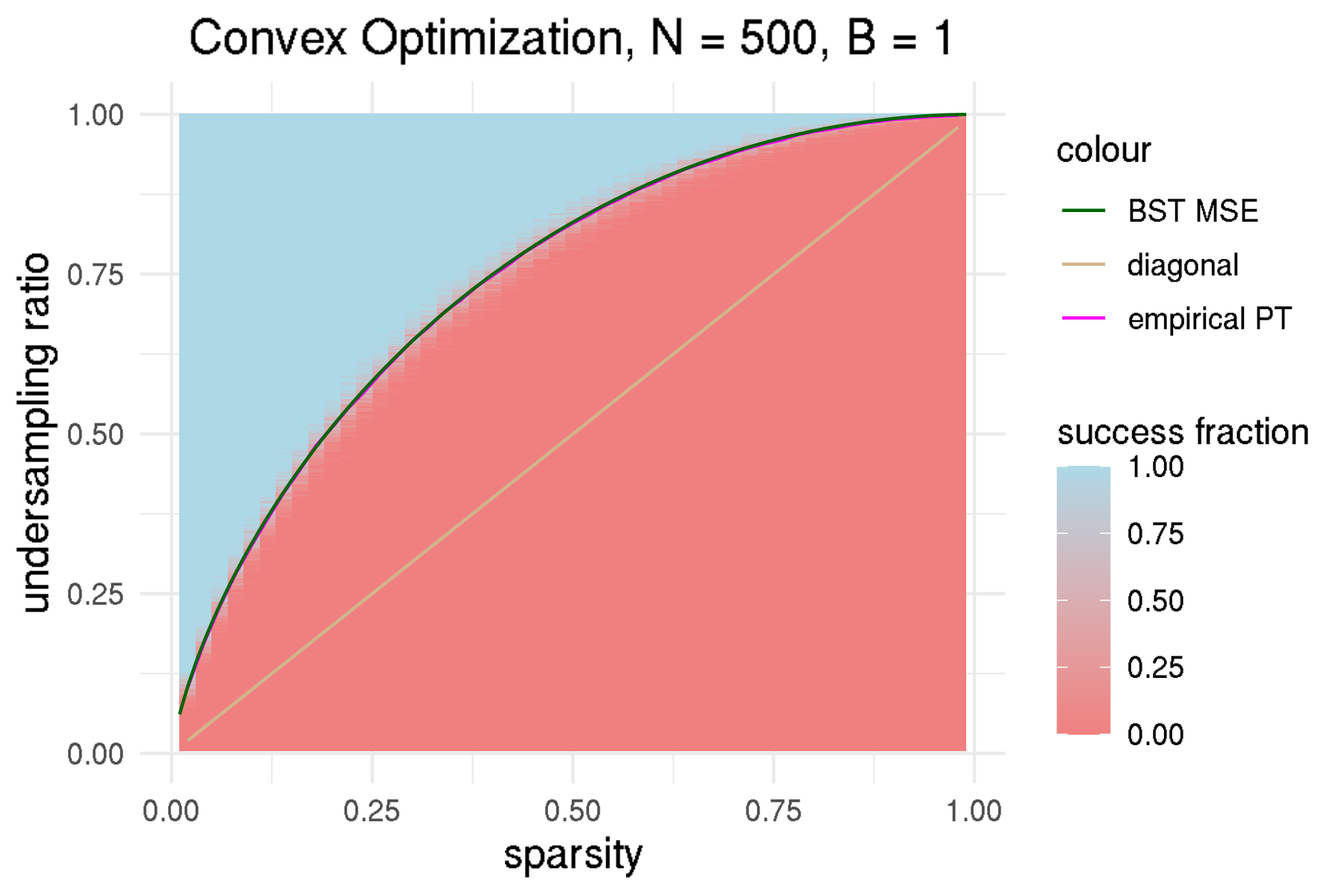}
    \caption{Convex optimization exhibits a phase transition when $B=1$. This reproduces the popularly known Donoho-Tanner Phase Transition \citep{donoho2009observed}. The nonzero scalars in the plot are chosen to be iid $N(0,1)$. The empirical phase transition curve matches the Minimax Risk of Soft Thresholding, which is the special case of Minimax Risk of BlockSoft Thresholding for $B=1$, to a high degree of accuracy. The sparsity level in x-axis varies in a fine grid $\{0.02,0.04,\cdots,0.98\}$. Each pixel depicts the fraction of successes for convex optimization from at least 25 Monte Carlo runs.}
    \label{fig:cvx_gaussian_B_1}
\end{figure}

\begin{figure}[htbp]
  \centering

  \begin{subfigure}[b]{0.9\textwidth}
    \includegraphics[width=\linewidth]{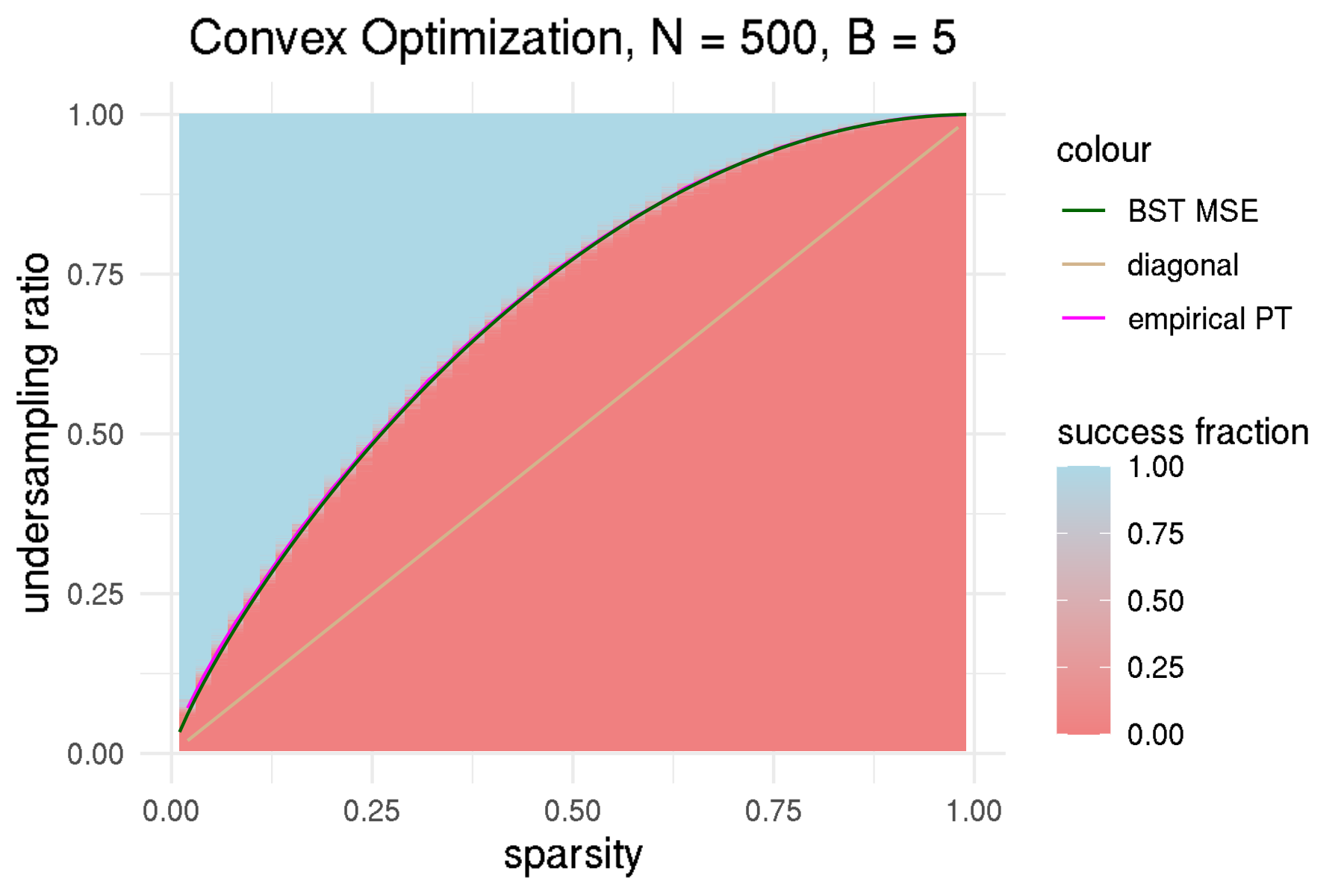}
  \end{subfigure}
  \vspace{1em}
  \begin{subfigure}[b]{0.9\textwidth}
    \includegraphics[width=\linewidth]{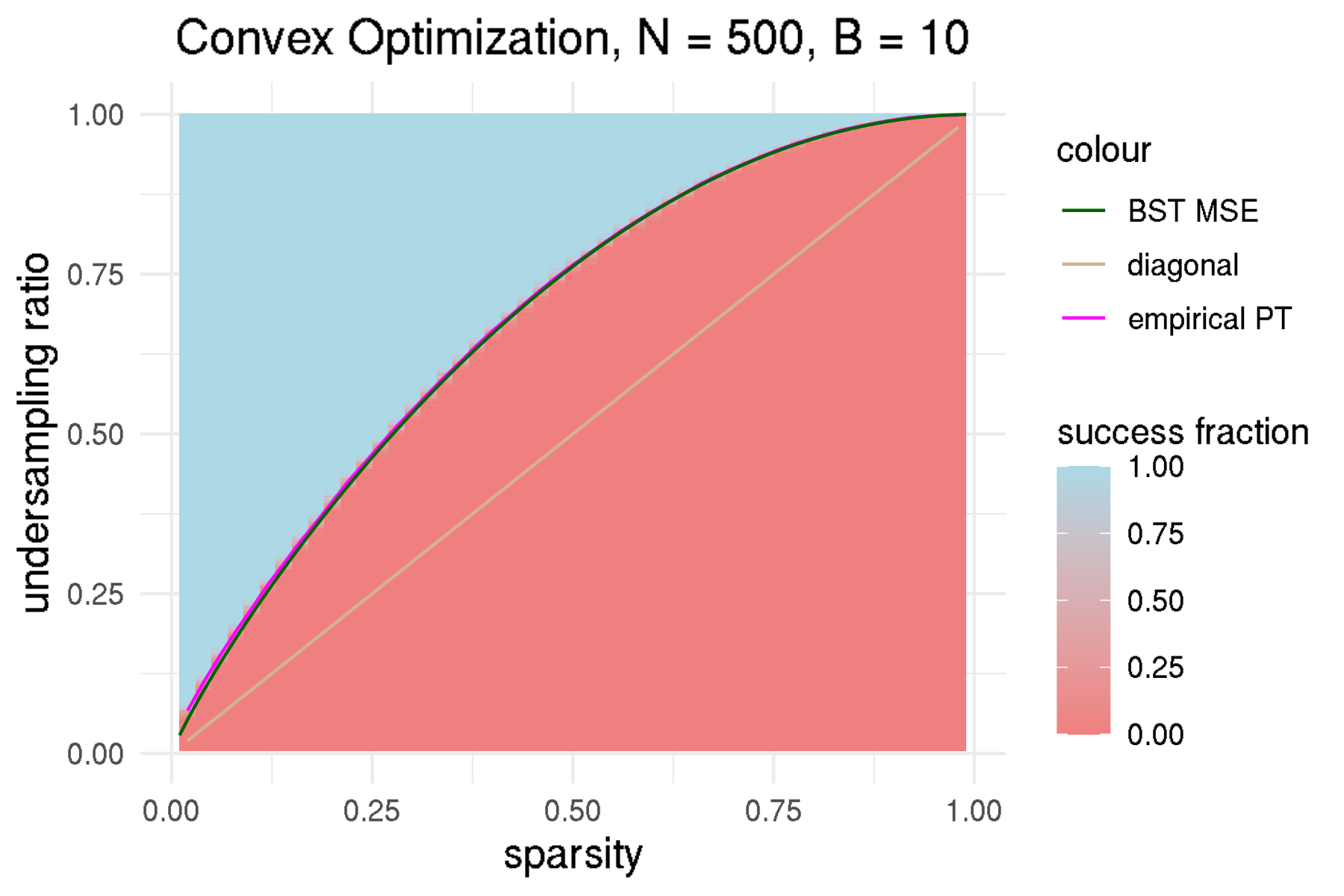}
  \end{subfigure}
  \caption{Convex optimization exhibits phase transitions for $B=5,10$. The entries in the nonzero vectors are chosen to be iid $N(0,1)$. The empirical phase transition curve matches the Minimax Risk of BlockSoft Thresholding to a high degree of accuracy. Each pixel depicts the fraction of successes for convex optimization from at least 25 Monte Carlo runs.}\label{fig:cvx_gaussian_B_5_10}
\end{figure}

\begin{figure}[htbp]
  \centering

  \begin{subfigure}[b]{0.9\textwidth}
    \includegraphics[width=\linewidth]{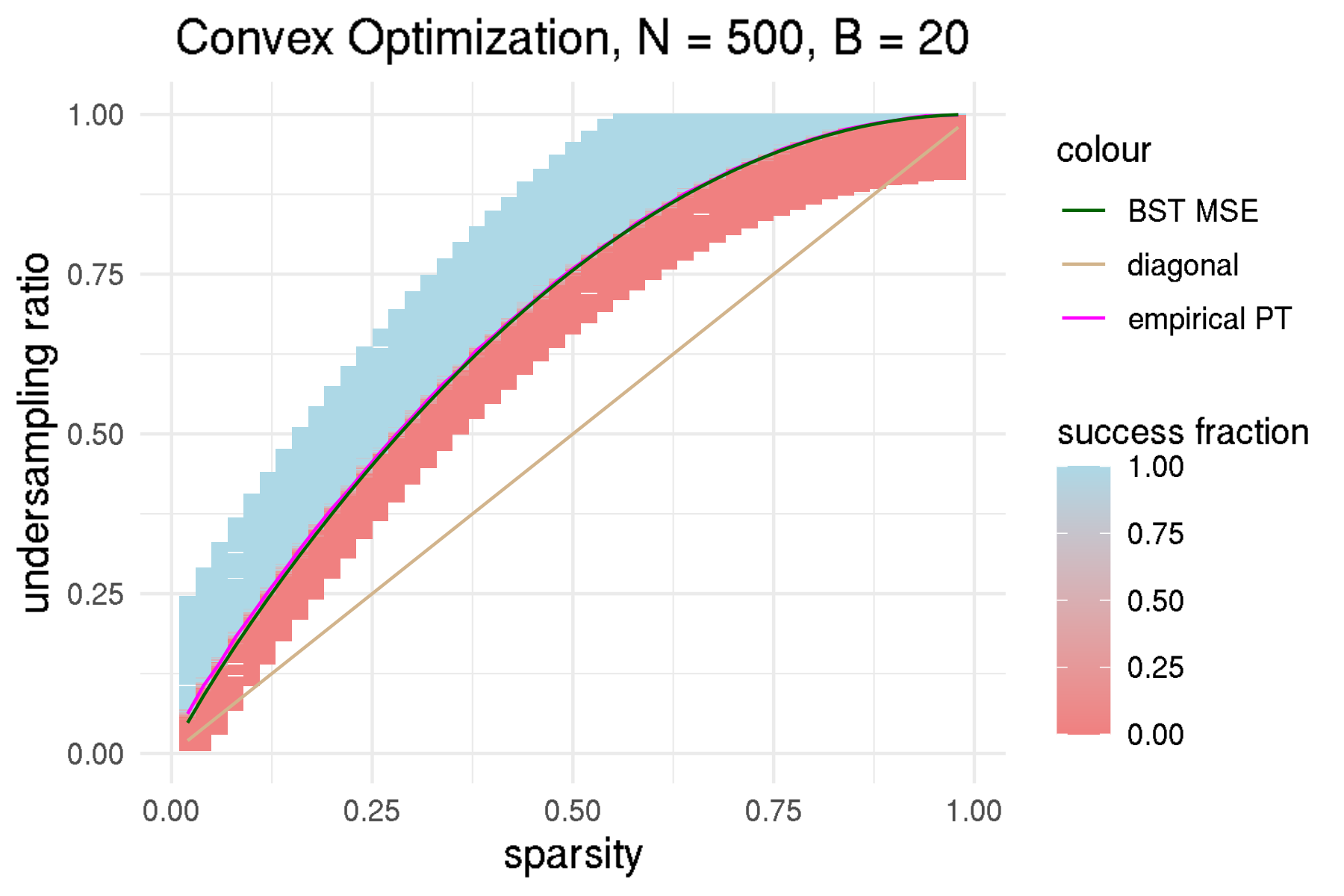}
  \end{subfigure}
  \vspace{1em}
  \begin{subfigure}[b]{0.9\textwidth}
    \includegraphics[width=\linewidth]{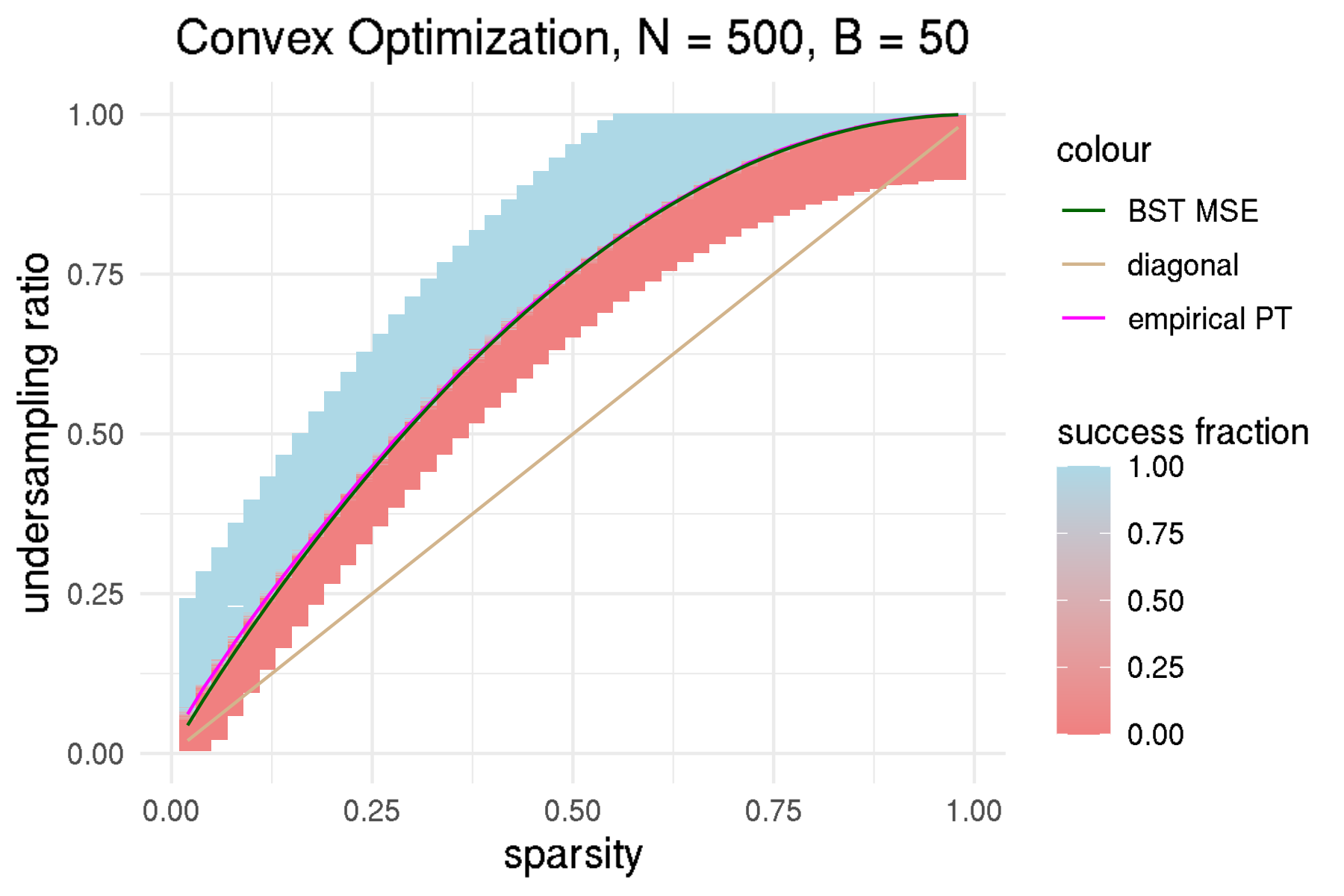}
  \end{subfigure}

  \caption{Convex optimization exhibits phase transitions for $B=20,50$. The entries in the nonzero vectors are chosen to be iid $N(0,1)$. The empirical phase transition curve matches the Minimax Risk of BlockSoft Thresholding to a high degree of accuracy. Each pixel depicts the fraction of successes for convex optimization from at least 25 Monte Carlo runs.}\label{fig:cvx_gaussian_B_20_50}
\end{figure}

\begin{figure}
    \centering
    \includegraphics[width=\linewidth]{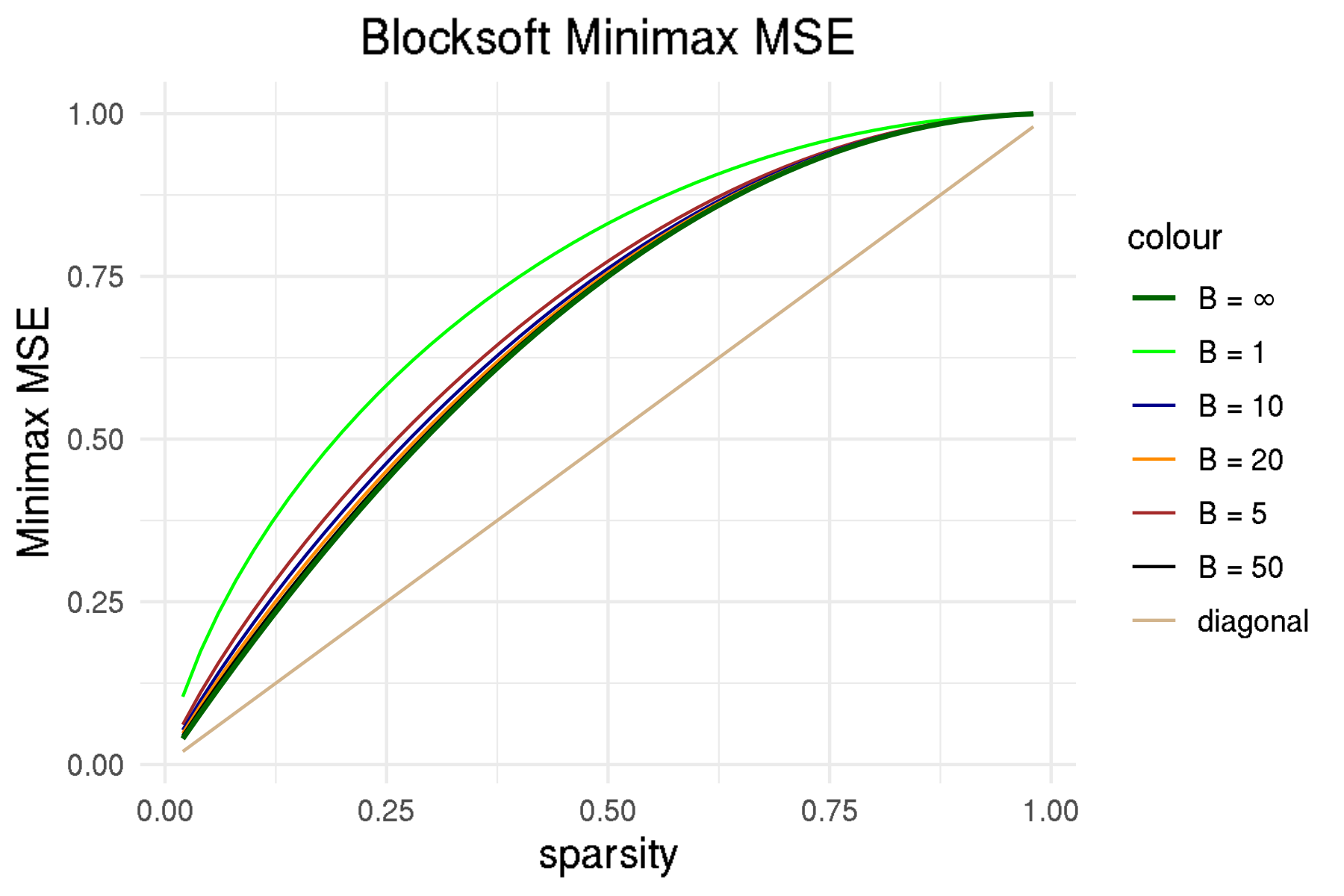}
    \caption{Numerically computed BlockSoft Thresholding minimax risk curves for different values of $\epsilon$ and $B$. As $B$ increases, the curves get lower, that is, the phase transitions of Convex Optimization improve. However, going beyond $B\geq 20$ offers negligible improvements. The theoretical $B=\infty$ curve equals $2\epsilon-\epsilon^2$. It forms the lower envelope for all the finite $B$ curves in the plot, and is still significantly away from the diagonal.}
    \label{fig:BST_MSE_vals}
\end{figure}

It is instructive to first study the performance of convex optimization (\ref{eqn:cvx_opt_B}) as $B$, the vector dimension, grows. Figures \ref{fig:cvx_gaussian_B_1}, \ref{fig:cvx_gaussian_B_5_10} and \ref{fig:cvx_gaussian_B_20_50} display the results. One notes the following:
\begin{enumerate}
    \item For each $\epsilon,B$, a phase transition exists, sharply demarcating success from failure. Namely, there exists a critical undersampling ratio value $\delta_{\cvx}(\epsilon,B)$ such that for undersampling ratio $\delta$ even a little bit above $\delta_{\cvx}(\epsilon,B)$, almost all experiments result in success, while for any $\delta$ a little bit below $\delta_{\cvx}(\epsilon,B)$, almost all experiments result in failure. We note in passing that the empirical phase transition corresponding to $B=1$ is classically known as the Donoho-Tanner phase transition curve \citep{donoho2009observed}.
    
    \item The phase transition is evident for $N$ \textit{just} in the hundreds; $N=500$ is enough.

    \item The empirical phase transition is accurately matched by a curve well understood in classical statistical decision theory, viz. minimax risk of BlockSoft Thresholding as a function of the sparsity, to be denoted as $M_\BST(\epsilon,B)$. These theoretical curves are plotted in Figure \ref{fig:BST_MSE_vals} to show their evolution with $B$.

    \item Figure \ref{fig:BST_MSE_vals} shows that the phase transitions \textit{improve} as $B$ increases. Consequently, for any sparsity $\epsilon$, convex optimization requires less undersampling for perfect recovery of $X_{i\star}$, as $B$ increases.

    \item Perhaps most importantly, we notice from Figure \ref{fig:BST_MSE_vals} that increasing $B$ beyond $B=20$ (say) results in very marginal benefits, since the phase transition curves do not seem to improve significantly.
\end{enumerate}

Building on point 5 above, it is reasonable to believe that Convex Optimization, despite improving with increasing $B$, is unable to offer benefits beyond a certain $B$. Results from more experiments with different distributions of nonzeros are shown in Section \ref{sec:appdx:cvx_more_experiments}, confirming this observation.

\paragraph{Significance of the diagonal.} We note that there remains a significant difference between the best-$B$ (in our case, $B=50$) phase transition curve and the diagonal. The diagonal has an important place in the phase diagram. It corresponds to an \textit{oracle} phase transition $\delta_{\text{oracle}}=\epsilon$; namely, the minimal undersampling ratio is equal to the sparsity in the presence of a support-aware oracle. Clearly, if one knows which $X_{i\star}\neq 0$, one only needs $n=k$ measurements, in fact, measure the $k$ vectors themselves. This corresponds to $\delta_{\text{oracle}}=\epsilon$. Thus the diagonal serves as a lower bound for any reasonable algorithm. Consequently, algorithms closer to the diagonal enjoy better (i.e. lower) phase transitions.

% \begin{figure}[H]
%     \centering
%     % First row: Two subfigures
%     \begin{subfigure}{0.49\textwidth}
%         \centering
%         \includegraphics[width=\linewidth]{figures/convex/convex_paper_B_1.png} 
%     \end{subfigure}
%     \hfill
%     \begin{subfigure}{0.49\textwidth}
%         \centering
%         \includegraphics[width=\linewidth]{figures/convex/convex_paper_B_5.png}
%     \end{subfigure}

%     \vspace{1em}  % Vertical space between rows

%     % Second row: Two subfigures
%     \begin{subfigure}{0.49\textwidth}
%         \centering
%         \includegraphics[width=\linewidth]{figures/convex/convex_paper_B_10.png}
%     \end{subfigure}
%     \hfill
%     \begin{subfigure}{0.49\textwidth}
%         \centering
%         \includegraphics[width=\linewidth]{figures/convex/convex_paper_B_20.png}
%     \end{subfigure}

%     \vspace{1em}  % Vertical space between rows

%     % Third row: Single centered subfigure
%     \begin{subfigure}{0.5\textwidth}  % Adjust width as needed
%         \centering
%         \includegraphics[width=\linewidth]{figures/convex/convex_paper_B_50.png}
%     \end{subfigure}

%     \caption{For each $B$, convex optimization exhibits a sharp phase transition even for very moderate values of $N$ and $B$. The empirical phase transition and theoretical BlockSoft Minimax Risk curves are very close to each other, almost indistinguishable, empirically confirming that the BlockSoft Minimax Risk $M_{\BST}(\epsilon,B)$, abbreviated as BST MSE in the figures, is the correct phase transition location.}
%     \label{fig:cvx_results}
% \end{figure}

% \begin{figure}
%     \centering
%     \includegraphics[width=\linewidth]{figures/convex/paper_cvx_N_500_B_5.png}
%     \caption{}
%     \label{fig:cvx_gaussian_B_5}
% \end{figure}

% \begin{figure}
%     \centering
%     \includegraphics[width=\linewidth]{figures/convex/paper_cvx_N_500_B_10.png}
%     \caption{Convex optimization exhibits a phase transition when $B=10$. The entries in the nonzero vectors are chosen to be iid $N(0,1)$. The empirical phase transition curve matches the Minimax Risk of BlockSoft Thresholding for $B=10$, to a high degree of accuracy. Each colored pixel depicts the fraction of successes for convex optimization from at least 25 Monte Carlo runs.}
%     \label{fig:cvx_gaussian_B_10}
% \end{figure}

% \begin{figure}
%     \centering
%     \includegraphics[width=\linewidth]{figures/convex/paper_cvx_N_500_B_20.png}
%     \caption{Convex optimization exhibits a phase transition when $B=20$. The entries in the nonzero vectors are chosen to be iid $N(0,1)$. The empirical phase transition curve matches the Minimax Risk of BlockSoft Thresholding for $B=20$, to a high degree of accuracy. Each colored pixel depicts the fraction of successes for convex optimization from at least 25 Monte Carlo runs.}
%     \label{fig:cvx_gaussian_B_20}
% \end{figure}

% \begin{figure}
%     \centering
%     \includegraphics[width=\linewidth]{figures/convex/paper_cvx_N_500_B_50.png}
%     \caption{Convex optimization exhibits a phase transition when $B=50$. The entries in the nonzero vectors are chosen to be iid $N(0,1)$. The empirical phase transition curve matches the Minimax Risk of BlockSoft Thresholding for $B=50$, to a high degree of accuracy. Each pixel depicts the fraction of successes for convex optimization from at least 25 Monte Carlo runs.}
%     \label{fig:cvx_gaussian_B_50}
% \end{figure}

\section{SoftSense}\label{sec:softsense}

Before we describe SteinSense, it is useful to consider an algorithm which we call SoftSense (see Algorithm \ref{alg:softsense}). We would like to think of SoftSense as a digital twin of Convex Optimization, and actually this is true in a certain case to be described in Section \ref{sec:array_cs}.

For $y\in\sR^B$, define the BlockSoft Thresholding denoiser by
\begin{align*}
    \eta_\BST(y;\tau) &= \left(1-\dfrac{\tau}{\|y\|}\right)_+\cdot y
\end{align*}BlockSoft Thresholding is the proximal operator for $\ell_2$ norm. That is,
\begin{align*}
    \eta_\BST(y;\tau) &= \argmin_{x\in\sR^B}\left(\|x\|_2 + \dfrac{1}{2\tau}\|y-x\|^2\right)
\end{align*}In statistics, BlockSoft Thresholding is perhaps the most popular generalization of soft thresholding to the case of block sparsity; see \citet{johnstone2002function} for a decision theoretic presentation, and see \citet{yuan2006model} for its use in grouped LASSO. Namely, consider the classical statistical problem of estimating $m_i\in\sR^B$ given $y_i\sim \gN_B(m_i,I_B)$ for $1\leq i \leq N$. If we have reason to believe that several $m_i=0$ (although we do not known which of them), then we may estimate each $m_i$ by $\hat m_i = \eta_\BST(y_i;\tau)$ for an appropriately chosen $\tau>0$.

In the above discussion, the covariance matrix of the $y_i$'s is assumed to be $I_B$. In the case when it is not $I_B$ but some positive definite matrix $\Sigma\in\sR^{B\times B}$, we define the \textit{Colored} BlockSoft Thresholding operator as:
\begin{align*}
    \eta_\CBST(y;\Sigma,\tau) &= \Sigma^{1/2}\eta_\BST(\Sigma^{-1/2}y;\tau)
\end{align*}In other words, we whiten $y$ using $\Sigma^{-1/2}$, apply BlockSoft Thresholding in the whitened coordinates, and then unwhiten. This Colored BlockSoft Thresholding denoiser forms a key piece in Algorithm \ref{alg:softsense}.

% Let $\gP(B)$ denote the class of all probability distributions on $\sR^B$. Consider the class of $\epsilon-$sparse distributions:
% \begin{align*}
%     \gF(\epsilon,B) &= \{\mu\in\gP(B):\mu(\{0\})\geq 1-\epsilon\}
% \end{align*}If $m$ has a prior $\mu\in\sR^B$, then

\begin{algorithm*}
\caption{SoftSense}\label{alg:softsense}
\begin{algorithmic}[1]
     \Require $A\in\sR^{n\times N}$, $Y\in\sR^{n\times B}$, $\{\tau_t\}_{t\geq 0}$ sequence of positive reals
    \State Start with $X^0=0\in\sR^{N\times B}$
    \For{$t\geq 0$}
    \begin{align*}
        R^t &= Y - AX^t + \dfrac{1}{\delta}R^{t-1}\cdot J_\CBST(H^t;S^{t-1},\tau_{t-1})\\
        S^t &= (R^t)^\top (I_n - J_n/n)R^t/n\\
        H^{t+1} &= X^t + A^\top R^t\\
        X^{t+1} &= \eta_\CBST(H^{t+1};S^t,\tau_t)
    \end{align*}
    \EndFor
\end{algorithmic}
\end{algorithm*}

In Algorithm \ref{alg:softsense}, any variable with negative superscript is automatically assumed to be $0$. Also, $\eta_\CBST(H^{t+1};S^t,\tau_t)$ is obtained by applying the denoiser $\eta_\CBST(\cdot;S^t,\tau_t)$ row-wise to $H^{t+1}$, and
\begin{align*}
    J_\CBST(H^t;S^{t-1},\tau_{t-1}) &= \dfrac{1}{N}\sum_{i=1}^N \text{Jac}(\eta_{\CBST})(H^t_{i\star};S^{t-1},\tau_{t-1})^\top
\end{align*}where $Jac(f)(v;\cdots)$ denotes the jacobian matrix of function $f:\sR^B\to\sR^B$ evaluated at $v\in\sR^B$, with $\cdots$ denoting additional parameters passed to $f$. Finally, $J_n$ denotes the $n\times n$ matrix of all $1$'s.

\begin{figure}[htbp]
  \centering

  \begin{subfigure}[b]{0.9\textwidth}
    \includegraphics[width=\linewidth]{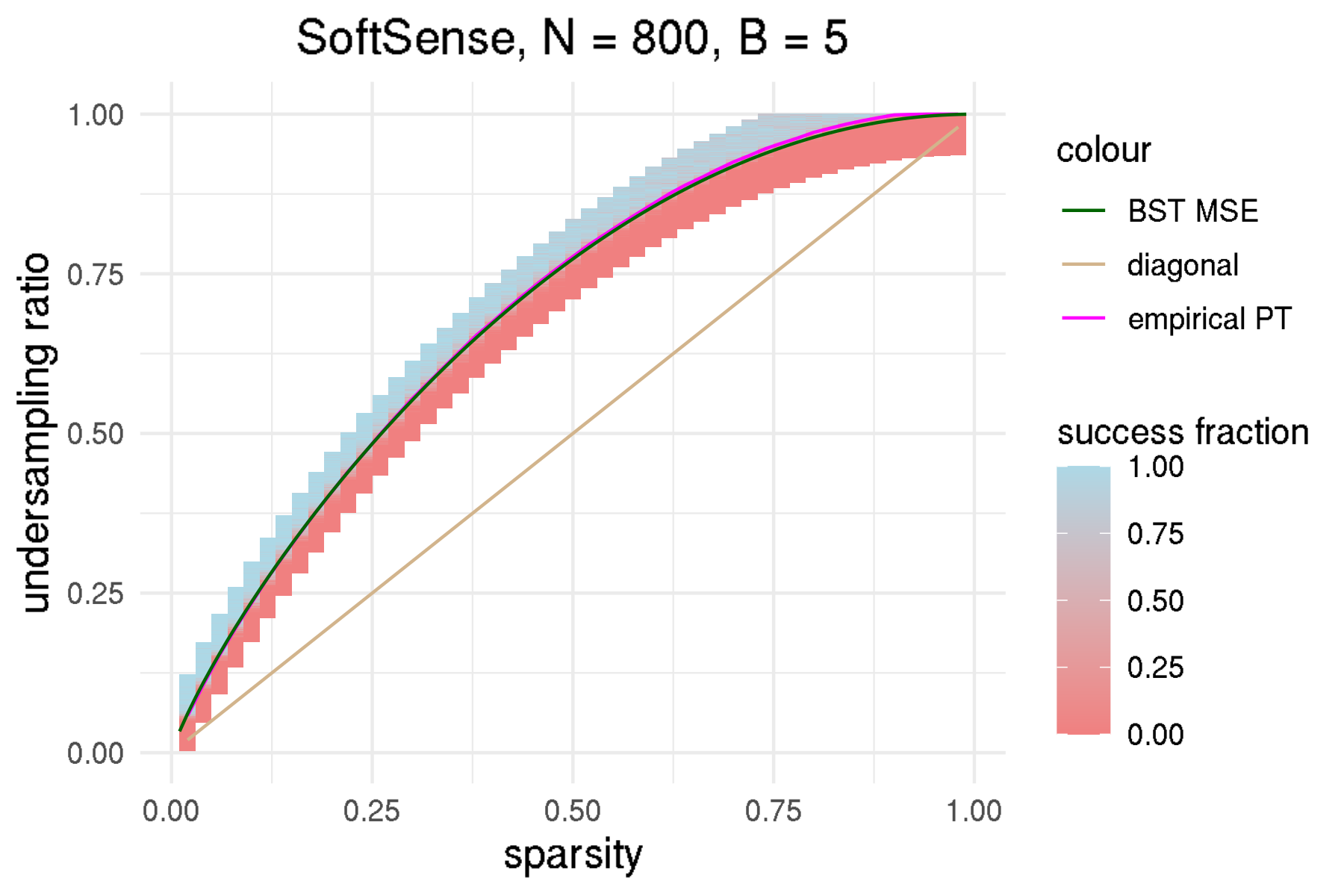}
  \end{subfigure}
  \vspace{1em}
  \begin{subfigure}[b]{0.9\textwidth}
    \includegraphics[width=\linewidth]{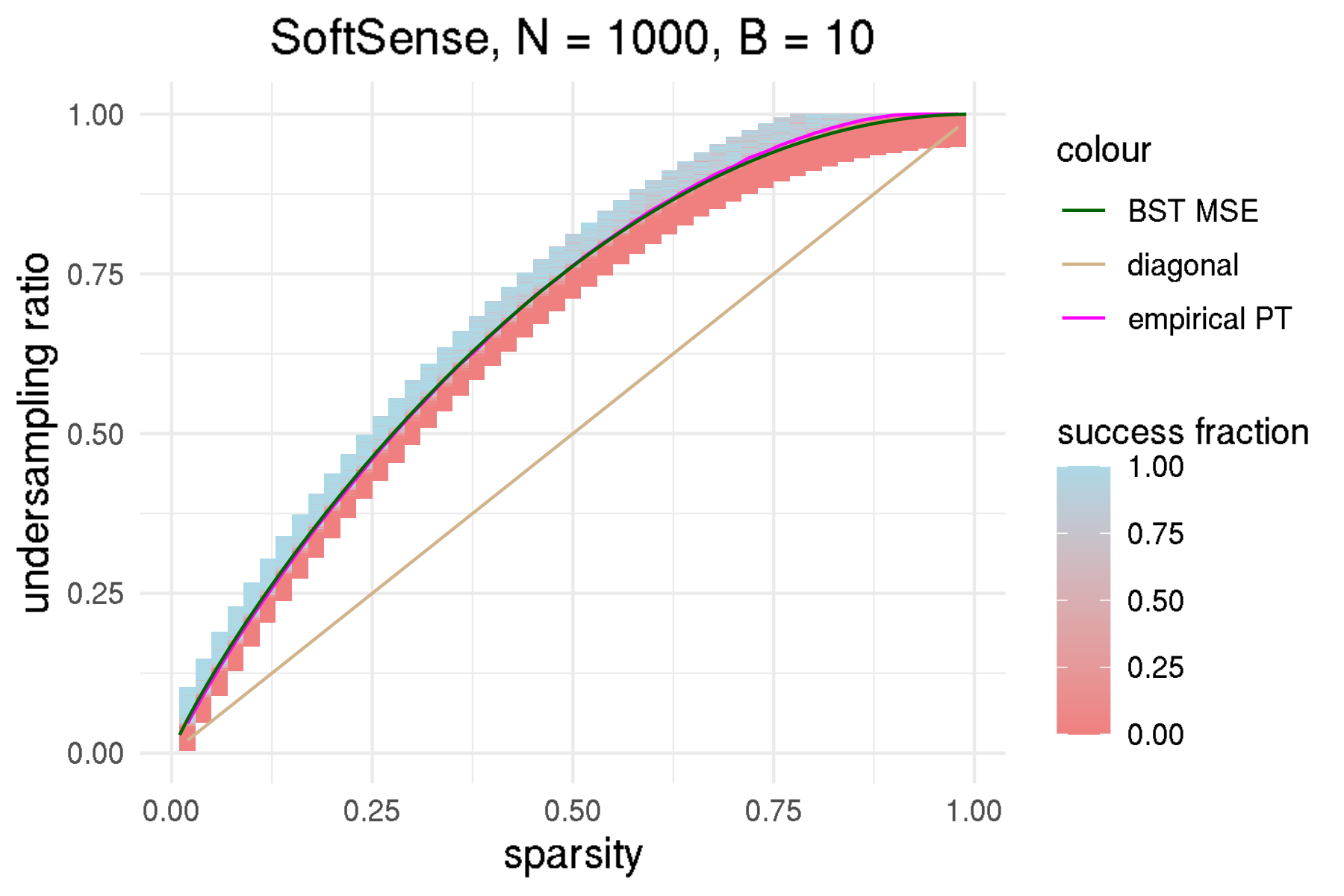}
  \end{subfigure}
  \caption{The nonzero entries in the vectors are chosen to be iid $N(0,1)$. Each colored pixel contains success fraction computed from at least 25 Monte Carlo runs. We see that the empirical phase transition is almost perfectly matching the BlockSoft minimax MSE curve, abbreviated as BST MSE in the figure. This is supported by Theorem \ref{thm:softsense_matching_BST_MSE}.}\label{fig:softsense_jit_normal_B_5_10}
\end{figure}

\begin{figure}[htbp]
  \centering

  \begin{subfigure}[b]{0.9\textwidth}
    \includegraphics[width=\linewidth]{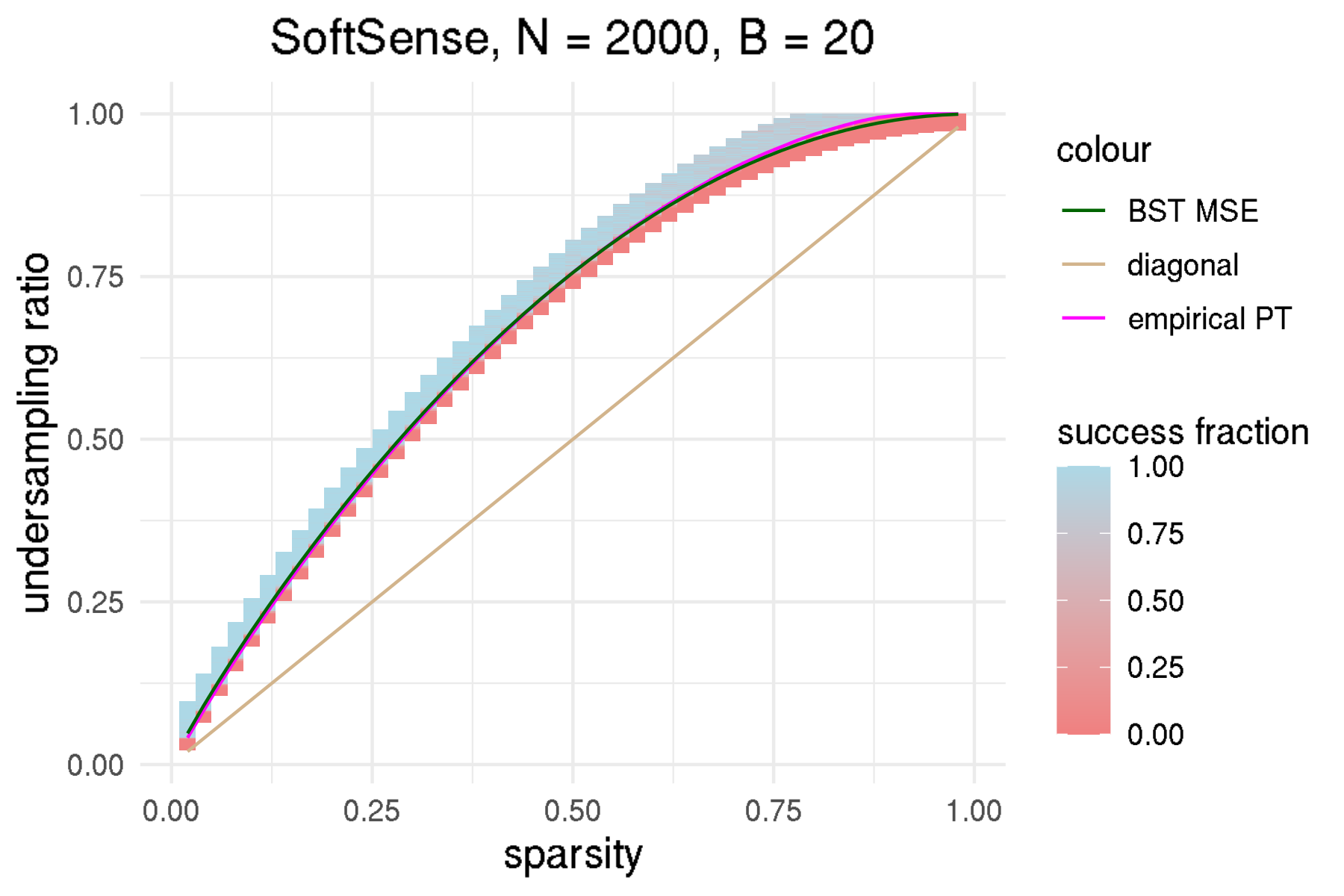}
  \end{subfigure}
  \vspace{1em}
  \begin{subfigure}[b]{0.9\textwidth}
    \includegraphics[width=\linewidth]{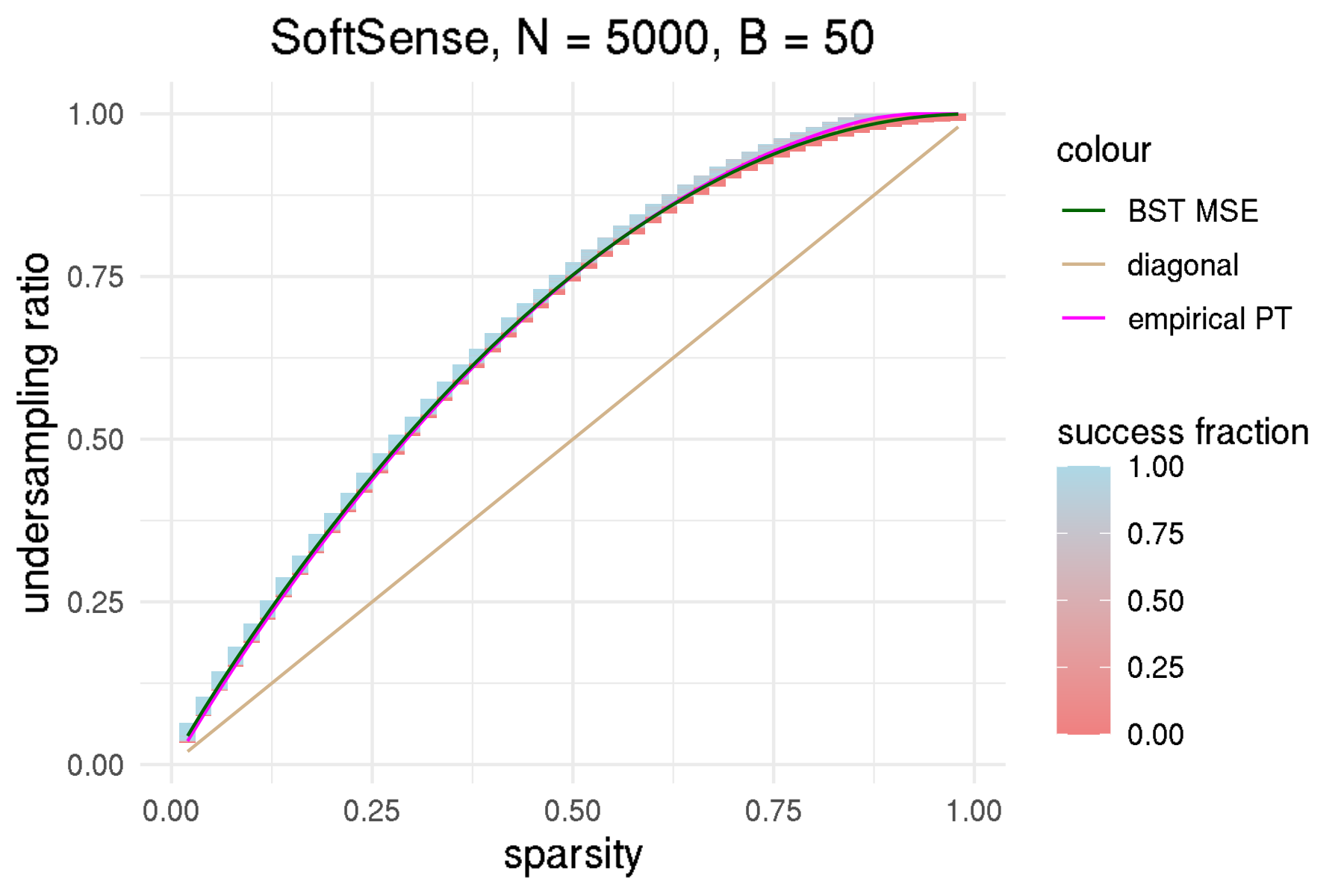}
  \end{subfigure}
  \caption{The nonzero entries in the vectors are chosen to be iid $N(0,1)$. Each colored pixel contains success fraction computed from at least 25 Monte Carlo runs. We see that the empirical phase transition is almost perfectly matching the BlockSoft minimax MSE curve, abbreviated as BST MSE in the figure. This is supported by Theorem \ref{thm:softsense_matching_BST_MSE}.\\
    Notice that there remains a significant gap from the diagonal, even when $B$ is this large.}\label{fig:softsense_jit_normal_B_20_50}
\end{figure}

Figures \ref{fig:softsense_jit_normal_B_5_10} and \ref{fig:softsense_jit_normal_B_20_50} show the performance of SoftSense for different values of $B$ when the nonzero entries in the vectors are taken to be iid $N(0,1)$. For every $B$, there is a clear match between the empirically computed phase transition and an analytically computed curve corresponding to the minimax risk of BlockSoft Thresholding, which we define as follows.
Theorem \ref{thm:softsense_matching_BST_MSE} confirms this, since the nonzero entries for the vectors in these examples come from symmetric exchangeable distributions.

\begin{figure}[htbp]
  \centering
  \begin{subfigure}[b]{0.9\textwidth}
    \includegraphics[width=\linewidth]{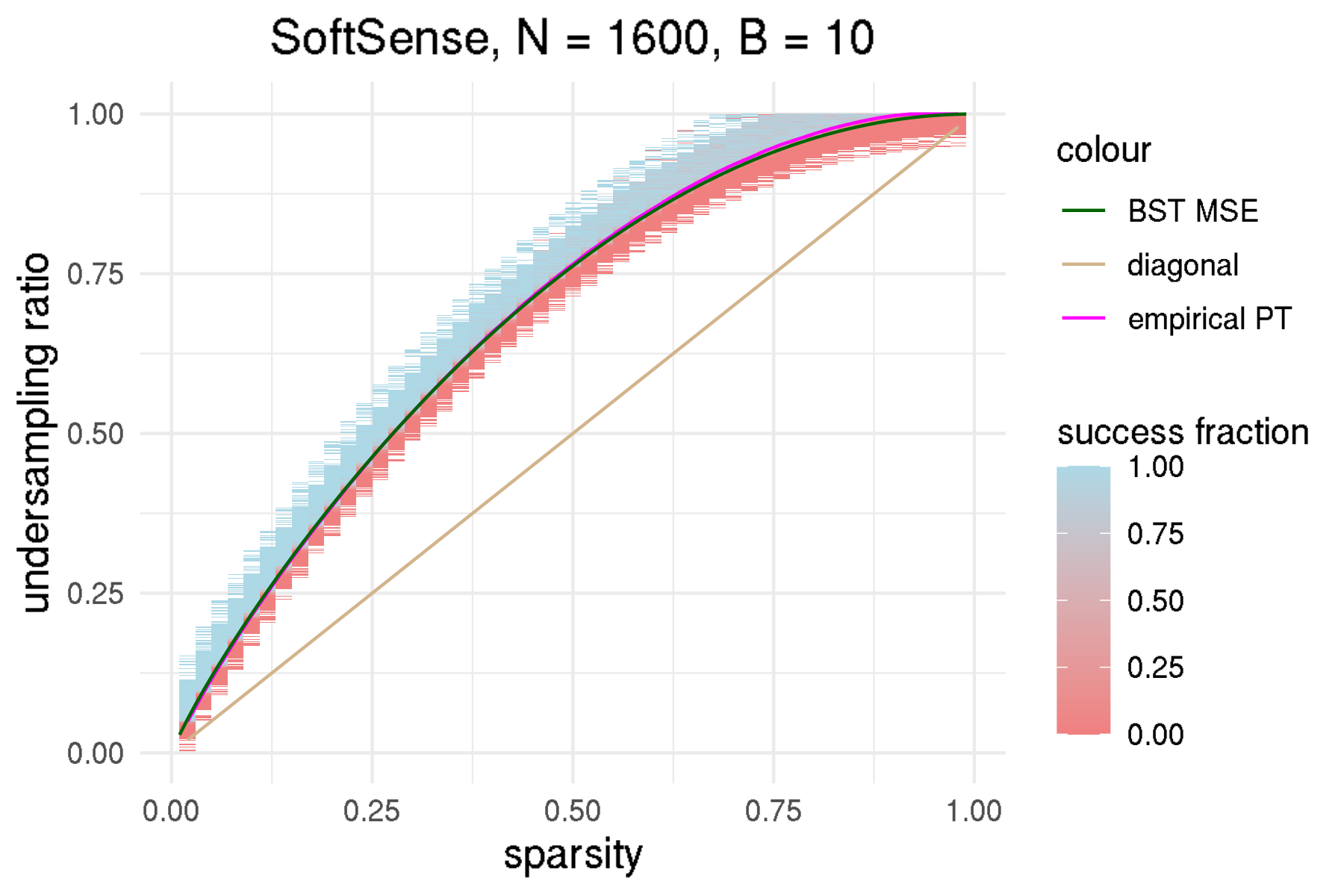}
  \end{subfigure}
  \vspace{1em}
  \begin{subfigure}[b]{0.9\textwidth}
    \includegraphics[width=\linewidth]{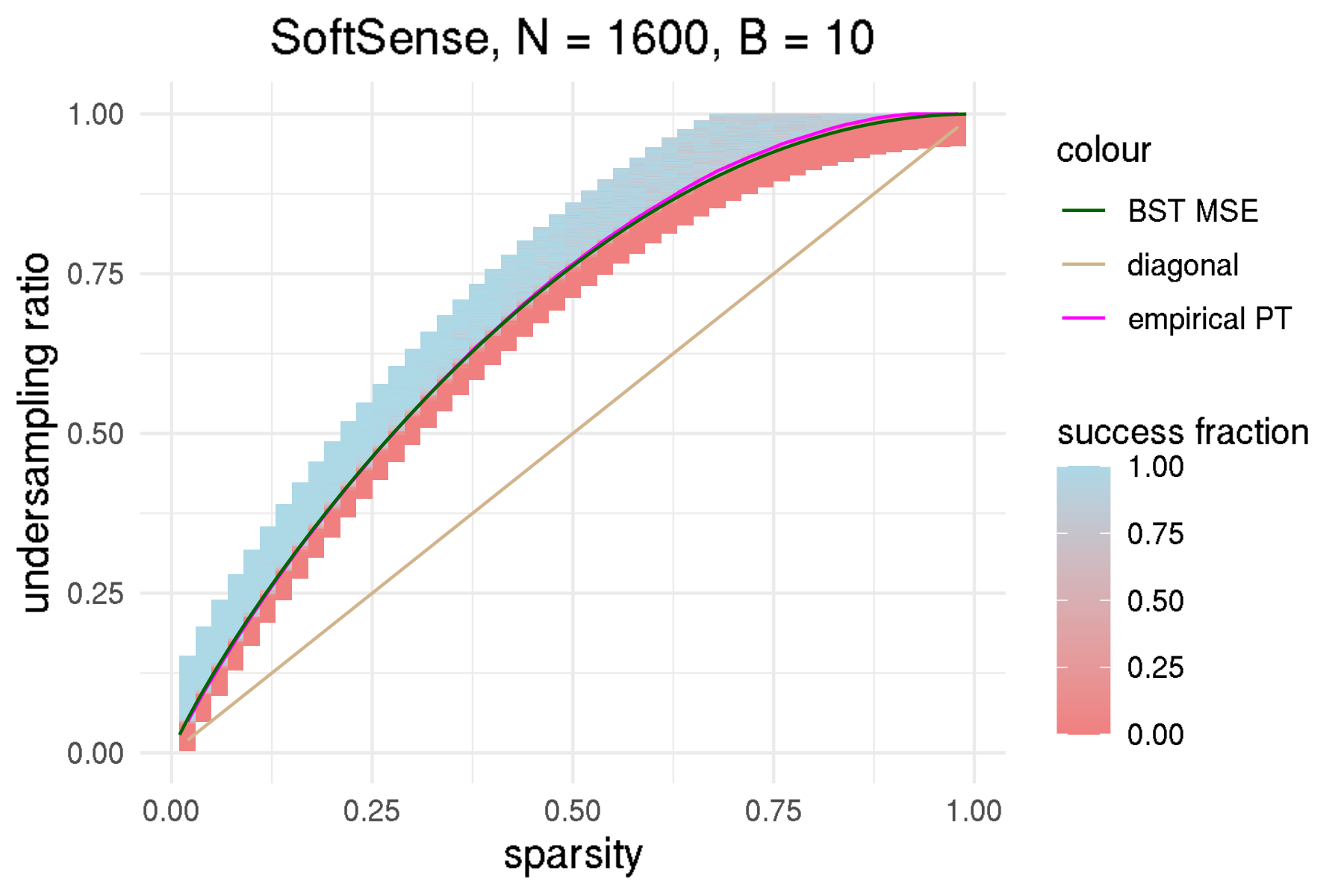}
  \end{subfigure}
  \caption{\textbf{Top.} The nonzero entries in the vectors are chosen to be absolute $N(0,1)$. The distribution of the non-zeros is exchangeable but not symmetric. \textbf{Bottom.} The nonzero entries in the vectors are chosen to be heterogeneous Poissons: $Poi(j)$ in column $j$. The distribution of the non-zeros is neither symmetric nor exchangeable. In both the cases, however, the empirical phase transitions match BlockSoft minimax risk curve to a high degree of accuracy.}\label{fig:softsense_jit_gaussian_absolute_poisson_hetero}
\end{figure}

Fascinatingly, going significantly beyond Theorem \ref{thm:softsense_matching_BST_MSE}, even when the nonzero entries \textbf{do not} come from symmetric exchangeable distributions, we find that the empirical phase transition delivered by SoftSense matches the same analytic BlockSoft minimax risk! This is shown in Figure \ref{fig:softsense_jit_gaussian_absolute_poisson_hetero}. More experiments for smaller $N$ and $B$ are shown in Section \ref{sec:appdx:cvx_more_experiments}.

% \begin{figure}
%     \centering
%     \includegraphics[width=\linewidth]{figures/softsense/paper_softsense_jit_N_1600_B_10_gaussian_absolute.png}
%     \caption{The nonzero entries in the vectors are chosen to be absolute $N(0,1)$. Each colored pixel contains success fraction computed from at least 25 Monte Carlo runs. We see that the empirical phase transition is almost perfectly matching the BlockSoft minimax MSE curve, abbreviated as BST MSE in the figure. This is an example where the conditions of Theorem \ref{thm:softsense_matching_BST_MSE} do not hold as the distribution of the nonzeros is not symmetric.}
%     \label{fig:paper_softsense_jit_N_1600_B_10_gaussian_absolute}
% \end{figure}

% \begin{figure}
%     \centering
%     \includegraphics[width=\linewidth]{figures/softsense/paper_softsense_jit_N_1600_B_10_poisson_hetero.png}
%     \caption{The nonzero entries in the vectors are chosen to be heterogeneous Poissons: $Poi(j)$ in column $j$. Each colored pixel contains success fraction computed from at least 25 Monte Carlo runs. We see that the empirical phase transition is almost perfectly matching the BlockSoft minimax MSE curve, abbreviated as BST MSE in the figure. This is an example where the conditions of Theorem \ref{thm:softsense_matching_BST_MSE} do not hold as the distribution of the non0zeros is neither symmetric nor exchangeable.}
%     \label{fig:paper_softsense_jit_N_1600_B_10_poisson_hetero}
% \end{figure}
\section{SteinSense, and Reaching the Diagonal}

From Theorem \ref{thm:softsense_matching_BST_MSE}, the performance of SoftSense (at least in the symmetric exchangeable case) can be attributed to the minimax risk of BlockSoft Thresholding, the denoiser employed by SoftSense. Therefore, it is conceivable that if we have a denoiser with a better minimax risk, we might be able to outperform SoftSense. Our main deliverable algorithm, SteinSense, which we present in Algorithm \ref{alg:steinsense}, indeed achieves this goal. It is a simple modification of SoftSense - it replaces the BlockSoft Thresholding denoiser by the James Stein denoiser $\eta_\JS:\sR^B\to\sR^B$ which is defined as follows:
\begin{align*}
    \eta_\JS(y) &= \left(1 - \dfrac{B-2}{\|y\|^2}\right)_+\cdot y
\end{align*}The James Stein estimator was developed about seventy years back in works of \citet{stein1956inadmissibility,james1961estimation}, and provides a uniformly better estimator than the maximum likelihood estimator $y\in\sR^B$ when estimating the mean $m\in\sR^B$ given $y\sim \gN_B(m,I_B)$ for $B>2$. Following the same principle outlined in the description of SoftSense, if the covariance matrix is not $I_B$ but some positive definite matrix $\Sigma$, we define the Colored James Stein denoiser:
\begin{align*}
    \eta_\CJS(y) &= \Sigma^{1/2}\eta_\JS(\Sigma^{-1/2}y)
\end{align*}

\begin{algorithm*}
\caption{SteinSense}\label{alg:steinsense}
\begin{algorithmic}[1]
     \Require $A\in\sR^{n\times N}$, $Y\in\sR^{n\times B}$
    \State Start with $X^0=0\in\sR^{N\times B}$
    \For{$t\geq 0$}
    \begin{align*}
        R^t &= Y - AX^t + \dfrac{1}{\delta}R^{t-1}\cdot J_\CJS(H^t;S^{t-1})\\
        S^t &= (R^t)^\top (I_n - J_n/n)R^t/n\\
        H^{t+1} &= X^t + A^\top R^t\\
        X^{t+1} &= \eta_\CJS(H^{t+1};S^t)
    \end{align*}
    \EndFor
\end{algorithmic}
\end{algorithm*}

In Algorithm \ref{alg:steinsense}, any variable with negative superscript is automatically assumed to be $0$. Also, $\eta_\CJS(H^{t+1};S^t)$ is obtained by applying the denoiser $\eta_\CJS(\cdot;S^t)$ row-wise to $H^{t+1}$, and
\begin{align*}
    J_\CJS(H^t;S^{t-1}) &= \dfrac{1}{N}\sum_{i=1}^N \text{Jac}(\eta_{\CJS})(H^t_{i\star};S^{t-1})^\top
\end{align*}

Clearly, SteinSense is a simple modification of SoftSense, both emplying very simple denoisers, and thus there is no concern for added computational complexity (which is, generally, a real concern when $N$ and $B$ are large) on going from SoftSense to SteinSense. 

Remarkably, for every sparsity value $\epsilon$, as $B$ grows, SteinSense not only outperforms SoftSense, but actually achieves oracle performance, that is, reaches the diagonal. Further, this happens without using any knowledge of the sparsity level, without any tuning parameter, without any knowledge of the distribution of the non-zero entries, and with simply an iid Gaussian measurement matrix.

\begin{figure}[htbp]
  \centering

  \begin{subfigure}[b]{0.9\textwidth}
    \includegraphics[width=\linewidth]{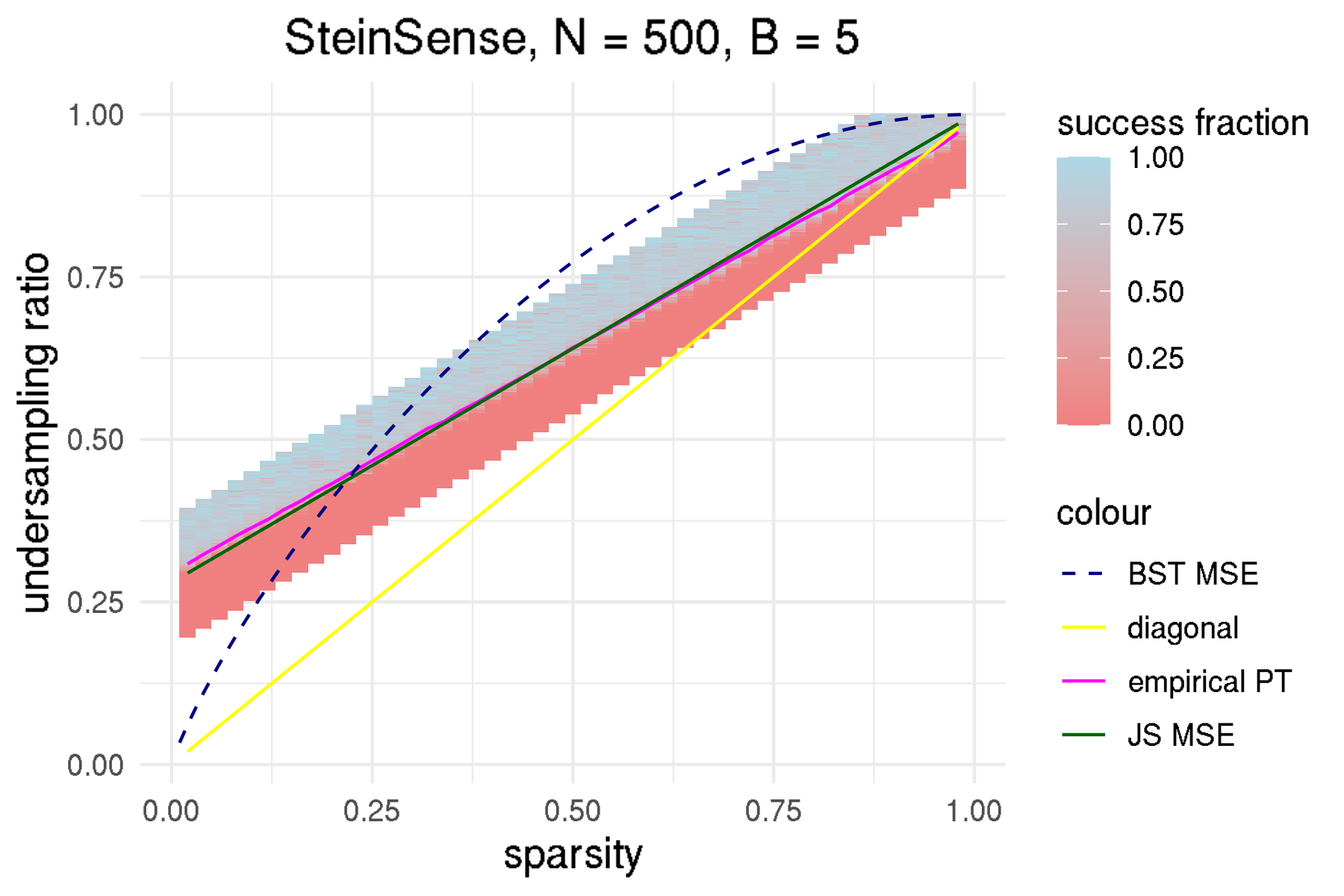}
  \end{subfigure}
  \vspace{1em}
  \begin{subfigure}[b]{0.9\textwidth}
    \includegraphics[width=\linewidth]{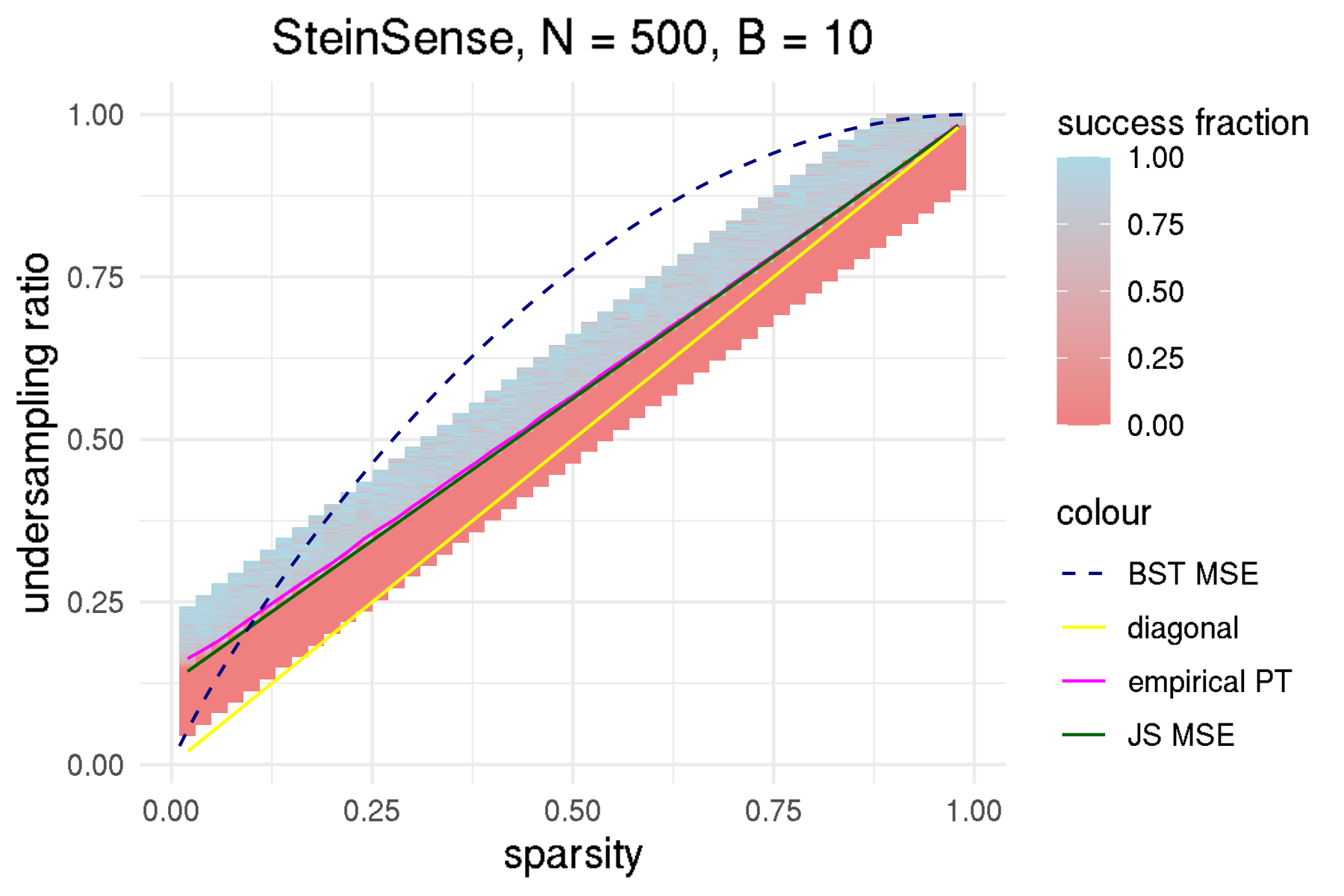}
  \end{subfigure}
  \caption{The nonzero entries in the vectors are chosen to be iid $N(0,1)$. Each colored pixel
contains success fraction computed from at least 25 Monte Carlo runs. The
empirical phase transition almost perfectly matches the James Stein minimax MSE curve,
abbreviated as JS MSE in the figure. The dashed curve corresponding to BlockSoft minimax risk is added for reference. SteinSense outperforms SoftSense for the majority of the sparsity values, and the region where SteinSense outperforms SoftSense enlarges as we go from $B=5$ to $B=10$.}\label{fig:steinsense_normal_B_5_10}
\end{figure}

\begin{figure}[htbp]
  \centering

  \begin{subfigure}[b]{0.9\textwidth}
    \includegraphics[width=\linewidth]{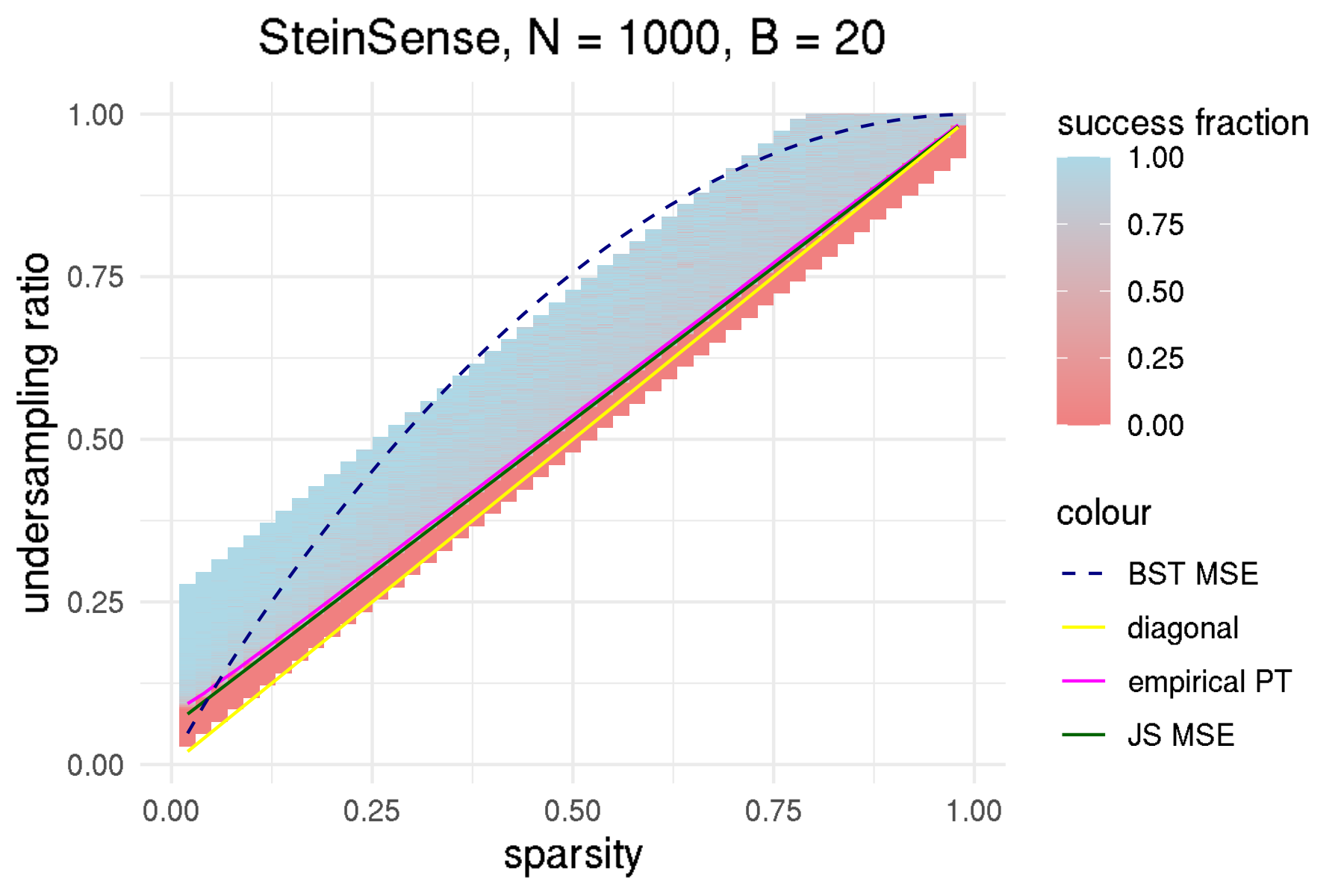}
  \end{subfigure}
  \vspace{1em}
  \begin{subfigure}[b]{0.9\textwidth}
    \includegraphics[width=\linewidth]{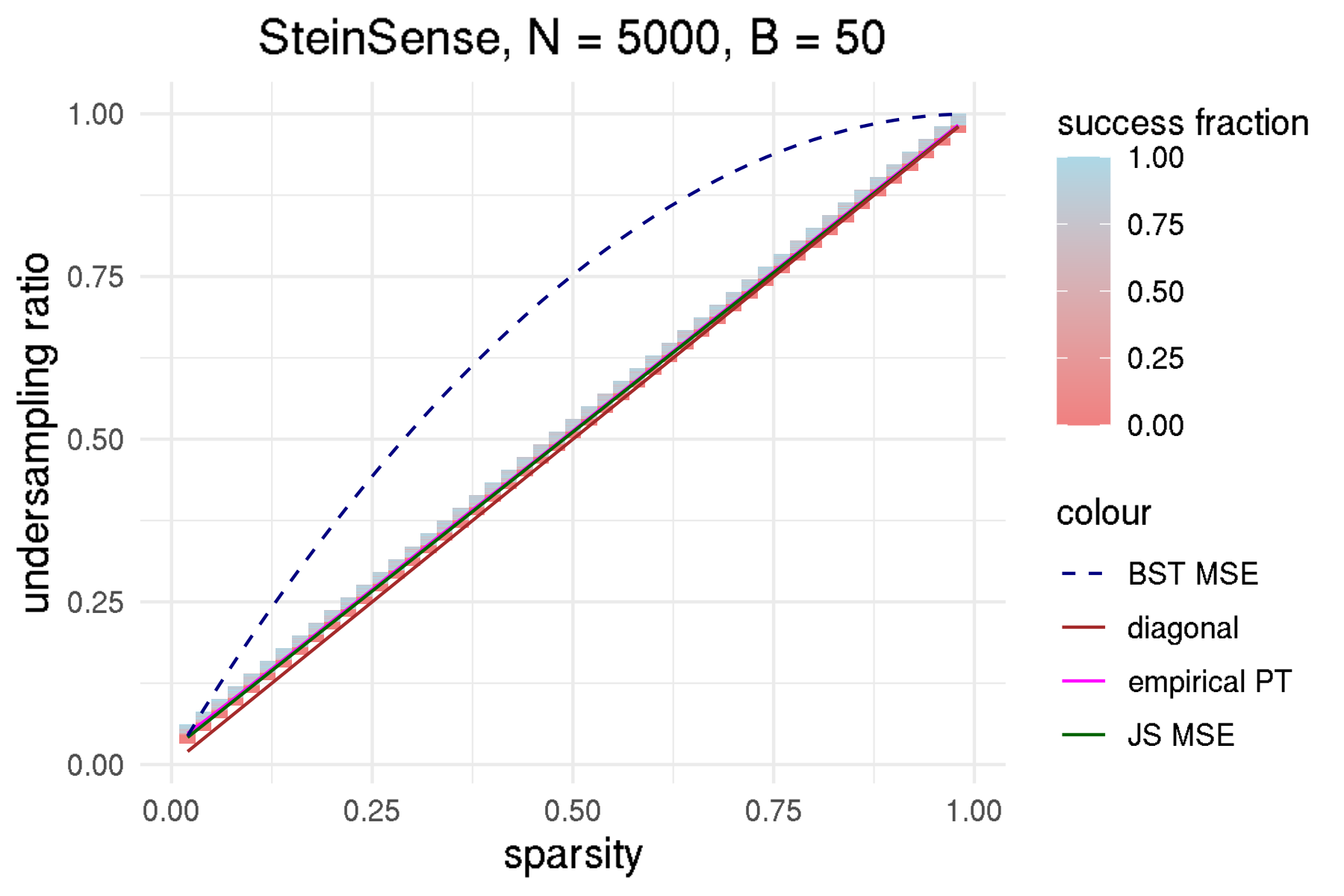}
  \end{subfigure}
  \caption{The nonzero entries in the vectors are chosen to be iid $N(0,1)$. Each colored pixel
contains success fraction computed from at least 25 Monte Carlo runs. We see that the
empirical phase transition is almost perfectly matching the James Stein minimax MSE curve. For the overwhelming majority of the sparsity values, SteinSense wins. For $B=50$, we have practically reached the diagonal.}\label{fig:steinsense_normal_B_20_50}
\end{figure}

Figures \ref{fig:steinsense_normal_B_5_10} and \ref{fig:steinsense_normal_B_20_50} present the performance of SteinSense when the nonzero entries of the vectors are chosen to be iid $N(0,1)$. These plots establish that the empirical phase transition of SteinSense matches the James Stein minimax risk to a high degree to accuracy, abbreviated as JS MSE in the plots, for every $B$ and for very moderate $N$. This is supported by Theorem \ref{thm:js_mse}. In particular, we have virtually reached the diagonal, for \textit{just} $B=50$. Appendix \ref{sec:appdx:cvx_more_experiments} contains more experimental results on SteinSense.

\begin{figure}[htbp]
  \centering

  \begin{subfigure}[b]{0.9\textwidth}
    \includegraphics[width=\linewidth]{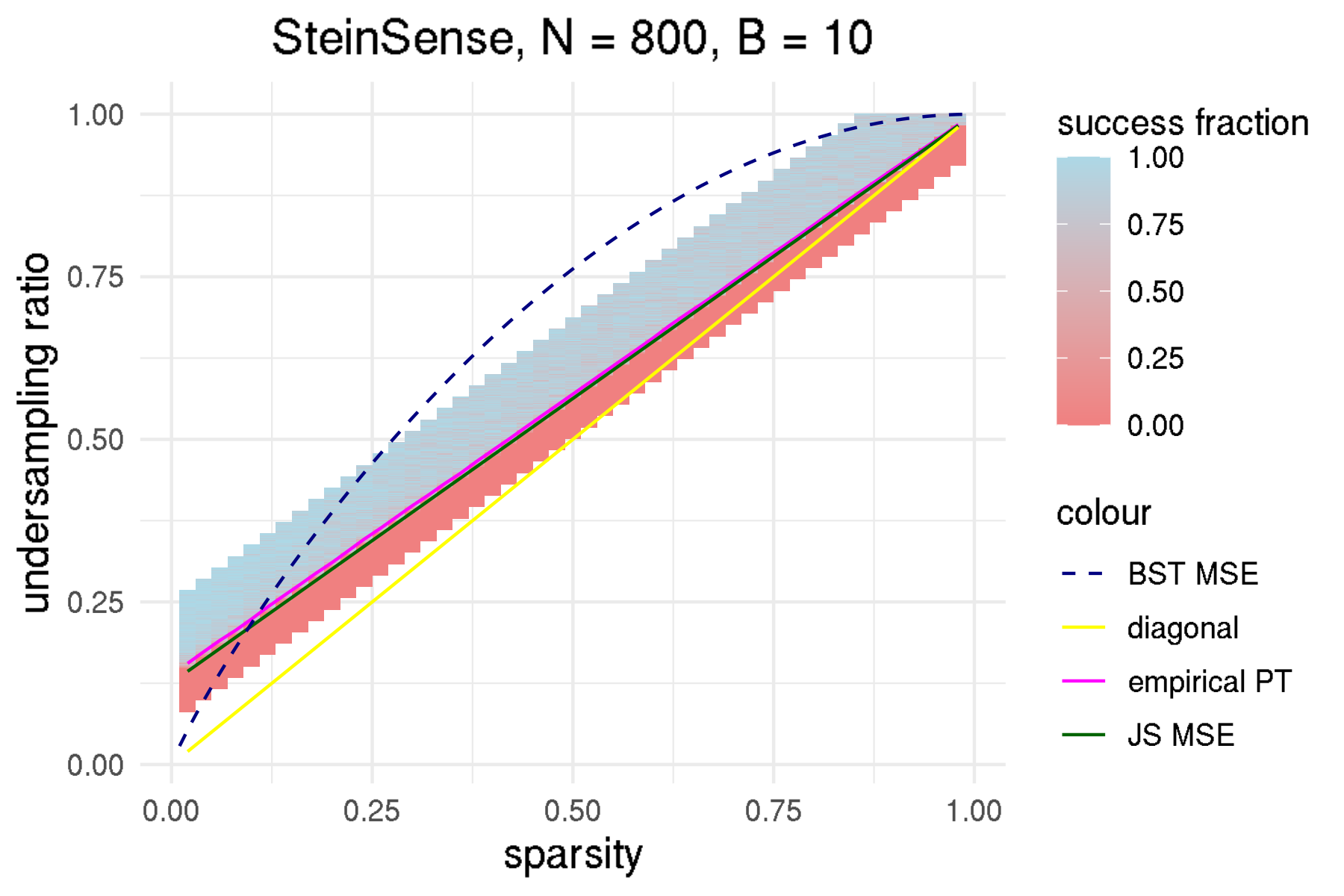}
  \end{subfigure}
  \vspace{1em}
  \begin{subfigure}[b]{0.9\textwidth}
    \includegraphics[width=\linewidth]{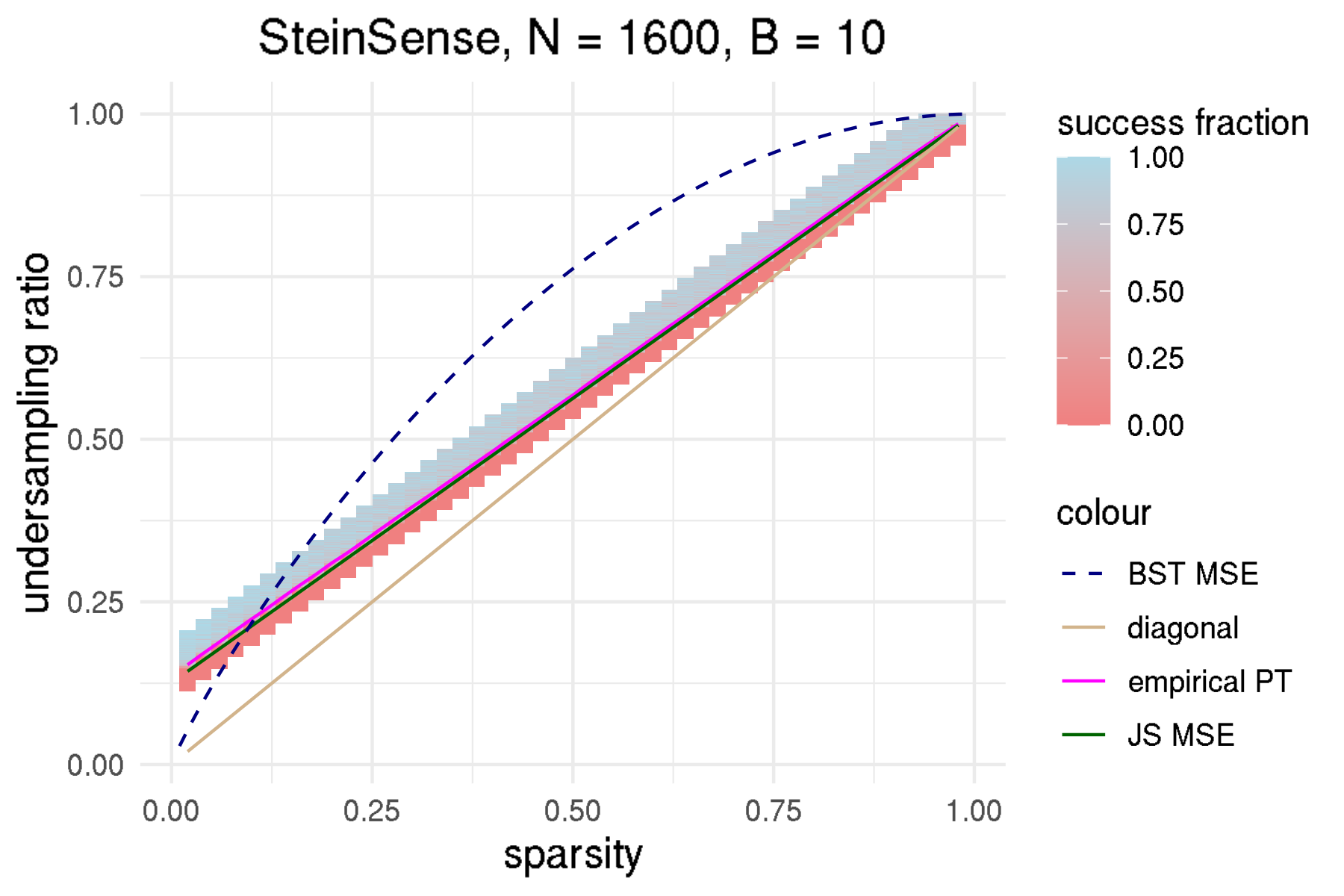}
  \end{subfigure}
  \caption{\textbf{Top.} The nonzero entries are iid absolute $N(0,1)$. \textbf{Bottom.} The nonzero entries are iid Exponential with rate $5$. In neither case are the nonzero entries symmetric. Still we find the empirical phase transition to closely match the James Stein minimax risk curve.}\label{fig:steinsense_gaussian_absolute_exponential_5}
\end{figure}

\begin{figure}[htbp]
  \centering

  \begin{subfigure}[b]{0.9\textwidth}
    \includegraphics[width=\linewidth]{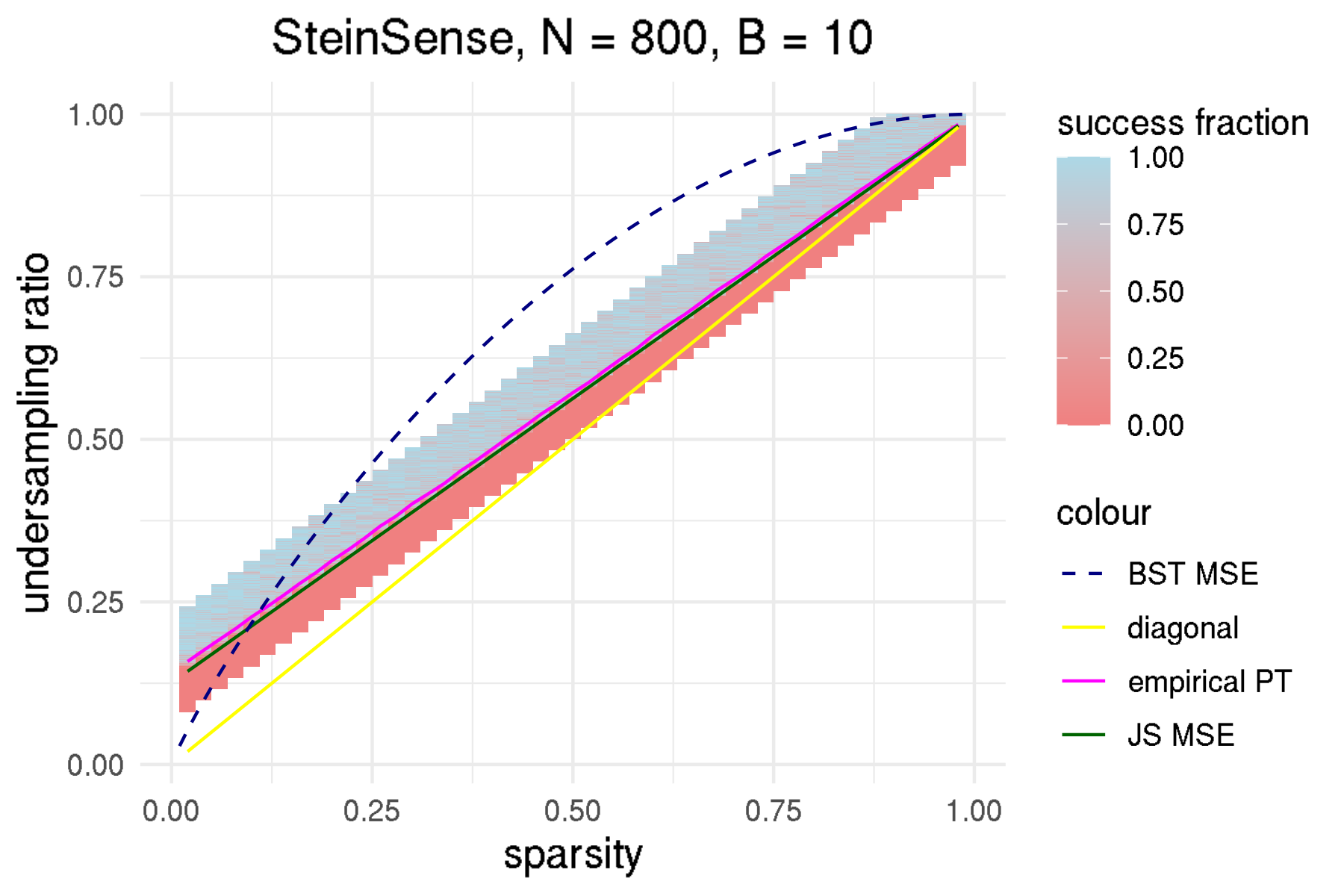}
  \end{subfigure}
  \vspace{1em}
  \begin{subfigure}[b]{0.9\textwidth}
    \includegraphics[width=\linewidth]{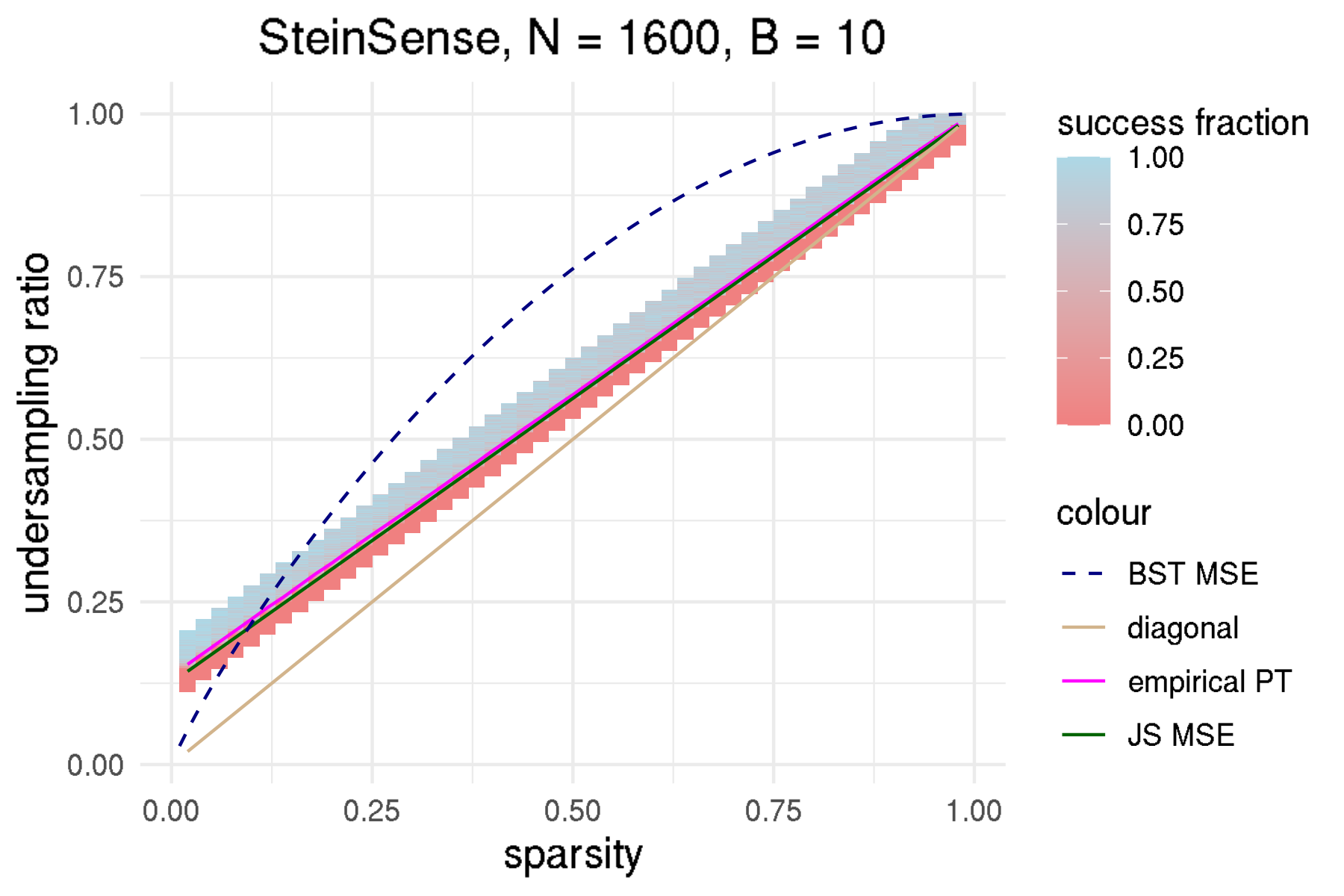}
  \end{subfigure}
  \caption{\textbf{Top.} The nonzero entries are chosen as heterogeneous Poisson: the $j$'th column has nonzeros drawn from $Poisson(j)$. \textbf{Bottom.} The nonzero entries in the 10 columns are accordingly $N(0,1)$,$\text{Logistic}(0,1)$, $\text{Laplace}(0,1)$, $t(5)$, $\text{Triangular}(-1,0,1)$, $N(0,50)$, $\text{Laplace}(0,100)$, $\text{Logistic}(0,10)$, $t(10)$ and $Triangular(-500,0,500)$. In either case, we see no significant difference between empirical phase transition and James Stein minimax risk.}\label{fig:steinsense_poisson_hetero_mix_match}
\end{figure}

We stress test SteinSense on situations that are not covered by Theorem \ref{thm:js_mse}. Once again, fascinatingly, SteinSense delivers the same phase transition each time; the empirical phase transition curves always match the James Stein minimax risk curves! Figures \ref{fig:steinsense_gaussian_absolute_exponential_5} and \ref{fig:steinsense_poisson_hetero_mix_match} certify this.

\paragraph{Achieving the diagonal for free.} The plots, particularly Figure \ref{fig:steinsense_normal_B_20_50}, show that SteinSense effectively reaches the diagonal and obtains oracle performance at very moderate values of $N$ and $B$. This point is worth a discussion; precise characterizations are deferred to Remark \ref{rmk:diagonal_achievement}. For $B=1$, using generic procedures and measurements, it is not possible to reach the diagonal. \cite{donoho2013information} is able to reach the diagonal through a rather specialized method, with the measurement matrix and denoiser specific to the distribution of nonzero entries. However, what we observe as $B$ grows, is that with very generic measurements (iid gaussian measurement matrix $A$) and by using a very generic denoiser that is completely oblivious to the distribution of the non-zeros, we can effectively reach the diagonal. 

It is certainly possible to do even better at finite $B$ by using the Bayes denoiser. But that would require one to know the distribution of the non-zeros precisely. Further, computation of the Bayes denoiser can add non-trivial complexity to the per iteration cost. There certainly are powerful deep learning based denoisers, but again, they would need knowledge of sparsity and can be computationally challenging to integrate into the iterative procedure. Moreover, it is difficult to get formal guarantees on such complex denoisers. Another so-far understated point is that we also need to compute a fairly large number of jacobians per iteration. Jacobian computation requires evaluation of the denoiser at multiple values, increasing manifold the computational hurdles if the denoiser is not simple enough. Indeed, in our computational experiments, we have observed this is a major computational bottleneck that needs to be overcome using advanced software.

The point of SteinSense is that it is a very lightweight procedure aimed to eliminate the need of any specialized or computationally heavy method, since when $B$ is large, it is impossible to beat SteinSense. Indeed, precisely when $B$ is large, other specialized methods are expected to falter. If one aims to use a bayes denoiser, one would need to estimate a $B$-dimensional density using $N$ samples, and that would be notoriously hard. When $B$ is large, if evaluations of the denoiser get more cumbersome, computation of many $B\times B$ jacobians will also suffer significantly.

\section{Real Data Experiments}\label{sec:real_exp}

The purpose of this section is to establish that SteinSense works beautifully on real data, which are clearly not at all covered by any theory. We demonstrate this through phase transition plots on several real datasets, broadly classified into two groups.

\subsection{Hyperspectral Image}

\begin{figure}
  \centering
  \begin{subfigure}[b]{0.6\textwidth}
    \includegraphics[width=\textwidth]{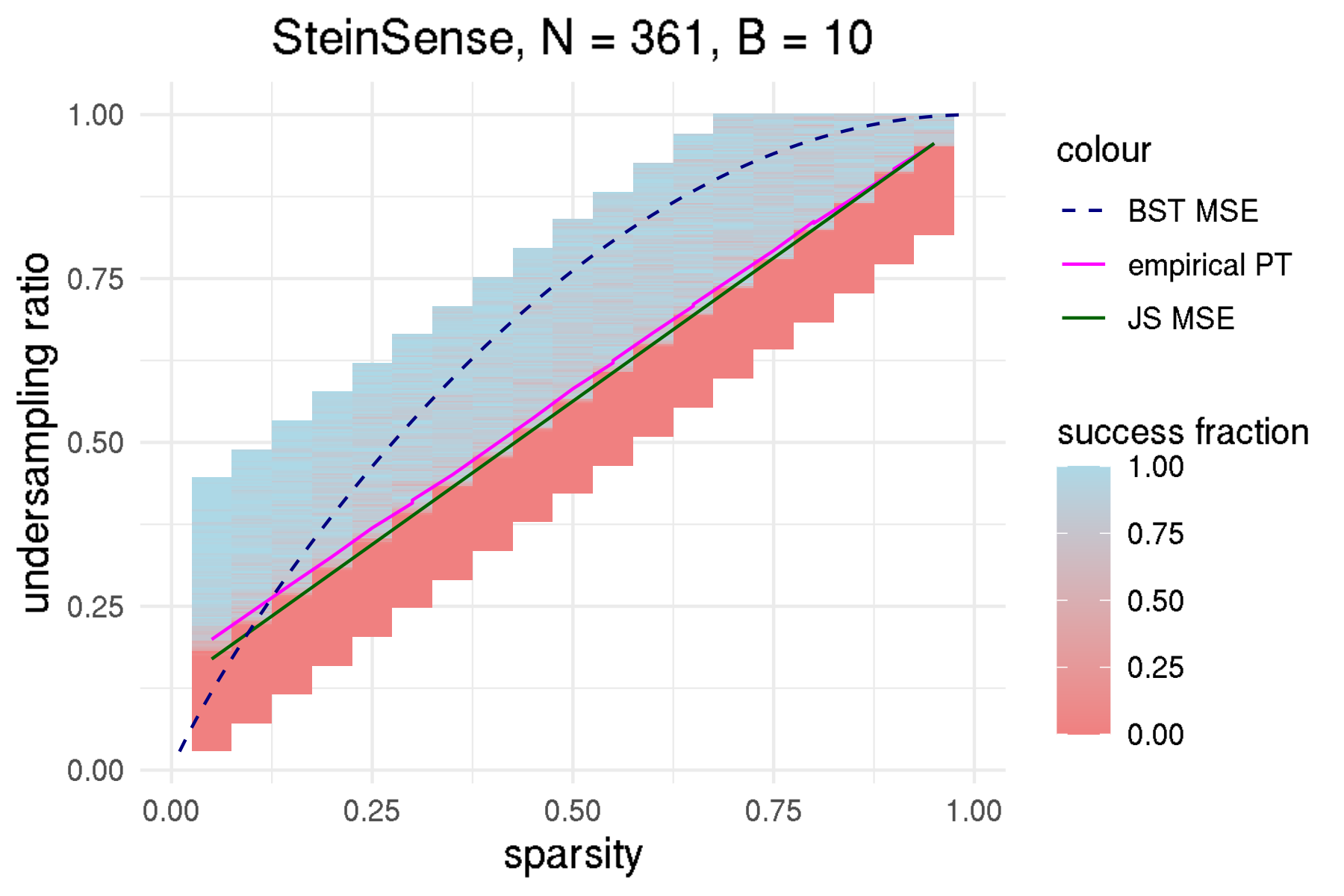}
  \end{subfigure}
  \hfill
  \begin{subfigure}[b]{0.6\textwidth}
    \includegraphics[width=\textwidth]{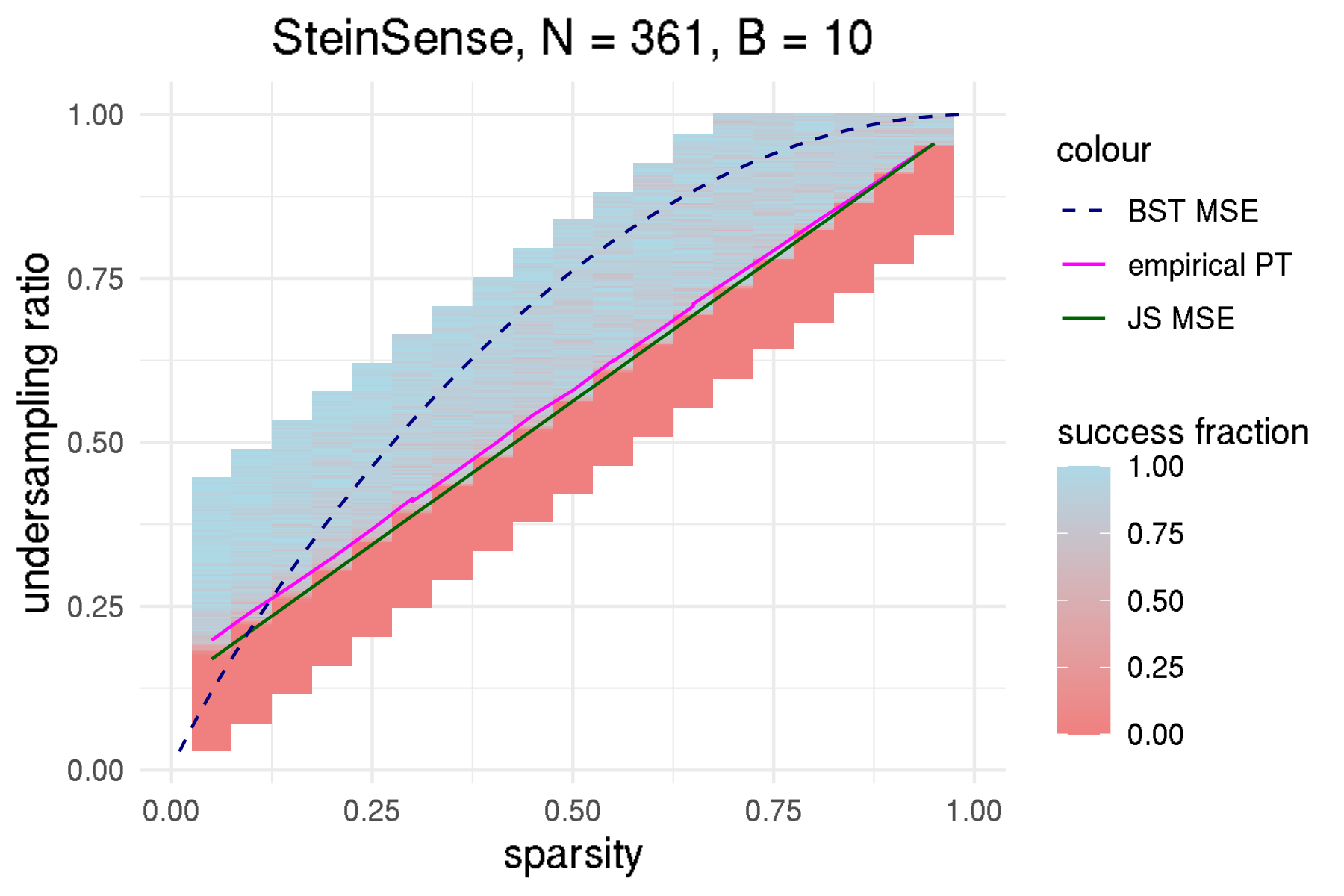}
  \end{subfigure}
  \hfill
  \begin{subfigure}[b]{0.6\textwidth}
    \includegraphics[width=\textwidth]{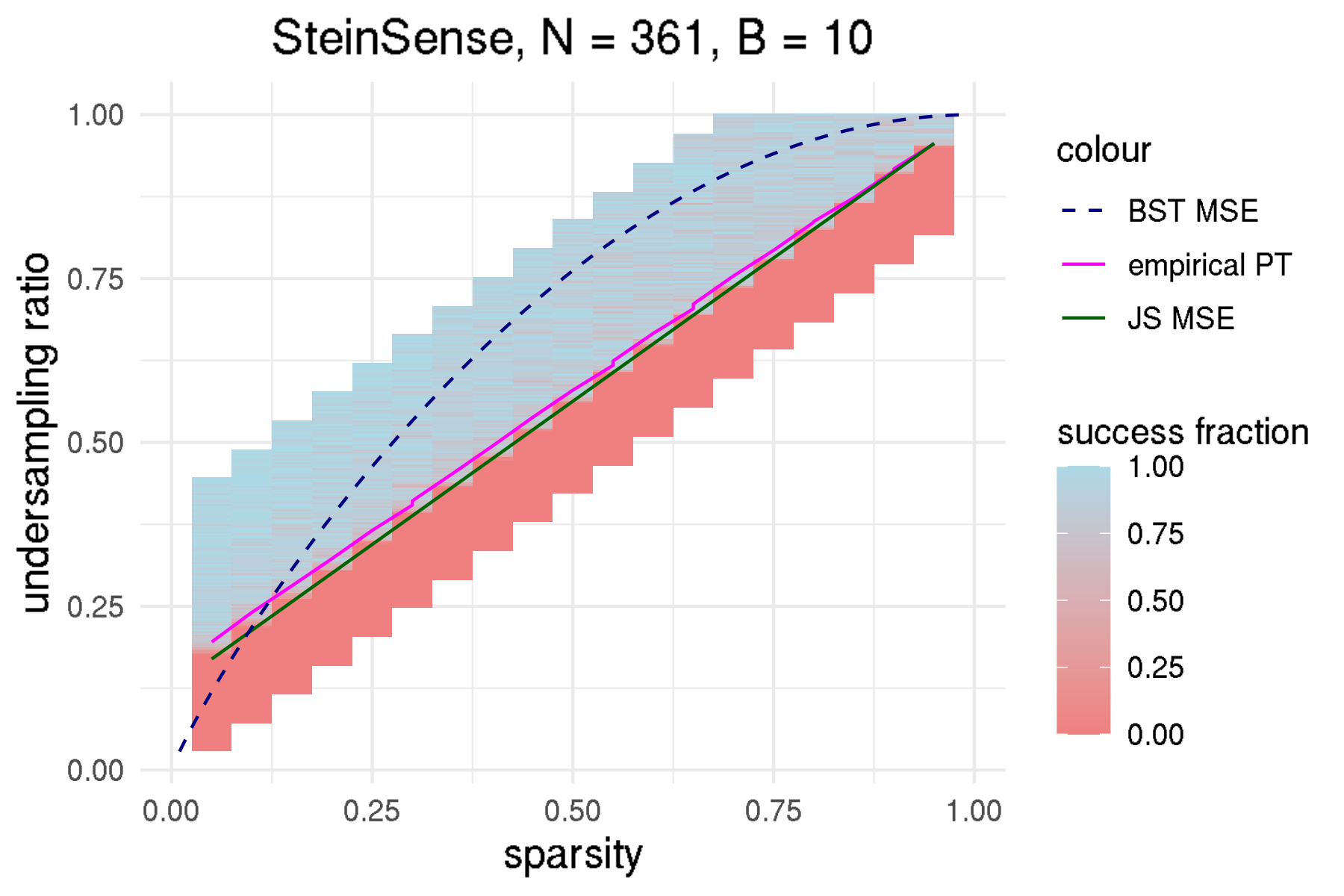}
  \end{subfigure}
  \caption{Plots showing phase transition of SteinSense applied to subbands 0, 1 and 2 of band 1 wavelet coefficients computed on the Indian Pines hyperspectral dataset. We see that the correspondence between the empirical phase transitions and James Stein minimax risk curves is pretty good already at $N=361$.}\label{fig:hyperspectral_band_1}
\end{figure}

\begin{figure}
  \centering
  \begin{subfigure}[b]{0.6\textwidth}
    \includegraphics[width=\textwidth]{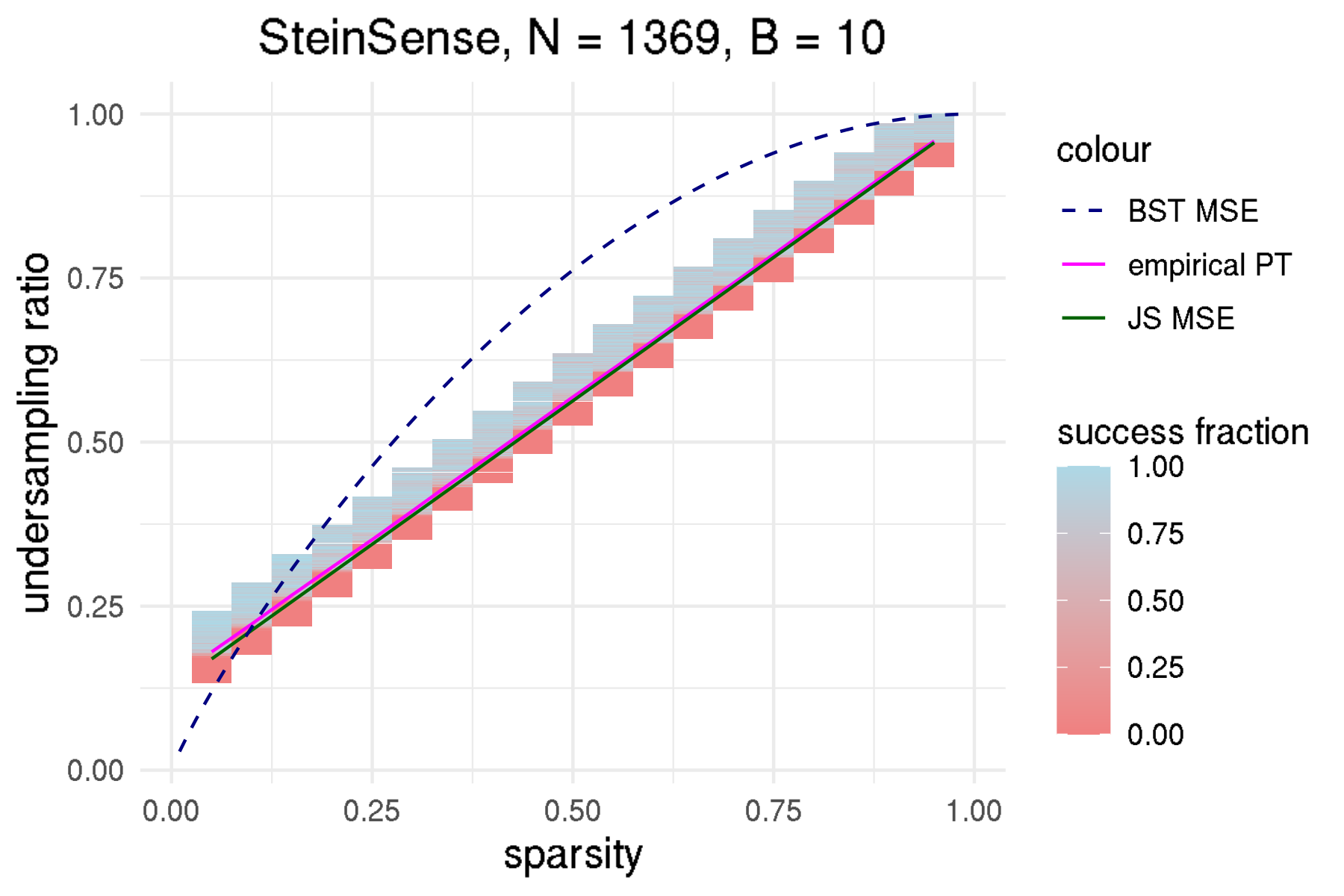}
  \end{subfigure}
  \hfill
  \begin{subfigure}[b]{0.6\textwidth}
    \includegraphics[width=\textwidth]{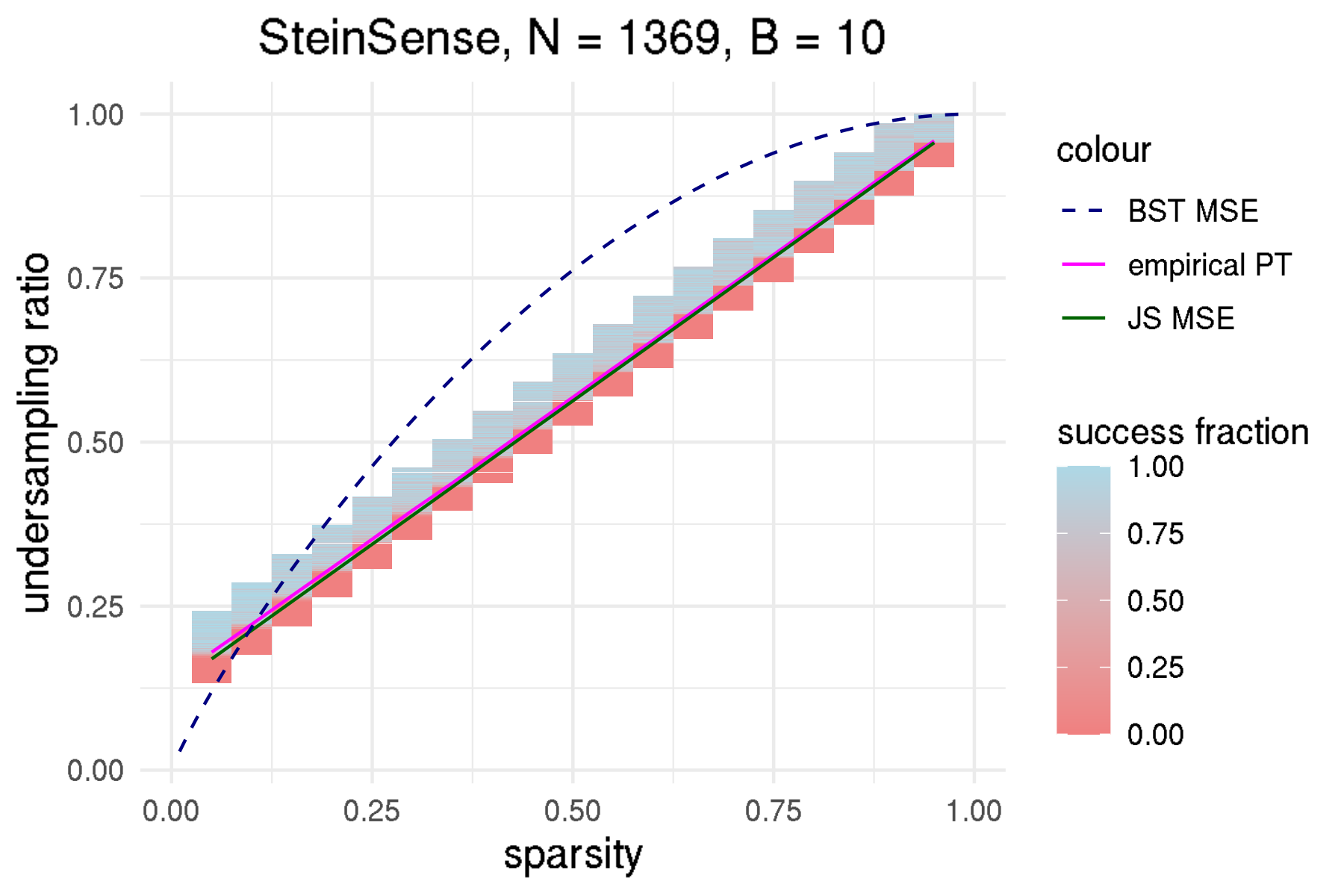}
  \end{subfigure}
  \hfill
  \begin{subfigure}[b]{0.6\textwidth}
    \includegraphics[width=\textwidth]{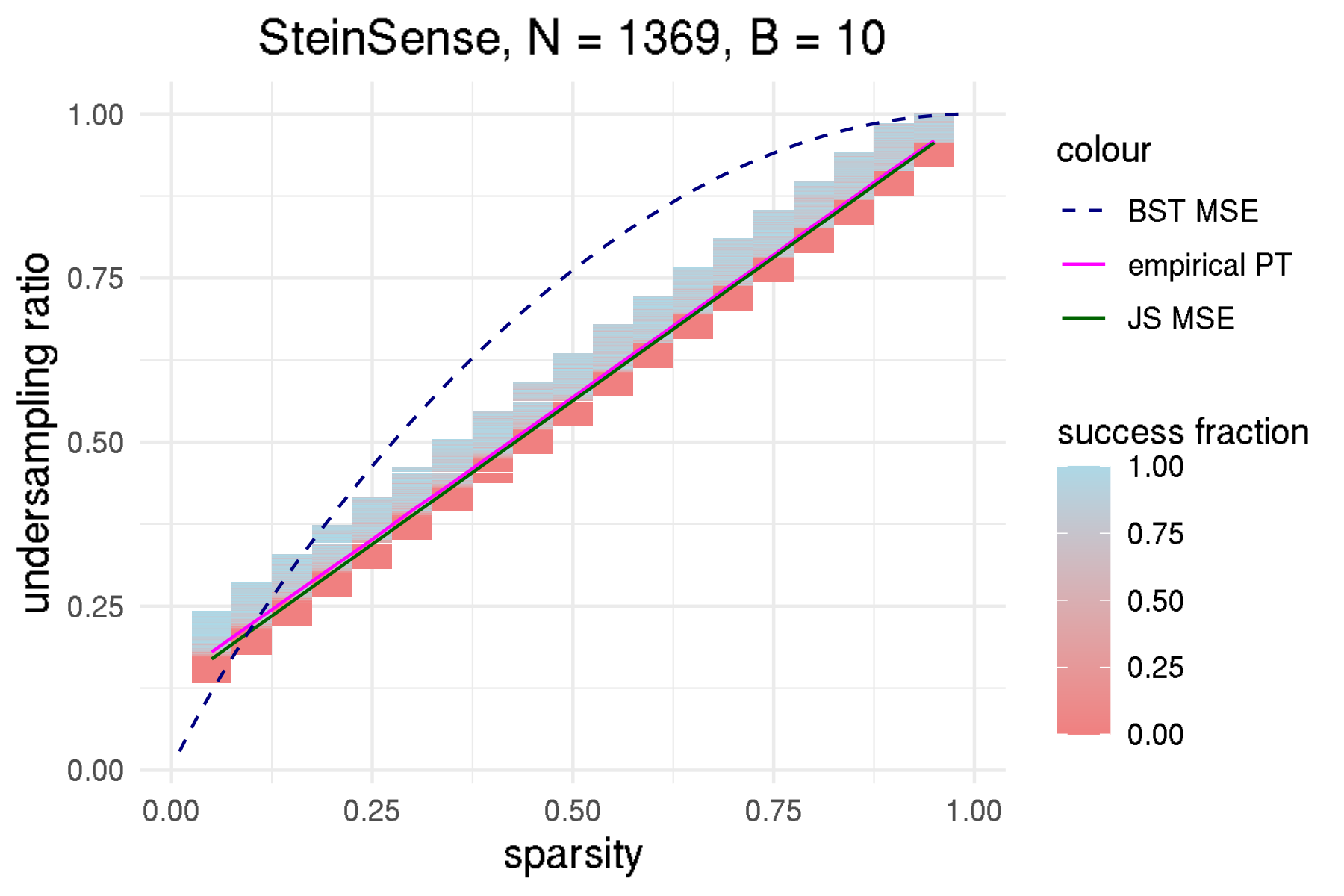}
  \end{subfigure}
  \caption{Plots showing phase transition of SteinSense applied to subbands 0, 1 and 2 of band 2 wavelet coefficients computed on the Indian Pines hyperspectral dataset. We now see that the empirical phase transition curves match the James Stein minimax risk curves to a high degree of accuracy.}\label{fig:hyperspectral_band_2}
\end{figure}

\begin{figure}
  \centering
  \begin{subfigure}[b]{0.6\textwidth}
    \includegraphics[width=\textwidth]{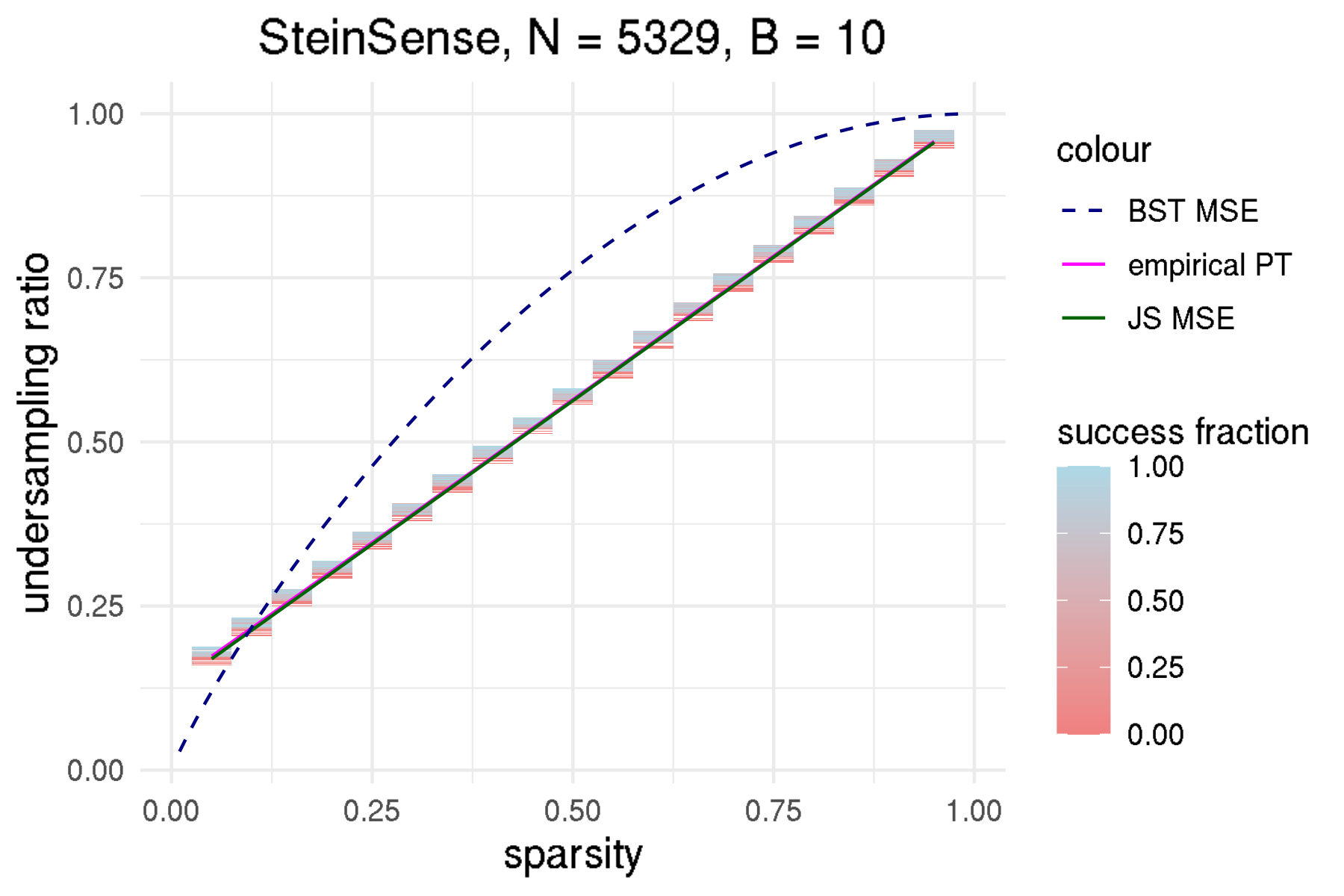}
  \end{subfigure}
  \hfill
  \begin{subfigure}[b]{0.6\textwidth}
    \includegraphics[width=\textwidth]{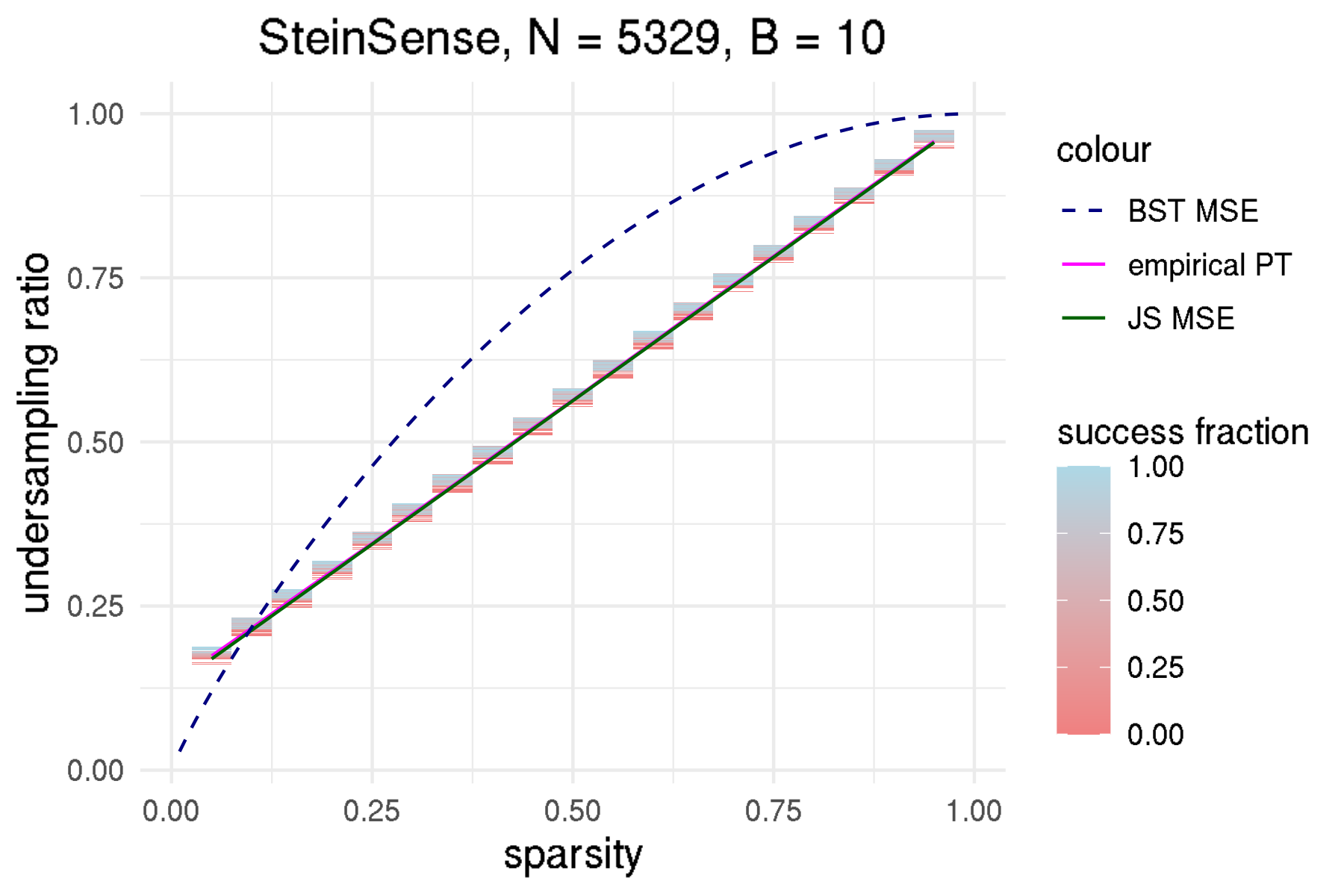}
  \end{subfigure}
  \hfill
  \begin{subfigure}[b]{0.6\textwidth}
    \includegraphics[width=\textwidth]{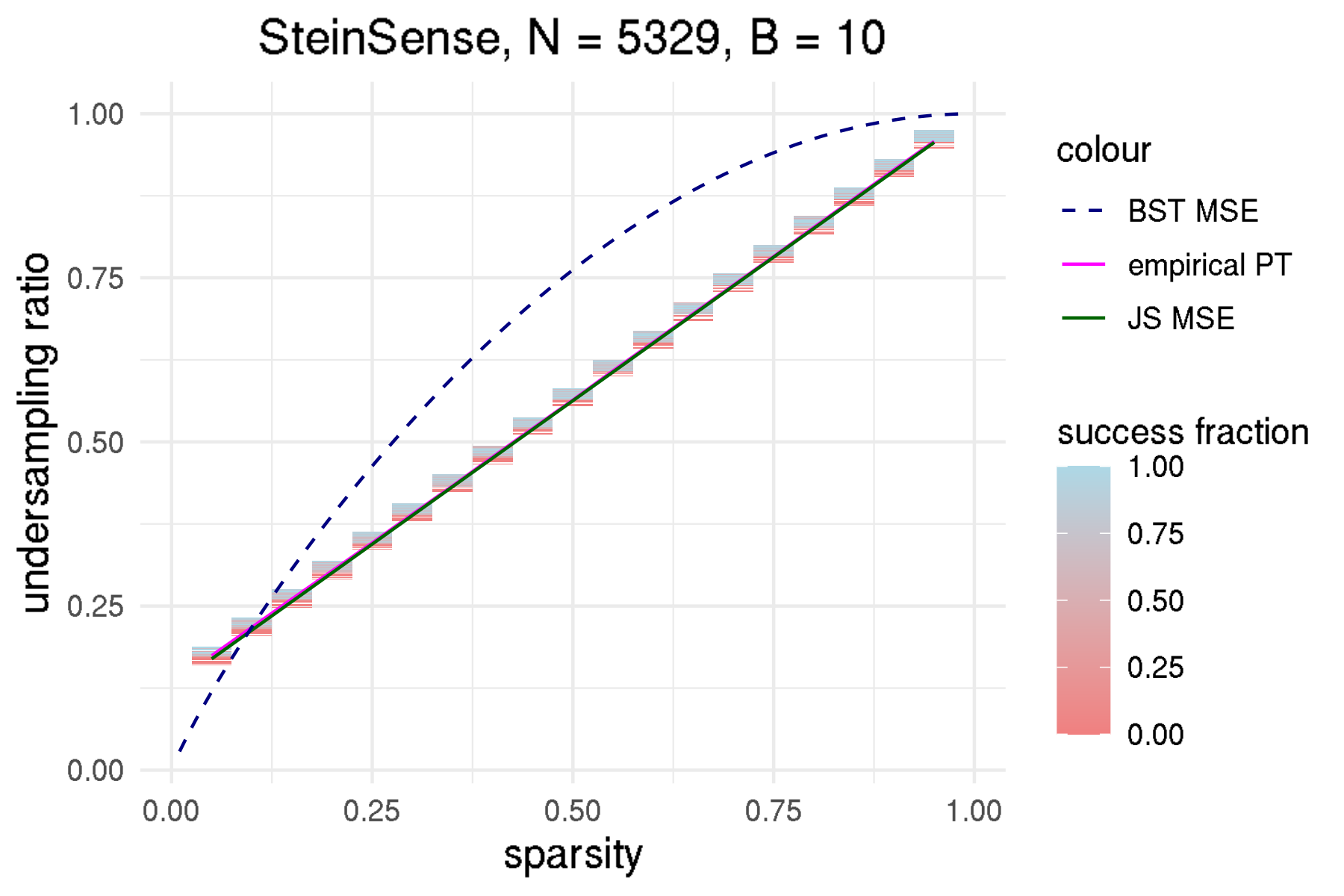}
  \end{subfigure}
  \caption{Plots showing phase transition of SteinSense applied to subbands 0, 1 and 2 of band 3 wavelet coefficients computed on the Indian Pines hyperspectral dataset. We see that the empirical phase transition curves match the James Stein minimax risk curves to a high degree of accuracy.}\label{fig:hyperspectral_band_3}
\end{figure}

We consider the publicly available Indian Pines hyperspectral dataset\footnote{https://paperswithcode.com/dataset/indian-pines}. It has 220 spectral bands, each containing a $145\times 145$ image. In natural images, sparsity is achieveable after some transformation. We take a pixel-wise Haar wavelet decomposition along the spectral direction and then perform a db2 wavelet decomposition at level 3 on each resulting slice. We perform the following experiment to get the phase transition for SteinSense on such data. At every subband of every band, we first randomly select $B=10$ spectra, keep the top \textit{sparsity} proportion of coefficients in magnitude (this sorting is done based on all the coefficients for that subband for thes selected spectra), zero out the rest of the coefficients, flatten each spectral face, and treat this as our matrix $X$ to be compressively sensed. Sparsity is varied in $\{0.05,0.1,\cdots, 0.95\}$. Different bands have different number of rows $N$: $N=361$ for band $1$, $N=1369$ for band $2$, and $N=5329$ for band $3$. At each sparsity value, we consider integers $n$ (corresponding to number of measurements) in a band around $N\delta_{\Stein}(\epsilon,B)$, where $\delta_\Stein(\epsilon,B)\equiv M_\JS(\epsilon,B)$ is the minimax risk of James Stein. Figures \ref{fig:hyperspectral_band_1}, \ref{fig:hyperspectral_band_2} and \ref{fig:hyperspectral_band_3} show the results. SteinSense performs exactly as predicted by Theorem \ref{thm:js_mse}, although the dataset is completely real now.

\subsection{RNASeq datasets}

\begin{figure}[htbp]
  \centering

  % Row 1
  \begin{subfigure}[b]{0.48\textwidth}
    \includegraphics[width=\linewidth]{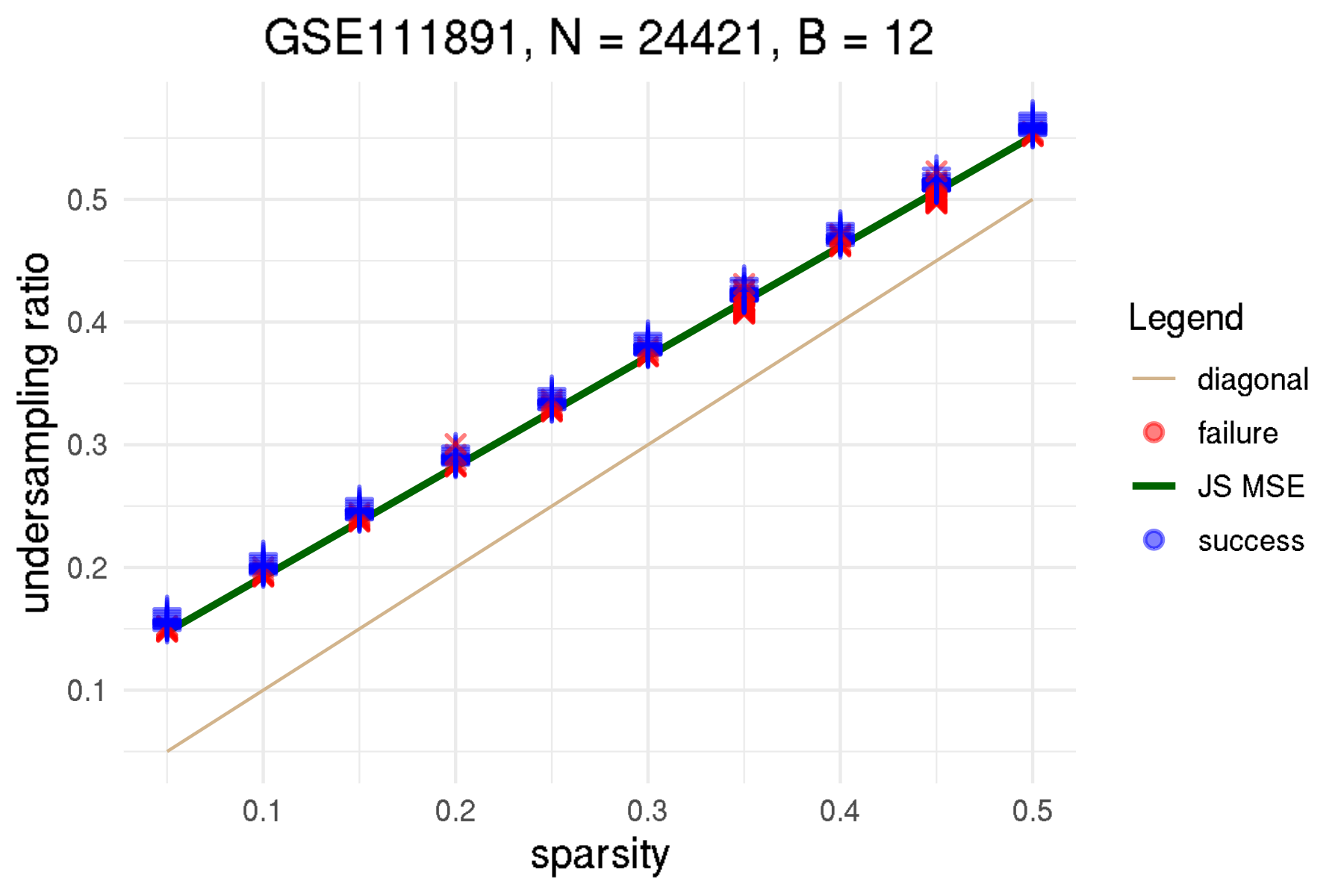}
  \end{subfigure}
  \hfill
  \begin{subfigure}[b]{0.48\textwidth}
    \includegraphics[width=\linewidth]{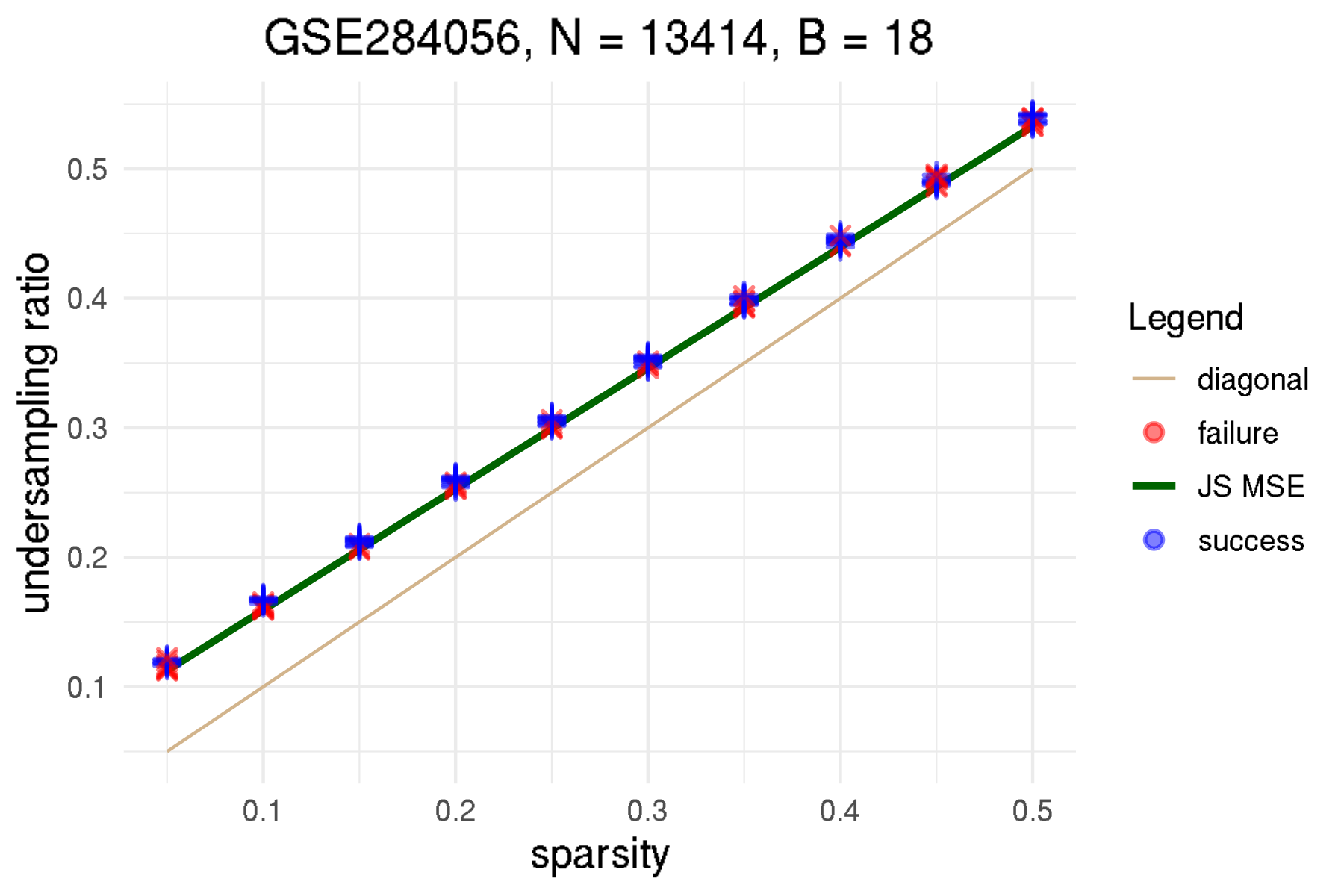}
  \end{subfigure}

  \vspace{1em}

  % Row 2
  \begin{subfigure}[b]{0.48\textwidth}
    \includegraphics[width=\linewidth]{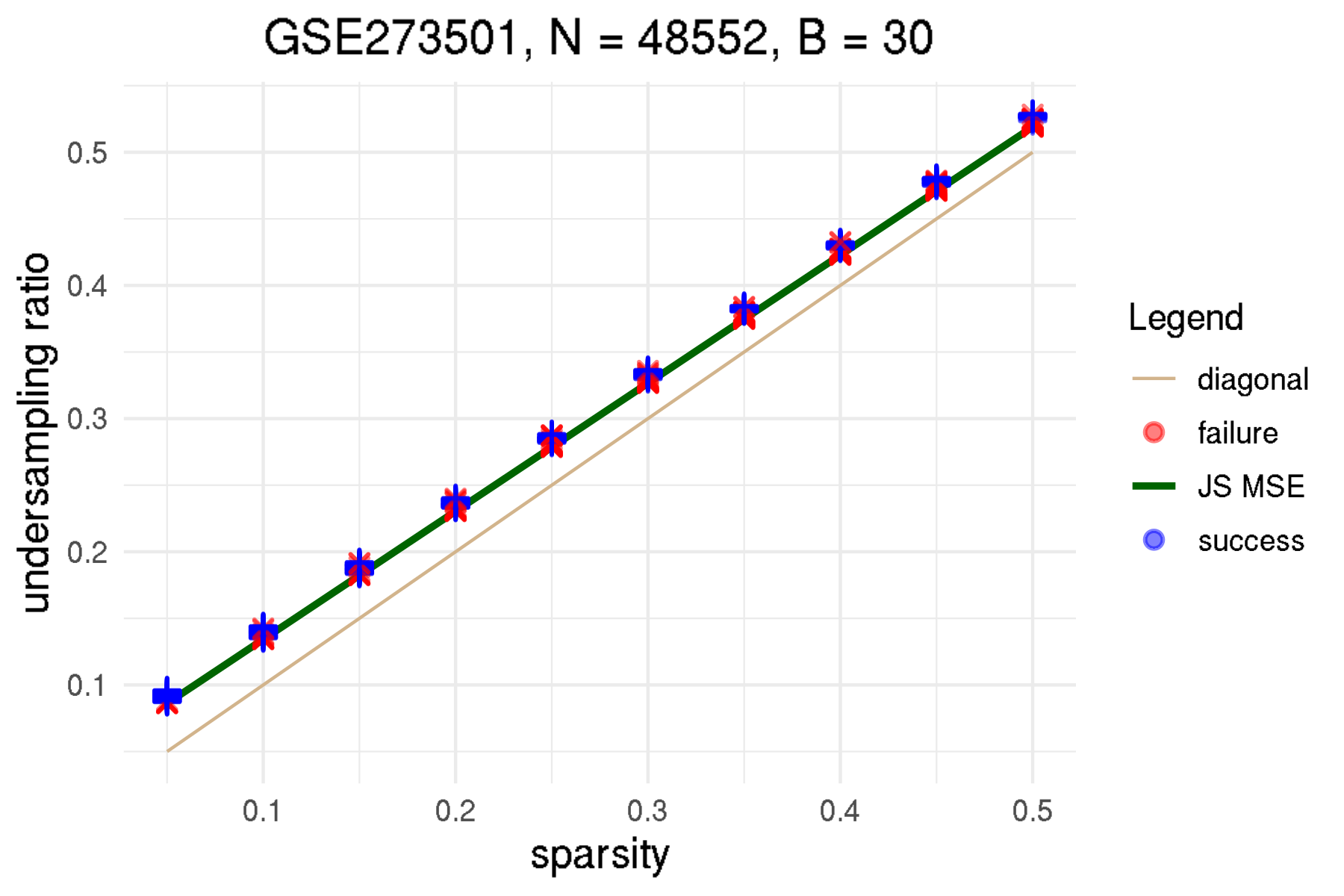}
  \end{subfigure}
  \hfill
  \begin{subfigure}[b]{0.48\textwidth}
    \includegraphics[width=\linewidth]{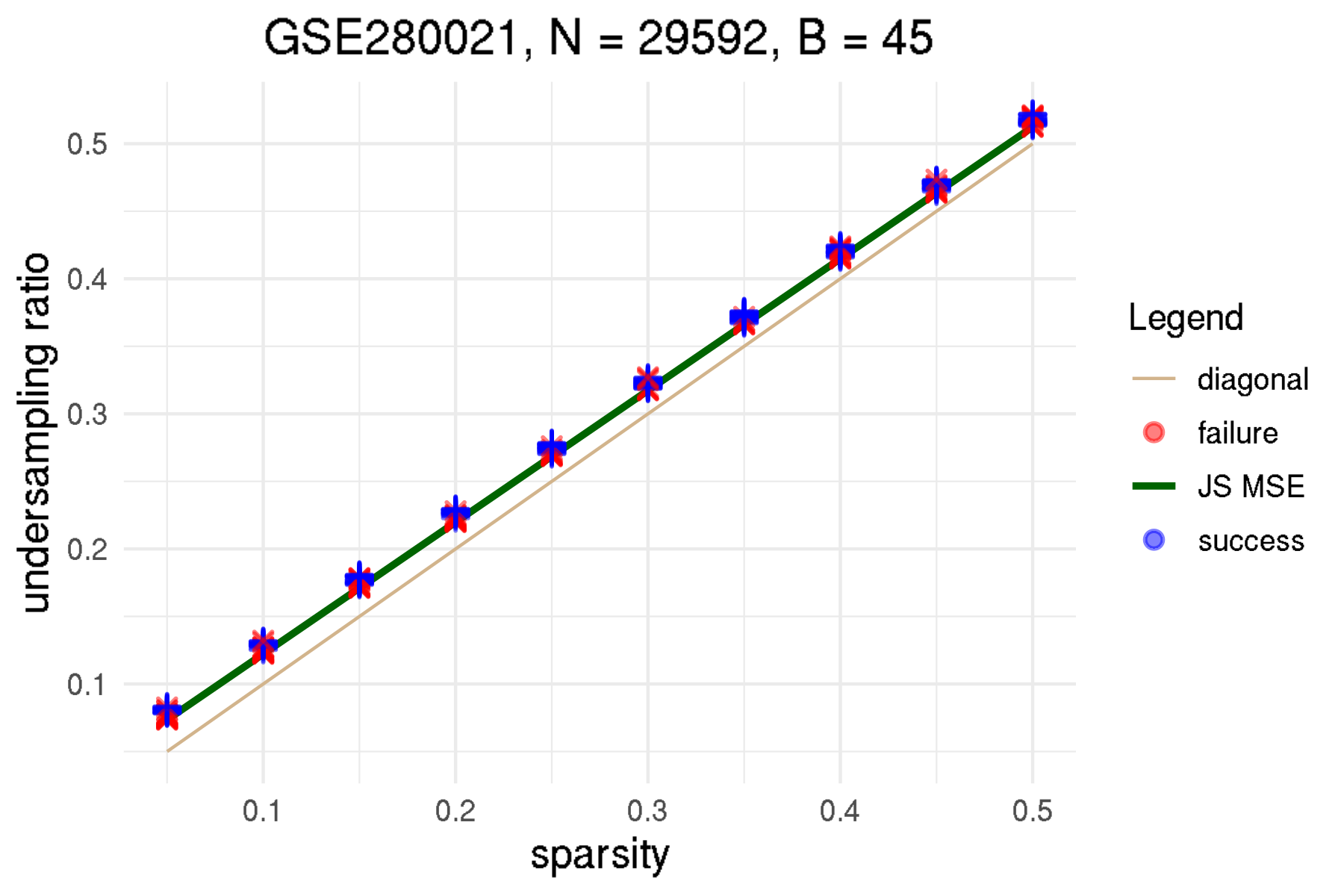}
  \end{subfigure}

  \vspace{1em}

  % Row 3
  \begin{subfigure}[b]{0.48\textwidth}
    \includegraphics[width=\linewidth]{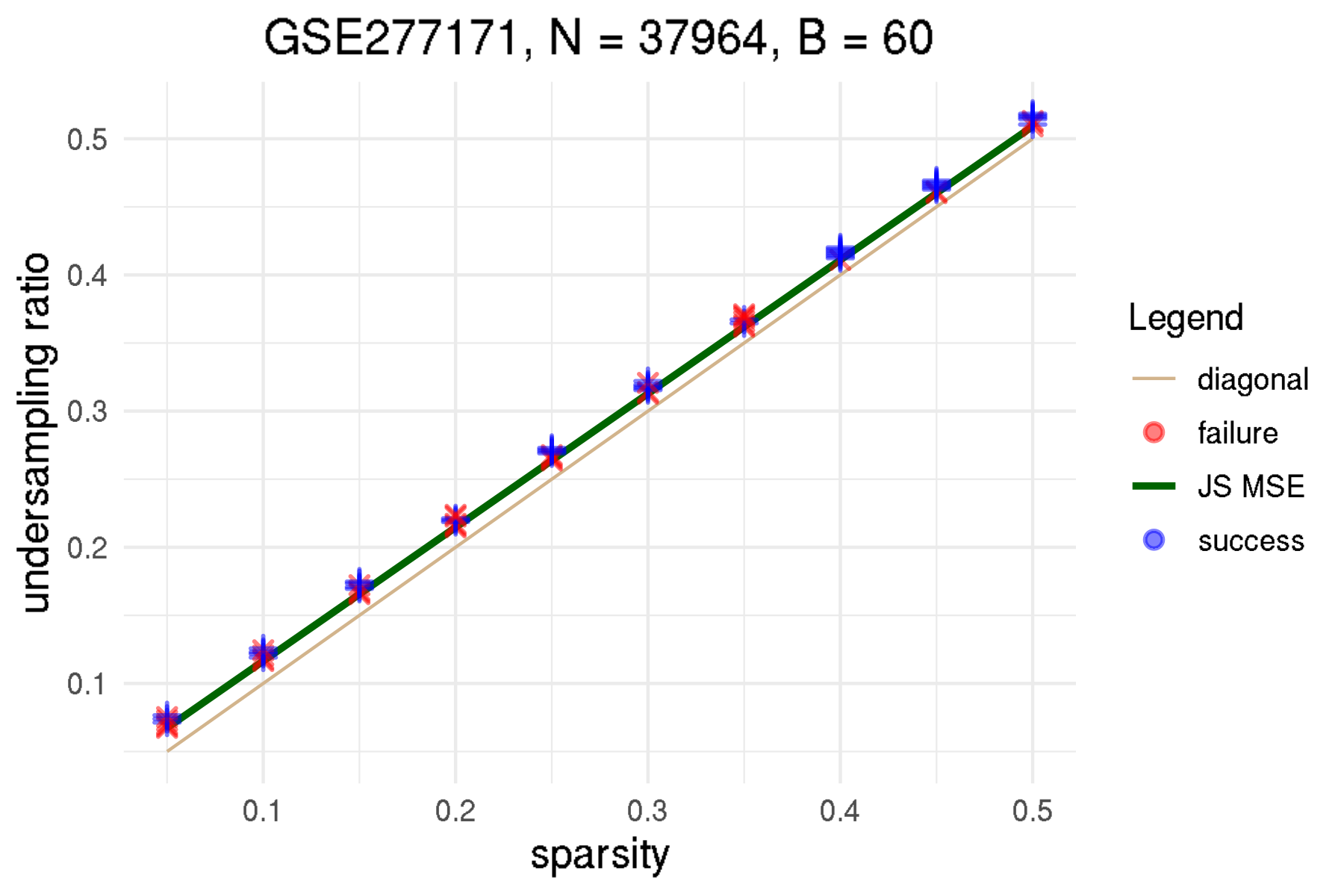}
  \end{subfigure}
  \hfill
  \begin{subfigure}[b]{0.48\textwidth}
    \includegraphics[width=\linewidth]{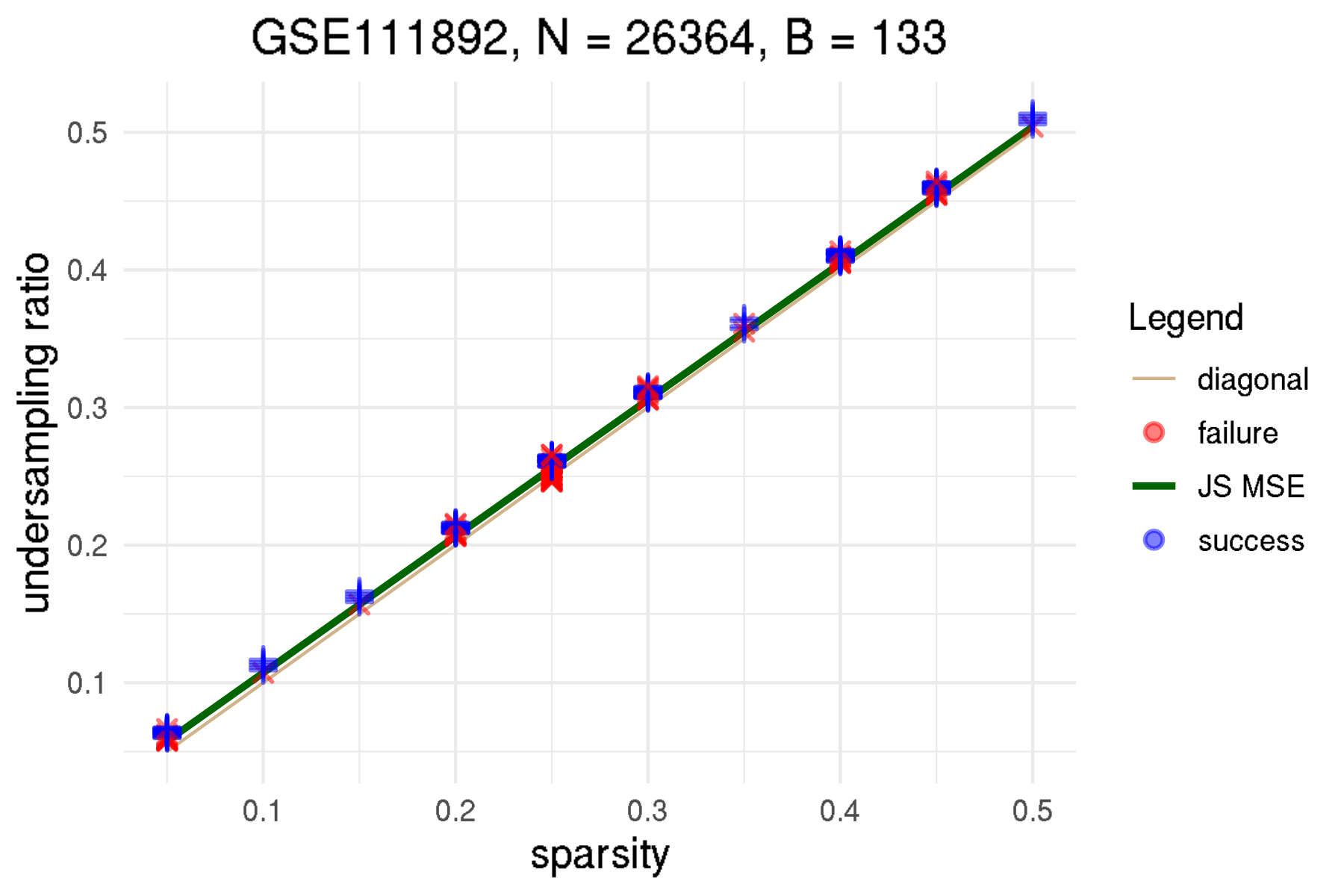}
  \end{subfigure}

  \caption{Performance of SteinSense on gene expression datasets. The plot titles contain the accession number, the number of rows and the number of columns in these datasets. For each run we record success if the relative error is smaller than $0.001$ and failure otherwise. We find that SteinSense begins succeeding just above the James Stein minimax risk curve.}\label{fig:RNASeq}
\end{figure}

Gene expression datasets form an example of naturally occurring real datasets where row sparsity is expected, since a sizeable fraction of genes show zero or negligible expressions. We thus consider 6 datasets from Gene Expression Omnibus\footnote{https://www.ncbi.nlm.nih.gov/geo/} with rows and columns. Rows usually denote genes and columns denote different experimental conditions. Genes that remain expressionless across different conditions contribute to zero (or negligible) rows. Although these datasets have characteristic inherent row sparsity, we vary the sparsity level $\epsilon\in\{0.05,0.1,\cdots,0.5\}$ by keeping the top $\epsilon$ fraction of rows and zero-ing out the others. This forms our signal matrix $X$. Since some of the cells have enormous counts, we use a $\log_2(\cdot+1)$ transform to the entries of $X$. Different datasets have different number of rows $N$ and different number of columns $B$. Now, $N$ is usually in the tens (and sometimes in the hundreds) of thousands, so we do not go for the full phase transition experiments with SteinSense as we have done before. We expect the phase transition to be given by $M_\JS)\epsilon,B)$ in accordance with Theorem \ref{thm:js_mse}. Thus we take undersampling values $\delta$ close to this theoretical curve, and we only perform $1$ or very few Monte Carlo runs at each $(\epsilon,\delta)$ point. The measurement matrix is generated with iid $N(0,1/n)$ entries, where $n=N\delta$. Figure \ref{fig:RNASeq} shows the results. We find that quite generally, successes start as $\delta$ gets even a little bit above $M_\JS(\epsilon,B)$.

\section{Array Compressed Sensing}\label{sec:array_cs}

In this section, we consider what we call Array Compressed Sensing, where the $N$ vectors $X_{1\star},\cdots, X_{N\star}$ are vertically stacked to form a long $NB-$dimensional vector $X_\arr$:
\begin{align*}
    X_\arr &= \begin{bmatrix}
        X_{1\star}\\
        \vdots \\
        X_{N\star}
    \end{bmatrix}
\end{align*}
Further, one employs a huge measurement matrix $A_\arr\in\sR^{n_\arr\times NB}$ made up of iid $N(0, 1/n_\arr)$ entries to sense $X_\arr$, recording measurements $$Y_\arr=A_\arr X_\arr\in\sR^{n_\arr}$$ Naturally, one asks how large should $n_\arr$ be in this setup for perfect recovery, and where a phase transition would occur in $\delta_\arr=n_\arr/NB$.

\citet{donoho2013accurate} use the traditional Approximate Message Passing algorithm for Array Compressed Sensing reconstruction, which for the convenience of the reader, is presented in Algorithm \ref{alg:array_cs}. Recall that for a function $f:\sR^{NB}\to \sR^{NB}$
\begin{align*}
    \text{div}(f)(v;\cdots) &= \dfrac{1}{NB}\sum_{i=1}^{NB}\dfrac{\partial f_i}{\partial v_i}(v;\cdots)
\end{align*}where $\cdots$ denote additional, fixed parameters passed to $f$.

\begin{algorithm*}
\caption{Array Compressed Sensing}\label{alg:array_cs}
\begin{algorithmic}[1]
     \Require $A\in\sR^{n_\arr\times NB}$, $Y\in\sR^{n_\arr}$, $\{\eta_t\}_{t\geq 0}$ sequence of denoisers
    \State Start with $X^0=0\in\sR^{NB}$
    \For{$t\geq 0$}
    \begin{align*}
        R^t &= Y - AX^t + \dfrac{NB}{n_\arr}R^{t-1}\cdot \text{div}(\eta_t)(H^t)\\
        H^{t+1} &= X^t + A^\top R^t\\
        X^{t+1} &= \eta_{t+1}(H^{t+1})
    \end{align*}
    \EndFor
\end{algorithmic}
\end{algorithm*}

Note that Algorithm \ref{alg:array_cs} has the usual divergence correction term from \citet{donoho2009message}. \citet{donoho2013accurate} have pointed out that using BlockSoft Thresholding and James Stein in place of the denoisers $\eta_t$ in Algorithm \ref{alg:array_cs} deliver phase transition located at the minimax risks of BlockSoft Thresholding and James Stein respectively. However, the purpose of the work was to connect theoretical phase transitions arising out of the (usual scalar) state evolution of Approximate message Passing and James Stein was one of the many denoisers considered there. Consequently, sufficient experimentation was not conducted to enable us to clearly see the differences in experimental performance between the Array Compressed Sensing and our Vector Compressed Sensing problems. As we will point out, there are important distinctions between the two problems, particularly from a computational persepctive.

\begin{figure}[htbp]
  \centering

  \begin{subfigure}[b]{0.6\textwidth}
    \includegraphics[width=\linewidth]{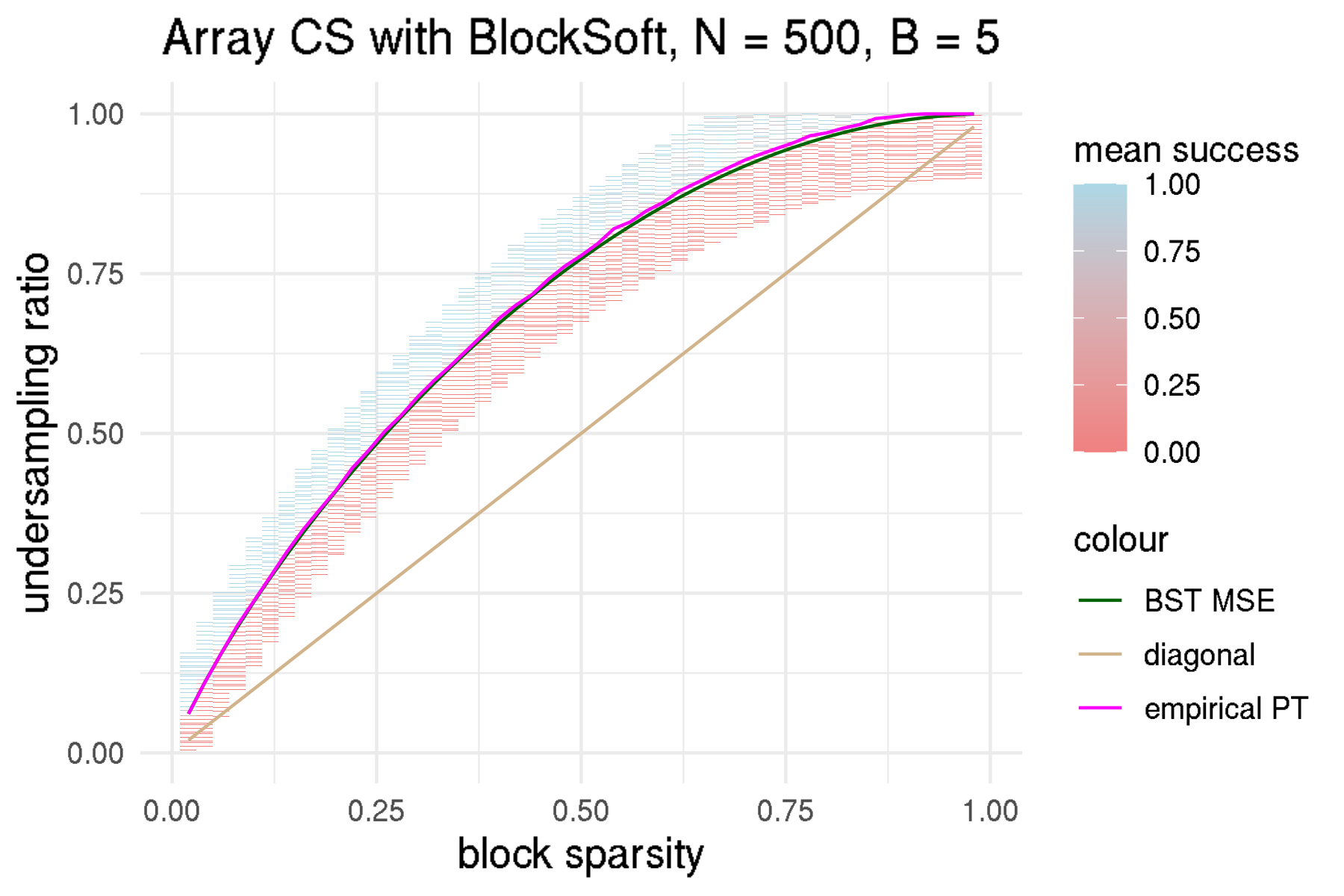}
  \end{subfigure}
  \vspace{1em}
  \begin{subfigure}[b]{0.6\textwidth}
    \includegraphics[width=\linewidth]{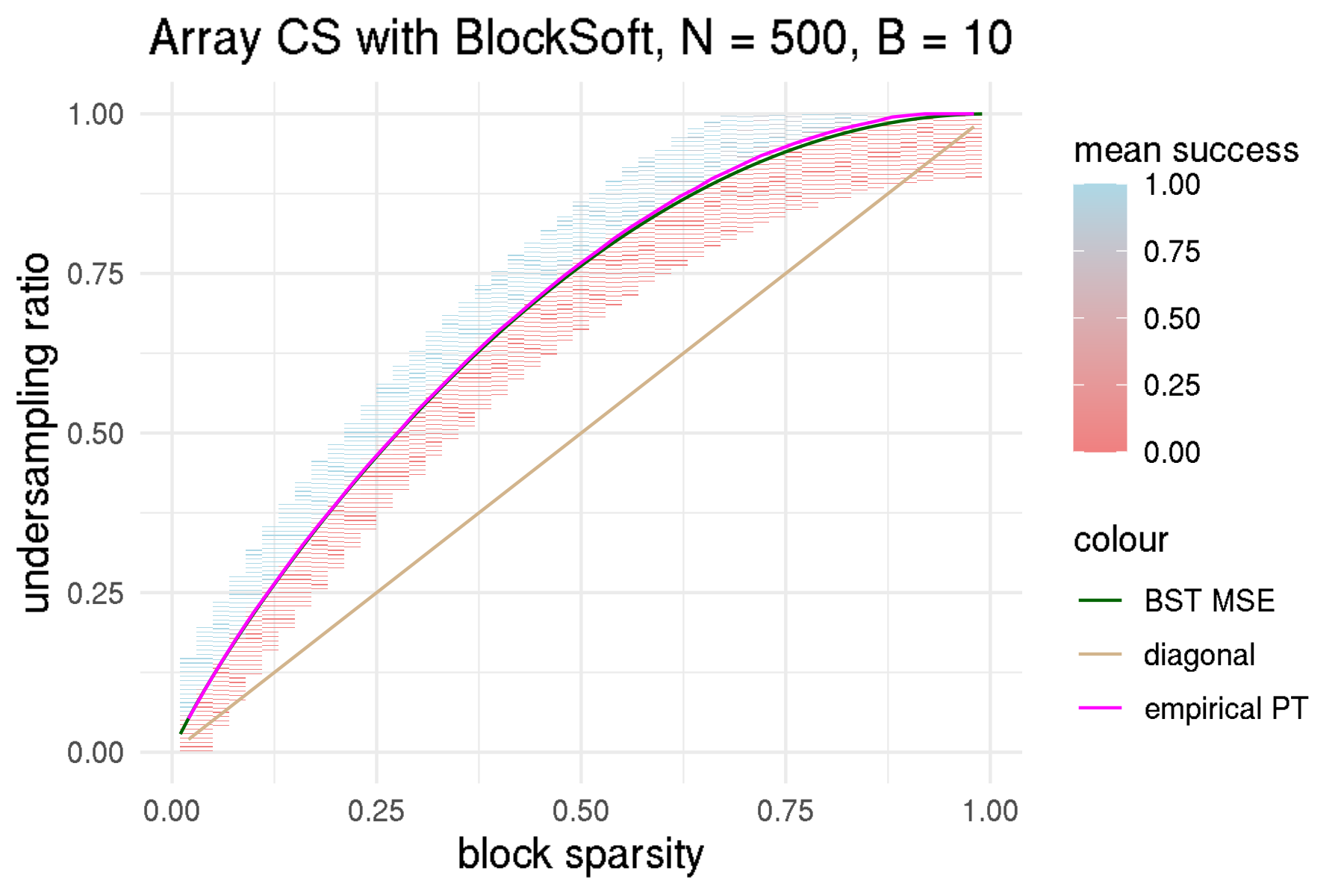}
  \end{subfigure}
    \vspace{1em}
  \begin{subfigure}[b]{0.6\textwidth}
    \includegraphics[width=\linewidth]{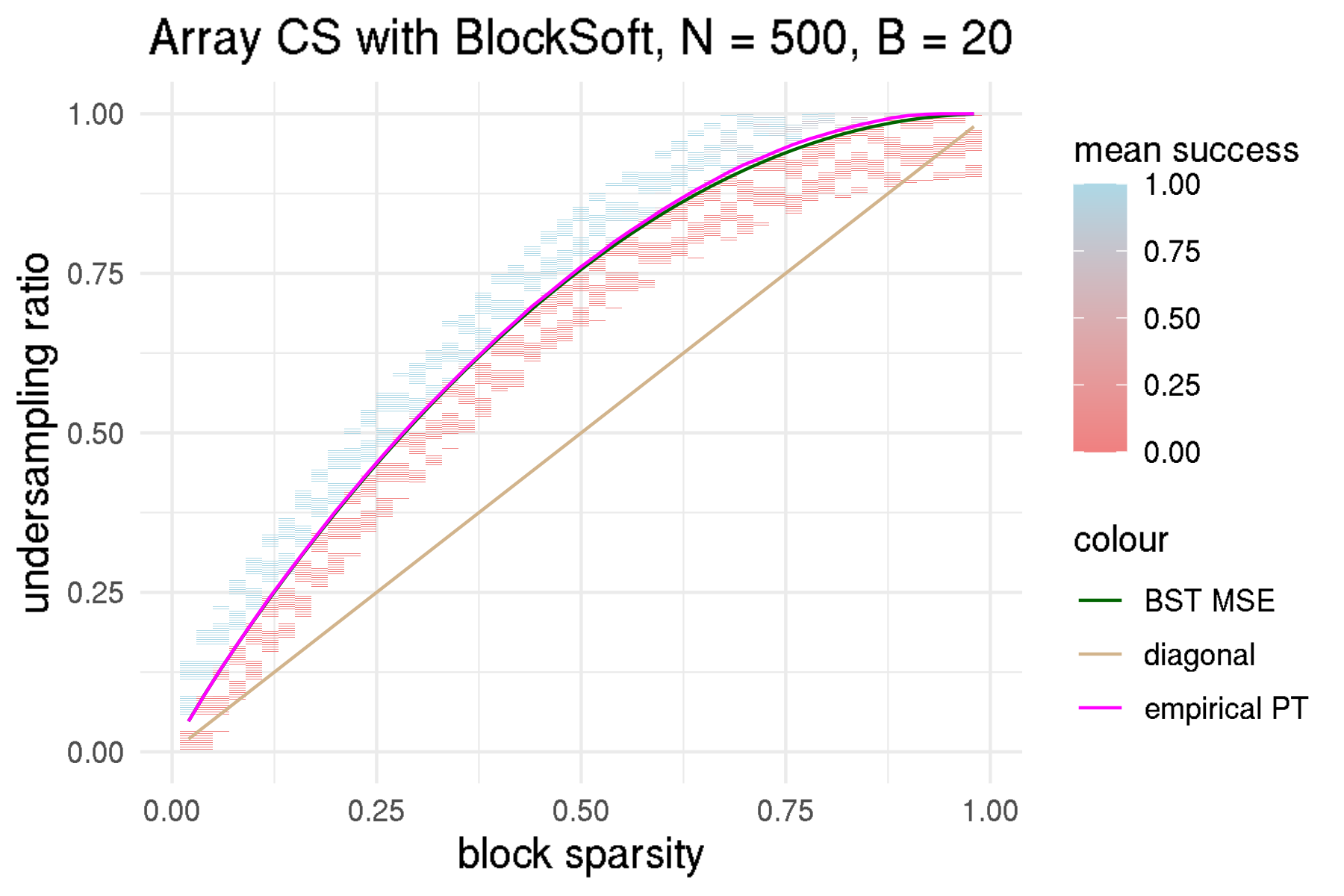}
  \end{subfigure}
  \caption{Signal non-zeros are chosen as iid $N(0,1)$. The empirical phase transition matches the minimax risk of BlockSoft Thresholding to a high degree of accuracy, confirming the predictions in \citet{donoho2013accurate}.}\label{fig:paper_softsense_concatenate}
\end{figure}

In Figure \ref{fig:paper_softsense_concatenate}, we show results on this Array Compressed Sensing problem with BlockSoft Thresholding. The experiments convincingly confirm that the phase transition appears again at the minimax risk curve of BlockSoft Thresholding - the same quantity we have seen appearing so far, confirming the results in \citet{donoho2013accurate}.

\begin{figure}[htbp]
  \centering

  \begin{subfigure}[b]{0.6\textwidth}
    \includegraphics[width=\linewidth]{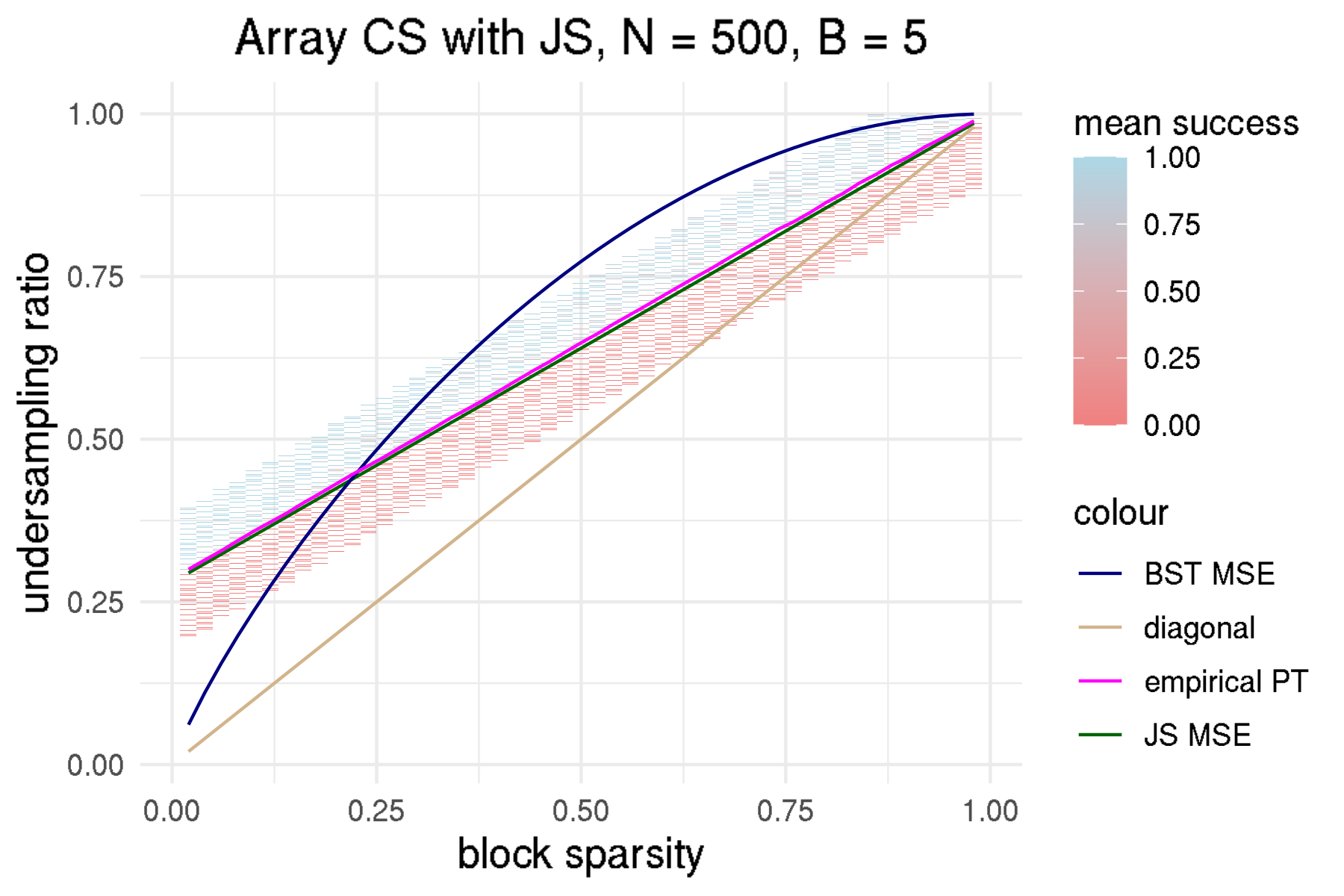}
  \end{subfigure}
  \vspace{1em}
  \begin{subfigure}[b]{0.6\textwidth}
    \includegraphics[width=\linewidth]{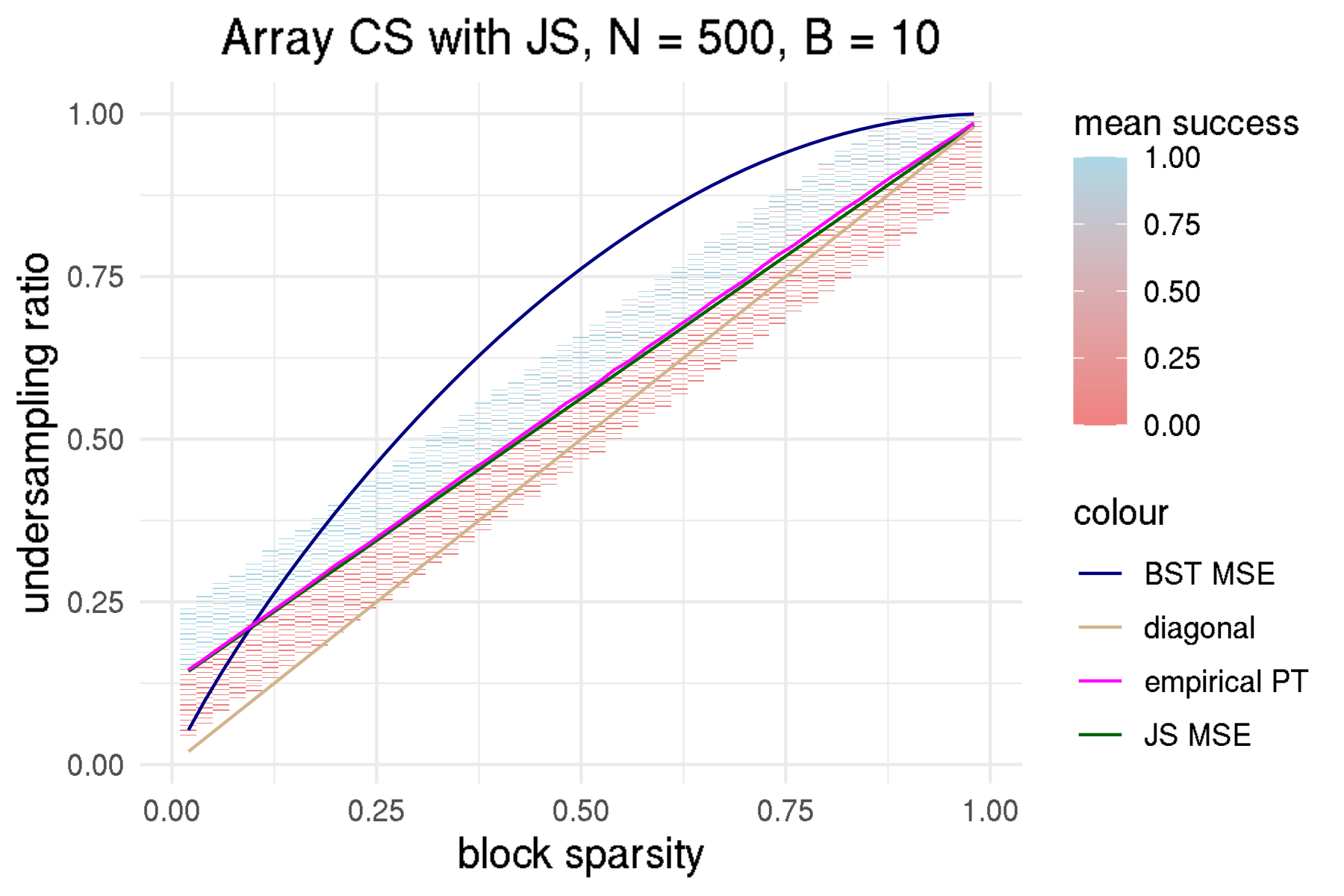}
  \end{subfigure}
    \vspace{1em}
  \begin{subfigure}[b]{0.6\textwidth}
    \includegraphics[width=\linewidth]{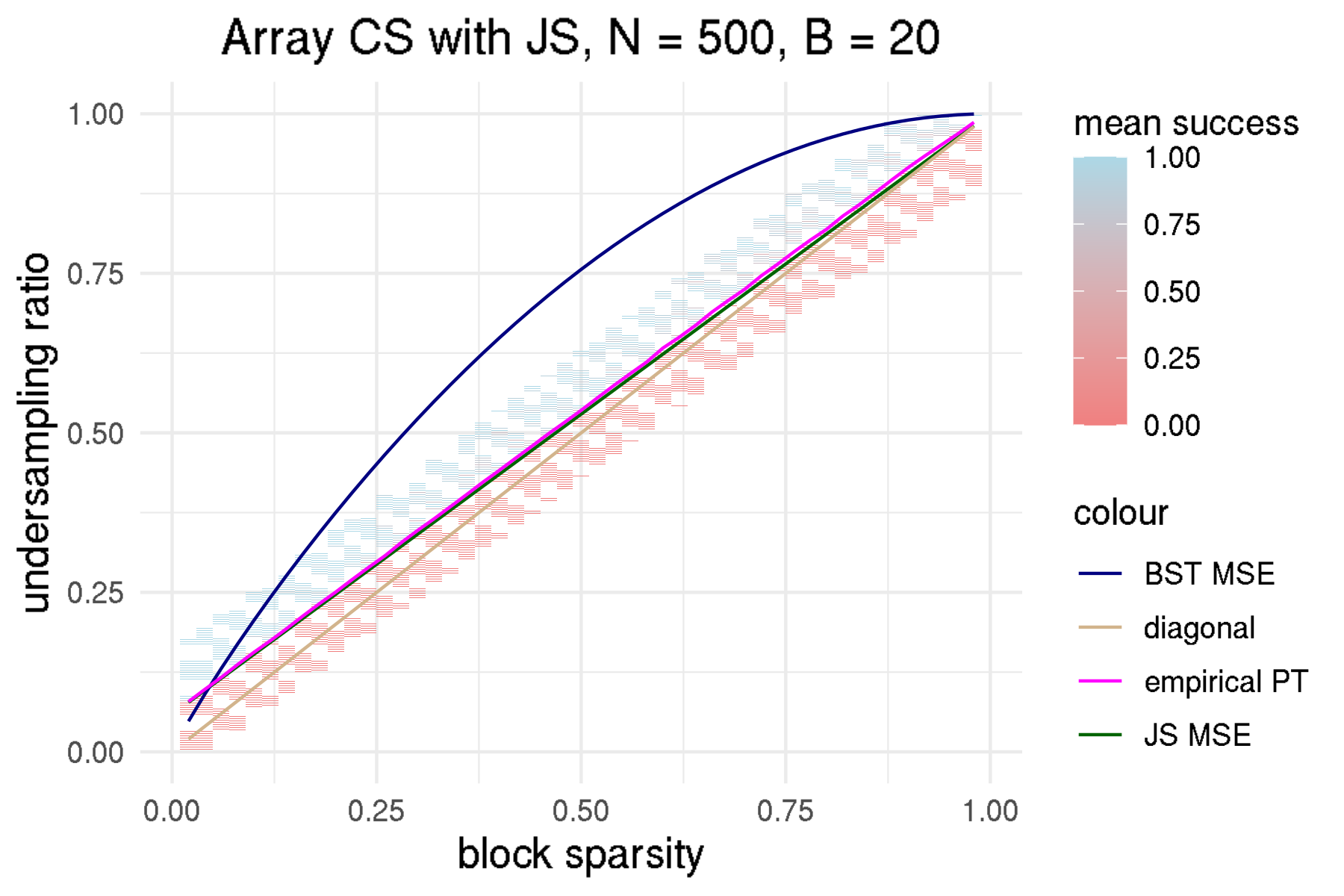}
  \end{subfigure}
  \caption{Signal non-zeros are chosen as iid $N(0,1)$. The empirical phase transition matches the minimax risk of BlockSoft Thresholding to a high degree of accuracy, confirming the predictions in \citet{donoho2013accurate}.}\label{fig:paper_steinsense_concatenate}
\end{figure}

Figure \ref{fig:paper_steinsense_concatenate} demonstrates the power of James Stein in this Array Compressed Sensing framework. Once again, the phase transition appears at the minimax risk of James Stein. Consequently, as $B$ gets larger, James Stein reaches the diagonal.

It is important to contrast the experimental results on SoftSense and SteinSense with those on Array Compressed Sensing.
\begin{enumerate}
    \item Algorithm \ref{alg:array_cs} is significantly easier to run. In Vector Compressed Sensing (Algorithms \ref{alg:softsense} and \ref{alg:steinsense}) involve computing large number of $B\times B$ Jacobian matrices, which involves computing $NB^2$ entries per iteration as opposed to just computing $NB$ entries per iteration for the divergence term in Algorithm \ref{alg:array_cs}. Consequently, Algorithms \ref{alg:softsense} and \ref{alg:steinsense} require specialized software for significant speed up.
    \item Algorithm \ref{alg:array_cs} involves scalar $s^t$ per iteration, while Algorithms \ref{alg:softsense} and \ref{alg:steinsense} require matrix $S^t$ to be fed into the denoisers. Following the definitions of $\eta_\CBST$ and $\eta_\CJS$, one can see that matrix inversions are required. We have noticed that as iterations progress, $S^t$ develops an essentially low-rank structure with large condition number, to the extent that often, numerically, it becomes rank deficient. Consequently, care needs to be taken in defining the denoisers $\eta_\CBST$ and $\eta_\CJS$, which has been done in the code. Such issues on numerical stability do not appear in Algorithm \ref{alg:array_cs}.
    \item Algorithm \ref{alg:array_cs} is, computationally, a \textit{cleaner} problem at very moderate problem sizes. A little bit above the expected theoretical phase transition, all the experiments result in success. For Algorithms \ref{alg:softsense} and \ref{alg:steinsense}, we find that one has to travel significantly above the phase transition to get all successes. Consequently, a lot more experiments need to be run in a wider band above the phase transition to get a reasonably accurate estimate of the phase transition location.
    \item While the above point is true, it implies a lot more \textit{embarrassingly parallel} experiments need to be run to track the phase transition accurately in So. The memory footprint of each experiment is relatively low, provided one uses appropriate software to speed up computations. Importantly, the measurement matrix $A$ only has $O(N^2)$ entries and this is reasonable for the matrix operations that SoftSense and SteinSense perform. For Algorithm \ref{alg:array_cs}, the measurement matrix $A_\arr$ consists of $O(N^2\color{red}B^2\color{black})$ entries! If $B$ is as small as $10$, one would need $100$ times the memory to store and operate on $A_\arr$ than what they would need for $A$. This becomes impractical for the large-$B$ problems in modern technology, e.g. hyperspectral images where $B$ is in the hundreds.
    \item A corollary from the Algorithm \ref{alg:array_cs} theoretical results is that the average coordinate-wise risk of the denoiser is the phase transition determining quantity. A primary focus of classical statistical theory has been average coordinate-wise squared error loss, and over decades denoisers have been developed with good risk properties \textit{under the average coordinate-wise squared error loss}, so one may leverage them to understand the performance of Algorithm \ref{alg:array_cs}. However, as will be explained in Section \ref{sec:theory}, Vector Compressed Sensing corresponds to matricial State Evolution, involving tracking full risk matrices of the denoisers under consideration, and classically there is very scanty literature on understanding properties of risk matrices. This makes the Vector Compressed Sensing problem we have studied in this paper, much more challenging to understand theoretically.
\end{enumerate}

% \begin{figure}
%     \centering
%     \includegraphics[width=\linewidth]{figures/array_CS/paper_softsense_concatenate_N_500_B_5.png}
%     \caption{Signal non-zeros are chosen as iid $N(0,1)$. The empirical Phase Transition matches the minimax risk of BlockSoft Thresholding to a high degree of accuracy, confirming the predictions in \citet{donoho2013accurate}.}
%     \label{fig:paper_softsense_concatenate_N_500_B_5}
% \end{figure}

% \begin{figure}
%     \centering
%     \includegraphics[width=\linewidth]{figures/array_CS/paper_softsense_concatenate_N_500_B_10.png}
%     \caption{Signal non-zeros are chosen as iid $N(0,1)$. The empirical Phase Transition matches the minimax risk of BlockSoft Thresholding to a high degree of accuracy, confirming the predictions in \citet{donoho2013accurate}.}
%     \label{fig:paper_softsense_concatenate_N_500_B_10}
% \end{figure}

% \begin{figure}
%     \centering
%     \includegraphics[width=\linewidth]{figures/array_CS/paper_softsense_concatenate_N_500_B_20.png}
%     \caption{Signal non-zeros are chosen as iid $N(0,1)$. The empirical Phase Transition matches the minimax risk of BlockSoft Thresholding to a high degree of accuracy, confirming the predictions in \citet{donoho2013accurate}.}
%     \label{fig:paper_softsense_concatenate_N_500_B_20}
% \end{figure}

% \begin{figure}
%     \centering
%     \includegraphics[width=\linewidth]{figures/array_CS/paper_steinsense_concatenate_N_500_B_5.png}
%     \caption{Signal non-zeros are chosen as iid $N(0,1)$. The empirical Phase Transition matches the minimax risk of BlockSoft Thresholding to a high degree of accuracy, confirming the predictions in \citet{donoho2013accurate}.}
%     \label{fig:paper_steinsense_concatenate_N_500_B_5}
% \end{figure}

% \begin{figure}
%     \centering
%     \includegraphics[width=\linewidth]{figures/array_CS/paper_steinsense_concatenate_N_500_B_10.png}
%     \caption{Signal non-zeros are chosen as iid $N(0,1)$. The empirical Phase Transition matches the minimax risk of BlockSoft Thresholding to a high degree of accuracy, confirming the predictions in \citet{donoho2013accurate}.}
%     \label{fig:paper_steinsense_concatenate_N_500_B_10}
% \end{figure}

% \begin{figure}
%     \centering
%     \includegraphics[width=\linewidth]{figures/array_CS/paper_steinsense_concatenate_N_500_B_20.png}
%     \caption{Signal non-zeros are chosen as iid $N(0,1)$. The empirical Phase Transition matches the minimax risk of BlockSoft Thresholding to a high degree of accuracy, confirming the predictions in \citet{donoho2013accurate}.}
%     \label{fig:paper_steinsense_concatenate_N_500_B_20}
% \end{figure}
\section{Tracking the Phase Transitions Analytically}\label{sec:theory}

This section contains theoretical results explaining why, at least in some cases, the empirical phase transitions of SoftSense and SteinSense match curves coming out of classical statistical calculations. To achieve this, it would be helpful to consider a general Vector Compressed Sensing reconstruction algorithm, of which SoftSense and SteinSense are special cases. As before, $\eta_t(M)$ implies application of $\eta_t(\cdot)$ row-wise to $M$.

\begin{algorithm*}
\caption{General Vector Compressed Sensing Algorithm}\label{alg:general_sense}
\begin{algorithmic}[1]
     \Require $A\in\sR^{n\times N}$, $Y\in\sR^{n\times B}$, $\{\eta_t\}_{t\geq 0}$ sequence of denoisers
    \State Start with $X^0=0\in\sR^{N\times B}$
    \For{$t\geq 0$}
    \begin{align*}
        R^t &= Y - AX^t + \dfrac{1}{\delta}R^{t-1}\cdot J_{\eta_t}(H^t)\\
        H^{t+1} &= X^t + A^\top R^t\\
        X^{t+1} &= \eta_{t+1}(H^{t+1})
    \end{align*}
    \EndFor
\end{algorithmic}
\end{algorithm*}
 Algorithm \ref{alg:general_sense} has been studied in Vector Compressed Sensing previously \citep{hara2022blind,hara2020grant} with specific nonzero distributions and denoisers. To the best of our knowledge, none of the prior works derive phase transitions for these algorithms, and the trends with increasing $B$ are not theoretically justified. The theoretical description of Algorithm \ref{alg:general_sense} relies heavily on the Generalized Approximate Message Passing framework \citep{javanmard2013state,rangan2011generalized}. Consider the following three assumptions.
\paragraph{Assumption 1.} The measurement matrix $A\in\sR^{n\times N}$ consists of iid $N(0,1/n)$ entries.\label{assm_1}

\paragraph{Assumption 2.} The empirical spectral distribution of $X_{1\star},\cdots, X_{N\star}$ converges weakly to a probability distribution $\mu\in\gF(\epsilon,B)$ with all moments finite:
\begin{align*}
    \dfrac{1}{N}\sum_{i=1}^N \delta_{X_{i\star}}\stackrel{\text{weakly}}{\to}\mu
\end{align*}where for any $j\geq 1$, $\int \|x\|^jd\mu(x)<\infty$. Further, the corresponding moments converge:
\begin{align*}
    \dfrac{1}{N}\sum_{i=1}^N \|X_{i\star}\|^j\to \int\|x\|^jd\mu(x), \hspace{0.1cm}j\geq 1
\end{align*}\label{assm_2}
\paragraph{Assumption 3.} The denoisers $\eta_t:\sR^B\to\sR^B$ are Lipschitz continuous.\label{assm_3}

Let $\gP(B)$ denote the class of all probability distributions on $\sR^B$. Suppose we aim to estimate the mean $m\in\sR^B$ given $y\sim \gN_B(m,\Sigma)$ where $\Sigma$ is a known positive definite matrix. Further suppose $m\sim \mu\in\gP(B)$. Define for an estimator $\eta\in\sR^B$, the average coordinate-wise risk
\begin{align*}
    R(\mu;\eta,\Sigma) &= \dfrac{1}{B}\E\|\eta(m+\Sigma^{1/2}z)-m\|^2
\end{align*}and the \textit{risk matrix}
\begin{align*}
    \gR(\mu;\eta,\Sigma) &= \E\left[(\eta(m+\Sigma^{1/2}z)-m)(\eta(m+\Sigma^{1/2}z)-m)^\top\right]
\end{align*}where the expectations are taken over both $m\sim\mu$ and $z\sim\gN_B(0,I_B)$ independent of each other. Notice the following simple identity:
\begin{align*}
    R(\mu;\eta,\Sigma) &= \dfrac{1}{B}\Tr(\gR(\mu;\eta,\Sigma))
\end{align*}where $\Tr(M)$ denotes the trace of a matrix $M$.

We record, for the convenience of the reader, the traditional theoretical result connecting Algorithm \ref{alg:array_cs} to a scalar state evolution. Recall that a function $\varphi:\sR^r\to\sR$ is called pseudo-Lipschitz of order $k$ if for any $x,y\in\sR^r$,
    \begin{align*}
        |\varphi(x)-\varphi(y)|\leq C(1+\|x\|^{k-1}+\|y\|^{k-1})\|x-y\|
    \end{align*}for a constant $C>0$.

\begin{theorem}[See \citet{donoho2009message,bayati2011dynamics,donoho2013accurate}] Make Assumptions 1, 2 and 3. Let $X^t\in\sR^{NB}$ denote the output of Algorithm \ref{alg:array_cs} after $t$ iterations. For any pseudo-Lipschitz function $\psi:\sR^B\times\sR^B\to\sR$,
\begin{align}\label{eqn:scalar_se}
    \dfrac{1}{N}\E[\psi(X^t_{i\star},X_{i\star})] \stackrel{\text{a.s.}}{=}\E[\psi(m+\sigma_tz,m)]
\end{align}where the expectation is taken over both $m\sim\mu$ independent of $z\sim\gN_B(0,I_B)$, and $\{\sigma^2_t\}_t$ follows a scalar dynamical system known as State Evolution, starting from $\sigma^2_0=\int \|x\|^2d\mu(x)/B$:
\begin{align*}
    \sigma^2_{t+1} &= R\left(\mu;\eta_t, (\sigma^2_t/\delta)I_B\right)
\end{align*}
    
\end{theorem}

We call the State Evolution in Equation \ref{eqn:scalar_se} a Scalar State Evolution, since it is a one-dimensional dynamical system tracking the scalar $\sigma^2_t$. Thus, Array Compressed Sensing gives rise to the traditional Scalar State Evolution that progresses by the average coordinate-wise risk of denoiser $\eta_t$. Next, we present the result for Algorithm \ref{alg:general_sense}.

\begin{theorem}\label{thm:general_vector_cs_se}
    Make Assumptions 1, 2 and 3. Let $X^t\in\sR^{N\times B}$ denote the output of Algorithm \ref{alg:general_sense} after $t$ iterations. For any pseudo-Lipschitz function $\psi:\sR^B\times \sR^B\to\sR$,
\begin{align}\label{eqn:matricial_se}
    \dfrac{1}{N}\E[\psi(X^t_{i\star},X_{i\star})] \stackrel{\text{a.s.}}{=}\E[\psi(m+\Sigma_t^{1/2}z, m)]
\end{align}where the expectation is taken over both $m\sim\mu$ independent of $z\sim\gN_B(0,I_B)$, and $\{\Sigma_t\}_t$ follows a matricial dynamical system known as State Evolution, starting from $\Sigma_0=\int xx^\top d\mu(x)$:
\begin{align*}
    \Sigma_{t+1} &= \gR\left(\mu;\eta_t, \Sigma_t/\delta\right)
\end{align*}
\end{theorem}
We will call the State Evolution in Equation \ref{eqn:matricial_se} a Matricial State Evolution, as now we need to track the risk matrices across iterations.

\begin{proof}[Proof of Theorem \ref{thm:general_vector_cs_se}]
    The result follows from a careful consideration of the case of symmetric $A$ in Theorem 1 in \citet{javanmard2013state} and necessary modifications to the rectangular case from discussions provided in \citet{javanmard2013state}.
\end{proof}

To the best of our knowledge, this Matricial State Evolution from Equation \ref{eqn:matricial_se} has not been previously pointed out this explicitly in the Signal Processing literature. Indeed, the Vector Compressed Sensing problem induces correlations among the different components of the iterates as the same measurement matrix $A$ is used to sense each column, and consequently one needs to track all the variances and covariances, not just the average coordinate-wise variance, as is common in Scalar State Evolution.

Now, we specialize to the class of $\epsilon-$sparse distributions. Define
\begin{align*}
    \gF(\epsilon,B)&=\{\mu\in\gP(B):\mu(\{0\})\geq 1-\epsilon\}
\end{align*}
Also define
\begin{align*}
    M_\BST(\epsilon,B) &= \inf_{\tau>0}\sup_{\mu\in\gF(\epsilon,B)}R(\mu;\eta_\BST,\tau, I_B)
\end{align*}and 
\begin{align*}
    M_\JS(\epsilon,B) &= \sup_{\mu\in\gF(\epsilon,B)}R(\mu;\eta_\JS,I_B)
\end{align*}to be the scalar minimax risks of BlockSoft Thresholding and James Stein respectively. It is well known \citep{johnstone2002function,lehmann2006theory} that the least favorable distribution $\mu_\LF$ places mass $(1-\epsilon)$ at $0\in\sR^B$ and the rest of its mass $\epsilon$ uniformly on a sphere of infinite radius. Also let $\tau(\epsilon,B)$ denote the minimax threshold for BlockSoft Thresholding, which is obtained by choosing the $\tau$ that works best when $\mu_\LF$ is used.

$M_\BST$ and $M_\JS$ can be computed analytically. Indeed, define
\begin{align*}
    h(\tau^2,B) &= \dfrac{\tau}{\E\left[\left(\sqrt{\chi^2_B}-\tau\right)_+\right]}\\
    g(\tau^2,B) &= \dfrac{\tau\E\left[\left(\sqrt{\chi^2_B}-\tau\right)^2_+\right]}{\E\left[\left(\sqrt{\chi^2_B}-\tau\right)_+\right]}
\end{align*}Then, $M_\BST(\epsilon,B)$ is given by 
\begin{align*}
    M_\BST(\epsilon,B) &= \dfrac{B+\tau^2(\epsilon,B) + g(\tau^2(\epsilon,B),B)}{B(1+h(\tau^2(\epsilon,B),B))}
\end{align*}where $\tau(\epsilon,B)$ is defined to be the solution $\tau$ to $1/(1+h(\tau^2))=\epsilon$. Details are available in \citet{donoho2013accurate,johnstone2002function}. An important aspect is the large $B$ behavior of the minimax risk. Indeed, \citet{donoho2013accurate} show that
\begin{align*}
    \lim_{B\to\infty} M_\BST(\epsilon,B) &= 2\epsilon-\epsilon^2
\end{align*}

The minimax risk of James Stein is simpler to compute.
\begin{align*}
    M_\JS(\epsilon,B) &= (1-\epsilon)R(\delta_0;\eta_\JS)+\epsilon
\end{align*}Now, $R(\delta_0;\eta_\JS)\leq 2/B$. Consequently, \begin{align*}
    \lim_{B\to\infty} M_\JS(\epsilon,B) &= \epsilon
\end{align*}Notice that $\epsilon<2\epsilon-\epsilon^2$ for any $\epsilon\in(0,1)$, and thus, as $B\to\infty$, James Stein becomes optimal.

We now describe the phase transition for SoftSense and SteinSense. For this, we will make one additional assumption.

\paragraph{Assumption 4.} The limiting distribution $\mu$ from Assumption 2 has symmetric exchangeable coordinates.

\begin{theorem}\label{thm:softsense_matching_BST_MSE}
    Make Assumptions 1,2, 3 and 4. Let $X^t$ denote the output of SoftSense with $\tau_t\equiv \tau(\epsilon,B)$ (the minimax threshold) for each $t$. If $\delta>M_\BST(\epsilon,B)$, 
    \begin{align*}
        \lim_{t\to\infty}\lim_{N\to\infty}\dfrac{1}{NB}\|X^t - X\|_F^2 &= 0
    \end{align*}Conversely, if $\delta<M_\BST(\epsilon,B)$, there exists $\mu\in\gF(\epsilon,B)$ symmetric exchangeable such that 
    \begin{align*}
        \liminf_{t\to\infty}\lim_{N\to\infty}\dfrac{1}{NB}\|X^t - X\|_F^2 &> 0
    \end{align*}
\end{theorem}

\begin{theorem}\label{thm:js_mse}
        Make Assumptions 1,2, 3 and 4. Let $X^t$ denote the output of SteinSense. If $\delta>M_\JS(\epsilon,B)$, 
    \begin{align*}
        \lim_{t\to\infty}\lim_{N\to\infty}\dfrac{1}{NB}\|X^t - X\|_F^2 &= 0
    \end{align*}Conversely, if $\delta<M_\JS(\epsilon,B)$, there exists $\mu\in\gF(\epsilon,B)$ symmetric exchangeable such that 
    \begin{align*}
        \liminf_{t\to\infty}\lim_{N\to\infty}\dfrac{1}{NB}\|X^t - X\|_F^2 &> 0
    \end{align*}
\end{theorem}

\begin{remark}[Reaching the diagonal]\label{rmk:diagonal_achievement}
    Theorem \ref{thm:js_mse} establishes that the phase transition of SteinSense occurs at $M_\JS(\epsilon,B)\approx \epsilon$ for large $B$. This the reason why SteinSense achieves the diagonal. To restate a point made earlier, for $B=1$, achieving the diagonal needs special care; see \citet{donoho2013information}. One would need to use a specialized measurement matrix and Bayes estimator, which are highly specific to $\mu$. However, SteinSense achieves the diagonal for large $B$ without any specialized knowledge! Further, this result establishes that there is absolutely no need to go for any computationally challenging denoiser, for example those based on deep learning. SteinSense, employing a very simple denoiser, will be essentially optimal.
\end{remark}

The argument for both theorems is similar, so we provide one proof covering both.

\begin{proof}[Proof of Theorems \ref{thm:softsense_matching_BST_MSE} and \ref{thm:js_mse}]
    Note that SoftSense and SteinSense both involve denoisers of the form
    \begin{align}\label{eqn:eta_type}
        \eta(y;\Sigma) &= c(y^\top\Sigma^{-1}y)\cdot y
    \end{align}For SoftSense, $c(x)=(1 - \tau/\sqrt{x})_+$ and for SteinSense, $c(x)=(1 - (B-2)/x)_+$. By Theorem \ref{thm:general_vector_cs_se}, Algorithm \ref{alg:general_sense} corresponds to Matricial State Evolutions:
    \begin{align*}
        \Sigma_{t+1} &= \gR(\mu;\eta,\Sigma_t/\delta)
    \end{align*}Since $\mu$ has symmetric exchangeable coordinates, the structure of $\eta$ enforces that $\Sigma_t$ is a multiple of the identity for every $t$. Consequently, $\Sigma_t=\sigma^2_tI_B$ and we reduce to the case of Scalar State Evolution:
    \begin{align*}
        \sigma^2_{t+1} &= R(\mu;\eta,(\sigma^2_t/\delta)I_B)
    \end{align*}This has been studied in detail for BlockSoft Thresholding and James Stein denoisers in \citet{donoho2013accurate}. If $\delta$ is larger than the minimax risk of the corresponding denoiser, then $\sigma^2_t\to 0$ as $t\to\infty$ geometrically fast. On the other hand, if $\delta$ is smaller than the minimax risk of the corresponding denoiser, then \citet{donoho2013accurate} points out that there exists a distribution $\nu\in\gF(\epsilon,B)$ (not necessarily symmetric or exchangeable) such that if Assumption 2 holds with $\mu$ replaced by $\nu$, then the corresponding State Evolution does not go to $0$, namely remains lower bounded. We will now show how to construct $\mu\in\gF(\epsilon,B)$ with symmetric exchangeable coordinates such that its risk under $\eta$ matches exactly that of $\nu$.
    
    Given any $s\in\{\pm1\}^B$ and permutation $\pi\in S_B$ (the group of permutations on $\{1,2,\cdots, B\}$), define $\nu(s,\pi)$ to be the following distribution. If $X\sim \nu$ then $(s\odot X)_\pi\sim \nu(s,\pi)$, where $\odot$ denotes Hadamard product and $x_\pi$ denotes, for a vector $x=(x_1,\cdots, x_B)$, the resulting vector $(x_{\pi(1)},\cdots,x_{\pi(B)})$. Then, define
    \begin{align*}
        \tilde\nu &= \dfrac{1}{2^BB!}\sum_{s\in\{\pm1\}^B}\sum_{\pi\in S_B}\nu(s,\pi)
    \end{align*}
    Then, $\tilde\nu$ is symmetric and exchangeable. Towards this, define for a Borel set $A\subseteq \sR^B$, for any sign vector $s\in\{\pm1\}^B$ and permutation $\pi\in S_B$,
    \begin{align*}
        A_\pi &= \{x_\pi:x\in A\},\\
        A_s &= \{s\odot x:x\in A\}
    \end{align*}
    Suppose $\tilde X\sim \tilde \nu$, then for any Borel $A$,
    \begin{align*}
        \sP(\tilde X\in A) &= \dfrac{1}{2^BB!}\sum_{s\in\{\pm1\}^B}\sum_{\pi\in S_B}\sP(X\in (A_{\pi^{-1}})_s)
    \end{align*}
    Take any permutation $\pi'\in S_B$, then
    \begin{align*}
        \sP(\tilde X_{\pi'}\in A) &=\sP(\tilde X\in A_{(\pi')^{-1}})\\
        &=\dfrac{1}{2^B B!}\sum_{s\in\{\pm1\}^B}\sum_{\pi\in S_B}\sP(X\in ((A_{(\pi')^{-1}})_{\pi^{-1}})_s)
    \end{align*}Notice that \begin{align*}
        (A_{(\pi')^{-1}})_{\pi^{-1}} &= \{x:x_\pi\in A_{(\pi')^{-1}}\}\\
        &= \{x:x_{\pi\circ \pi'}\in A\}\\
        &= A_{(\pi\circ \pi')^{-1}}
    \end{align*}Thus, \begin{align*}
        \sum_{\pi\in S_B}\sP(X\in ((A_{(\pi')^{-1}})_{\pi^{-1}})_s) &= \sum_{\pi\in S_B}\sP(X\in (A_{\pi^{-1}})_s)
    \end{align*}which finally concludes that $\sP(\tilde X_{\pi'}\in A)=\sP(\tilde X\in A)$. This shows exchangeability. To see symmetry, take a sign vector $s'\in\{\pm1\}^B$. Then,
    \begin{align*}
        \sP(\tilde X_{s'}\in A) &= \sP(\tilde X \in A_{s'})\\
        &= \dfrac{1}{2^BB!}\sum_{s\in\{\pm1\}^B}\sum_{\pi\in S_B}\sP(X\in ((A_{s'})_{\pi^{-1}})_s)
    \end{align*}Observe that for any permutation $\pi$ and sign vector $s$, 
    \begin{align*}
        (A_s)_{\pi^{-1}} &= \{(s\odot x)_{\pi^{-1}}:x\in A\}\\
        &= \{s_{\pi^{-1}}\odot x_{\pi^{-1}}:x\in A\}\\
        &= \{x_{\pi^{-1}}:x\in A\}_{s_{\pi^{-1}}}\\
        &= (A_{\pi^{-1}})_{s_{\pi^{-1}}}
    \end{align*}Thus, $((A_{s'})_{\pi^{-1}})_s=(A_{\pi^{-1}})_{s'_{\pi^{-1}}s}$ which implies
    \begin{align*}
        \sum_{s\in\{\pm1\}^B}\sP(X\in ((A_{s'})_{\pi^{-1}})_s) &= \sum_{s\in\{\pm1\}^B}\sP(X\in (A_{\pi^{-1}})_{s'_{\pi^{-1}}s})\\
        &= \sum_{s\in{\pm1\}^B}}\sP(X\in (A_{\pi^{-1}})_s)
    \end{align*}and once again this concludes $\sP(\tilde X_{\pi'}\in A)=\sP(\tilde X\in A)$. This shows symmetry.
    Thus, $\tilde \nu$ is symmetric exchangeable. Finally, we want to show that for $\eta$ of the type \ref{eqn:eta_type}, for any $\sigma^2$, $R(\nu;\eta,\sigma^2I)=R(\tilde\nu;\eta,\sigma^2 I)$. To see this, 
    \begin{align*}
        R(\tilde\nu;\eta,\sigma^2 I) &= \dfrac{1}{2^BB!}\sum_{\pi\in S_B}\sum_{s\in\{\pm1\}^B}\dfrac{1}{B}\E\|\eta((m\odot s)_\pi+\sigma z;\sigma^2 I)-(m\odot s)_\pi\|^2\\
        &= \dfrac{1}{2^BB!}\sum_{\pi\in S_B}\sum_{s\in\{\pm1\}^B}\dfrac{1}{B}\E\|\eta((m\odot s)_\pi+\sigma (z\odot s)_\pi;\sigma^2 I)-(m\odot s)_\pi\|^2\\
        &= \dfrac{1}{B}\E\|\eta(m+\sigma z;\sigma^2 I)-m\|^2\\
        &= R(\nu;\eta,\sigma^2 I)
    \end{align*}Consequently, we have a distribution $\mu=\tilde \nu\in\gF(\epsilon,B)$ which is symmetric exchangeable, such that the Scalar State Evolution produced by Algorithm \ref{alg:array_cs} on $\nu$ is exactly identical to the Matricial State Evolution produced by Algorithm \ref{alg:general_sense} on $\mu$, iteration by iteration. Since Scalar State Evolution does not decay to $0$ for $\nu$, Matricial State Evolution also does not decay to $0$ for $\mu$. The conclusion then follows by applying Theorem \ref{thm:general_vector_cs_se} with the pseudo-Lipschitz function $\psi(x^t,x)=\|x^t-x\|^2/B$.
\end{proof}

\section{Optimality of BlockSoft Thresholding at Extreme Sparsity}\label{sec:blocksoft_optimality_extreme_sparsity}

We have seen that for any fixed $\epsilon\in(0,1)$,
\begin{align*}
    \lim_{B\to\infty} M_\JS(\epsilon,B) = \epsilon < 2\epsilon-\epsilon^2=\lim_{B\to\infty} M_\BST(\epsilon,B)
\end{align*}
This result might enable one to feel that SteinSense should be the go-to algorithm. The following result shows that BlockSoft Thresholding is minimax optimal in the limit of extreme sparsity, and thus when the nonzero entries of the vectors are symmetric and exchangeable, SoftSense will be minimax optimal over the class $\gF(\epsilon,B)$.

\begin{theorem}\label{thm:extreme_sparsity_minimax}
    Consider the Gaussian vector mean denoising problem where the goal is to estimate $m\in\sR^B$ given data $y\sim \gN_B(m,I_B)$ with $m\sim\mu\in\gF(\epsilon,B)$. Define the (global) minimax risk
    \begin{align*}
        M_\MM(\epsilon,B) &= \inf_\eta \sup_{\mu\in\gF(\epsilon,B)} R(\mu;\eta)
    \end{align*}where the infimum is taken over all possible denoisers $\eta:\sR^B\to\sR^B$ and as before,
    \begin{align*}
        R(\mu;\eta) &= \dfrac{1}{B}\E\|\eta(y)-m\|^2
    \end{align*}denotes the average coordinate-wise square error risk. Then, as $\epsilon\to 0$,
    \begin{align*}
        M_\BST(\epsilon,B) = M_\MM(\epsilon,B)(1+o_\epsilon(1))
    \end{align*}
\end{theorem}

\begin{remark}
    The result is well known for $B=1$; see \citet{johnstone2002function}. Happily, Theorem \ref{thm:extreme_sparsity_minimax} continues to be true for any $B\geq 1$.
\end{remark}

Towards proving Theorem \ref{thm:extreme_sparsity_minimax}, we need a few lemmas. Define, for $\tau\geq 1$ and $b\in\sR$,
\begin{align*}
    I(\tau^2;b) &= \int_{\tau^2}^\infty x^{b/2-1}e^{-x/2}dx
\end{align*}Note that since $\tau\geq 1$, $I(\tau^2;b)<\infty$ for any $b\in\sR$. Also, for $b>0$, denoting by $\chi^2_b$ a chi-squared random variable with $b$ degrees of freedom,
\begin{align}\label{eqn:chi_i_relation}
    \sP(\chi^2_b>\tau^2) &= \dfrac{I(\tau^2;b)}{2^{b/2}\Gamma(b/2)}
\end{align}

\begin{lemma}\label{lemma:I_tail}
    As $\tau\to\infty$, for any $b\in\sR$,
    \begin{align*}
        I(\tau^2;b) &= 2\tau^{b-2}e^{-\tau^2/2}(1 + O(\tau^{-2}))
    \end{align*}
\end{lemma}

\begin{proof}[Proof of Lemma \ref{lemma:I_tail}]
Integrating by parts, we get, for any $b\in\sR$,
\begin{align}\label{eqn:by_parts}
    I(\tau^2;b) &= 2\tau^{b-2}e^{-\tau^2/2}+(b-2)I(\tau^2;b-2)
\end{align}Note that for $\nu\leq 2$, $I(\tau^2;\nu)\leq 2\tau^{\nu-2}e^{-\tau\\^2/2}$. If $b\leq 2$, then taking $\nu=b-2$, we get
\begin{align*}
    I(\tau^2;b) &= 2\tau^{b-2}e^{-\tau^2/2}+O(\tau^{b-4}e^{-\tau^2/2})
\end{align*}which proves the claim. If $b>2$, then take $k(b)$ to be the unique positive integer such that $b-2k(b)\in(0,2]$. We know that $I(\tau^2;b-2k(b))=O(\tau^{b-2k(b)-2}e^{-\tau^2/2})$, and thus, using Equation \ref{eqn:by_parts},
\begin{align*}
    I(\tau^2;b-2k(b)+2) &= 2\tau^{b-2k(b)}e^{-\tau^2/2} + O(\tau^{b-2k(b)-2}e^{-\tau^2/2})\\
    &= 2\tau^{b-2k(b)}e^{-\tau^2/2}(1 + 
    O(\tau^{-2}))
\end{align*}and thus, in particular, $I(\tau^2;b-2k(b)+2) = O(\tau^{b-k(b)}e^{-\tau^2/2})$. Iterating this $k(b)$ times, we get the desired result.
\end{proof}

Using Lemma 3.2 in \citet{donoho2013accurate}, given the sparsity level $\epsilon$, the minimax threshold $\tau(\epsilon)$ is given by the solution $\tau$ to $h(\tau)=1/\epsilon-1$, where
\begin{align*}
    h(\tau^2) &= \dfrac{\tau}{\E\left[\left(\sqrt{\chi^2_B}-\tau\right)_+\right]}
\end{align*}Consequently, it is important to derive the asymptotic behavior of the denominator.

\begin{lemma}\label{lemma:asymp_deno}
As $\epsilon\to 0$, 
\begin{align*}
    \E\left[\left(\sqrt{\chi^2_B}-\tau\right)_+\right] &= \dfrac{2\tau^{B-3}e^{-\tau^2/2}}{2^{B/2}\Gamma(B/2)}\left(1+O(\tau^{-2})\right)
\end{align*}
\end{lemma}

\begin{proof}[Proof of Lemma \ref{lemma:asymp_deno}]
    Using Equation \ref{eqn:chi_i_relation}, \begin{align*}
    \E\left[\left(\sqrt{\chi^2_B}-\tau\right)_+\right] &= \dfrac{1}{2^{B/2}\Gamma(B/2)}\left(I(\tau^2;B+1)-\tau I(\tau^2;B)\right)
\end{align*}Using Equation \ref{eqn:by_parts} repeatedly and some algebra yields
\begin{align*}
    I(\tau^2;B+1)-\tau I(\tau^2;B) &= 2\tau^{B-3}e^{-\tau^2/2}+ (B-1)(B-3)I(\tau^2;B-3) - (B-2)(B-4)\tau I(\tau^2;B-4)
\end{align*}By Lemma \ref{lemma:I_tail} applied to $b=B-3$ and $b=B-4$, we get $I(\tau^2;B-3)=O(\tau^{B-5}e^{-\tau^2/2})$ and $I(\tau^2;B-4)=O(\tau^{B-6}e^{-\tau^2/2})$, and therefore
\begin{align*}
    \E\left[\left(\sqrt{\chi^2_B}-\tau\right)_+\right] &= \dfrac{1}{2^{B/2}\Gamma(B/2)}\left[2\tau^{B-3}e^{-\tau^2/2}+O(\tau^{B-5}e^{-\tau^2/2})\right]\\
    &= \dfrac{2\tau^{B-3}e^{-\tau^2/2}}{2^{B/2}\Gamma(B/2)}(1 + O(\tau^{-2}))
\end{align*}
\end{proof}

The following lemma provides an asymptotic characterization of the minimax threshold $\tau(\epsilon;B)$ used for BlockSoft Thresholding, as $\epsilon\to 0$.

\begin{lemma}\label{lemma:BST_minimax_thresh}
    Let $\tau(\epsilon)$ denote the minimax threshold for BlockSoft Thresholding. Then, $\tau(\epsilon)=\sqrt{2\log(1/\epsilon)}(1+o_\epsilon(1))$ as $\epsilon\to 0$.
\end{lemma}

\begin{proof}[Proof of Lemma \ref{lemma:BST_minimax_thresh}]
    Recall that $\tau(\epsilon)$ is the solution to 
    \begin{align}\label{eqn:minimax_solution}
        \dfrac{\tau}{\E\left[\left(\sqrt{\chi^2_B}-\tau\right)_+\right]} &= \dfrac{1}{\epsilon}-1
    \end{align}
    First of all, perhaps by passing to a subsequence, we show that $\tau(\epsilon)\to\infty$ as $\epsilon\to 0$. If $\tau(\epsilon)\leq C$ for a constant $C$,
    \begin{align*}
        \E\left[\left(\sqrt{\chi^2_B}-\tau\right)_+\right] &\geq \E\left[\left(\sqrt{\chi^2_B}-C\right)_+\right]
    \end{align*}implying the left side of \Eqref{eqn:minimax_solution} is bounded while the right side becomes unboundedly large as $\epsilon\to 0$. This is a contradiction, hence $\tau(\epsilon)\to\infty$.

    Denote $\tau_\epsilon:=\sqrt{2\log(1/\epsilon)}$ as the posited correct asymptotic behavior of $\tau(\epsilon)$. We will show that $\tau^2(\epsilon)/\tau_\epsilon^2\to 1$ as $\epsilon\to 0$.

    Using Lemma \ref{lemma:asymp_deno}, as $\epsilon\to 0$, $\tau(\epsilon)$ satisfies
    \begin{align}\label{eqn:tau_solution_minimax}
        \tau(\epsilon)^{4-B}\exp(\tau^2(\epsilon)/2) &= C_Bg(\tau(\epsilon))\left(\dfrac{1}{\epsilon}-1\right)
    \end{align}for a constant $C_B>0$ depending only on $B$, and a function $g(\tau(\epsilon))=1+O(\tau(\epsilon)^{-2})$. Consequently, taking logs,
    \begin{align}\label{eqn:final_minimax_equation}
        (4-B)\log(\tau(\epsilon))+\dfrac{1}{2}\tau^2(\epsilon) - \log\left(\dfrac{1}{\epsilon}\right)-\log(1-\epsilon)-\log(C_B)-\log(g(\tau^2(\epsilon))) &= 0
    \end{align}
    If $\tau^2(\epsilon)\geq \tau_\epsilon^2(1+\alpha)$, then the left side of Equation \ref{eqn:final_minimax_equation} eventually exceeds $\alpha\log(1/\epsilon)/2$ as $\epsilon\to 0$, while the right side is zero. If $\tau^2(\epsilon)\leq \tau_\epsilon^2(1-\alpha)$, then the left side is eventually at most $-\alpha \log(1/\epsilon)/2$, while the right side is zero. Consequently, $\tau(\epsilon)/\tau_*\to 1$ as $\epsilon\to 0$ must hold.
\end{proof}

We put these elements together to conclude the precise asymptotic behavior of the minimax risk of BlockSoft Thresholding in the limit of extreme sparsity.

\begin{lemma}\label{lemma:BST_minimax_risk}
As $\epsilon\to 0$, \begin{align*}
    M_\BST(\epsilon,B)=\dfrac{2\epsilon\log(1/\epsilon)}{B}\left(1 + o(1)\right)
\end{align*}
\end{lemma}

\begin{proof}[Proof of Lemma \ref{lemma:BST_minimax_risk}]
    Using Lemma 3.2 in \citet{donoho2013accurate}, it can be derived that
    \begin{align*}
        M_\BST(\epsilon;B) &= \epsilon\left(1 + \dfrac{\tau^2(\epsilon)}{B} + \dfrac{1}{B}\left(\dfrac{1}{\epsilon}-1\right)\E\left(\left(\sqrt{\chi^2_B}-\tau\right)_+^2\right)\right) \\
        &= \epsilon + \dfrac{\epsilon\tau^2(\epsilon)}{B} + \dfrac{1-\epsilon}{B}\E\left(\left(\sqrt{\chi^2_B}-\tau(\epsilon)\right)_+^2\right)
    \end{align*}Write
    \begin{align*}
        \E\left(\left(\sqrt{\chi^2_B}-\tau(\epsilon)\right)_+^2\right) &= \dfrac{1}{2^{B/2}\Gamma(B/2)}\left(I(\tau^2(\epsilon);B+2)-2\tau(\epsilon) I(\tau^2(\epsilon);B+1) + \tau^2(\epsilon) I(\tau^2(\epsilon);B)\right)
    \end{align*}Using Equation \ref{eqn:by_parts} and Lemma \ref{lemma:I_tail} repeatedly, one gets that 
    \begin{align*}
        \E\left(\left(\sqrt{\chi^2_B}-\tau(\epsilon)\right)_+^2\right) &= O(e^{-\tau^2(\epsilon)/2}\tau(\epsilon)^{B-4})=O(\epsilon)
    \end{align*}where the last equality follows from Equation \ref{eqn:tau_solution_minimax}. Thus, $M_\BST(\epsilon;B)=\epsilon\tau^2(\epsilon)/B+O(\epsilon)$ and the result follows by using the asymptotic form of $\tau(\epsilon)$ from Lemma \ref{lemma:BST_minimax_thresh}.
\end{proof}

We need a final lemma to complete the proof of Theorem \ref{thm:extreme_sparsity_minimax}.

\begin{lemma}\label{lemma:minimax_risk_global}
We have, as $\epsilon\to 0$,
    \begin{align*}
    M_{\MM}(\epsilon,B) &\geq \dfrac{2\epsilon\log(1/\epsilon)}{B}\left(1+o(1)\right)
\end{align*}
\end{lemma}

\begin{remark}
    It is well-known \citep{johnstone2002function} that computing the (global) minimax risk on $\gF(\epsilon,B)$ boils down to computing the Bayes risk with respect to the least favorable prior on $\gF(\epsilon,B)$. Unfortunately, even for $B=1$, exact knowledge of this least favorable prior is difficult, and constitutes the infamous Mallow's Conjecture \citep{mallows1978minimizing}; see \citet{johnstone1994minimax} for a partial resolution. However, when $\epsilon$ is small, it is possible to find an approximately least favorable prior, whose Bayes risk, as we will show, is at least $2\epsilon\log(1/\epsilon)(1+o(1))/B$. The methodology follows Section 8.5 in \citet{johnstone2002function}.
\end{remark}

\begin{proof}[Proof of Lemma \ref{lemma:minimax_risk_global}]
    Recall that without loss of generality, we may consider the mean $\mu\in\sR^B$ to be of the form $\|\mu\|e_1$ with $e_1=(1,0,\cdots,0)\in\sR^B$ being the first elementary vector. The approximate least favorable prior for $\mu$ is a two-point prior 
\begin{align*}
    \pi_{\epsilon,a} = (1-\epsilon)\delta_0+\epsilon\delta_{ae_1}
\end{align*}with a carefully chosen value of $a$. The posterior distribution for $\mu$ given $Y\sim \gN_B(\mu,I_B)$ is supported on $\{0,ae_1\}$ as well, with posterior probability
\begin{align*}
    \pi(ae_1|Y) &= \dfrac{\epsilon \phi(Y_1-a)\prod_{j=2}^B \phi(Y_j)}{(1-\epsilon)\prod_{j=1}^B\phi(Y_j) + \epsilon\phi(Y_1-a)\prod_{j=2}^B \phi(Y_j)}\\
    &= \dfrac{\epsilon\phi(Y_1-a)}{(1-\epsilon)\phi(Y_1)+\epsilon\phi(Y_1-a)}\\
    &= \dfrac{1}{1+m(Y_1)}
\end{align*}where $m(x)=(1-\epsilon)\phi(x)/\epsilon\phi(x-a)$, $\phi(\cdot)$ denoting the standard Normal density function. Note that with the definition $\lambda_\epsilon=\sqrt{2\log((1-\epsilon)/\epsilon)}$, and writing $Y_1=a+Z_1$, we conclude that
\begin{align*}
    m(Y_1) &= \exp\left(\dfrac{\lambda_\epsilon^2}{2}-\dfrac{a^2}{2}-aZ_1\right)
\end{align*}Choose $a^2 = \lambda_\epsilon^2 - 2\lambda_\epsilon^{3/2}$. Then, 
\begin{align*}
    m(Y_1) &= \exp\left(\lambda_\epsilon^{3/2} -aZ_1\right)
\end{align*}With this choice of $a$, the Bayes estimator equals the posterior mean for $\pi_{\epsilon,\mu}$:
\begin{align*}
    \eta_{\text{Bayes}}(Y) &= \dfrac{a}{1+m(Y_1)}e_1
\end{align*}The Bayes risk is then lower bounded by the contribution to the risk at $ae_1$:
\begin{align*}
    R_{\text{Bayes}}(\pi_{\epsilon,a}) &\geq \epsilon\times r(ae_1;\eta_{\text{Bayes}})\\
    &= \dfrac{\epsilon a^2}{B}\E\left(\dfrac{1}{1+\exp(aZ_1-\lambda_\epsilon^{3/2})}\right)^2
\end{align*}where the expectation is taken over $Z_1\sim \gN(0,1)$. Notice that only the first coordinate contributes; the risk at the other coordinates is zero. Now, the function $z\mapsto (1 + \exp(az-\lambda_\epsilon^{3/2})^{-2}$ is uniformly bounded by $1$, and $a\leq \lambda_\epsilon<<\lambda_\epsilon^{3/2}$ as $\epsilon\to 0$. Thus, for any $z\in\sR$, $az-\lambda_\epsilon^{3/2}\to -\infty$ as $\epsilon\to 0$, since $\lambda_\eps\to\infty$. Therefore, by Bounded Convergence Theorem, 
\begin{align*}
    \E\left(\dfrac{1}{1+\exp(aZ_1-\lambda_\epsilon^{3/2})}\right)^2 \to 1
\end{align*}Thus,
\begin{align*}
    M_{\MM}(\epsilon,B)\geq R_{\text{Bayes}}(\pi_{\epsilon,a}) \geq \dfrac{2\epsilon\log(1/\epsilon)}{B}(1+o(1))
\end{align*}
\end{proof}

\begin{proof}[Proof of Theorem \ref{thm:extreme_sparsity_minimax}]
    Since $M_\BST(\epsilon;B)\geq M_\MM(\epsilon;B)$ for any $\epsilon$ and $B$, Lemma \ref{lemma:BST_minimax_risk} and Lemma \ref{lemma:minimax_risk_global} together conclude the proof.
\end{proof}

\section{Conclusion}

We have presented SteinSense - an essentially optimal, lightweight Compressed Sensing algorithm for reconstructing high dimensional vectors from undersampled measurements. SteinSense is proposed as a scalable alternative to Convex Optimization, which overcomes the fundamental performance barrier that Convex Optimization suffers from. The efficacy of SteinSense has been demonstrated through a wide variety of computational experiments on both real and synthetic datasets. SteinSense enjoys the best of both worlds - it is easily scalable for large $B$, and enjoys firm theoretical guarantees coming from the theory of generalized Approximate Message Passing (AMP). We have discovered, through massive experimentation, that SteinSense is fascinatingly robust; the performance of SteinSense remains practically unchanged no matter what distribution is used, no matter if the conditions of the theory hold or not. The experimental data collected so far has provided unprecedented fine-grained insights into the real performance and computational issues associated with Approximate Message Passing algorithms applied to Multiple Measurement Vector recovery problems. This marks the start of our explorations with SteinSense - more experiments will be conducted, more data will be collected, and more plots will be generated, to be ultimately visible in \href{https://vector-cs-plots-apratim.streamlit.app/}{https://vector-cs-plots-apratim.streamlit.app/}.

\bibliographystyle{conference}  % 1. Set the bibliography style

                % 2. Input the precompiled bibliography
% \bibliography{mybib}          % 3. Comment this out for arXiv

% \bibliography{mybib}
% % \input{main.bbl} # for arxiv
% \bibliographystyle{conference}

\appendix

\newpage
\section{More experiments}\label{sec:appdx:cvx_more_experiments}

It is important to remember that SoftSense and SteinSense (Algorithms \ref{alg:softsense} and \ref{alg:steinsense}) deliver the corresponding minimax risks of BlockSoft Thresholding and James Stein under conditions outlined by Theorems \ref{thm:softsense_matching_BST_MSE} and \ref{thm:js_mse}. In particular, we must have $B$ fixed and $N\to\infty$ to see expected results. The reason is that the Approximate Message Passing algorithm provides asymptotic (in $N$) guarantees, and at the end of the day, SoftSense and SteinSense are built on the theoretical grounding of Approximate Message Passing. In real applications, however, we have a fixed $N$ and a fixed $B$. As we see in Section \ref{sec:real_exp}, particularly in Figure \ref{fig:hyperspectral_band_1}, when one performs wavelet decompositions at sufficiently high level, there could be subbands where $B$ is preserved but $N$ gets small. In Figure \ref{fig:hyperspectral_band_1}, $N=361$ while $B=10$ for example. How do our algorithms behave in such cases? Can we trust the theory there as well? If yes, to what extent? Unfortunately, Theorems \ref{thm:softsense_matching_BST_MSE} and \ref{thm:js_mse} do not answer such questions.

Therefore, we have performed a large number of experiments at various small values of $N$ (relative to $B$) until we get a very close match between the empirical and theoretical phase transition curves, for all of SteinSense, Convex Optimization and SoftSense. The results show that indeed, if $N$ is small (relative to $B$), SteinSense and SoftSense may not accurately reflect the minimax risk curves. Sometimes, extreme sparsity is affected. Thankfully, all these problems steadily reduce as $N$ grows, and eventually disappear once $N$ is sufficiently large. Finally, as shown in the plots in the main text, $N$ does not need to be exorbitantly large for high quality fits between empirical phase transition curves and minimax risk curves.

For real applications, it is important to understand the trends displayed by the phase transition curves, in $N$ and $B$. Thus, any anomalous behavior that is experimentally captured needs to be revealed. The captions in the respective figures will explain the essential points in the plots.

\begin{figure}[htbp]
  \centering

  % Row 1
  \begin{subfigure}[b]{0.48\textwidth}
    \includegraphics[width=\linewidth]{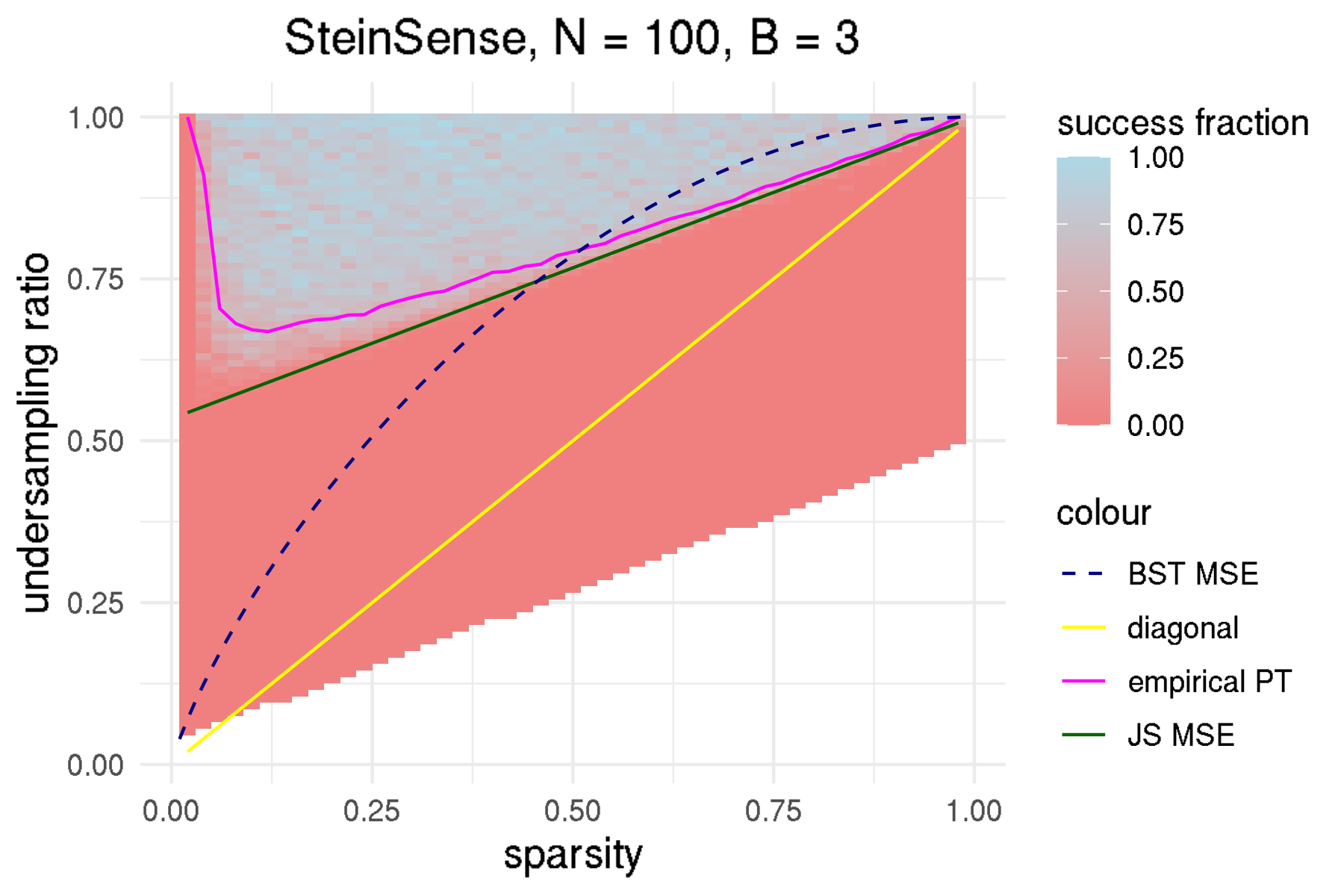}
  \end{subfigure}
  \hfill
  \begin{subfigure}[b]{0.48\textwidth}
    \includegraphics[width=\linewidth]{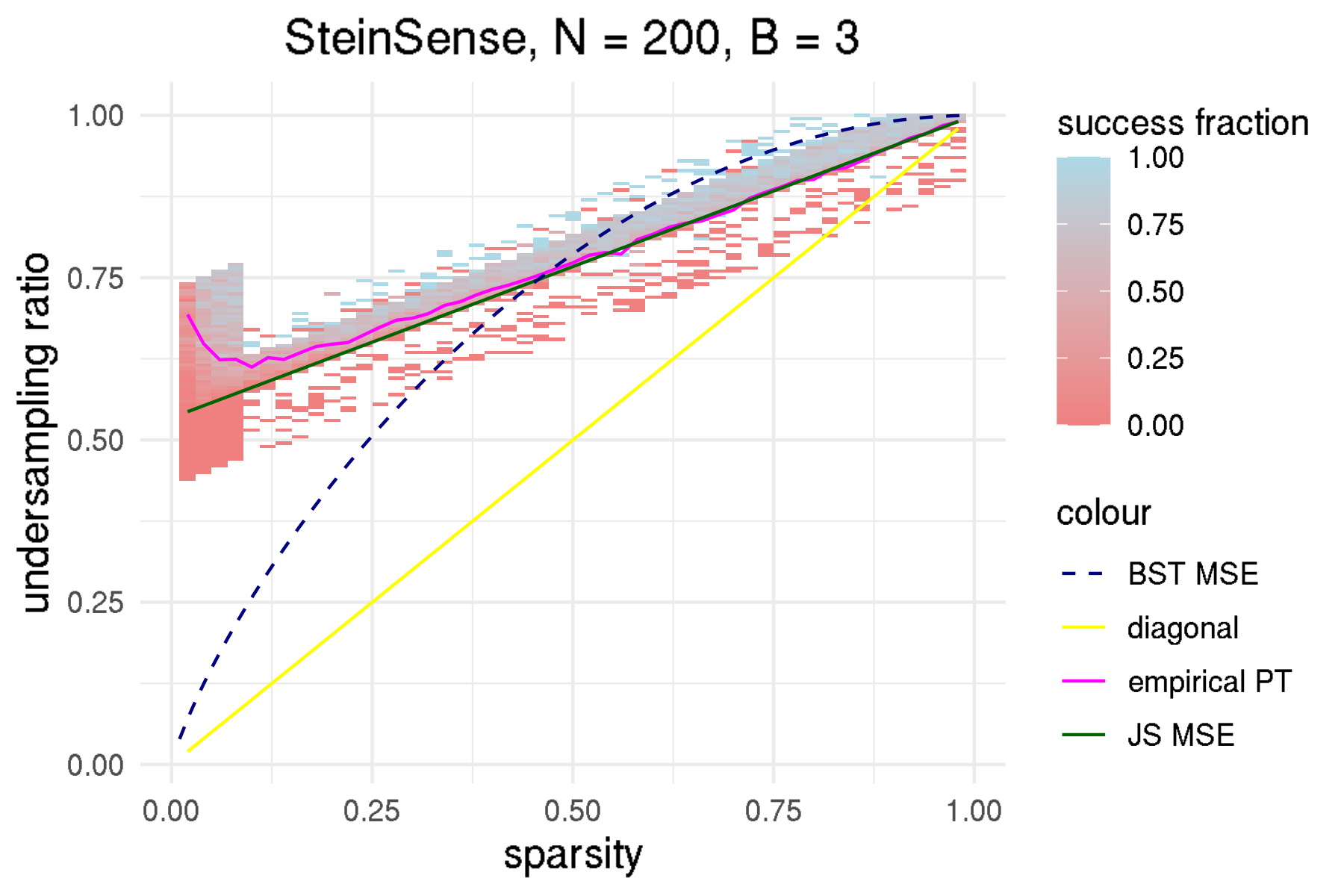}
  \end{subfigure}

  \vspace{1em}

  % Row 2
  \begin{subfigure}[b]{0.48\textwidth}
    \includegraphics[width=\linewidth]{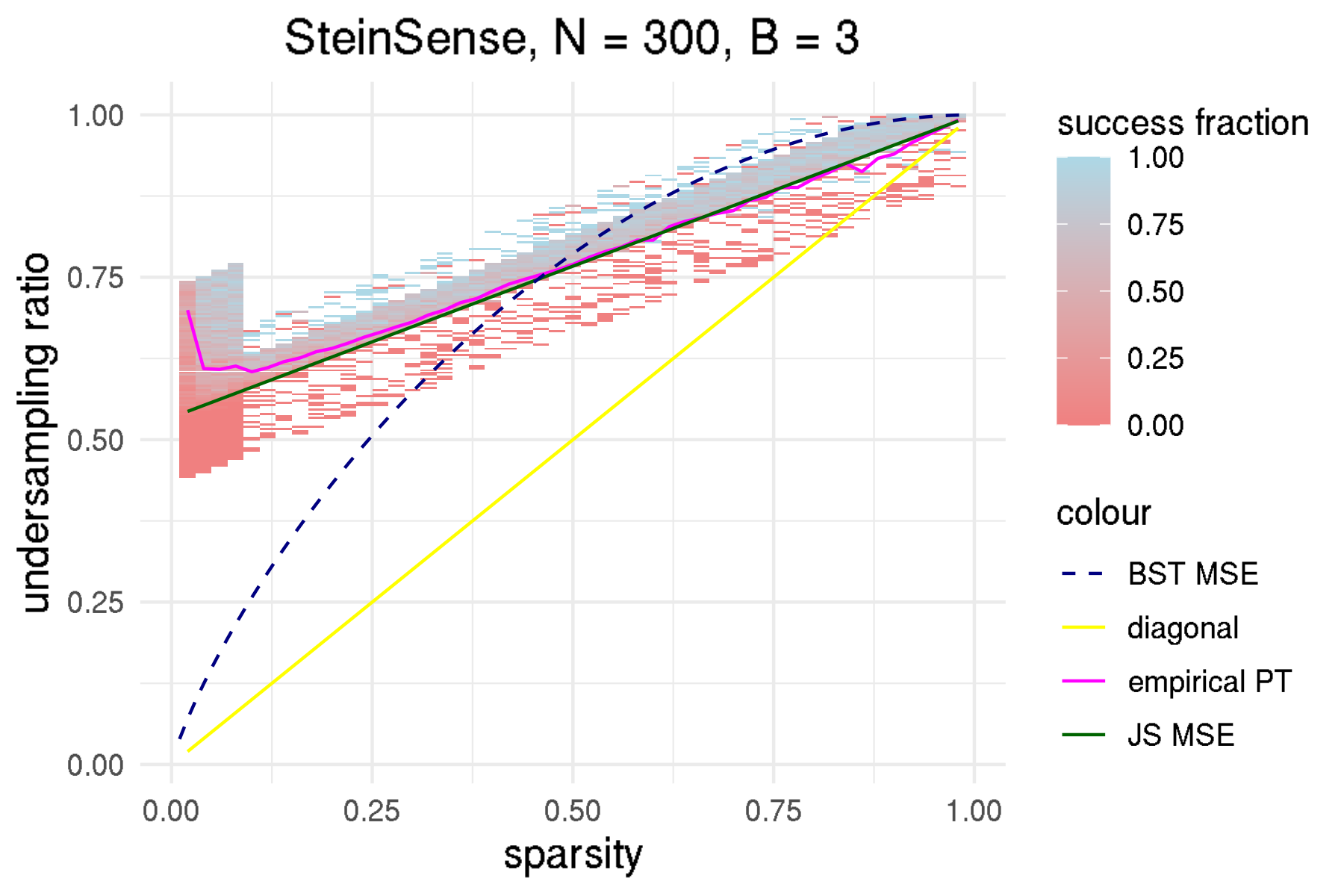}
  \end{subfigure}
  \hfill
  \begin{subfigure}[b]{0.48\textwidth}
    \includegraphics[width=\linewidth]{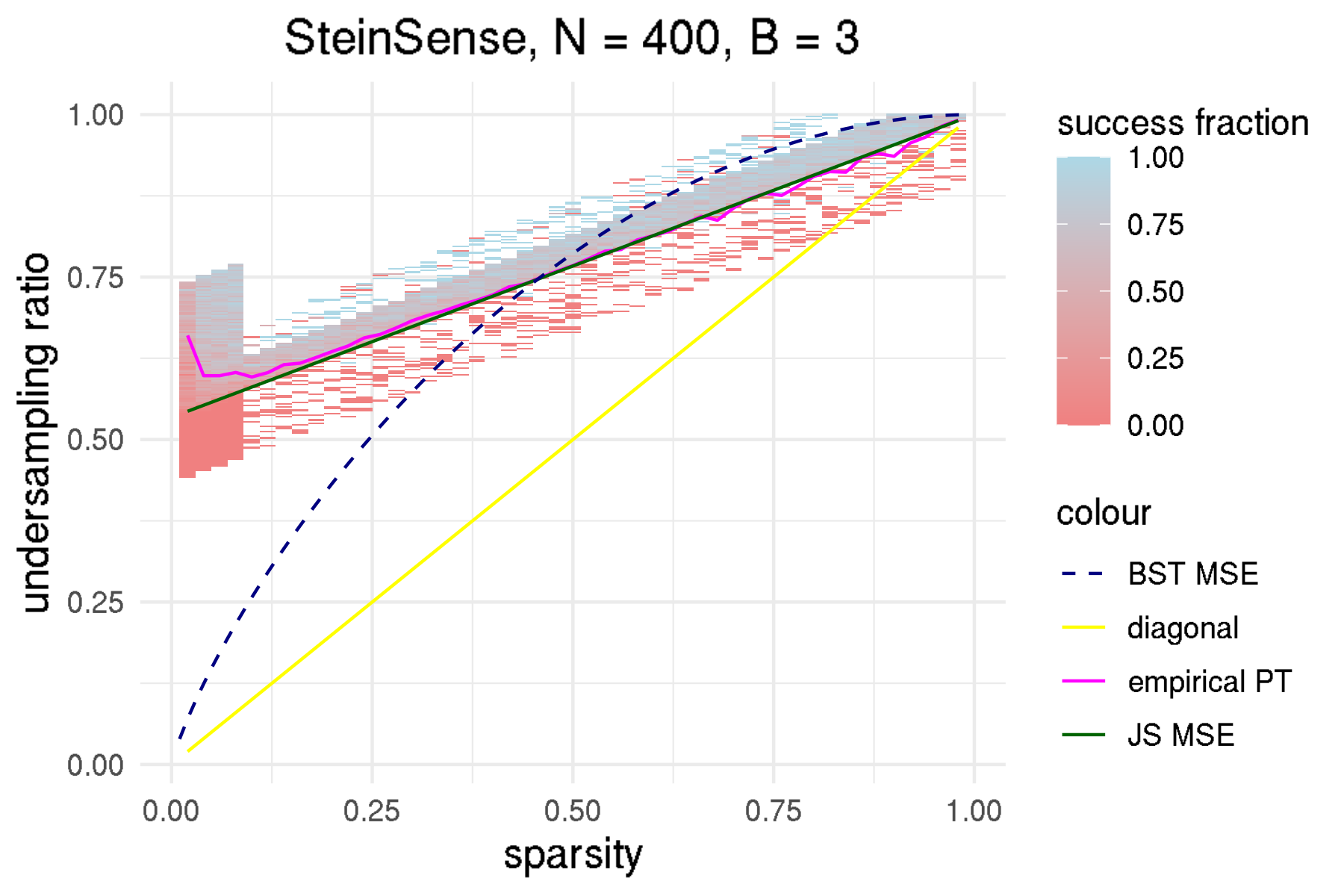}
  \end{subfigure}

  \caption{The nonzero entries are iid $N(0,1)$. The empirical phase transition is fairly bad at low $N$, although it steadily improves as $N$ grows. For $N=100$, practically none of the experiments are successful at extreme sparsity i.e. very small $\epsilon$. Poor performance at extreme sparsity persists even for larger $N$.}
\end{figure}

% \begin{figure}[htbp]
%     \centering
%     \includegraphics[width=\linewidth]{figures/steinsense/paper_steinsense_N_100_B_3.png}
%     \caption{The nonzero entries are iid $N(0,1)$. The empirical phase transition is fairly bad at such low $N$. Practically none of the experiments were successful at extreme sparsity i.e. very small $\epsilon$.}
%     \label{fig:steinsense_normal_N_100_B_3}
% \end{figure}

% \begin{figure}[htbp]
%     \centering
%     \includegraphics[width=\linewidth]{figures/steinsense/paper_steinsense_N_200_B_3.png}
%     \caption{The nonzero entries are iid $N(0,1)$. The empirical phase transition plot has come closer to the James Stein minimax risk curve. Extreme sparsity still suffers.}
%     \label{fig:steinsense_normal_N_200_B_3}
% \end{figure}

% \begin{figure}[htbp]
%     \centering
%     \includegraphics[width=\linewidth]{figures/steinsense/paper_steinsense_N_400_B_3.png}
%     \caption{The nonzero entries are iid $N(0,1)$. The empirical phase transition curve is certainly a bit more aligned, }
%     \label{fig:steinsense_normal_N_400_B_3}
% \end{figure}

\begin{figure}[htbp]
  \centering

  % Row 1
  \begin{subfigure}[b]{0.6\textwidth}
    \includegraphics[width=\linewidth]{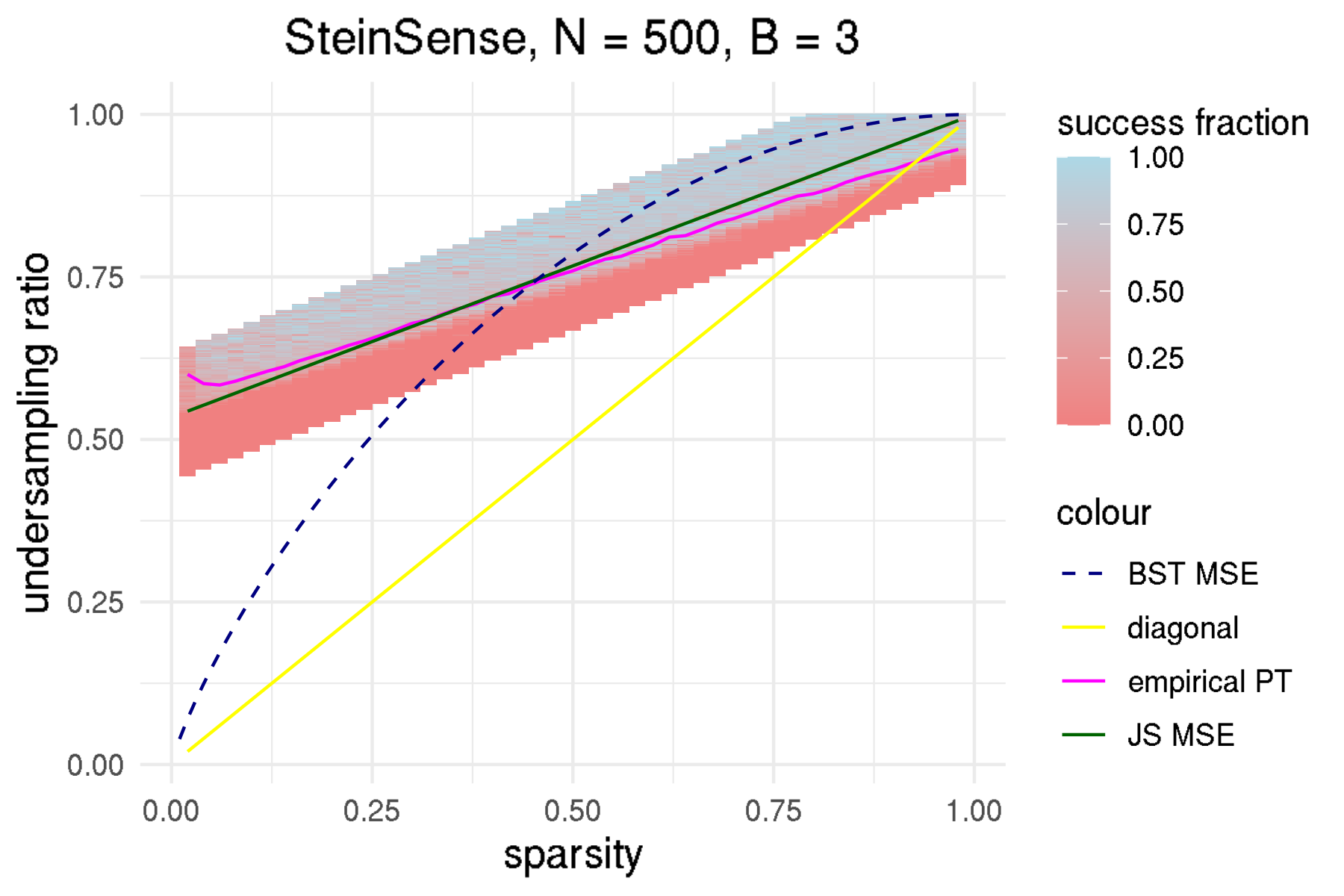}
  \end{subfigure}
  \vspace{1em}
  \begin{subfigure}[b]{0.6\textwidth}
    \includegraphics[width=\linewidth]{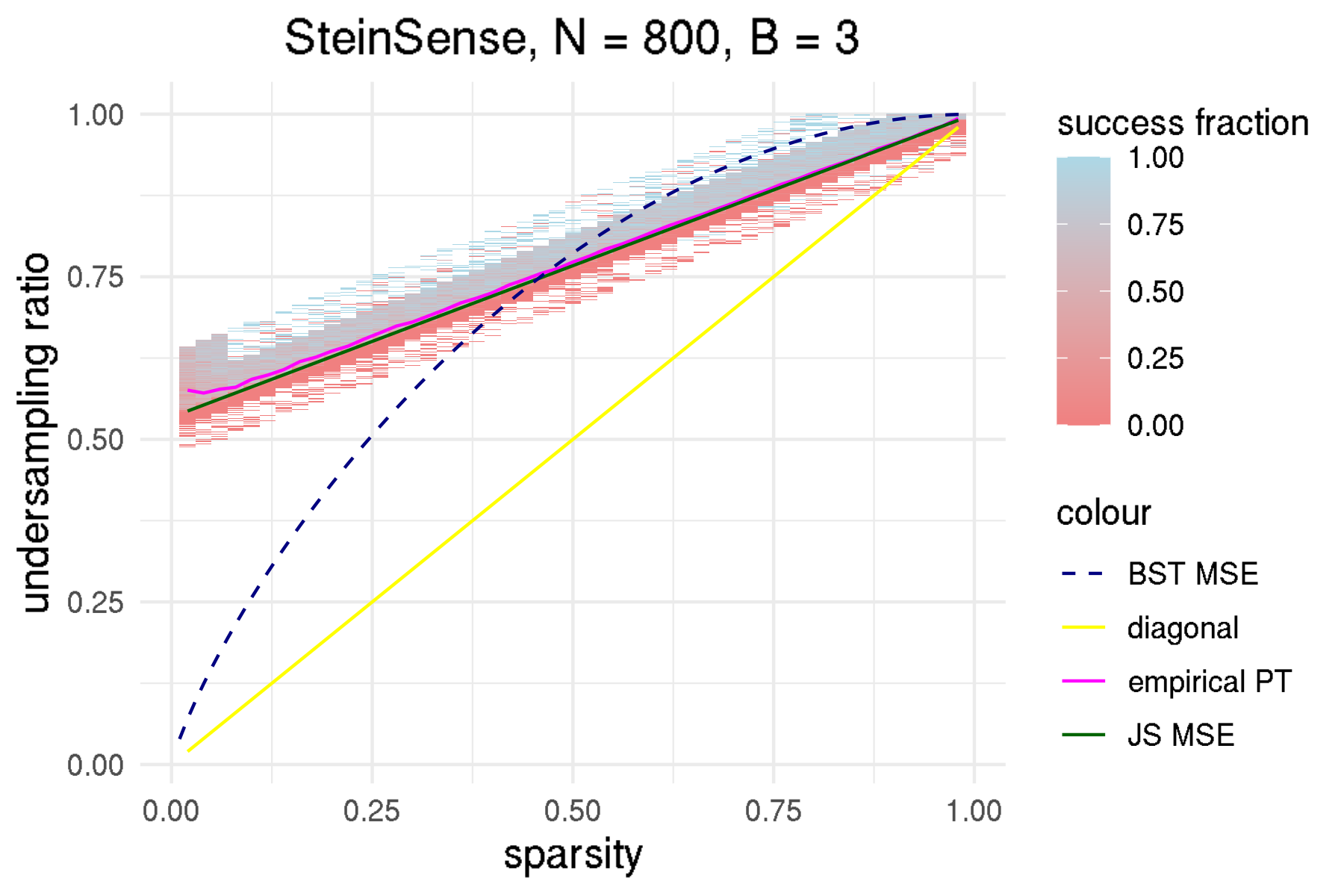}
  \end{subfigure}

  \vspace{1em}

  % Row 2
  \begin{subfigure}[b]{0.6\textwidth}
    \includegraphics[width=\linewidth]{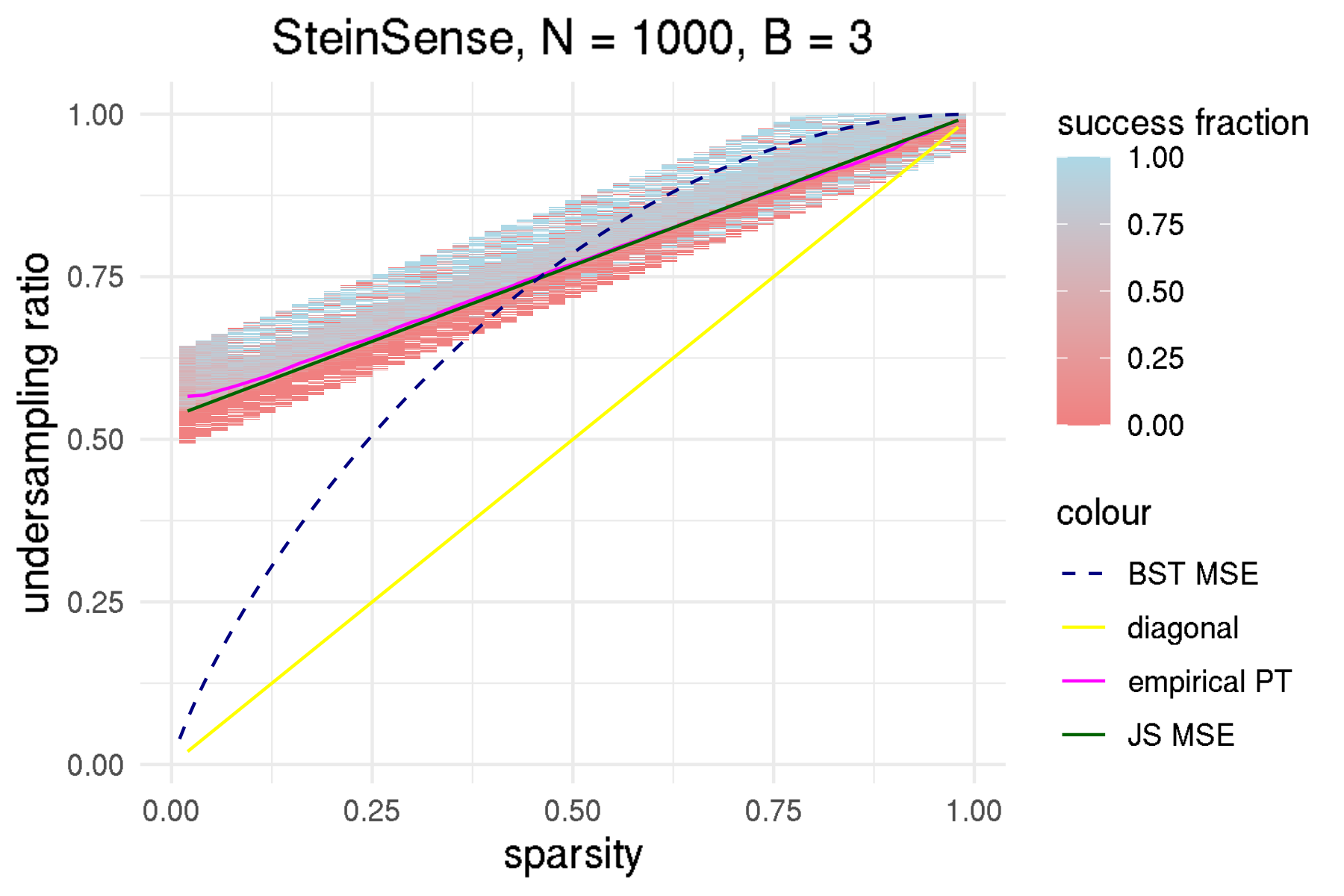}
  \end{subfigure}

  \caption{The nonzero entries are iid $N(0,1)$. The poor performance at low sparsity certainly improves as $N$ gets large, but does not completely disappear. Even for $N=1000$, we see the deterioration at extreme sparsity.}
\end{figure}

% \begin{figure}[htbp]
%     \centering
%     \includegraphics[width=\linewidth]{figures/steinsense/paper_steinsense_N_500_B_3.png}
%     \caption{Caption}
%     \label{fig:steinsense_normal_N_500_B_3}
% \end{figure}

% \begin{figure}[htbp]
%     \centering
%     \includegraphics[width=\linewidth]{figures/steinsense/paper_steinsense_N_800_B_3.png}
%     \caption{Caption}
%     \label{fig:steinsense_normal_N_800_B_3}
% \end{figure}

% \begin{figure}[htbp]
%     \centering
%     \includegraphics[width=\linewidth]{figures/steinsense/paper_steinsense_N_1000_B_3.png}
%     \caption{Caption}
%     \label{fig:steinsense_normal_N_1000_B_3}
% \end{figure}

\begin{figure}[htbp]
  \centering

  % Row 1
  \begin{subfigure}[b]{0.48\textwidth}
    \includegraphics[width=\linewidth]{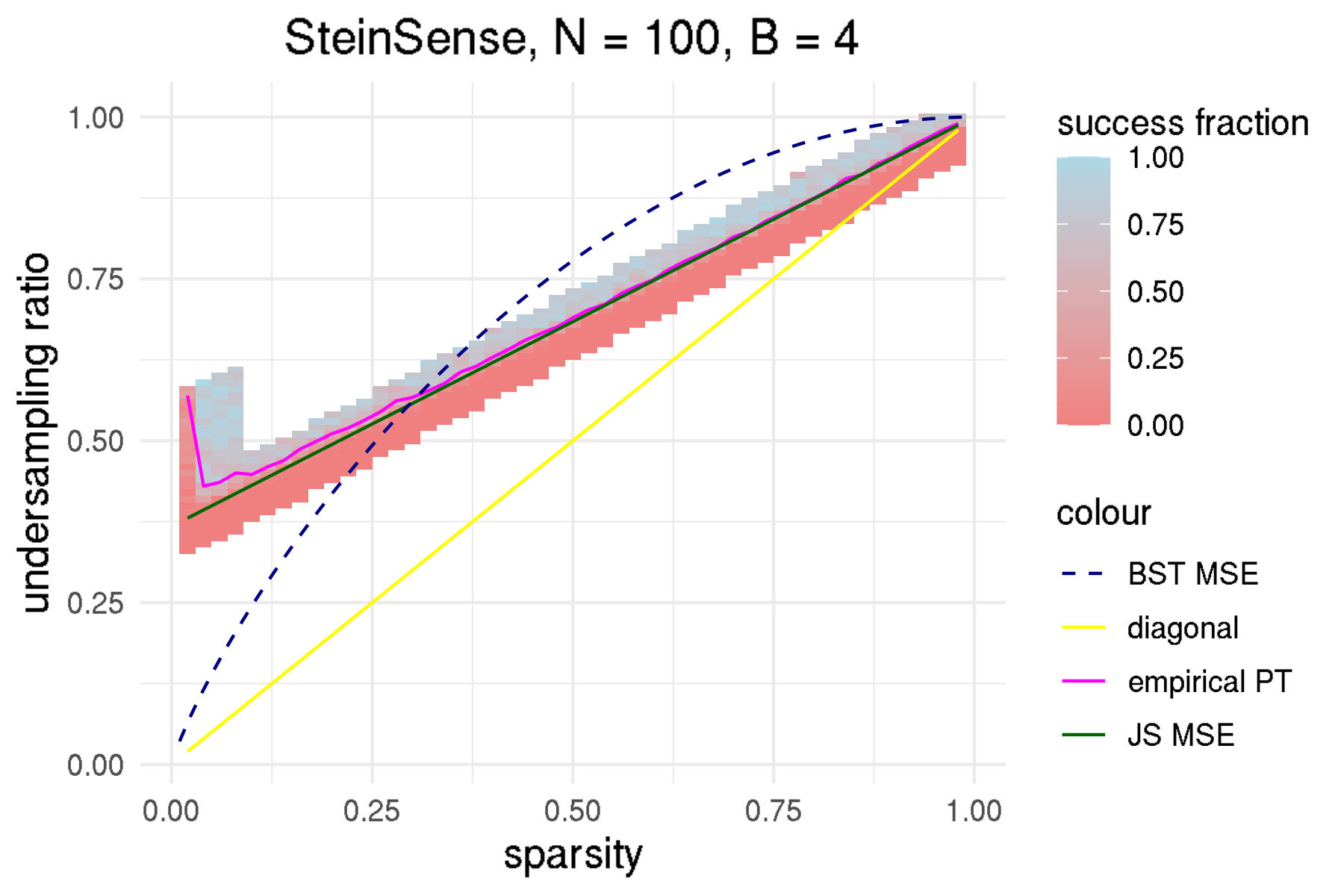}
  \end{subfigure}
  \hfill
  \begin{subfigure}[b]{0.48\textwidth}
    \includegraphics[width=\linewidth]{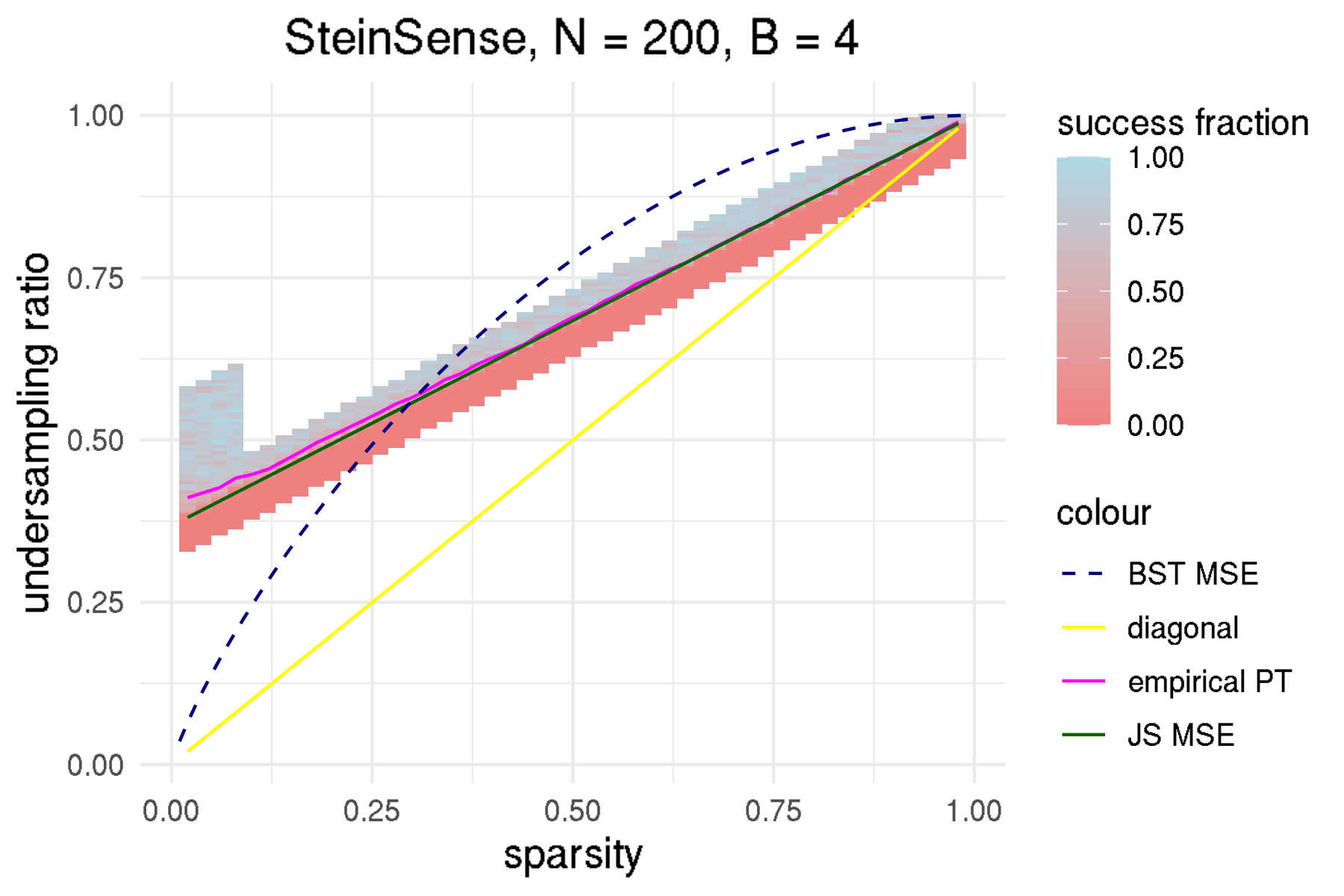}
  \end{subfigure}

  \vspace{1em}

  % Row 2
  \begin{subfigure}[b]{0.48\textwidth}
    \includegraphics[width=\linewidth]{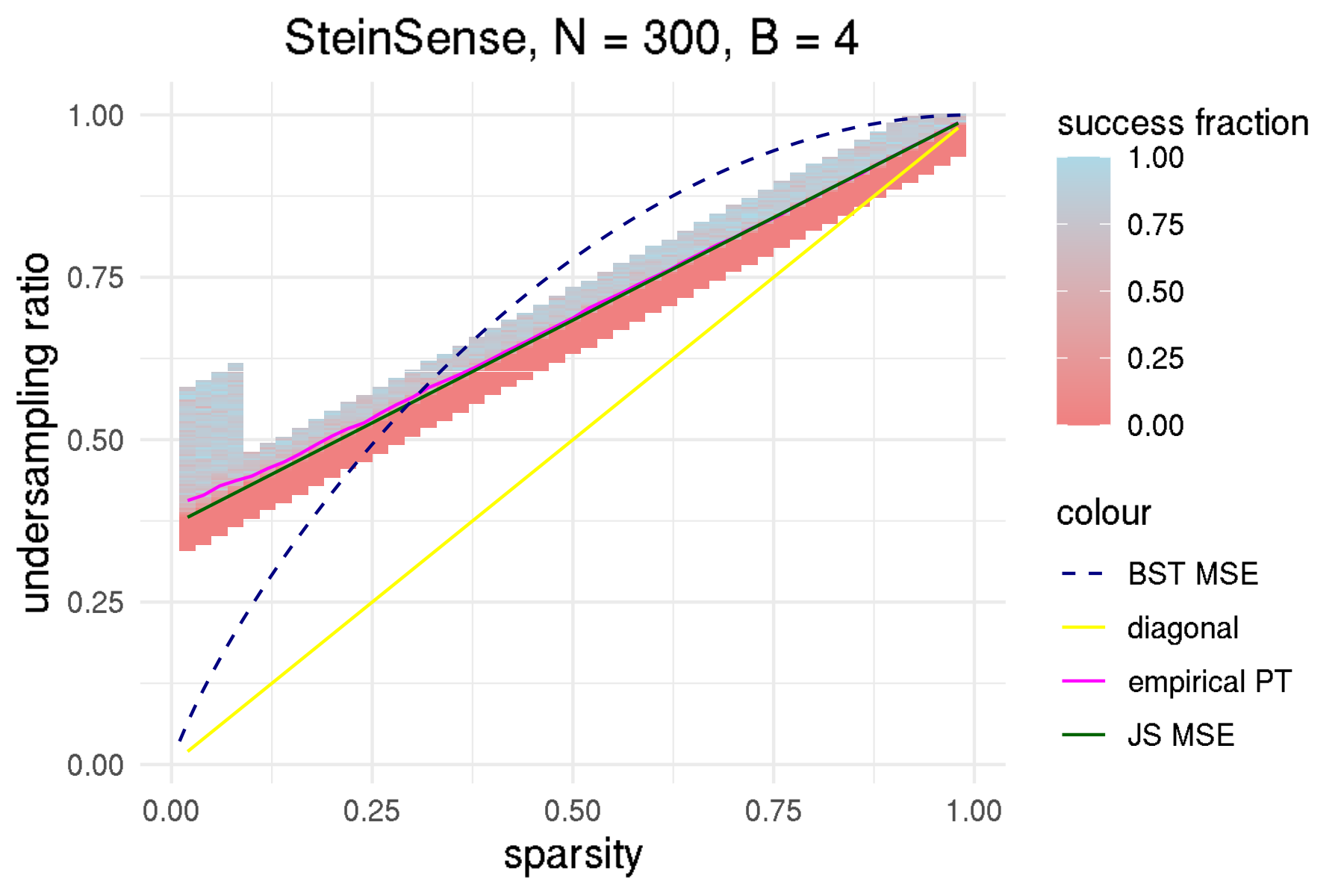}
  \end{subfigure}
    \hfill
  \begin{subfigure}[b]{0.48\textwidth}
    \includegraphics[width=\linewidth]{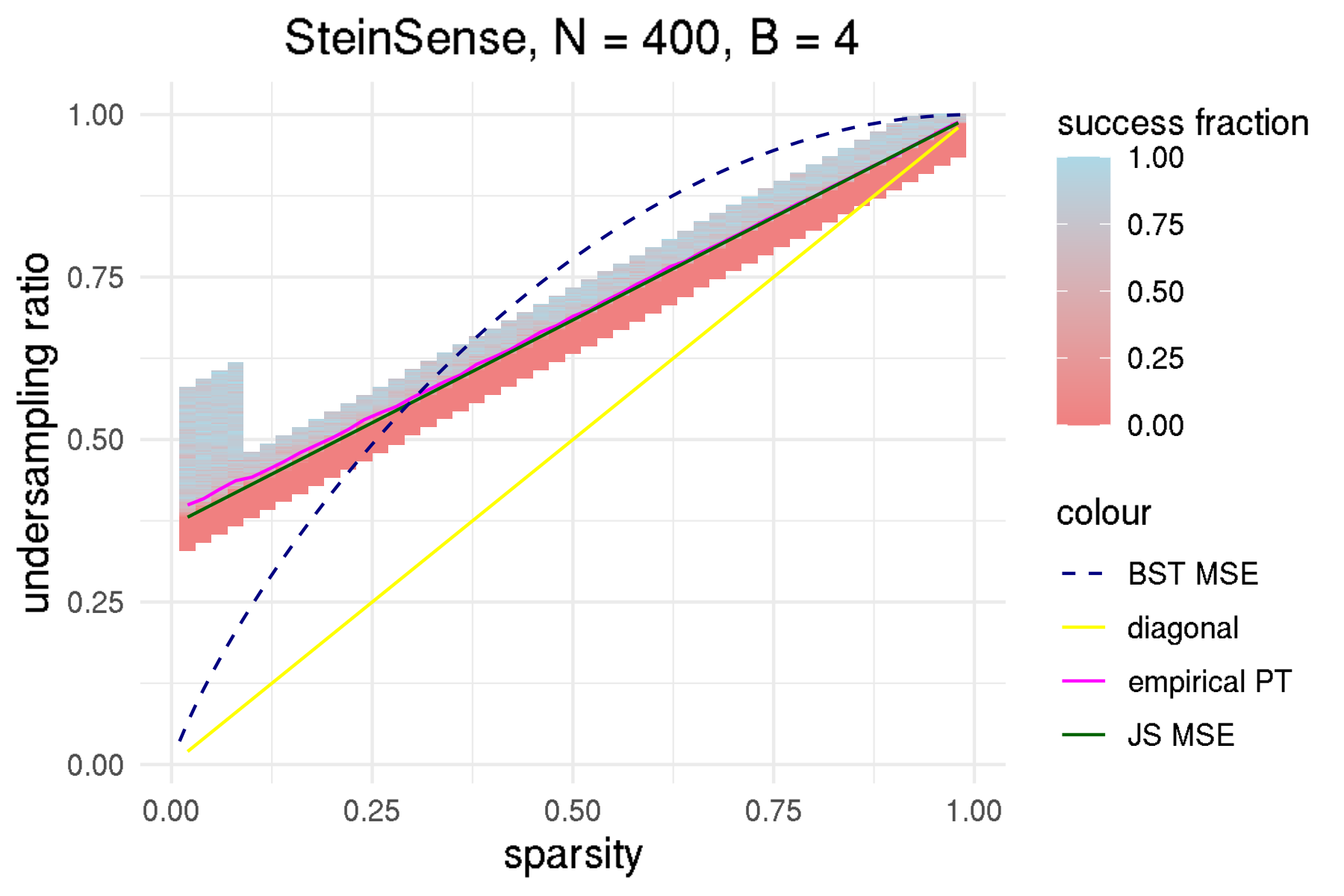}
  \end{subfigure}

  \caption{The nonzero entries are iid $N(0,1)$. The empirical phase transitions match the James Stein minimax risk pretty accurately, except, again, at extreme sparsity.}
\end{figure}

% \begin{figure}[htbp]
%     \centering
%     \includegraphics[width=\linewidth]{figures/steinsense/paper_steinsense_N_100_B_4.png}
%     \caption{Caption}
%     \label{fig:steinsense_normal_N_100_B_4}
% \end{figure}

% \begin{figure}[htbp]
%     \centering
%     \includegraphics[width=\linewidth]{figures/steinsense/paper_steinsense_N_200_B_4.png}
%     \caption{Caption}
%     \label{fig:steinsense_normal_N_200_B_4}
% \end{figure}

% \begin{figure}[htbp]
%     \centering
%     \includegraphics[width=\linewidth]{figures/steinsense/paper_steinsense_N_300_B_4.png}
%     \caption{Caption}
%     \label{fig:steinsense_normal_N_300_B_4}
% \end{figure}

% \begin{figure}[htbp]
%     \centering
%     \includegraphics[width=\linewidth]{figures/steinsense/paper_steinsense_N_400_B_4.png}
%     \caption{Caption}
%     \label{fig:steinsense_normal_N_400_B_4}
% \end{figure}

\begin{figure}[htbp]
  \centering

  % Row 1
  \begin{subfigure}[b]{0.6\textwidth}
    \includegraphics[width=\linewidth]{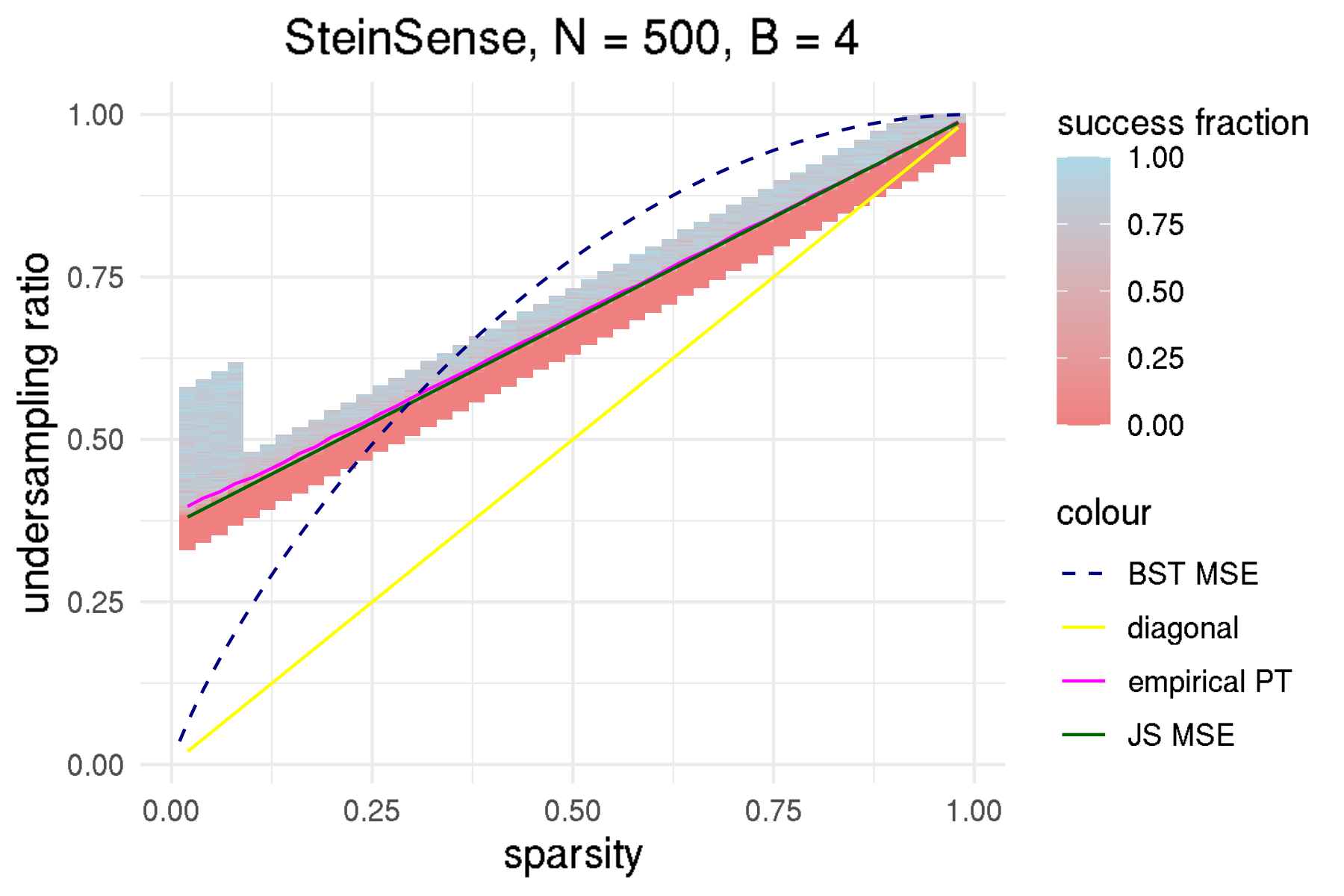}
  \end{subfigure}
  \vspace{1em}
  \begin{subfigure}[b]{0.6\textwidth}
    \includegraphics[width=\linewidth]{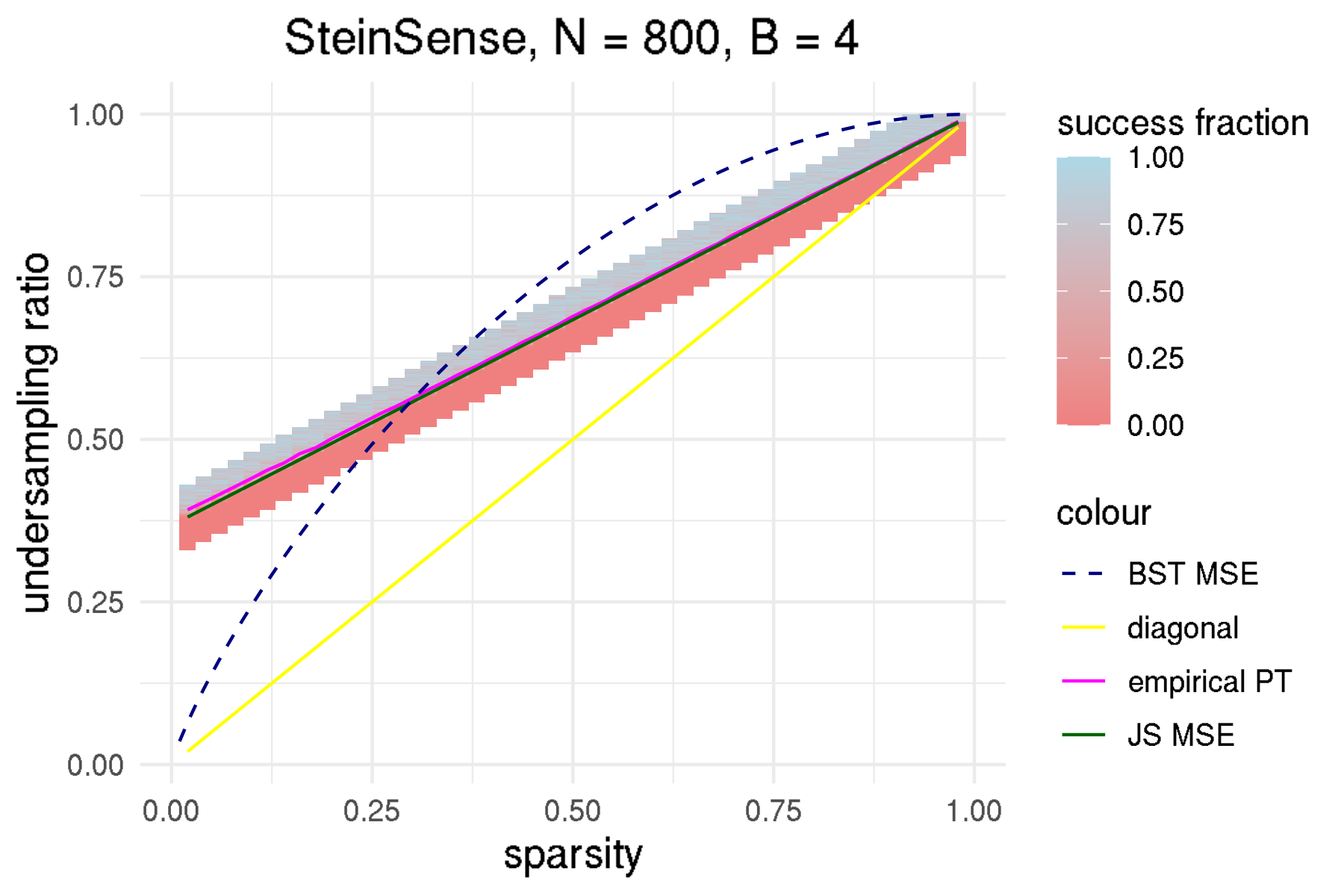}
  \end{subfigure}

  \vspace{1em}

  % Row 2
  \begin{subfigure}[b]{0.6\textwidth}
    \includegraphics[width=\linewidth]{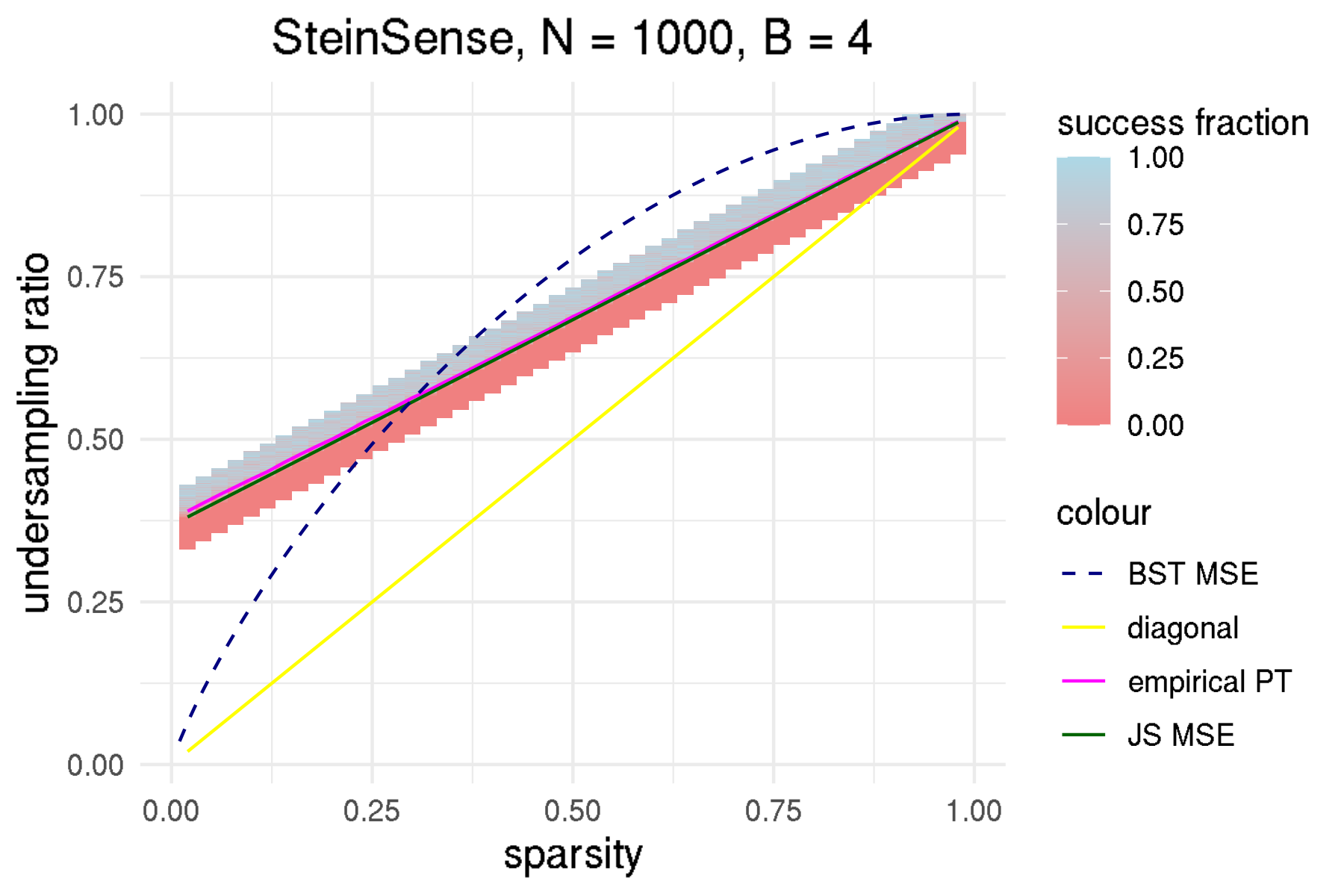}
  \end{subfigure}

  \caption{The nonzero entries are iid $N(0,1)$. The empirical phase transitions match the James Stein minimax risk pretty accurately. For $N\geq 800$, the deterioration at extreme sparsity also disappears.}
\end{figure}

\begin{figure}[htbp]
  \centering

  % Row 1
  \begin{subfigure}[b]{0.9\textwidth}
    \includegraphics[width=\linewidth]{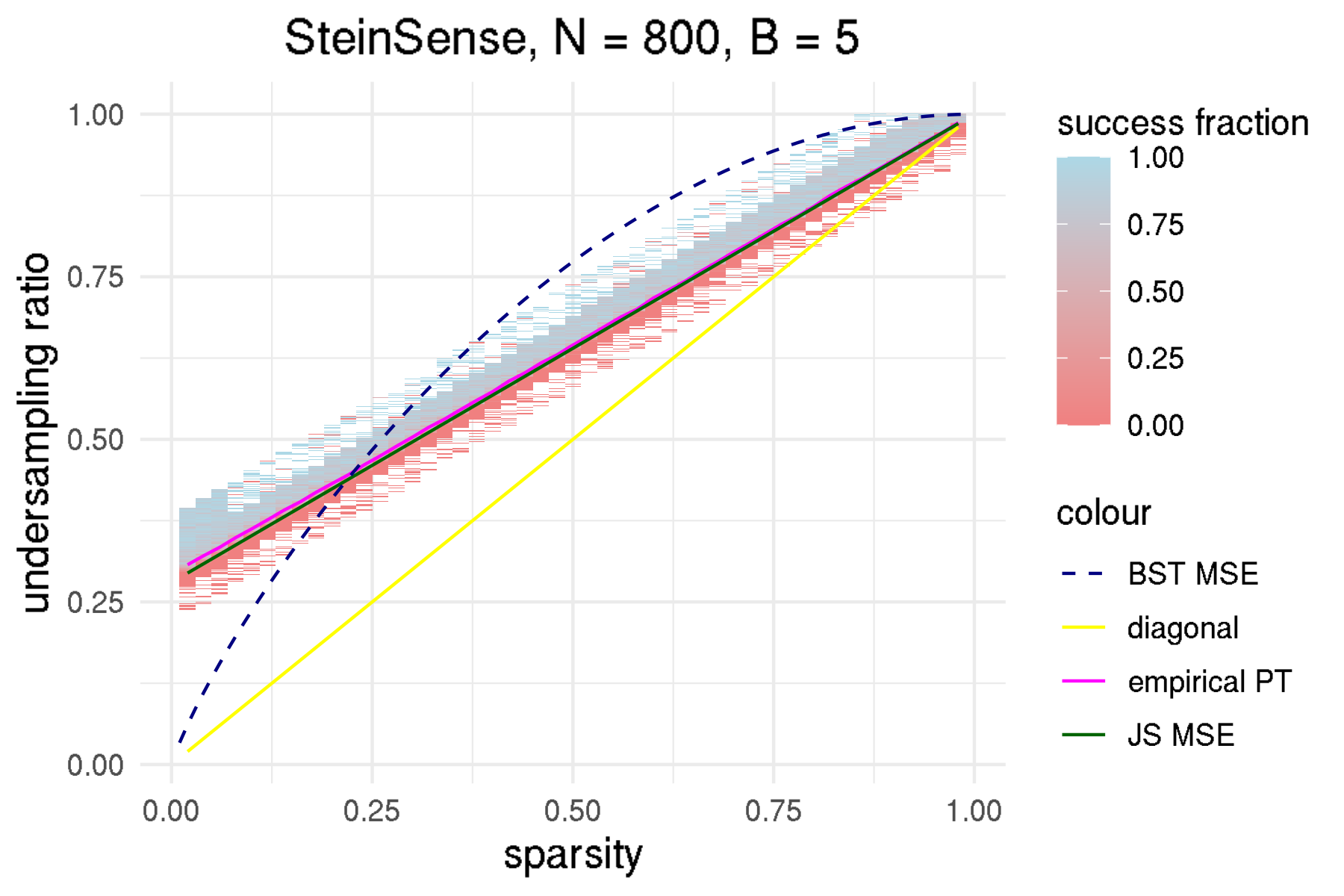}
  \end{subfigure}
    \vspace{1em}
  \begin{subfigure}[b]{0.9\textwidth}
    \includegraphics[width=\linewidth]{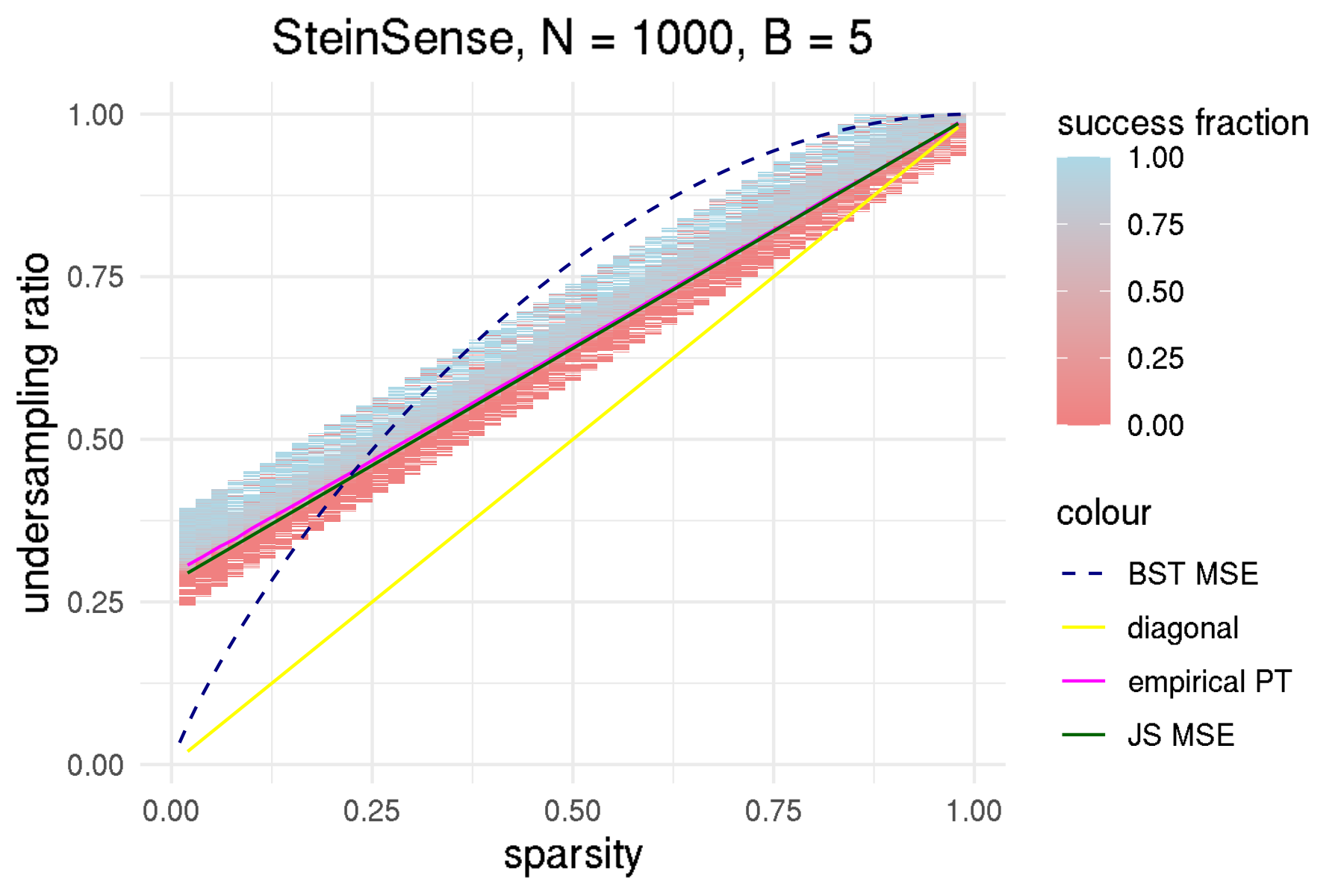}
  \end{subfigure}
  \caption{The nonzero entries are iid $N(0,1)$. The phase transitions match James Stein minimax risk to a high degree of accuracy.}
\end{figure}

% \begin{figure}
%     \centering
%     \includegraphics[width=\linewidth]{figures/steinsense/paper_steinsense_N_800_B_5.png}
%     \caption{Caption}
%     \label{fig:steinsense_normal_N_800_B_5}
% \end{figure}

% \begin{figure}
%     \centering
%     \includegraphics[width=\linewidth]{figures/steinsense/paper_steinsense_N_1000_B_5.png}
%     \caption{Caption}
%     \label{fig:steinsense_normal_N_1000_B_5}
% \end{figure}
% \clearpage

\begin{figure}[htbp]
  \centering

  % Row 1
  \begin{subfigure}[b]{0.9\textwidth}
    \includegraphics[width=\linewidth]{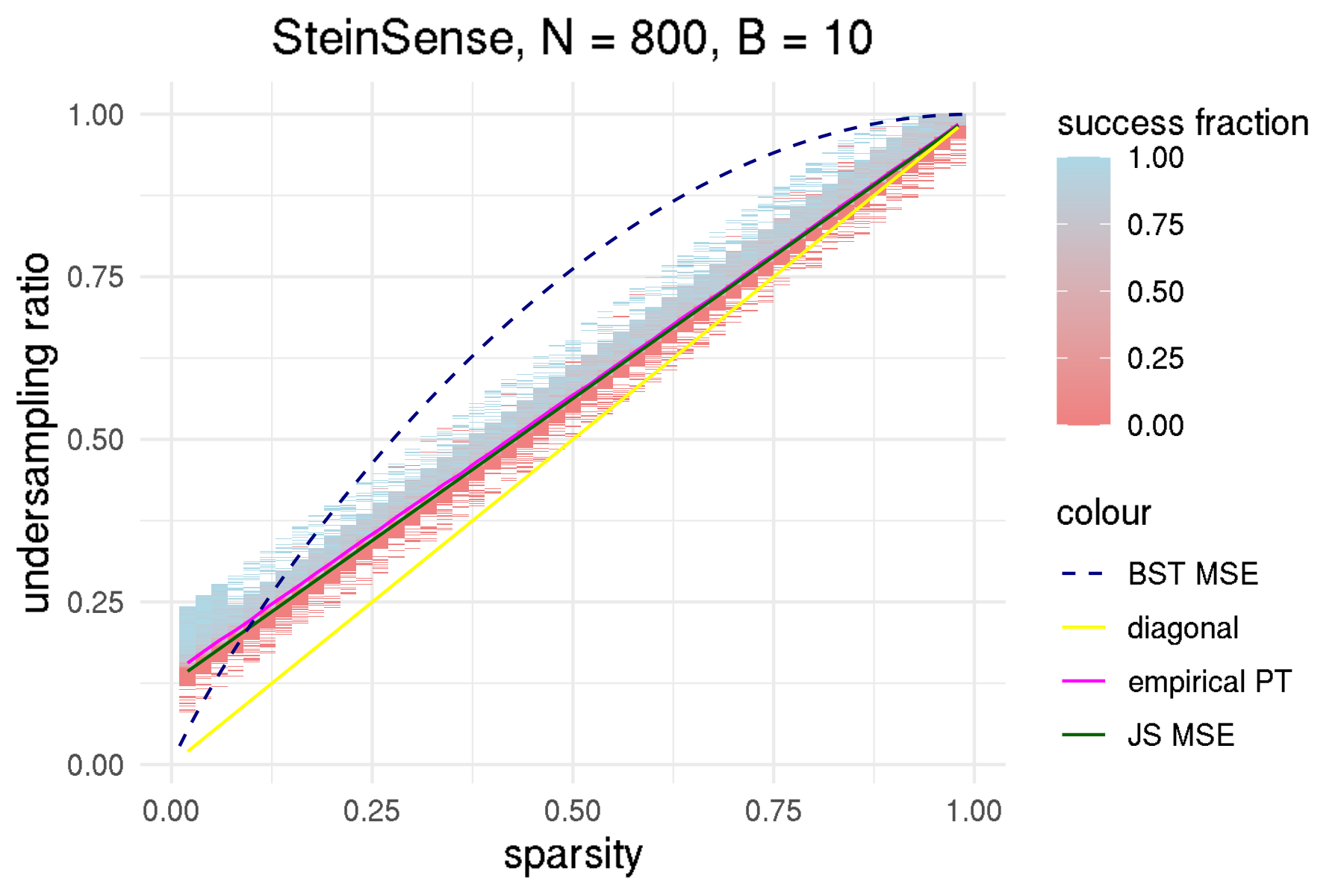}
  \end{subfigure}
  \vspace{1em}
  \begin{subfigure}[b]{0.9\textwidth}
    \includegraphics[width=\linewidth]{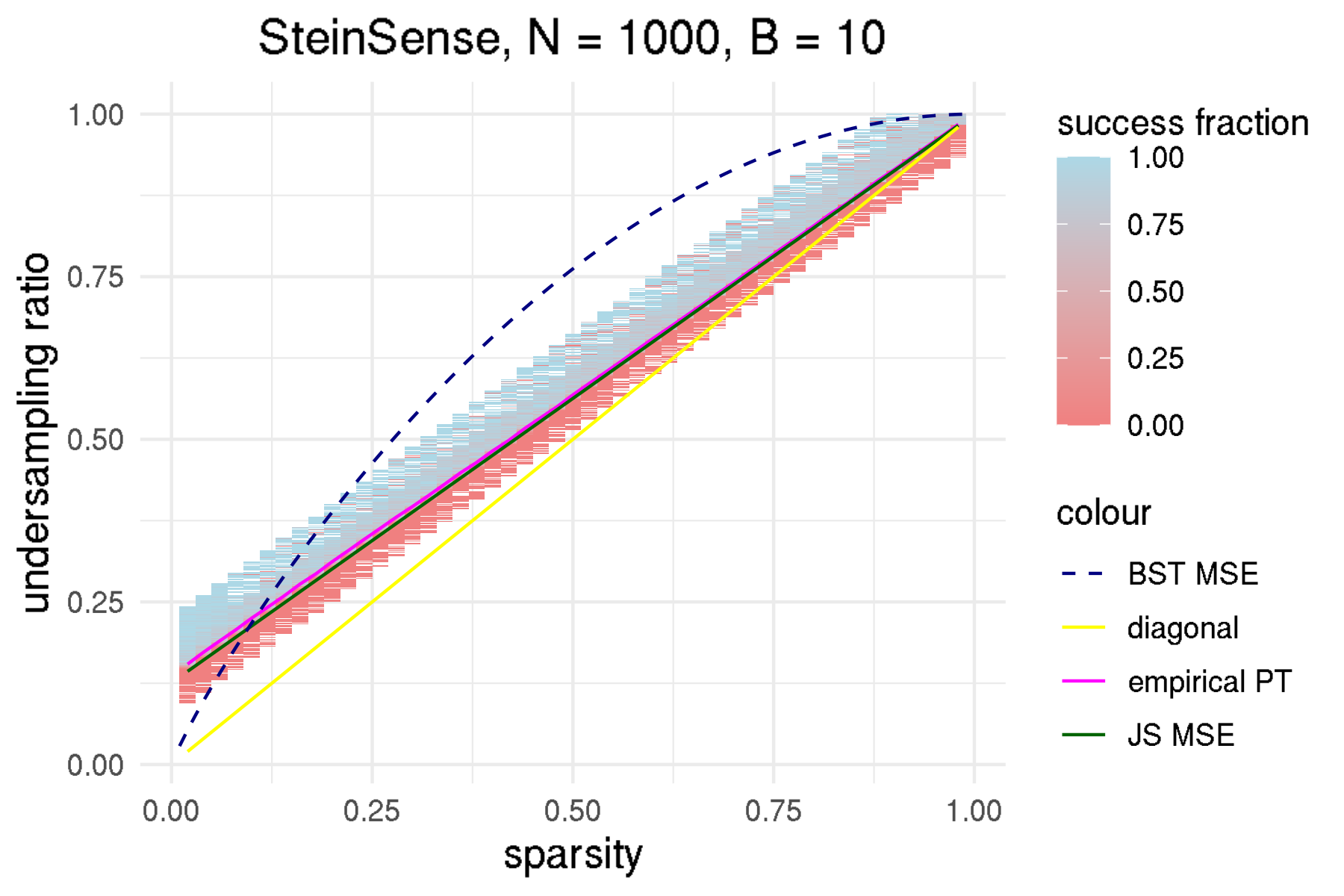}
  \end{subfigure}
  \caption{The nonzero entries are iid $N(0,1)$. The phase transitions match James Stein minimax risk to a high degree of accuracy.}
\end{figure}

% \begin{figure}[htbp]
%     \centering
%     \includegraphics[width=\linewidth]{figures/steinsense/paper_steinsense_N_100_B_10.png}
%     \caption{Caption}
%     \label{fig:steinsense_normal_N_100_B_10}
% \end{figure}

% \begin{figure}[htbp]
%     \centering
%     \includegraphics[width=\linewidth]{figures/steinsense/paper_steinsense_N_200_B_10.png}
%     \caption{Caption}
%     \label{fig:steinsense_normal_N_200_B_10}
% \end{figure}

% \begin{figure}[htbp]
%     \centering
%     \includegraphics[width=\linewidth]{figures/steinsense/paper_steinsense_N_300_B_10.png}
%     \caption{Caption}
%     \label{fig:steinsense_normal_N_300_B_10}
% \end{figure}

% \begin{figure}[htbp]
%     \centering
%     \includegraphics[width=\linewidth]{figures/steinsense/paper_steinsense_N_400_B_10.png}
%     \caption{Caption}
%     \label{fig:steinsense_normal_N_400_B_10}
% \end{figure}

% \begin{figure}[htbp]
%     \centering
%     \includegraphics[width=\linewidth]{figures/steinsense/paper_steinsense_N_800_B_10.png}
%     \caption{Caption}
%     \label{fig:steinsense_normal_N_800_B_10}
% \end{figure}

% \begin{figure}[htbp]
%     \centering
%     \includegraphics[width=\linewidth]{figures/steinsense/paper_steinsense_N_1000_B_10.png}
%     \caption{Caption}
%     \label{fig:steinsense_normal_N_1000_B_10}
% \end{figure}

\begin{figure}[htbp]
    \centering
    \includegraphics[width=\linewidth]{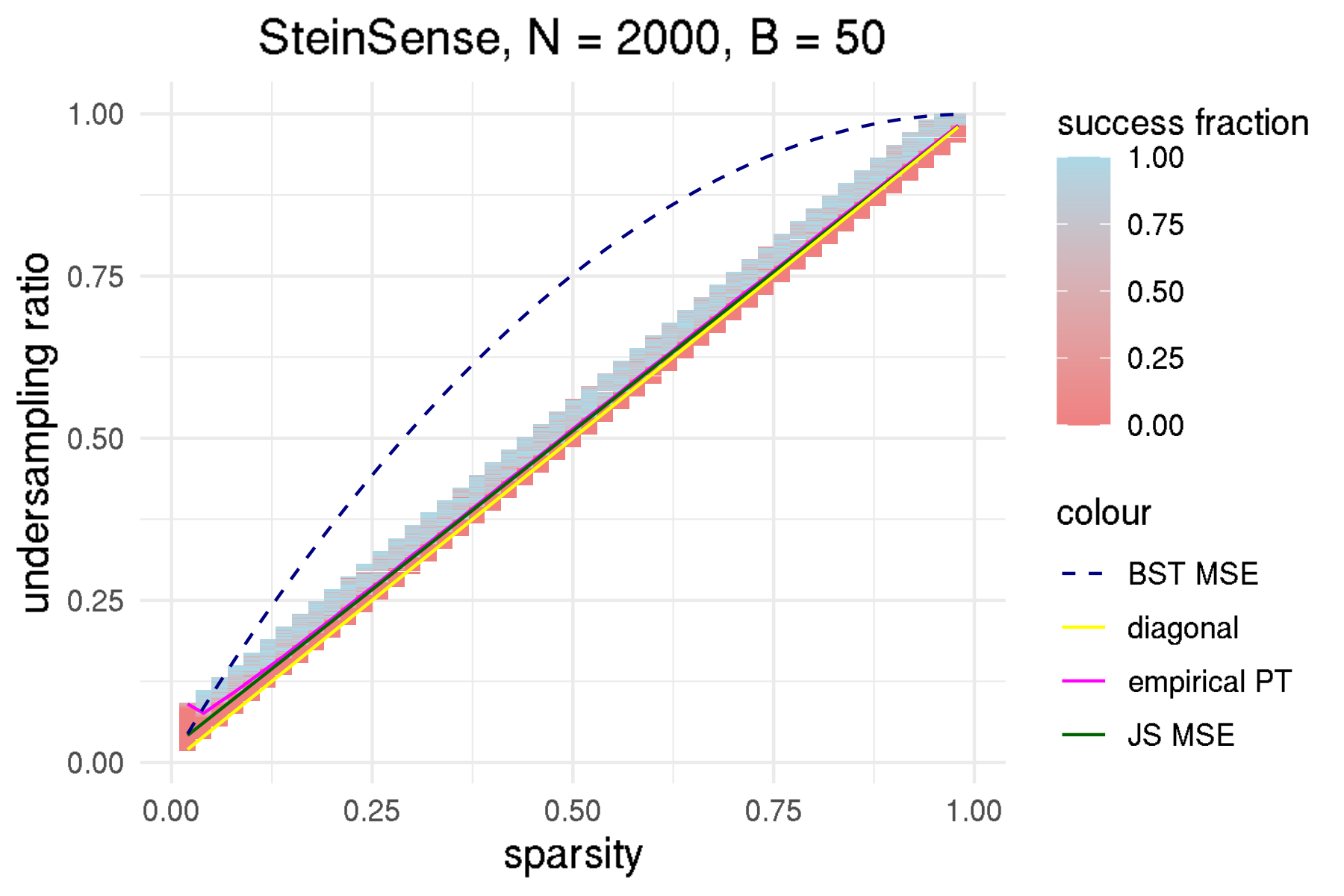}
    \caption{The nonzeros are iid $N(0,1)$. For $B=50$, we find even $N=2000$ is not large enough to subdue the deterioration at extreme sparsity. Consequently, going for $N=5000$ as displayed in Figure \ref{fig:steinsense_normal_B_20_50}, is important.}
    \label{fig:steinsense_normal_N_2000_B_50}
\end{figure}

\begin{figure}
    \centering
    \includegraphics[width=\linewidth]{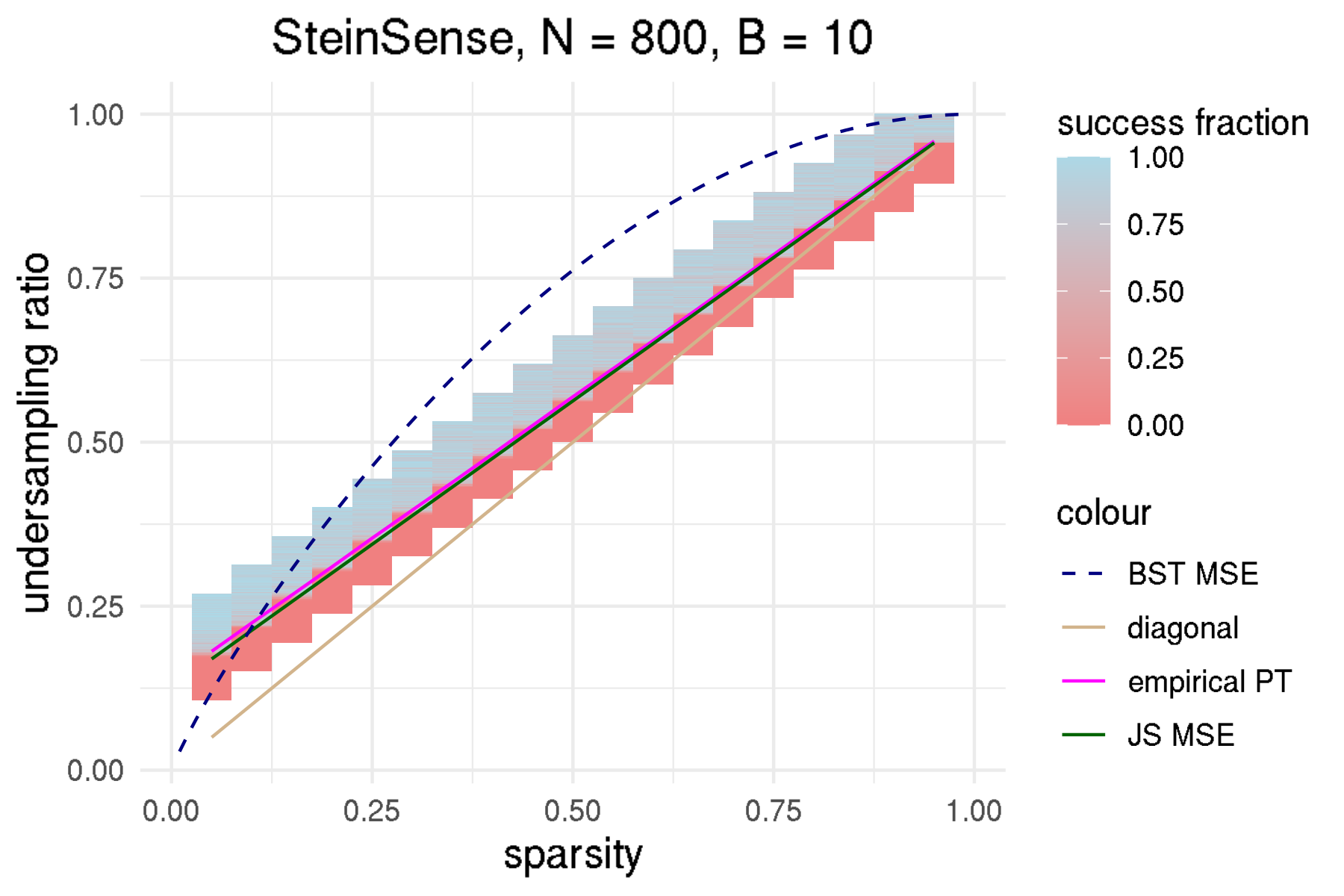}
    \caption{The nonzero entries are iid from $Poisson(2)$. This distribution is exchangeable but not symmetric and thus is outside the purview of Theorem \ref{thm:js_mse}. Still, SteinSense provides the same phase transition at the James Stein minimax risk.}
    \label{fig:steinsense_N_800_B_10_poisson_2}
\end{figure}

\begin{figure}[htbp]
  \centering
  % Row 1
  \begin{subfigure}[b]{0.6\textwidth}
    \includegraphics[width=\linewidth]{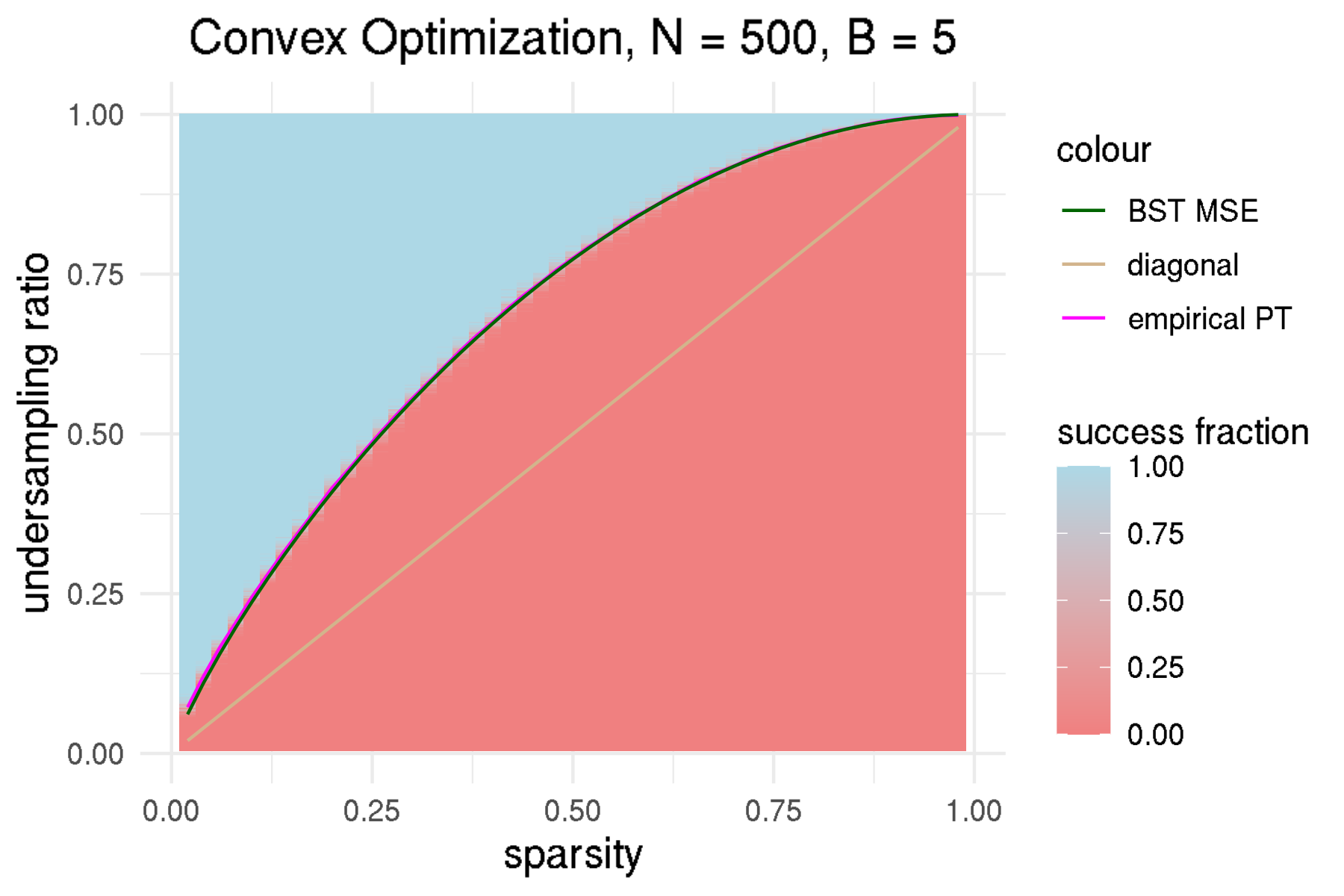}
  \end{subfigure}
  \vspace{1em}
  \begin{subfigure}[b]{0.6\textwidth}
    \includegraphics[width=\linewidth]{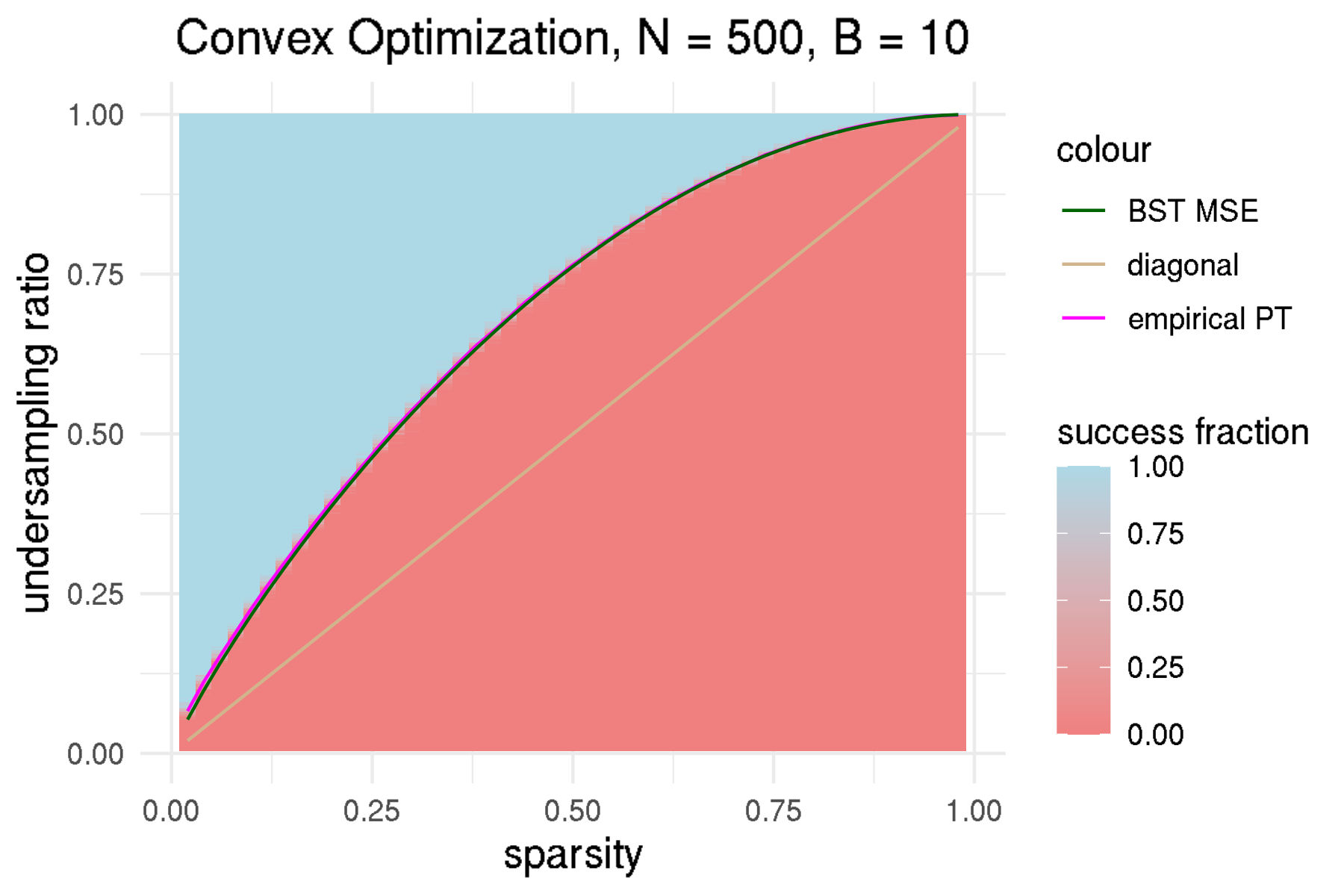}
  \end{subfigure}
  \vspace{1em}
    \begin{subfigure}[b]{0.6\textwidth}
    \includegraphics[width=\linewidth]{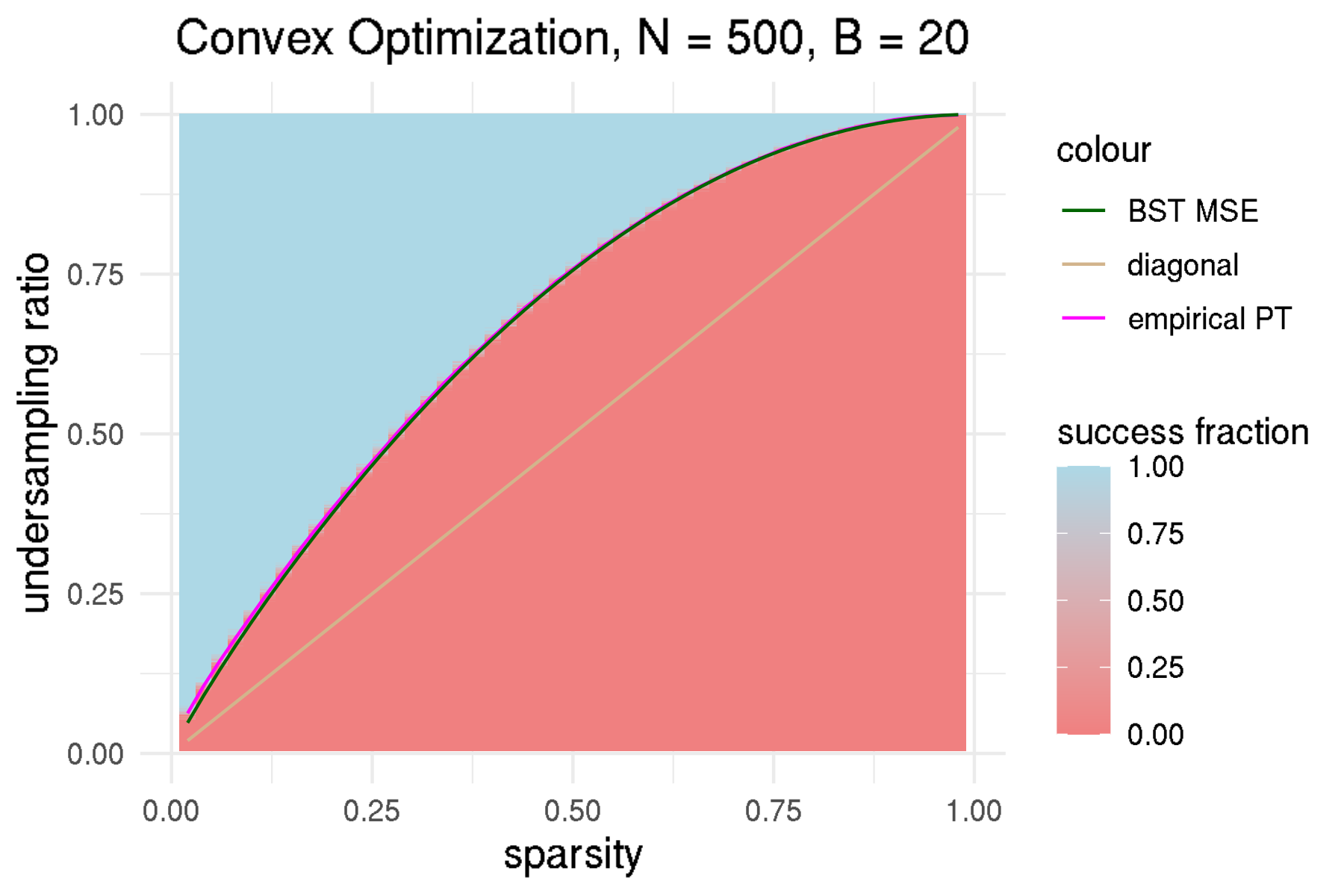}
  \end{subfigure}
  \caption{The nonzero entries in the vectors are chosen to be $\pm1/2$ with probability $1/2$ each. We get the BlockSoft minimax risk as the location of the phase transition.}
\end{figure}

% \begin{figure}
%     \centering
%     \includegraphics[width=\linewidth]{figures/convex/paper_cvx_N_500_B_5_binary_centered_p_0.5.png}
%     \caption{The nonzero entries in the vectors are chosen to be $\pm1/2$ with probability $1/2$ each.}
%     \label{fig:paper_cvx_N_500_B_5_binary_centered_p_0.5}
% \end{figure}

% \begin{figure}[htbp]
%     \centering
%     \includegraphics[width=\linewidth]{figures/convex/paper_cvx_N_500_B_10_binary_centered_p_0.5.png}
%     \caption{The nonzero entries in the vectors are chosen to be $\pm1/2$ with probability $1/2$ each.}
%     \label{fig:paper_cvx_N_500_B_10_binary_centered_p_0.5}
% \end{figure}

% \begin{figure}[htbp]
%     \centering
%     \includegraphics[width=\linewidth]{figures/convex/paper_cvx_N_500_B_20_binary_centered_p_0.5.png}
%     \caption{The nonzero entries in the vectors are chosen to be $\pm1/2$ with probability $1/2$ each.}
%     \label{fig:paper_cvx_N_500_B_20_binary_centered_p_0.5}
% \end{figure}

\begin{figure}[htbp]
    \centering
    \includegraphics[width=\linewidth]{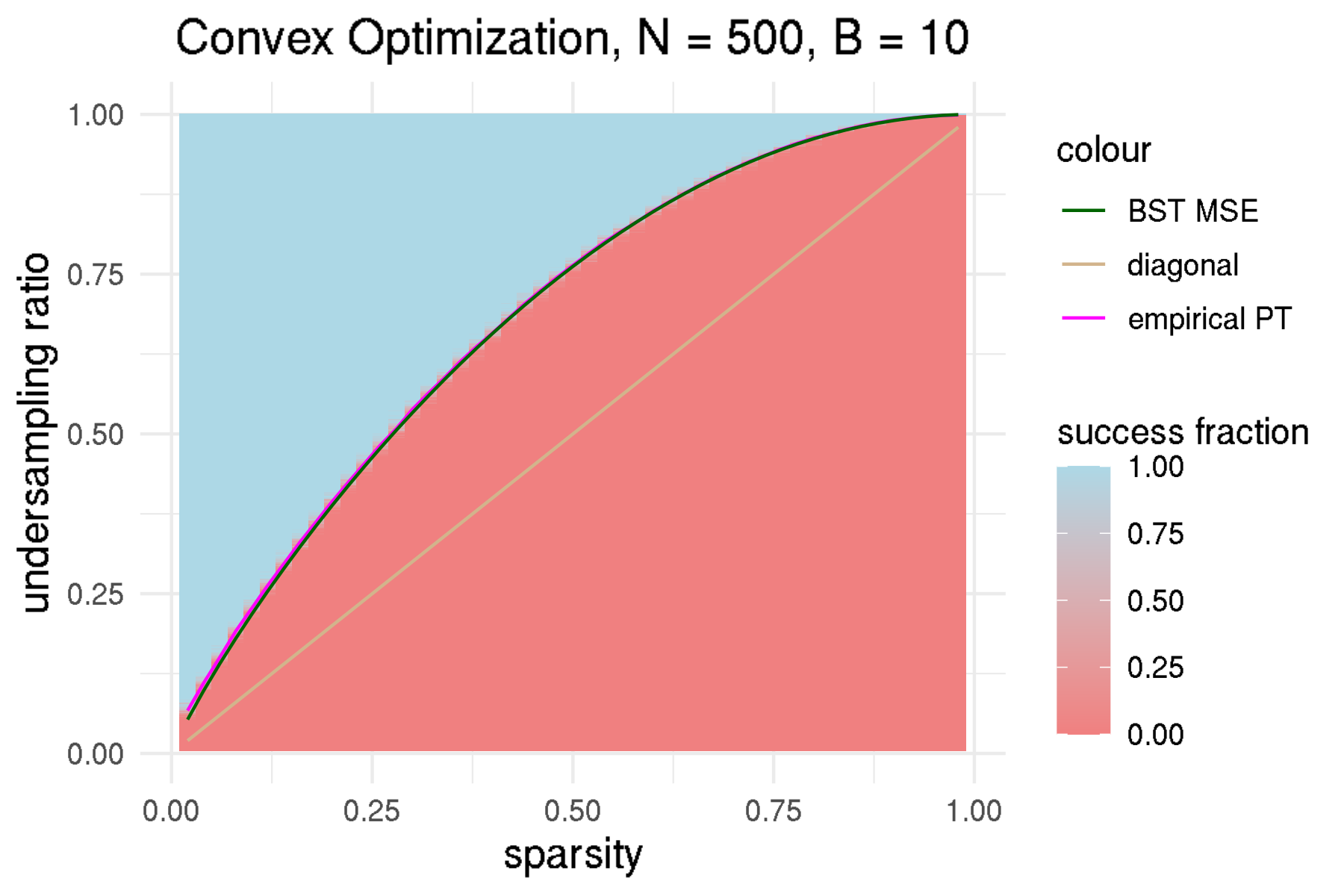}
    \caption{The nonzero entries in the vectors are chosen to be $0$ with probability $1/2$ and $\pm1/2$ with probability $1/4$ each.}
    \label{fig:paper_cvx_N_500_B_5_binary_random_sign}
\end{figure}

\begin{figure}[htbp]
  \centering
  % Row 1
  \begin{subfigure}[b]{0.48\textwidth}
    \includegraphics[width=\linewidth]{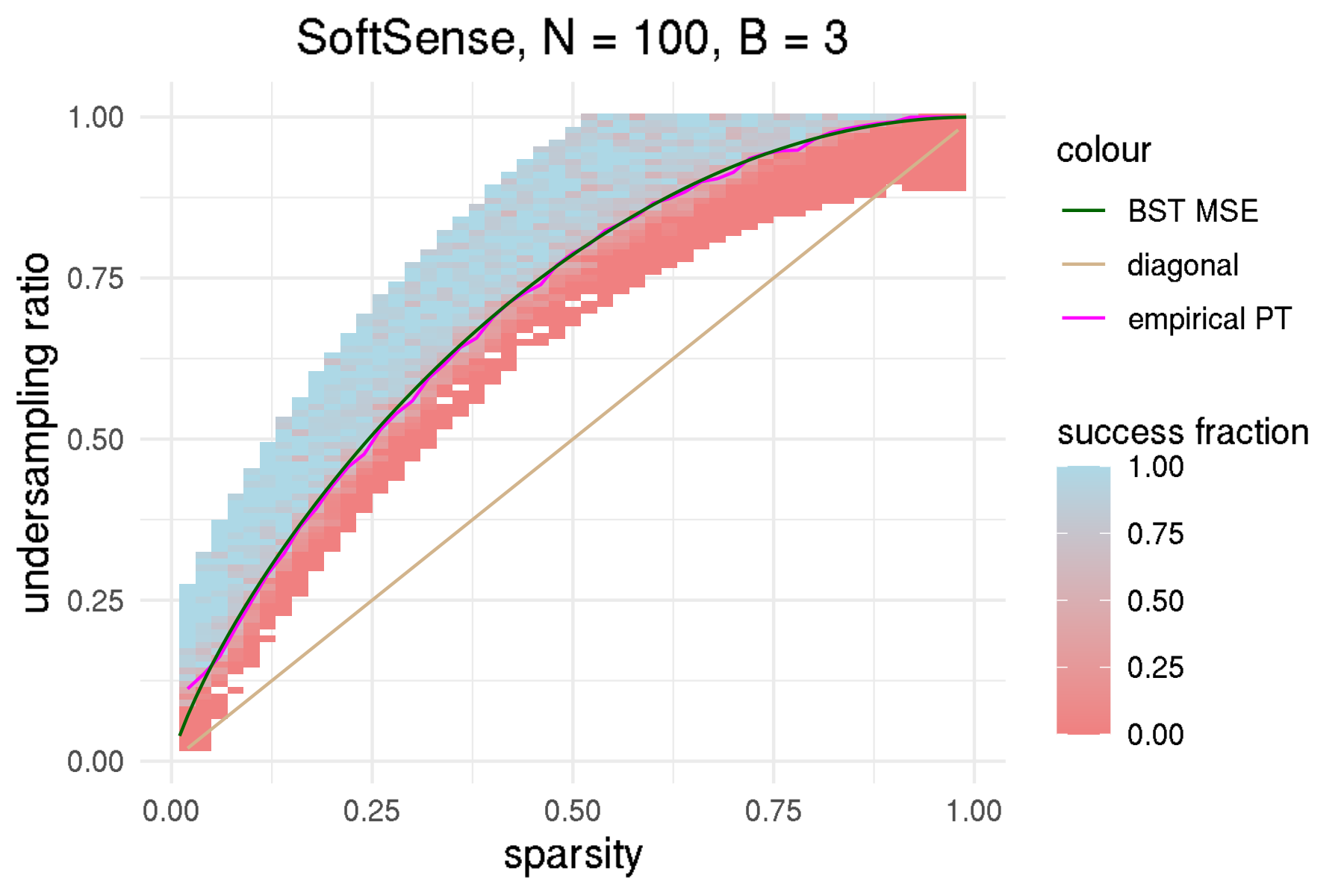}
  \end{subfigure}
  \hfill
  \begin{subfigure}[b]{0.48\textwidth}
    \includegraphics[width=\linewidth]{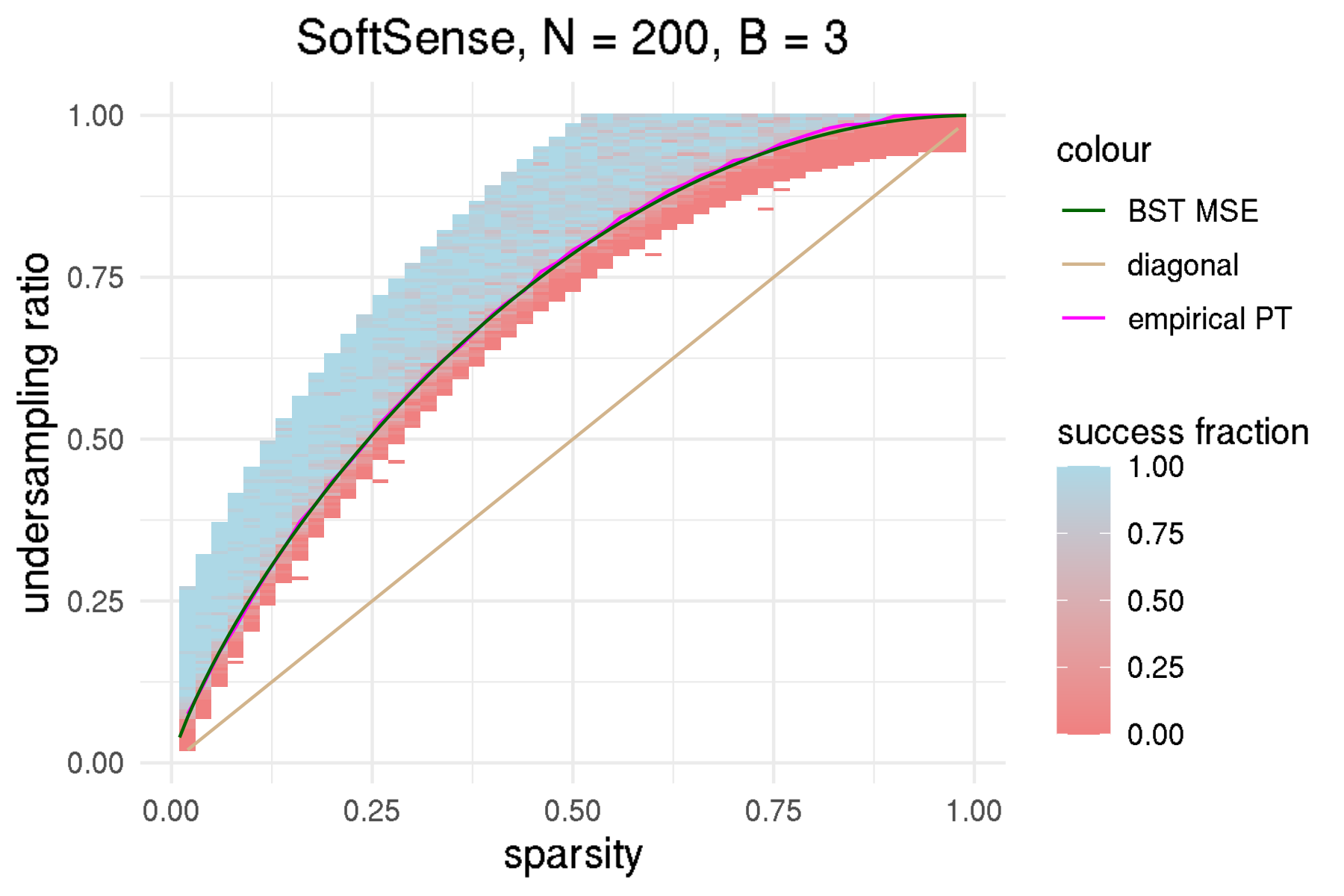}
  \end{subfigure}
  \vspace{1em}
    \begin{subfigure}[b]{0.48\textwidth}
    \includegraphics[width=\linewidth]{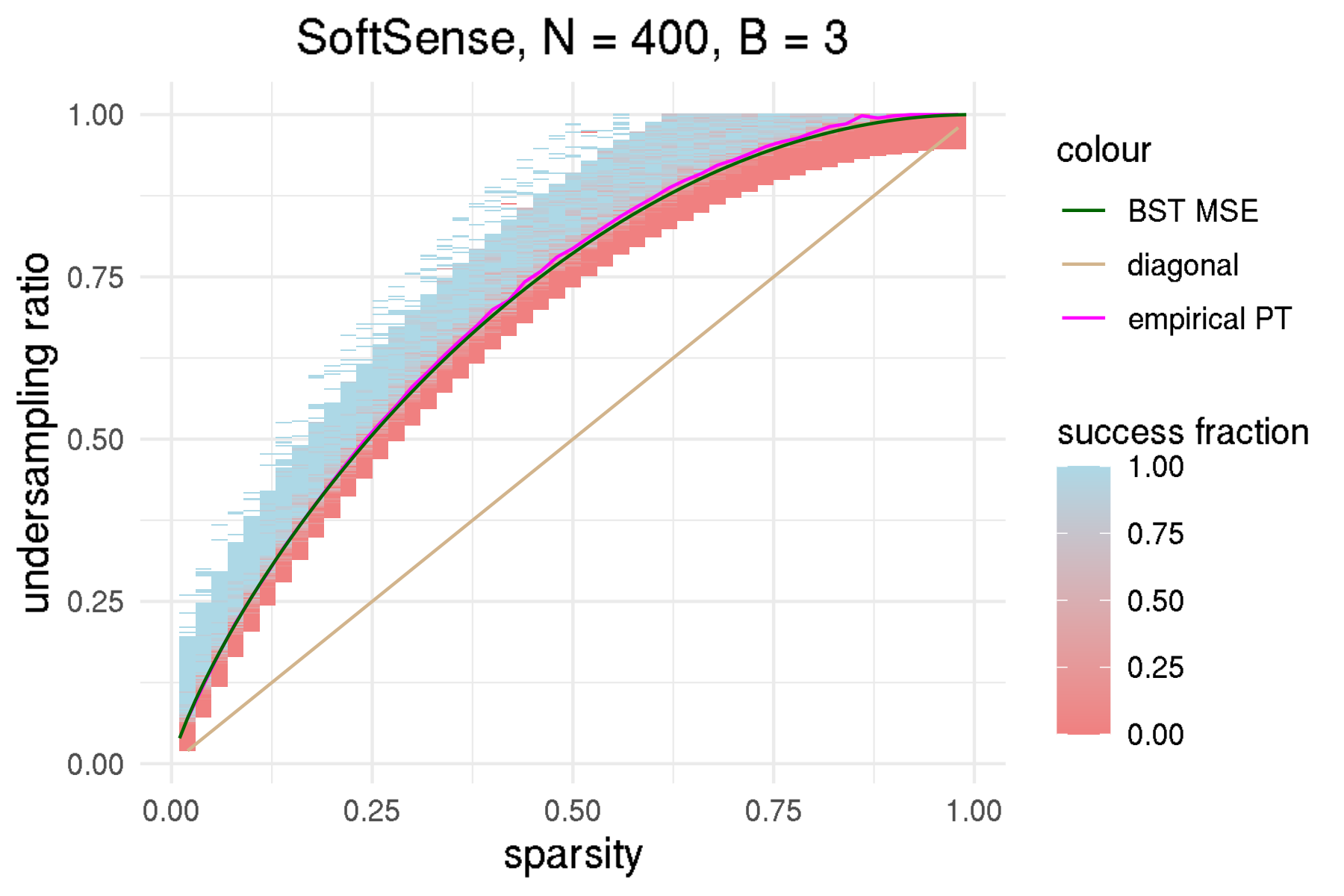}
  \end{subfigure}
  \hfill
    \begin{subfigure}[b]{0.48\textwidth}
    \includegraphics[width=\linewidth]{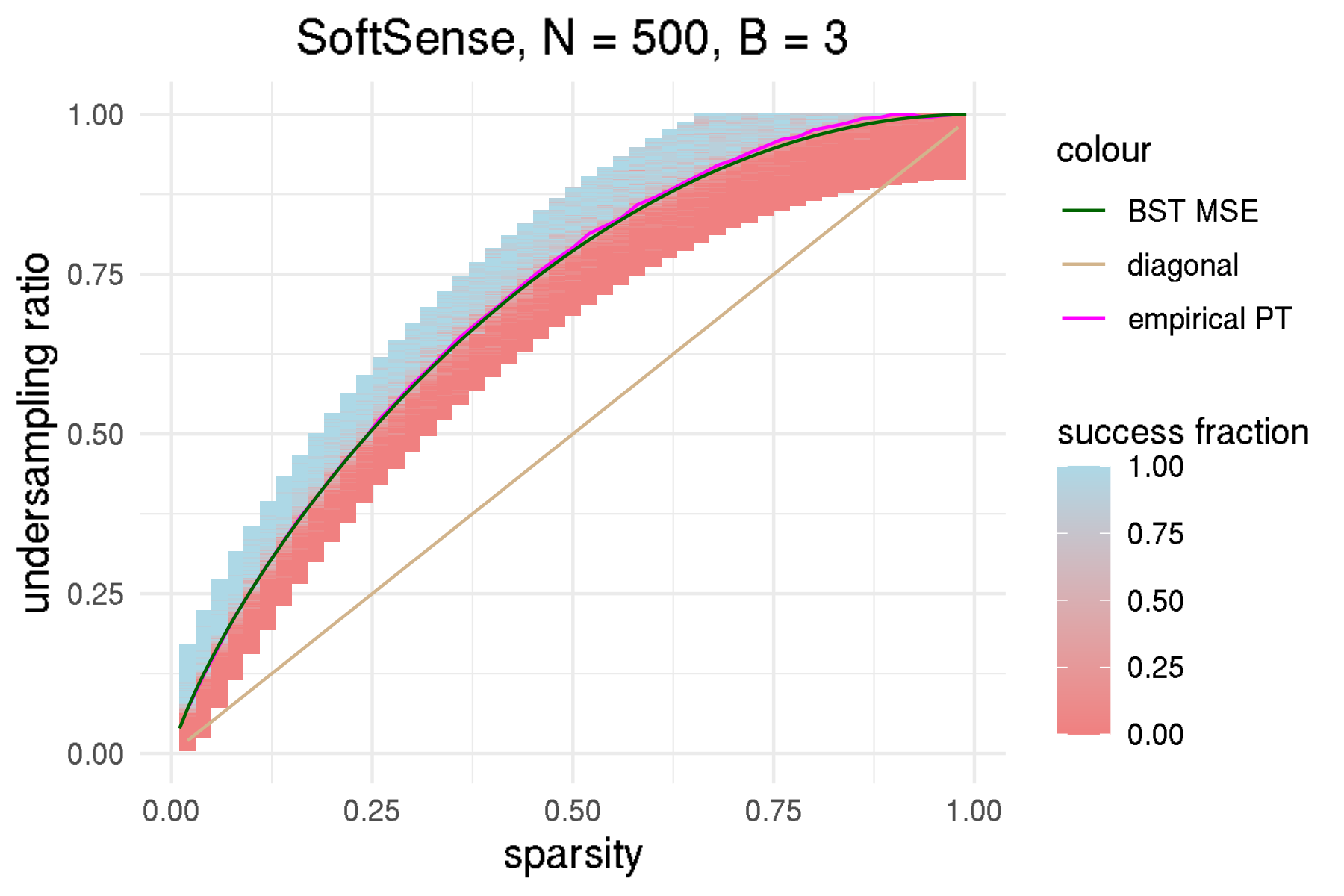}
  \end{subfigure}
  \caption{The nonzero entries are iid $N(0,1)$. Unlike SteinSense, SoftSense is pretty robust to small $B$. Already at $N=400$ there is a very good match between the empirical phase transition and BlockSoft minimax risk curve.}
\end{figure}

% \begin{figure}[htbp]
%     \centering
%     \includegraphics[width=\linewidth]{figures/softsense/paper_softsense_jit_N_100_B_3.png}
%     \caption{Nonzeros are iid $N(0,1)$.}
%     \label{fig:softsense_jit_normal_N_100_B_3}
% \end{figure}

% \vspace{1em}

% \begin{figure}[htbp]
%     \centering
%     \includegraphics[width=\linewidth]{figures/softsense/paper_softsense_jit_N_200_B_3.png}
%     \caption{Caption}
%     \label{fig:softsense_jit_normal_N_200_B_3}
% \end{figure}

% \vspace{1em}

% \begin{figure}[htbp]
%     \centering
%     \includegraphics[width=\linewidth]{figures/softsense/paper_softsense_jit_N_400_B_3.png}
%     \caption{Caption}
%     \label{fig:softsense_jit_normal_N_400_B_3}
% \end{figure}

% \vspace{1em}

% Now we get the next one

% \begin{figure}[htbp]
%     \centering
%     \includegraphics[width=\linewidth]{figures/softsense/paper_softsense_jit_N_500_B_3.png}
%     \caption{Caption}
%     \label{fig:softsense_jit_normal_N_500_B_3}
% \end{figure}

\begin{figure}[htbp]
  \centering
  % Row 1
  \begin{subfigure}[b]{0.9\textwidth}
    \includegraphics[width=\linewidth]{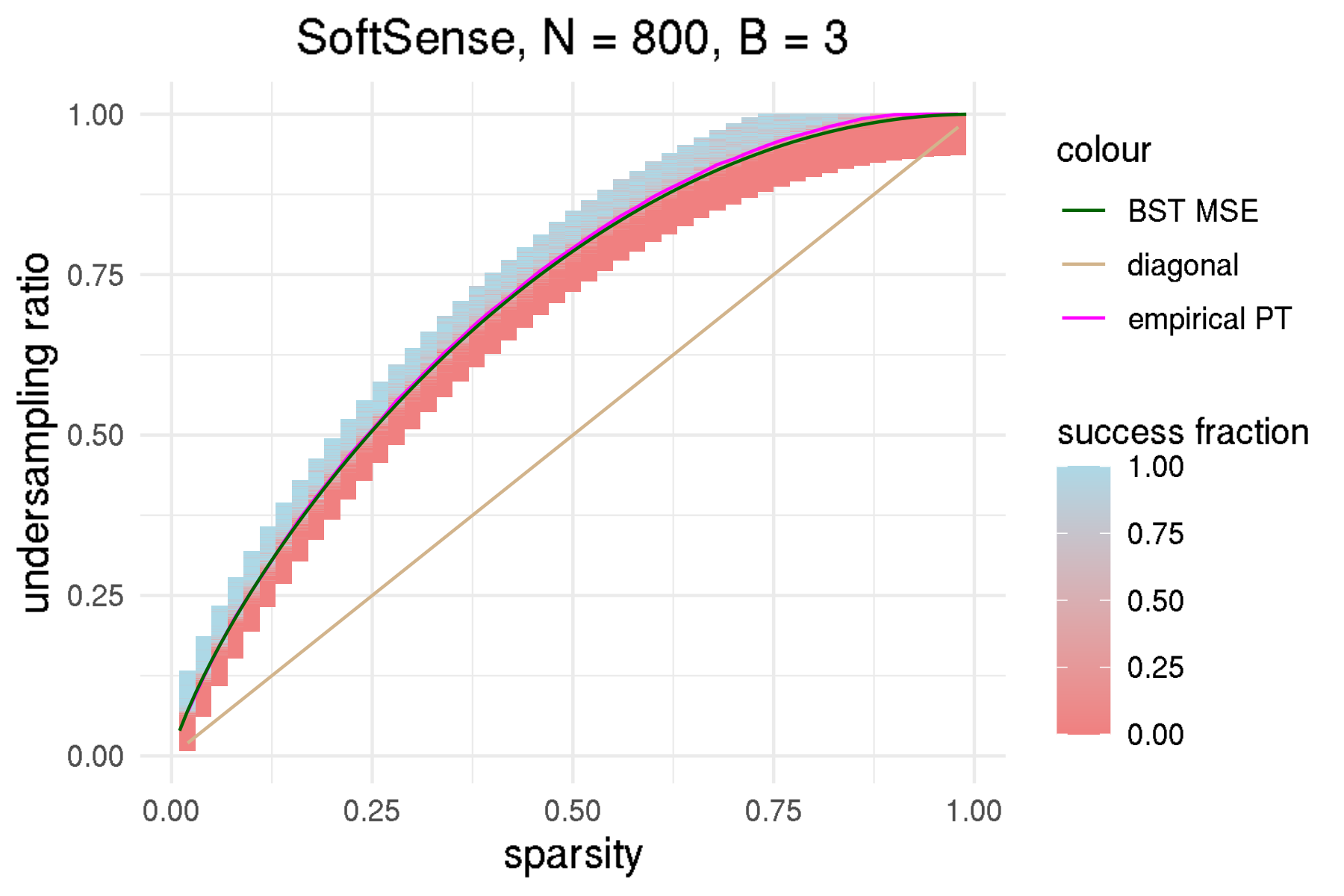}
  \end{subfigure}
  \vspace{1em}
  \begin{subfigure}[b]{0.9\textwidth}
    \includegraphics[width=\linewidth]{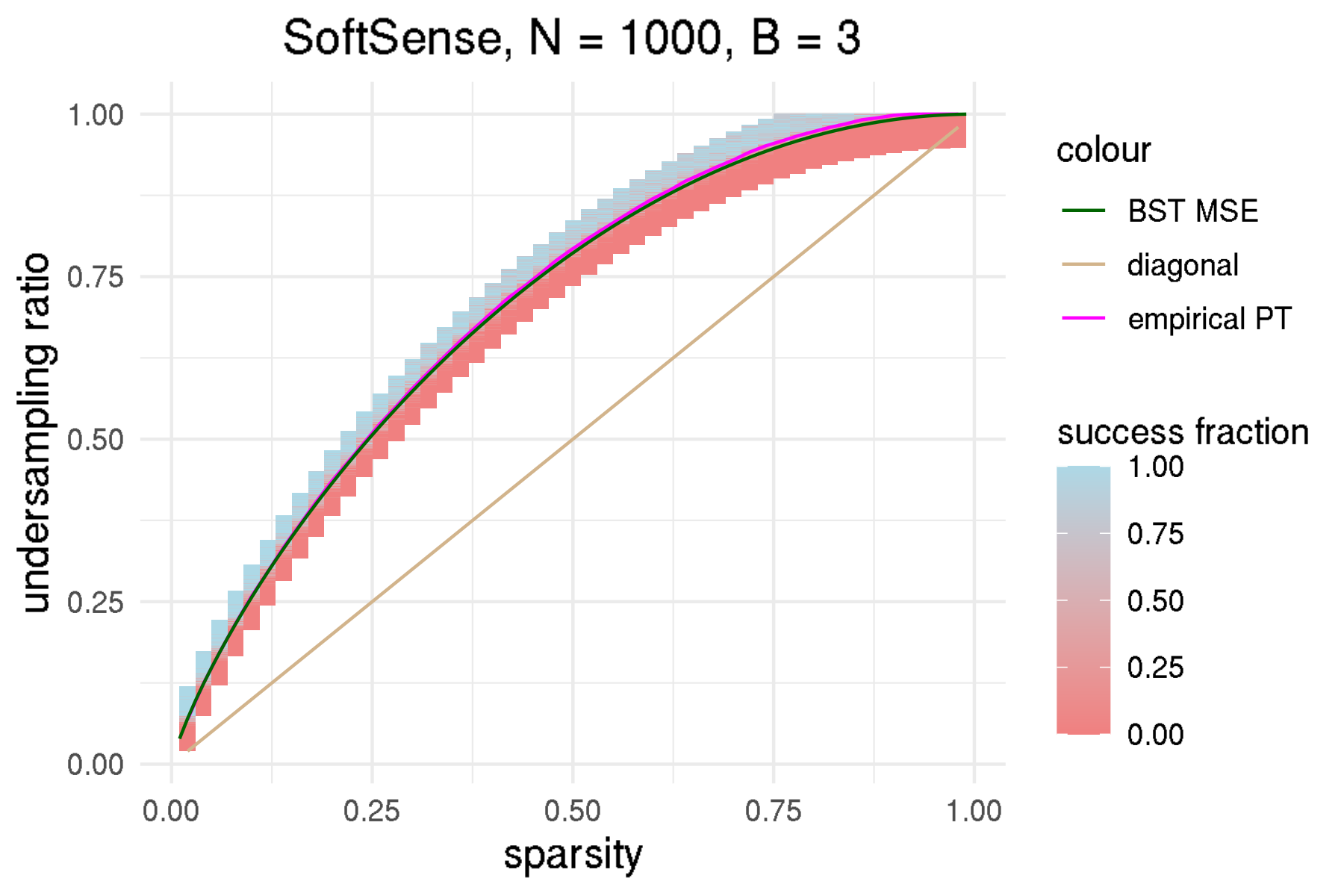}
  \end{subfigure}
  \caption{The nonzero entries are iid $N(0,1)$. There is a very good match between the empirical phase transition and BlockSoft minimax risk curve.}
\end{figure}

% \begin{figure}[htbp]
%     \centering
%     \includegraphics[width=\linewidth]{figures/softsense/paper_softsense_jit_N_800_B_3.png}
%     \caption{Caption}
%     \label{fig:softsense_jit_normal_N_800_B_3}
% \end{figure}

% \vspace{1em}

% \begin{figure}[htbp]
%     \centering
%     \includegraphics[width=\linewidth]{figures/softsense/paper_softsense_jit_N_1000_B_3.png}
%     \caption{Caption}
%     \label{fig:softsense_jit_normal_N_1000_B_3}
% \end{figure}

\begin{figure}[htbp]
  \centering
  % Row 1
  \begin{subfigure}[b]{0.6\textwidth}
    \includegraphics[width=\linewidth]{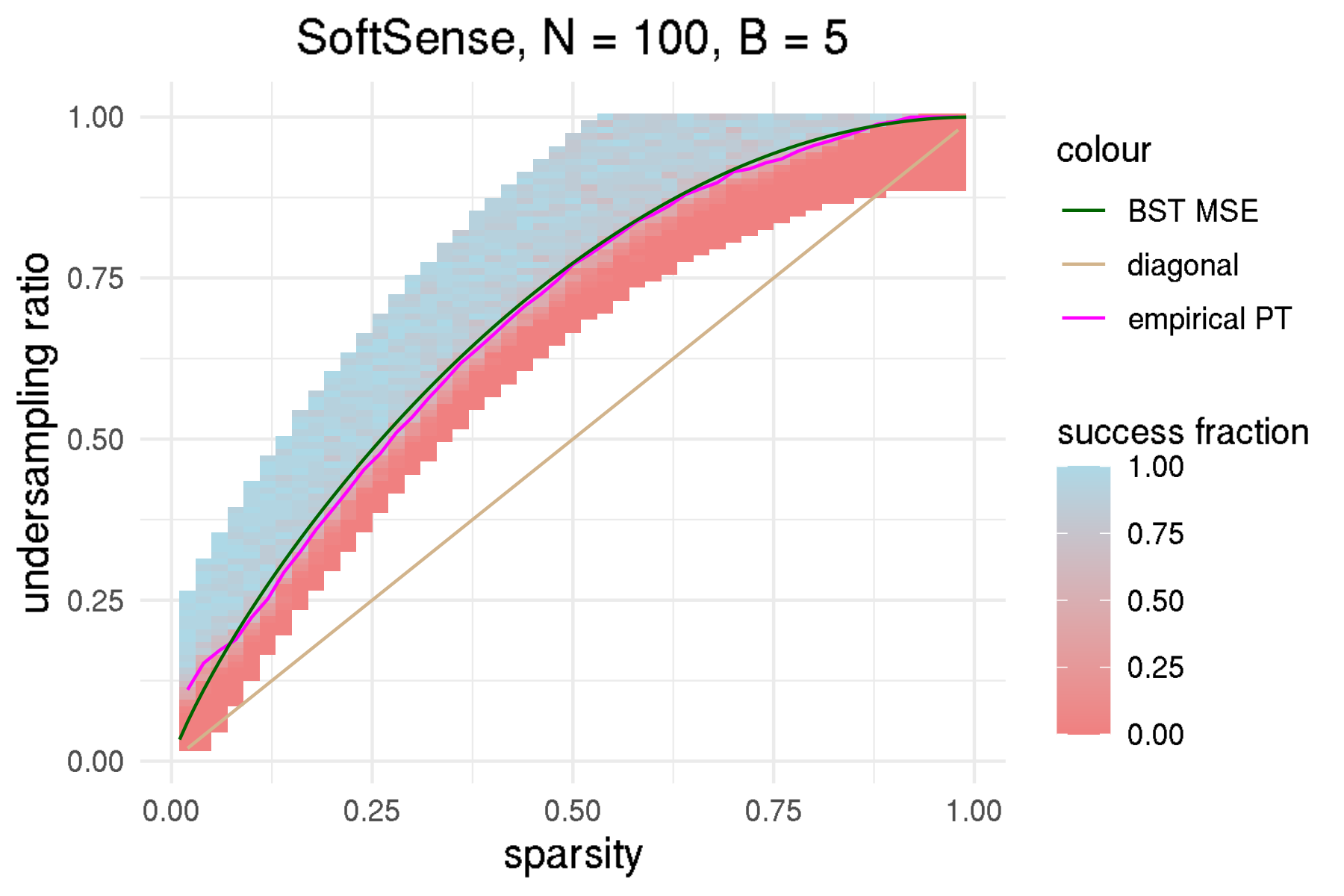}
  \end{subfigure}
  \vspace{1em}
  \begin{subfigure}[b]{0.6\textwidth}
    \includegraphics[width=\linewidth]{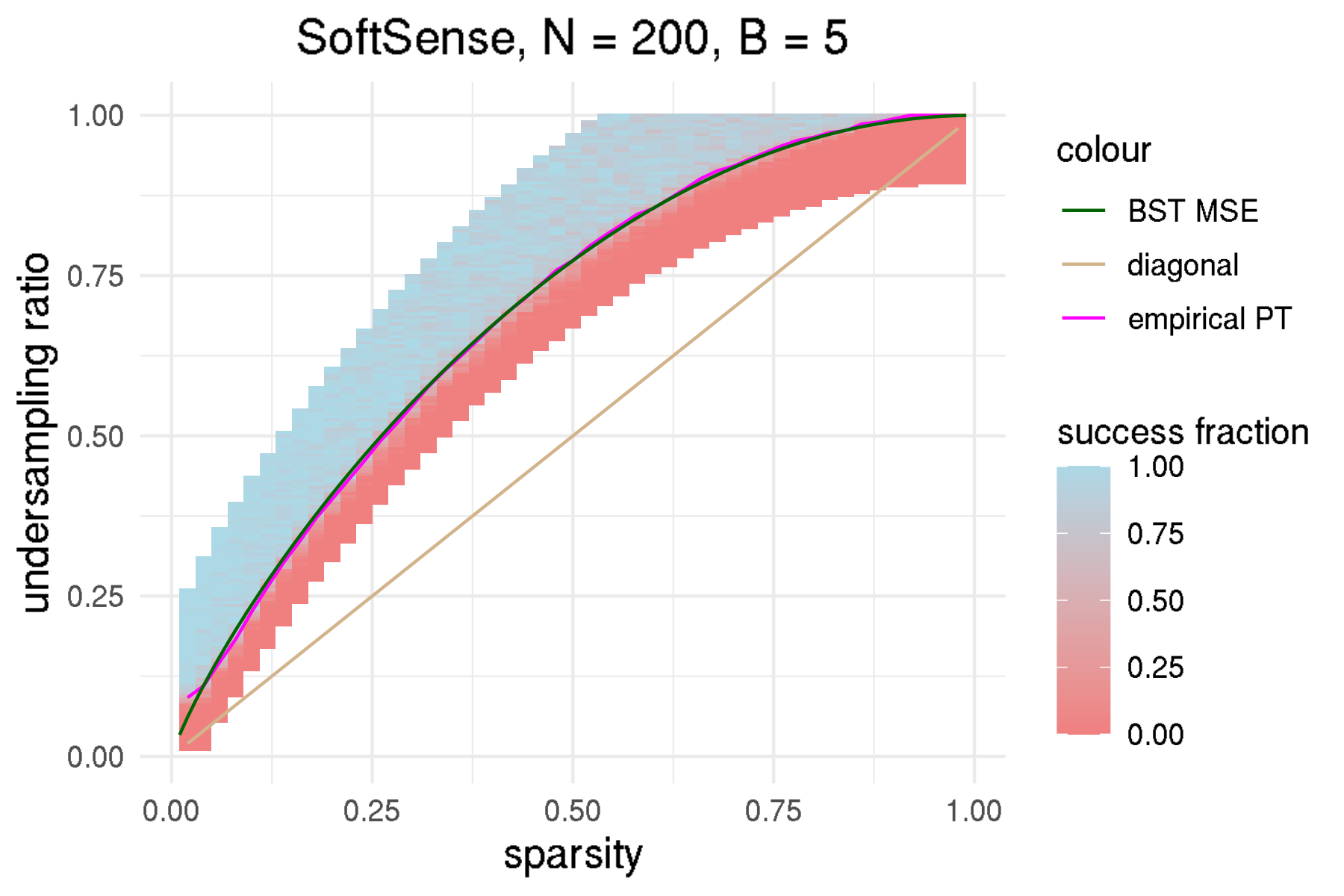}
  \end{subfigure}
  \vspace{1em}
    \begin{subfigure}[b]{0.6\textwidth}
    \includegraphics[width=\linewidth]{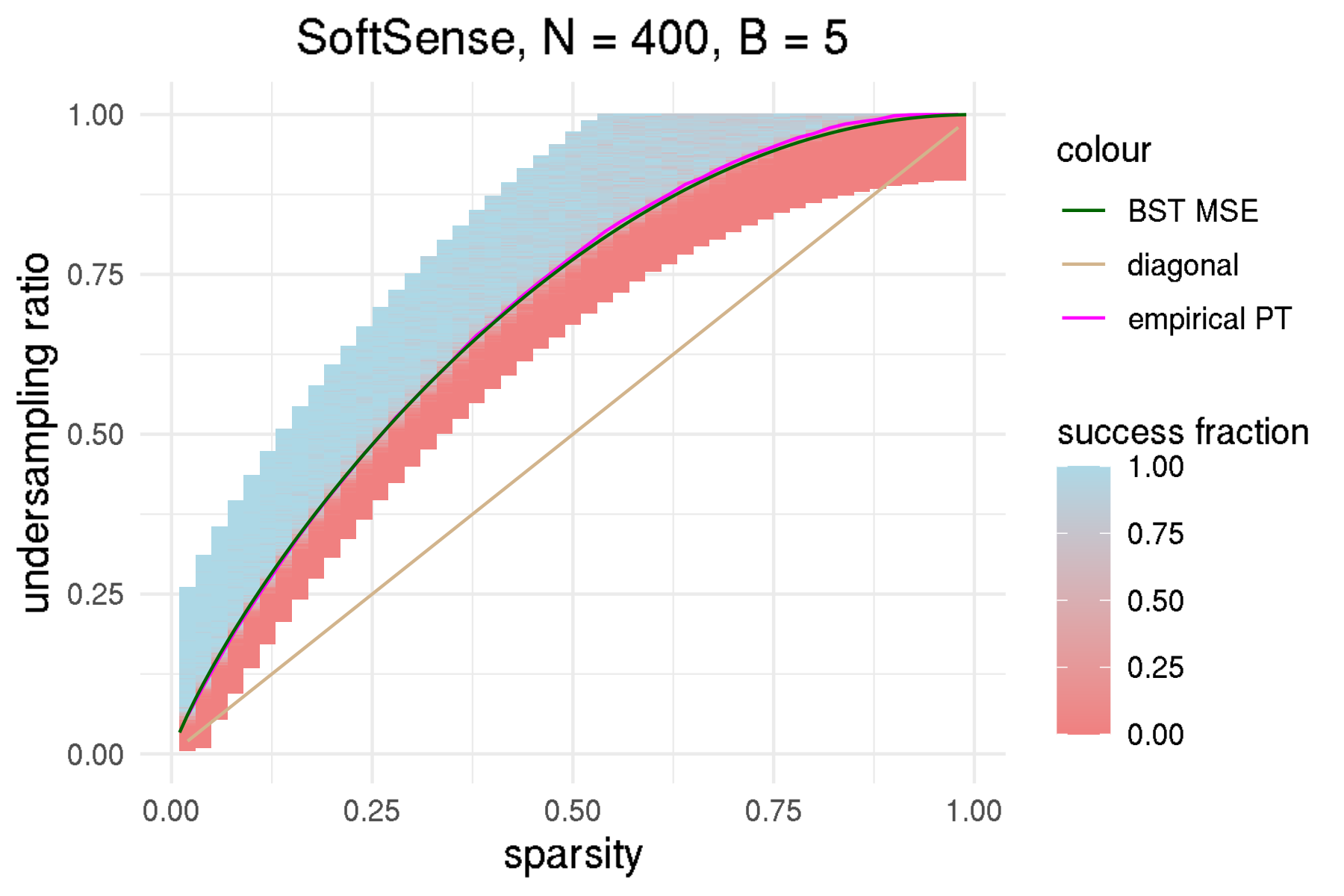}
  \end{subfigure}
  \caption{The nonzero entries are iid $N(0,1)$. Already at $N=400$ there is a very good match between the empirical phase transition and BlockSoft minimax risk curve. For smaller $N$ we see extreme sparsity suffering a bit.}
\end{figure}

% \begin{figure}[htbp]
%     \centering
%     \includegraphics[width=\linewidth]{figures/softsense/paper_softsense_jit_N_100_B_5.png}
%     \caption{Caption}
%     \label{fig:softsense_jit_normal_N_100_B_5}
% \end{figure}

% \vspace{1em}

% \begin{figure}[htbp]
%     \centering
%     \includegraphics[width=\linewidth]{figures/softsense/paper_softsense_jit_N_200_B_5.png}
%     \caption{Caption}
%     \label{fig:softsense_jit_normal_N_200_B_5}
% \end{figure}

% \vspace{1em}

% \begin{figure}[htbp]
%     \centering
%     \includegraphics[width=\linewidth]{figures/softsense/paper_softsense_jit_N_400_B_5.png}
%     \caption{Caption}
%     \label{fig:softsense_jit_normal_N_400_B_5}
% \end{figure}

% \vspace{1em}

\begin{figure}[htbp]
  \centering
  % Row 1
  \begin{subfigure}[b]{0.9\textwidth}
    \includegraphics[width=\linewidth]{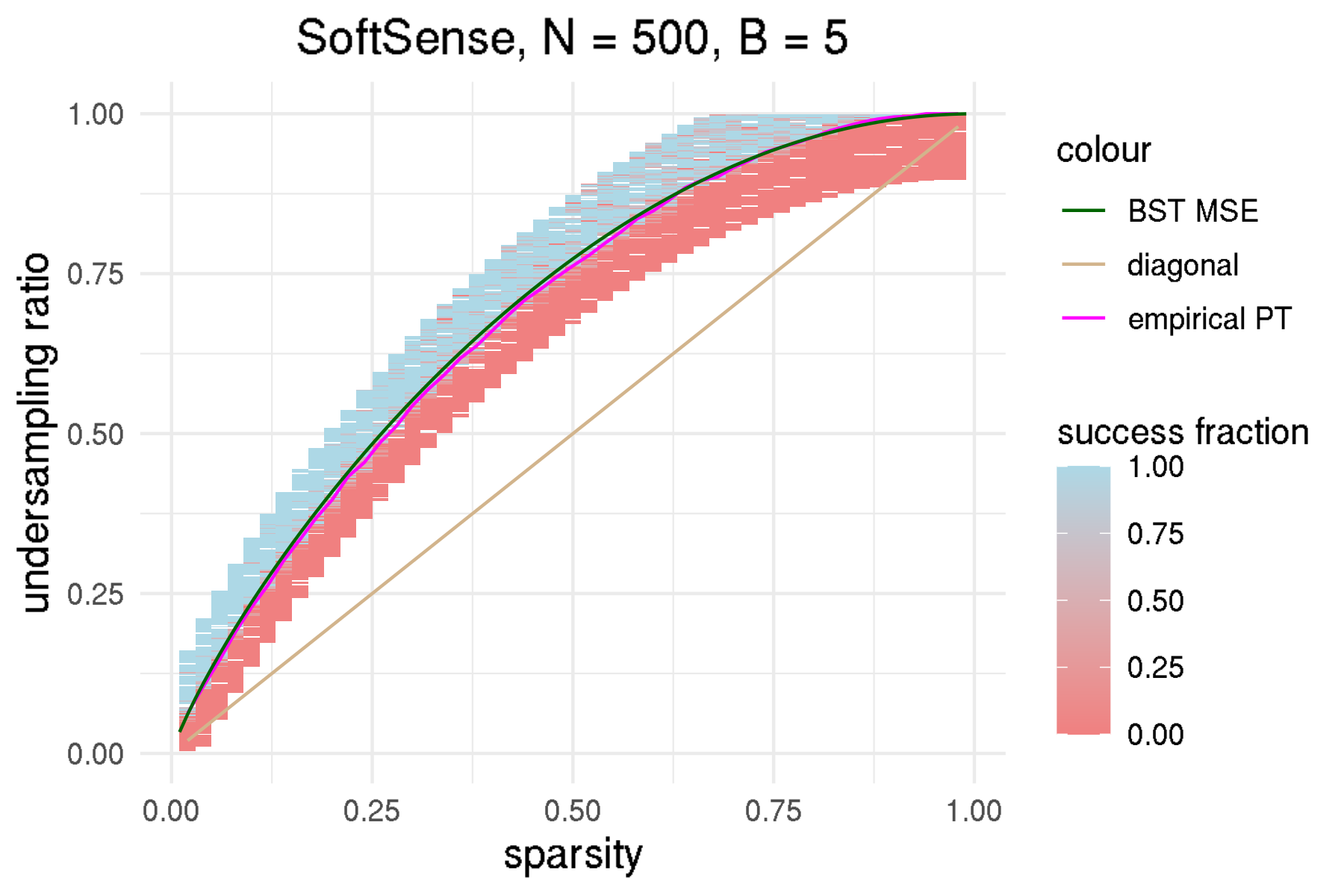}
  \end{subfigure}
  \vspace{1em}
  \begin{subfigure}[b]{0.9\textwidth}
    \includegraphics[width=\linewidth]{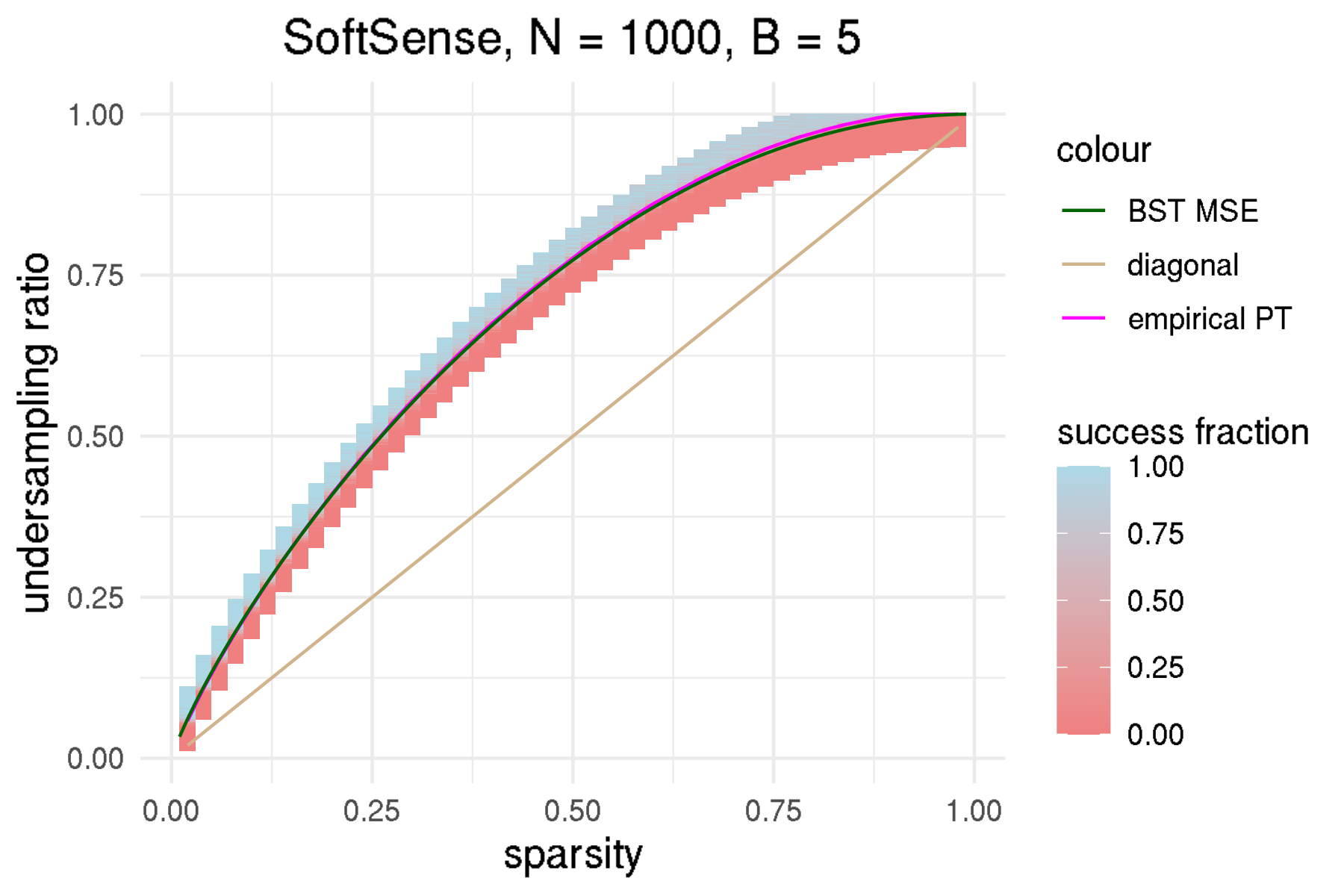}
  \end{subfigure}
  \caption{The nonzero entries are iid $N(0,1)$. The empirical phase transition matches the BlockSoft minimax risk almost perfectly.}
\end{figure}

% \begin{figure}[htbp]
%     \centering
%     \includegraphics[width=\linewidth]{figures/softsense/paper_softsense_jit_N_500_B_5.png}
%     \caption{Caption}
%     \label{fig:softsense_jit_normal_N_500_B_5}
% \end{figure}

% \begin{figure}[htbp]
%     \centering
%     \includegraphics[width=\linewidth]{figures/softsense/paper_softsense_jit_N_1000_B_5.png}
%     \caption{Caption}
%     \label{fig:softsense_jit_normal_N_1000_B_5}
% \end{figure}

\begin{figure}[htbp]
  \centering
  % Row 1
  \begin{subfigure}[b]{0.9\textwidth}
    \includegraphics[width=\linewidth]{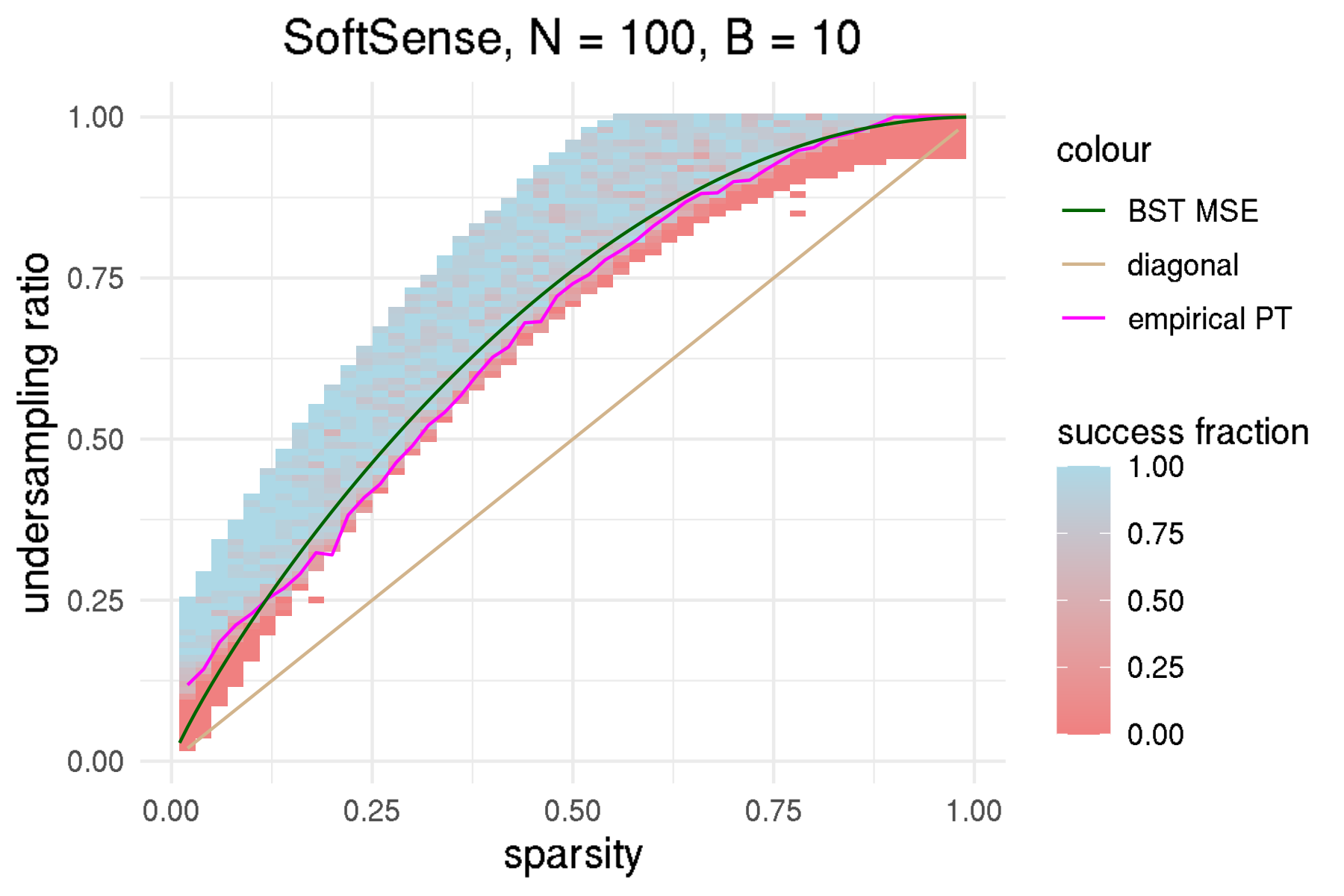}
  \end{subfigure}
  \vspace{1em}
  \begin{subfigure}[b]{0.9\textwidth}
    \includegraphics[width=\linewidth]{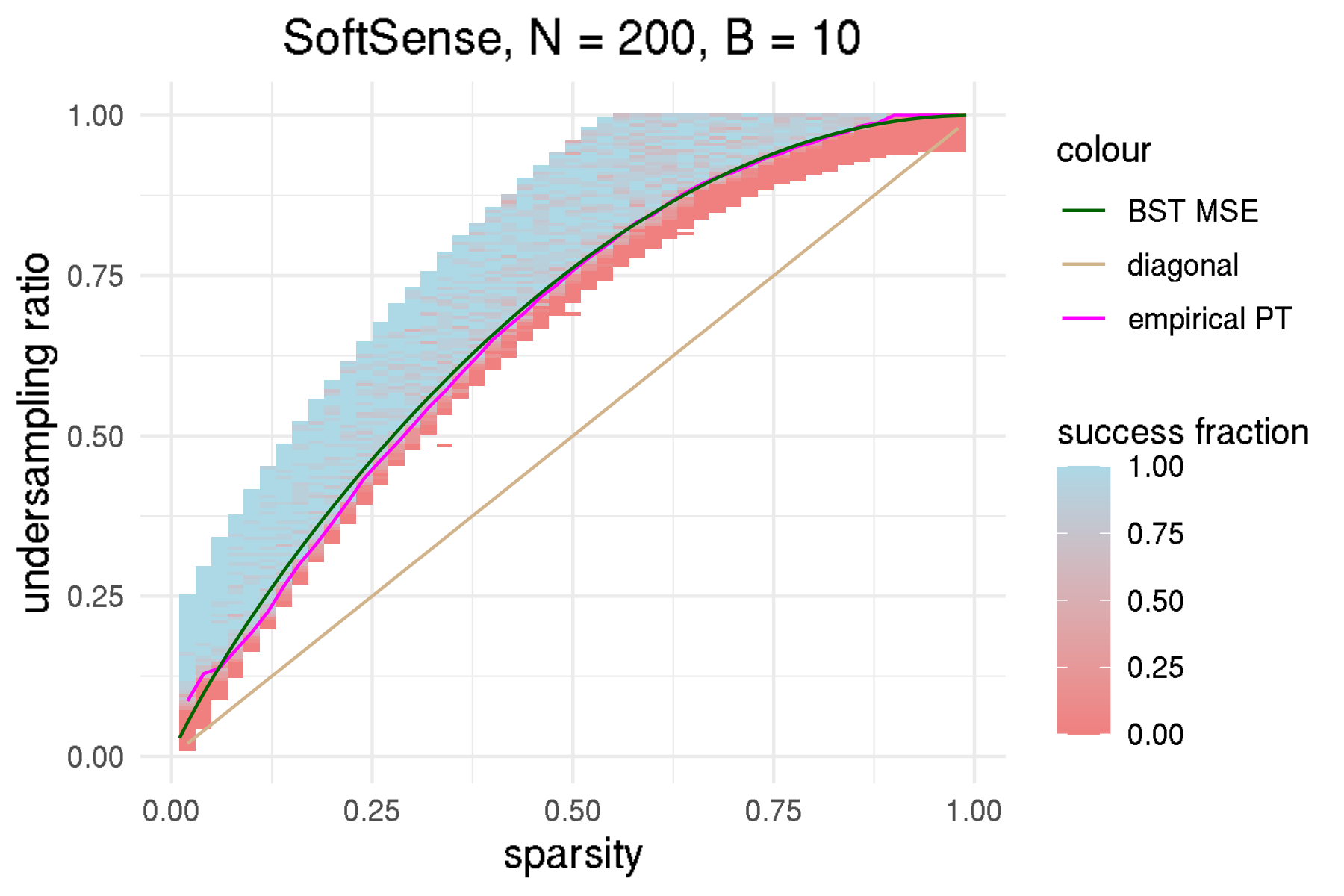}
  \end{subfigure}
  \caption{The nonzero entries are iid $N(0,1)$. While the empirical phase transition qualitatively agrees with the BlockSoft minimax risk, some weirdness prevails due to $B$ being large compared to $N$.}
\end{figure}

\begin{figure}[htbp]
  \centering
  % Row 1
  \begin{subfigure}[b]{0.9\textwidth}
    \includegraphics[width=\linewidth]{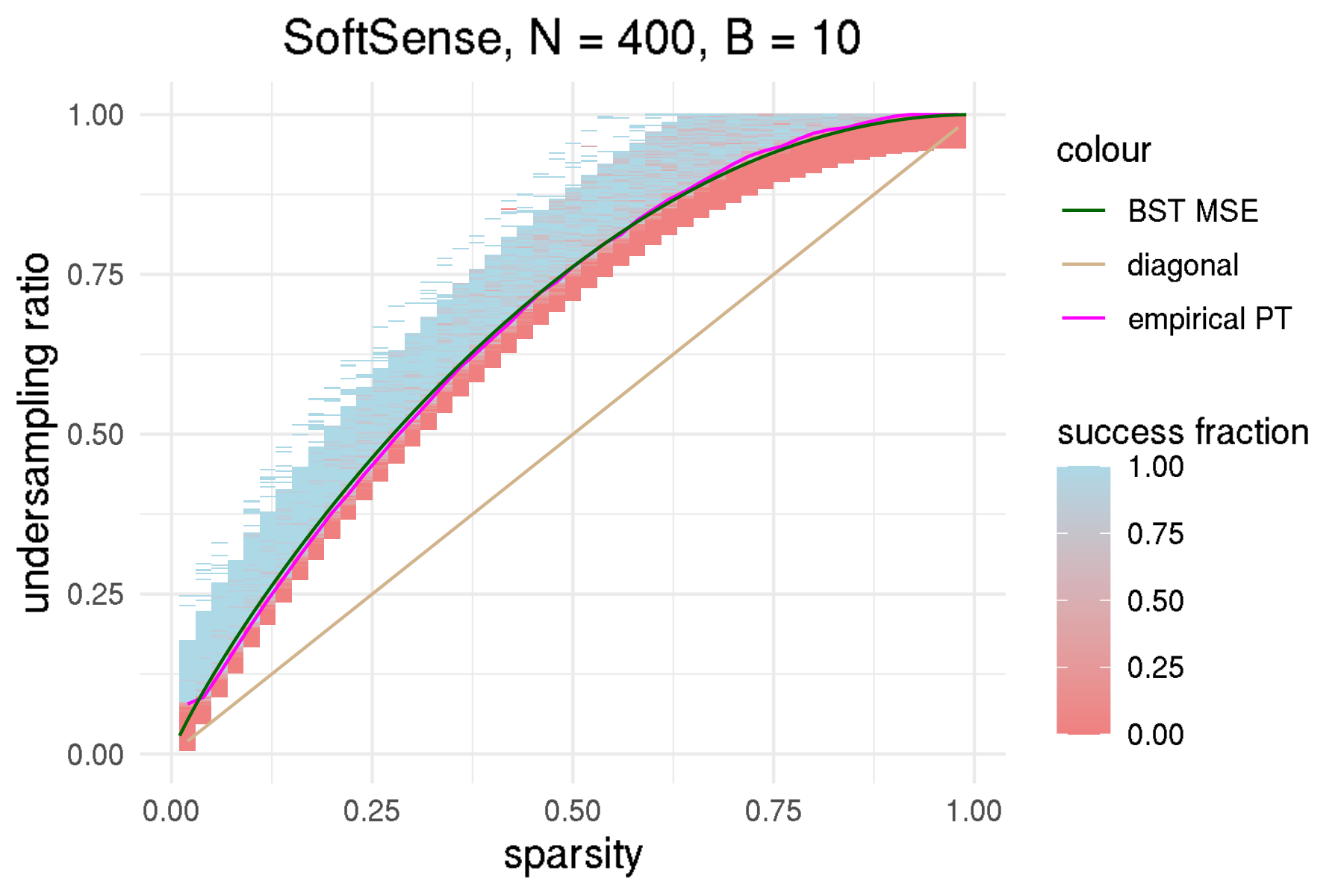}
  \end{subfigure}
  \vspace{1em}
  \begin{subfigure}[b]{0.9\textwidth}
    \includegraphics[width=\linewidth]{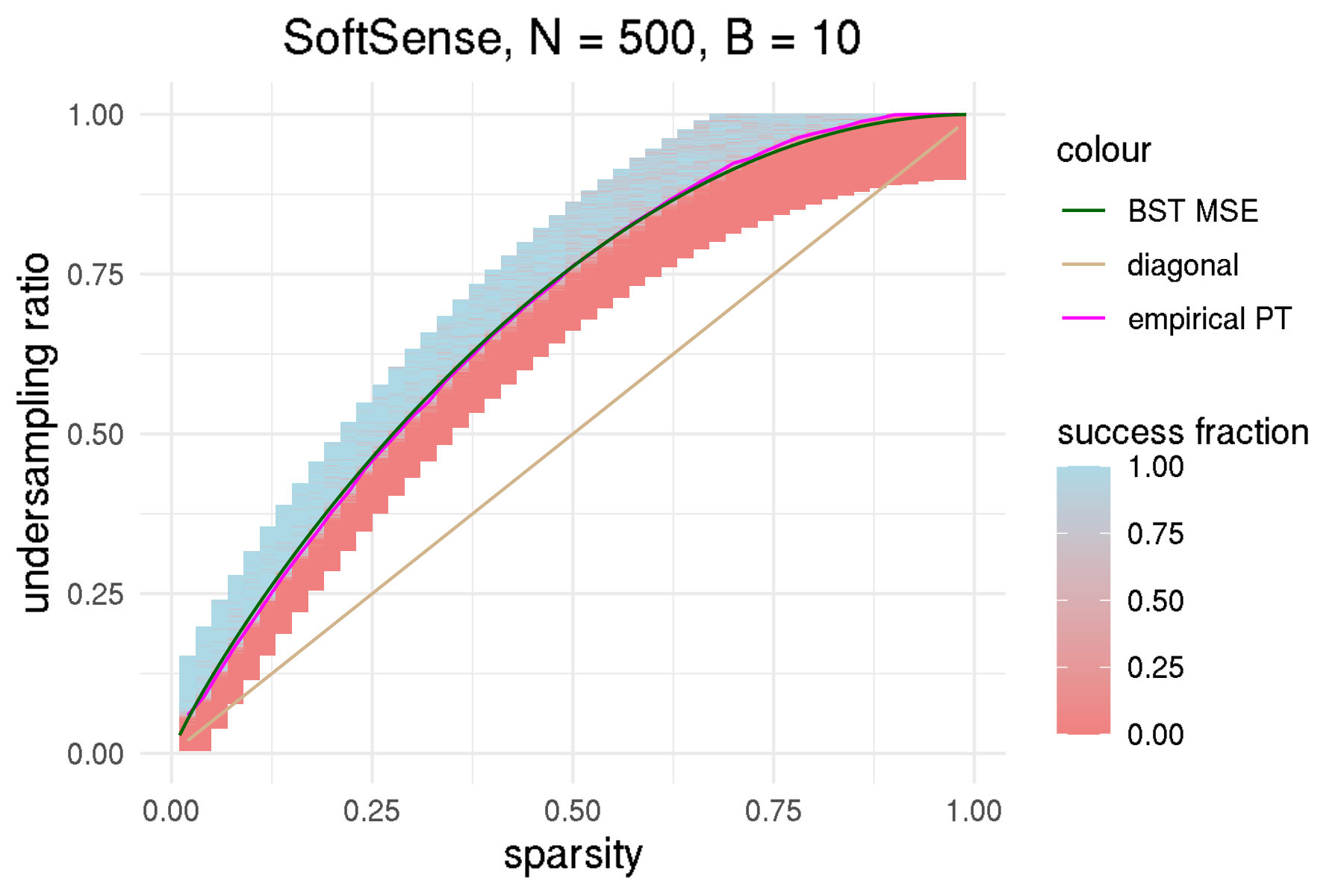}
  \end{subfigure}
  \caption{The nonzero entries are iid $N(0,1)$. WWe see that now the empirical phase transition pretty accurately matches the BlockSoft minimax risk. For $N=400$ we see extreme sparsity suffering, but that reduces when $N=500$.}
\end{figure}

% \begin{figure}[htbp]
%     \centering
%     \includegraphics[width=\linewidth]{figures/softsense/paper_softsense_jit_N_100_B_10.png}
%     \caption{Caption}
%     \label{fig:softsense_jit_normal_N_100_B_10}
% \end{figure}

% \begin{figure}[htbp]
%     \centering
%     \includegraphics[width=\linewidth]{figures/softsense/paper_softsense_jit_N_200_B_10.png}
%     \caption{Caption}
%     \label{fig:softsense_jit_normal_N_200_B_10}
% \end{figure}

% \begin{figure}[htbp]
%     \centering
%     \includegraphics[width=\linewidth]{figures/softsense/paper_softsense_jit_N_400_B_10.png}
%     \caption{Caption}
%     \label{fig:softsense_jit_normal_N_400_B_10}
% \end{figure}

% \begin{figure}[htbp]
%     \centering
%     \includegraphics[width=\linewidth]{figures/softsense/paper_softsense_jit_N_500_B_10.png}
%     \caption{Caption}
%     \label{fig:softsense_jit_normal_N_500_B_10}
% \end{figure}

\begin{figure}[htbp]
  \centering
  % Row 1
  \begin{subfigure}[b]{0.9\textwidth}
    \includegraphics[width=\linewidth]{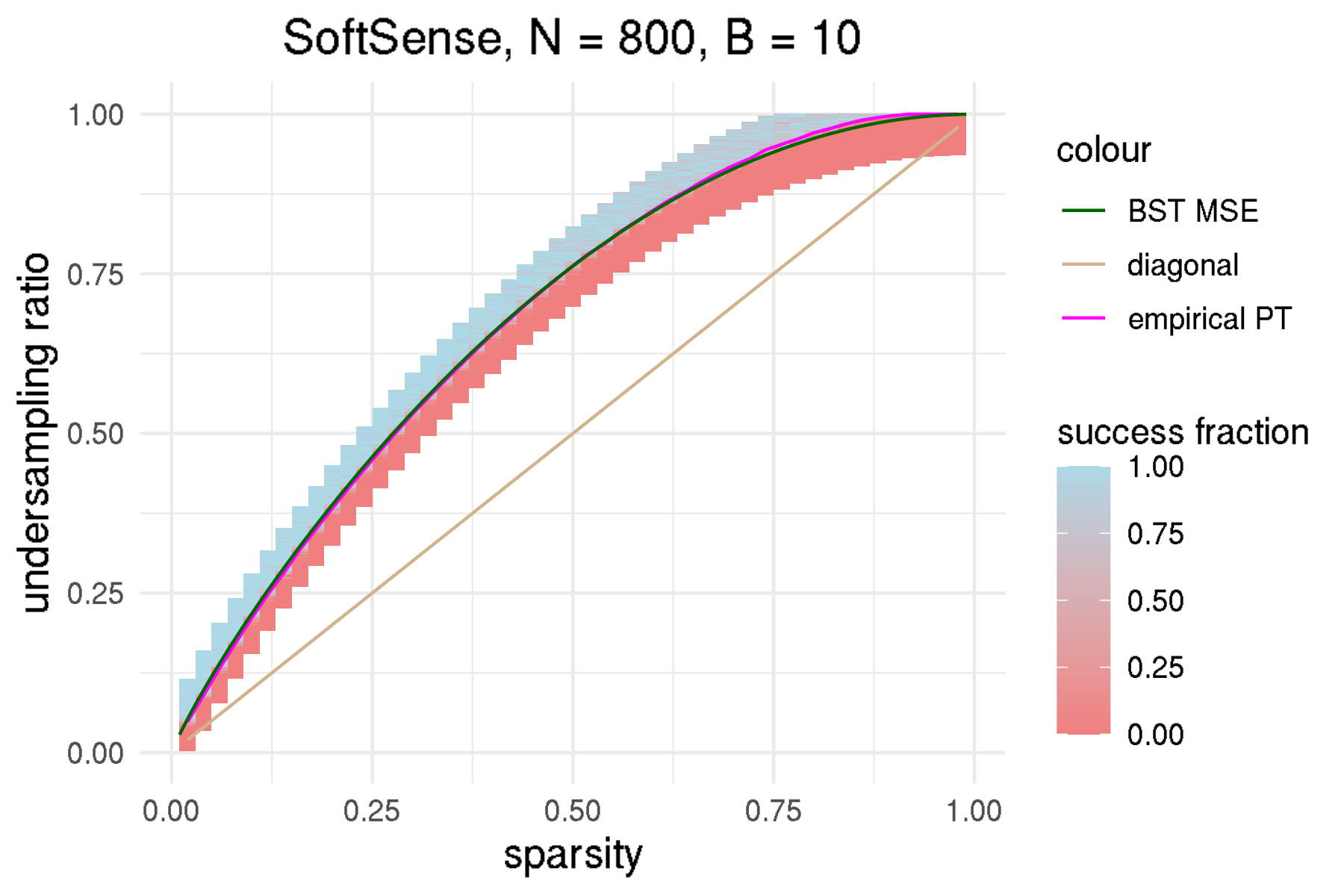}
  \end{subfigure}
  \vspace{1em}
  \begin{subfigure}[b]{0.9\textwidth}
    \includegraphics[width=\linewidth]{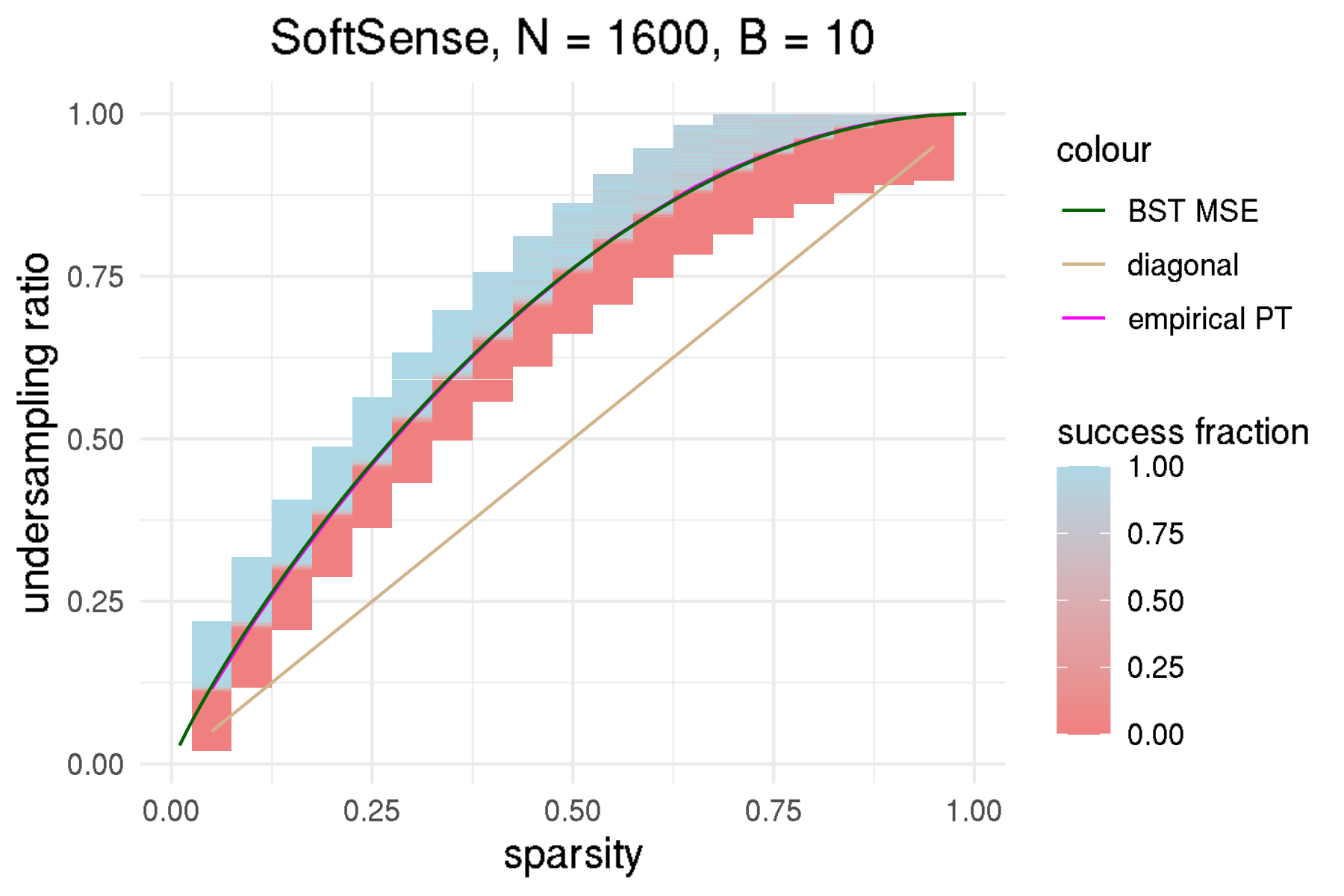}
  \end{subfigure}
  \caption{The nonzero entries are iid $N(0,1)$. We see an almost perfect match between the empirical phase transition and BlockSoft minimax risk.}
\end{figure}

% \begin{figure}[htbp]
%     \centering
%     \includegraphics[width=\linewidth]{figures/softsense/paper_softsense_jit_N_800_B_10.png}
%     \caption{Caption}
%     \label{fig:softsense_jit_normal_N_800_B_10}
% \end{figure}

% \begin{figure}[htbp]
%     \centering
%     \includegraphics[width=\linewidth]{figures/softsense/paper_softsense_jit_N_1600_B_10.png}
%     \caption{Caption}
%     \label{fig:softsense_jit_normal_N_1600_B_10}
% \end{figure}

% \vspace{1em}

\begin{figure}[htbp]
  \centering
  % Row 1
  \begin{subfigure}[b]{0.9\textwidth}
    \includegraphics[width=\linewidth]{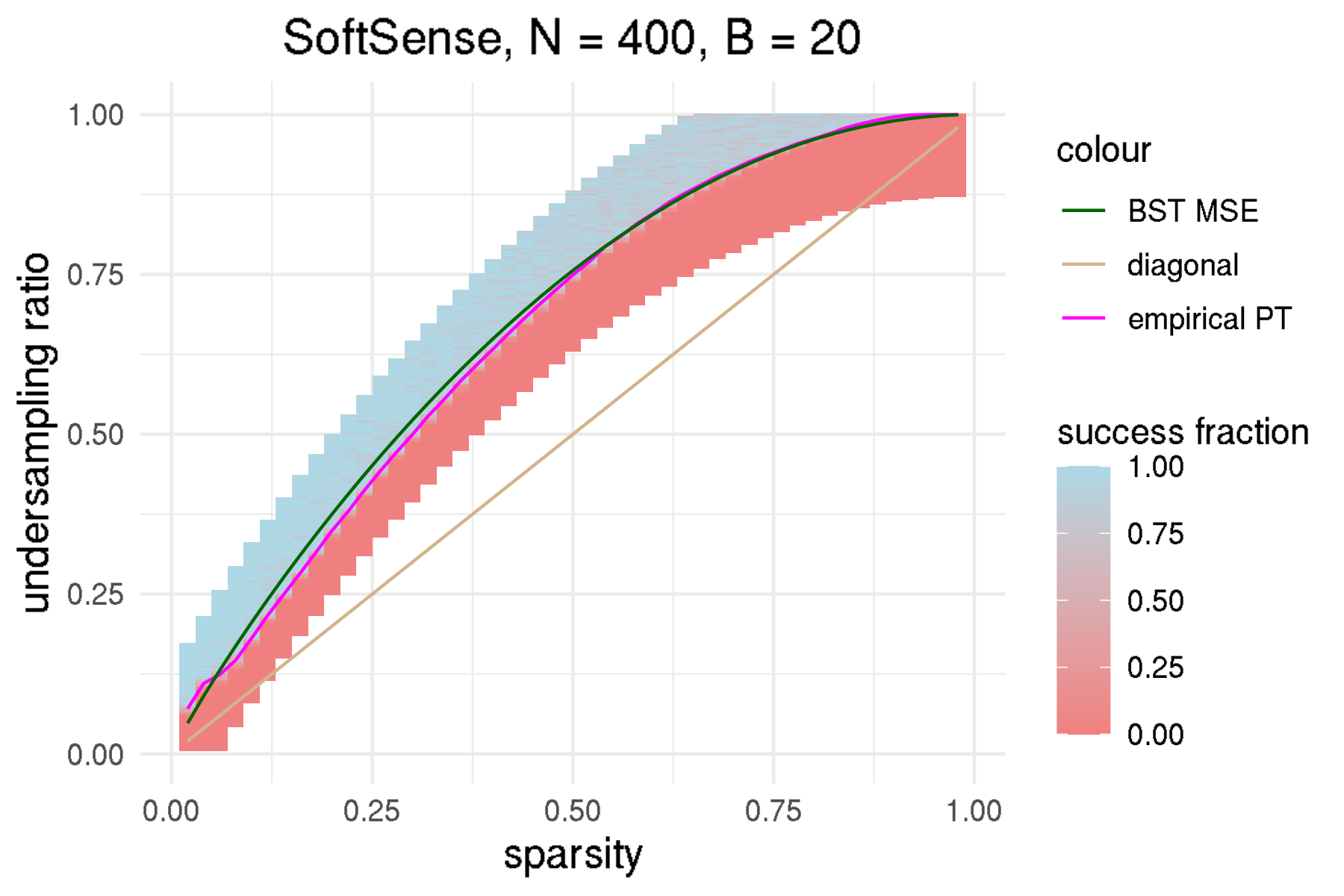}
  \end{subfigure}
  \vspace{1em}
  \begin{subfigure}[b]{0.9\textwidth}
    \includegraphics[width=\linewidth]{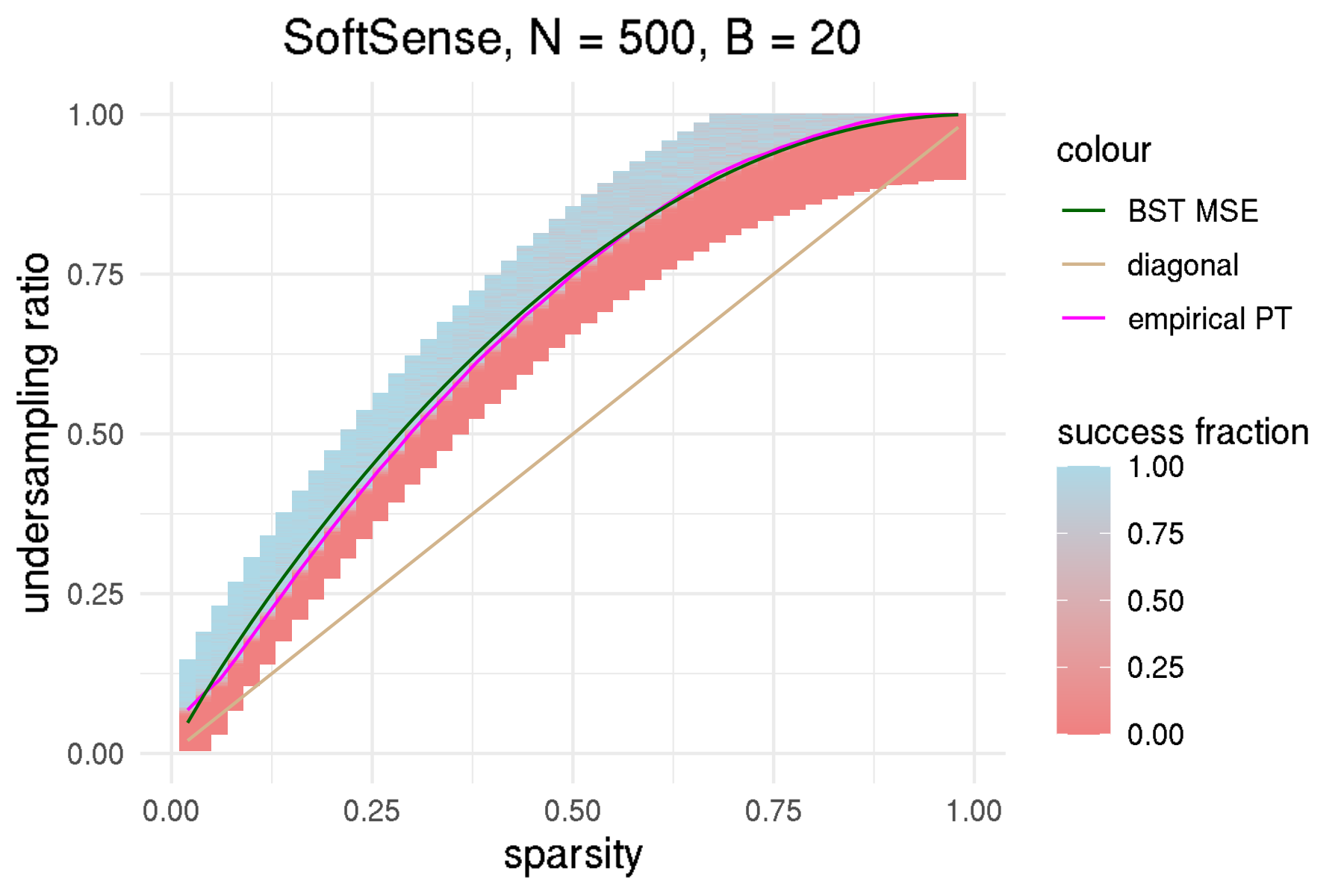}
  \end{subfigure}
  \caption{The nonzero entries are iid $N(0,1)$. At this large $B$, this range of $N$ is not sufficient to guarantee close numerical match between empirical phase transition and BlockSoft minimax risk.}
\end{figure}

% \begin{figure}[htbp]
%     \centering
%     \includegraphics[width=\linewidth]{figures/softsense/paper_softsense_jit_N_400_B_20.png}
%     \caption{Caption}
%     \label{fig:softsense_jit_normal_N_400_B_20}
% \end{figure}

% \vspace{1em}

% \begin{figure}[htbp]
%     \centering
%     \includegraphics[width=\linewidth]{figures/softsense/paper_softsense_jit_N_500_B_20.png}
%     \caption{Caption}
%     \label{fig:softsense_jit_normal_N_500_B_20}
% \end{figure}

% \vspace{1em}

\begin{figure}[htbp]
  \centering
  % Row 1
  \begin{subfigure}[b]{0.9\textwidth}
    \includegraphics[width=\linewidth]{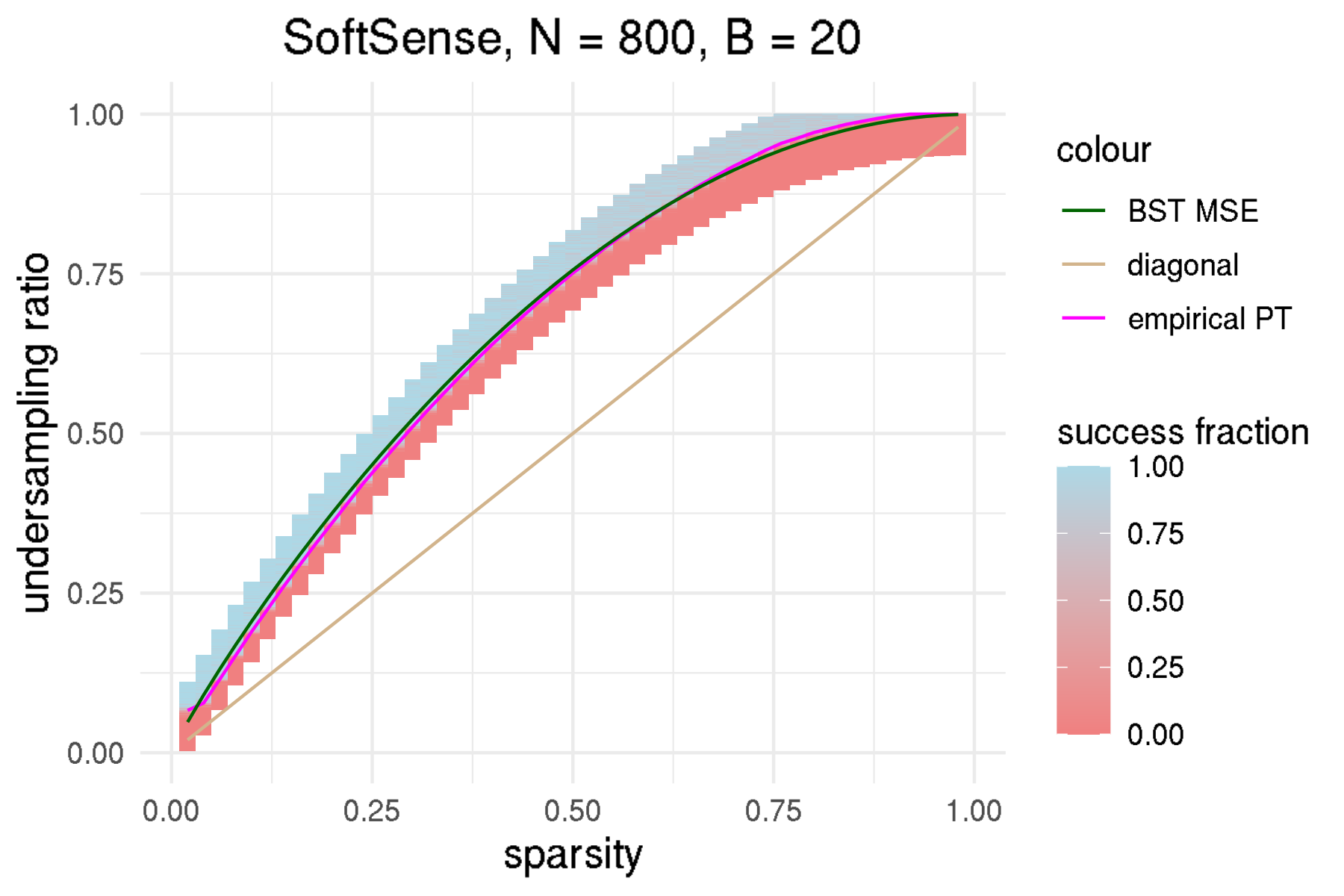}
  \end{subfigure}
  \vspace{1em}
  \begin{subfigure}[b]{0.9\textwidth}
    \includegraphics[width=\linewidth]{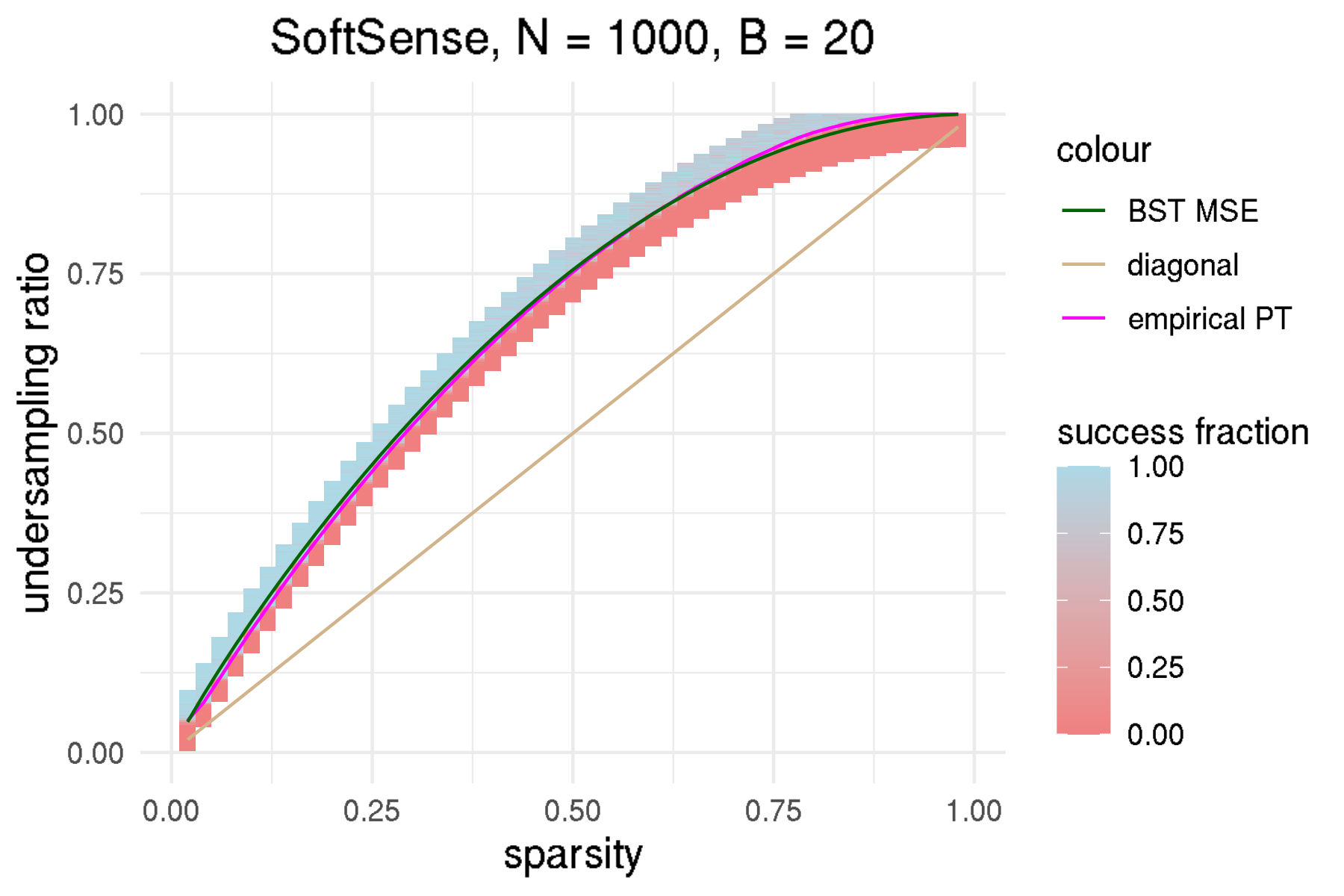}
  \end{subfigure}
  \caption{The nonzero entries are iid $N(0,1)$. The match between empirical phase transition and BlockSoft minimax risk definitely improves, but even at $N=1000$ we see a little bit of strange behavior (very faint). This seems to completely disappear when $N=2000$ as shown in Figure \ref{fig:softsense_jit_normal_B_20_50} in the main text.}
\end{figure}

% \begin{figure}[htbp]
%     \centering
%     \includegraphics[width=\linewidth]{figures/softsense/paper_softsense_jit_N_800_B_20.png}
%     \caption{Caption}
%     \label{fig:softsense_jit_normal_N_800_B_20}
% \end{figure}

% \vspace{1em}

% \begin{figure}[htbp]
%     \centering
%     \includegraphics[width=\linewidth]{figures/softsense/paper_softsense_jit_N_1000_B_20.png}
%     \caption{Caption}
%     \label{fig:softsense_jit_normal_N_1000_B_20}
% \end{figure}

% \vspace{1em}

\begin{figure}[htbp]
  \centering
  % Row 1
  \begin{subfigure}[b]{0.9\textwidth}
    \includegraphics[width=\linewidth]{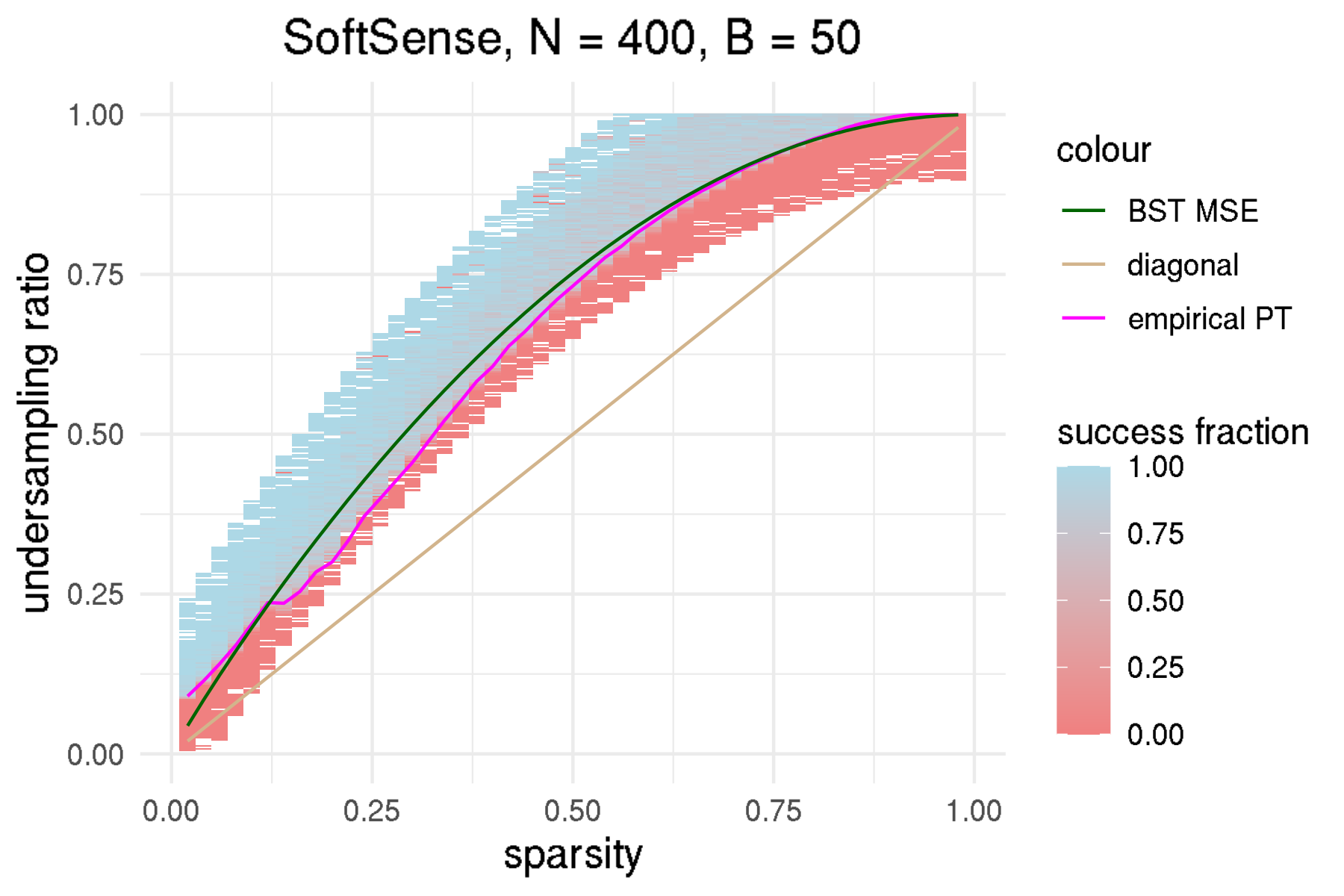}
  \end{subfigure}
  \vspace{1em}
  \begin{subfigure}[b]{0.9\textwidth}
    \includegraphics[width=\linewidth]{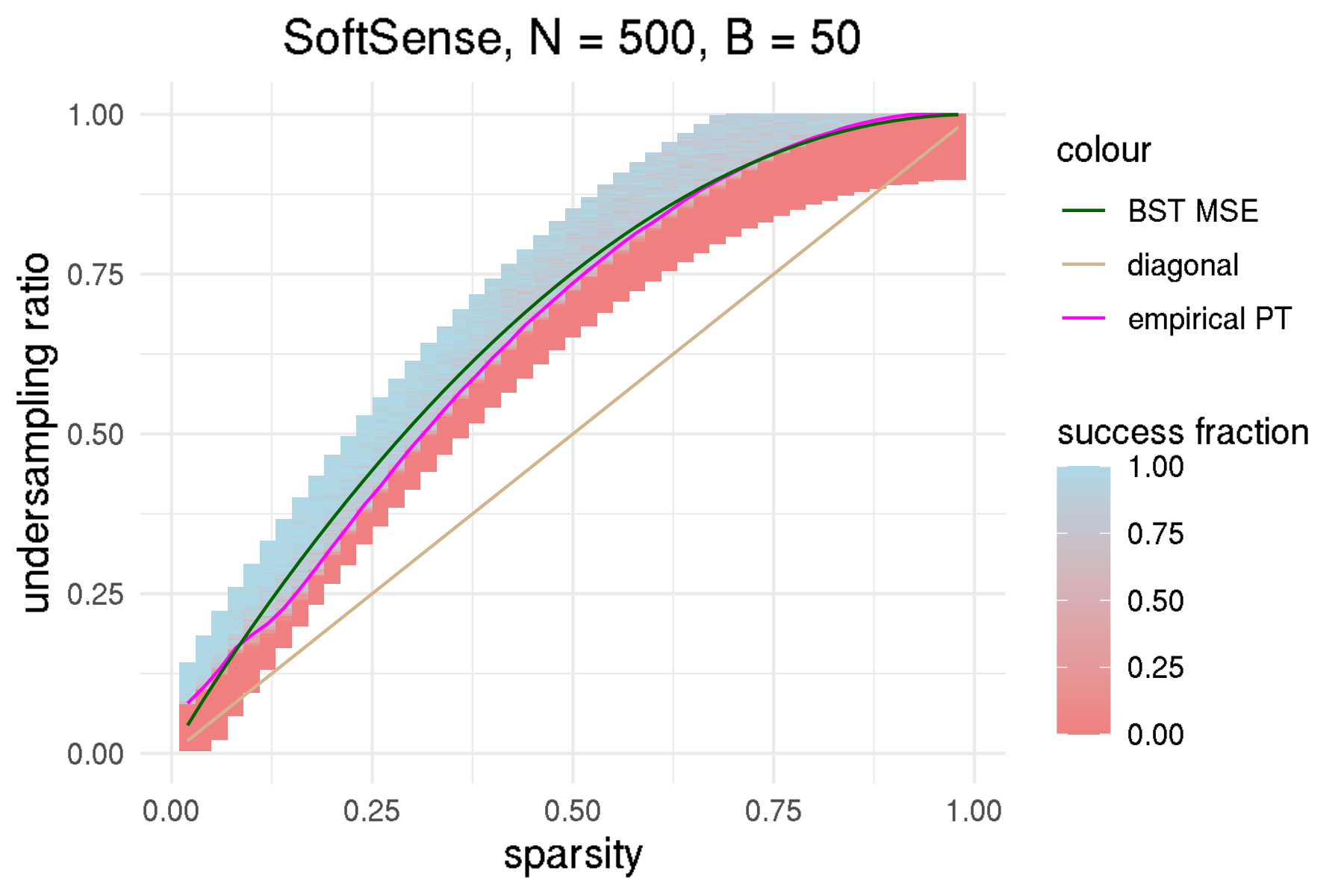}
  \end{subfigure}
  \caption{The nonzero entries are iid $N(0,1)$. The mismatch between empirical phase transition and BlockSoft minimax risk exists, as expected, as $B=50$ now.}
\end{figure}

% \begin{figure}[htbp]
%     \centering
%     \includegraphics[width=\linewidth]{figures/softsense/paper_softsense_jit_N_400_B_50.png}
%     \caption{Caption}
%     \label{fig:softsense_jit_normal_N_400_B_50}
% \end{figure}

% \vspace{1em}

% \begin{figure}[htbp]
%     \centering
%     \includegraphics[width=\linewidth]{figures/softsense/paper_softsense_jit_N_500_B_50.png}
%     \caption{Caption}
%     \label{fig:softsense_jit_normal_N_500_B_50}
% \end{figure}

% \vspace{1em}

\begin{figure}[htbp]
  \centering
  % Row 1
  \begin{subfigure}[b]{0.9\textwidth}
    \includegraphics[width=\linewidth]{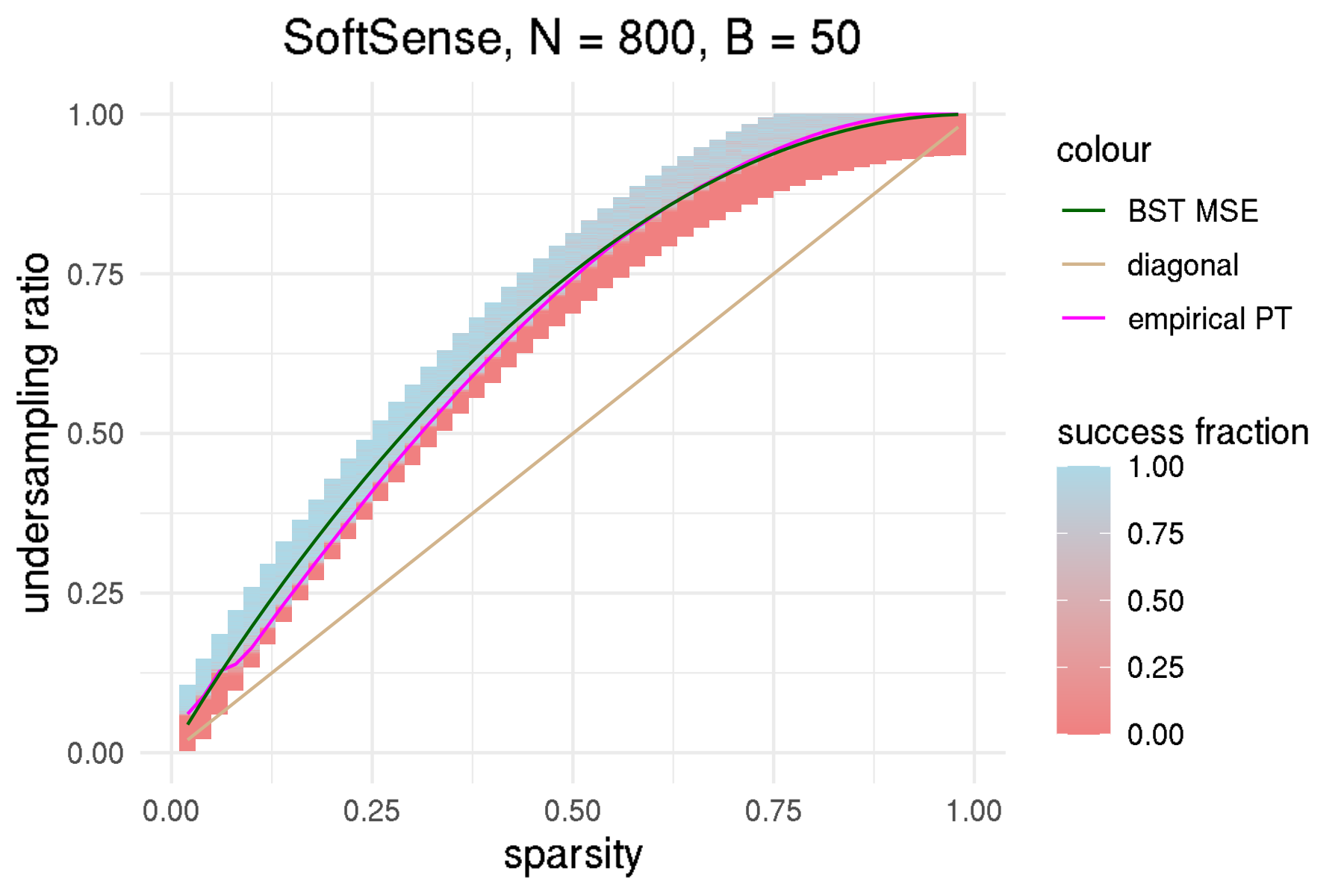}
  \end{subfigure}
  \vspace{1em}
  \begin{subfigure}[b]{0.9\textwidth}
    \includegraphics[width=\linewidth]{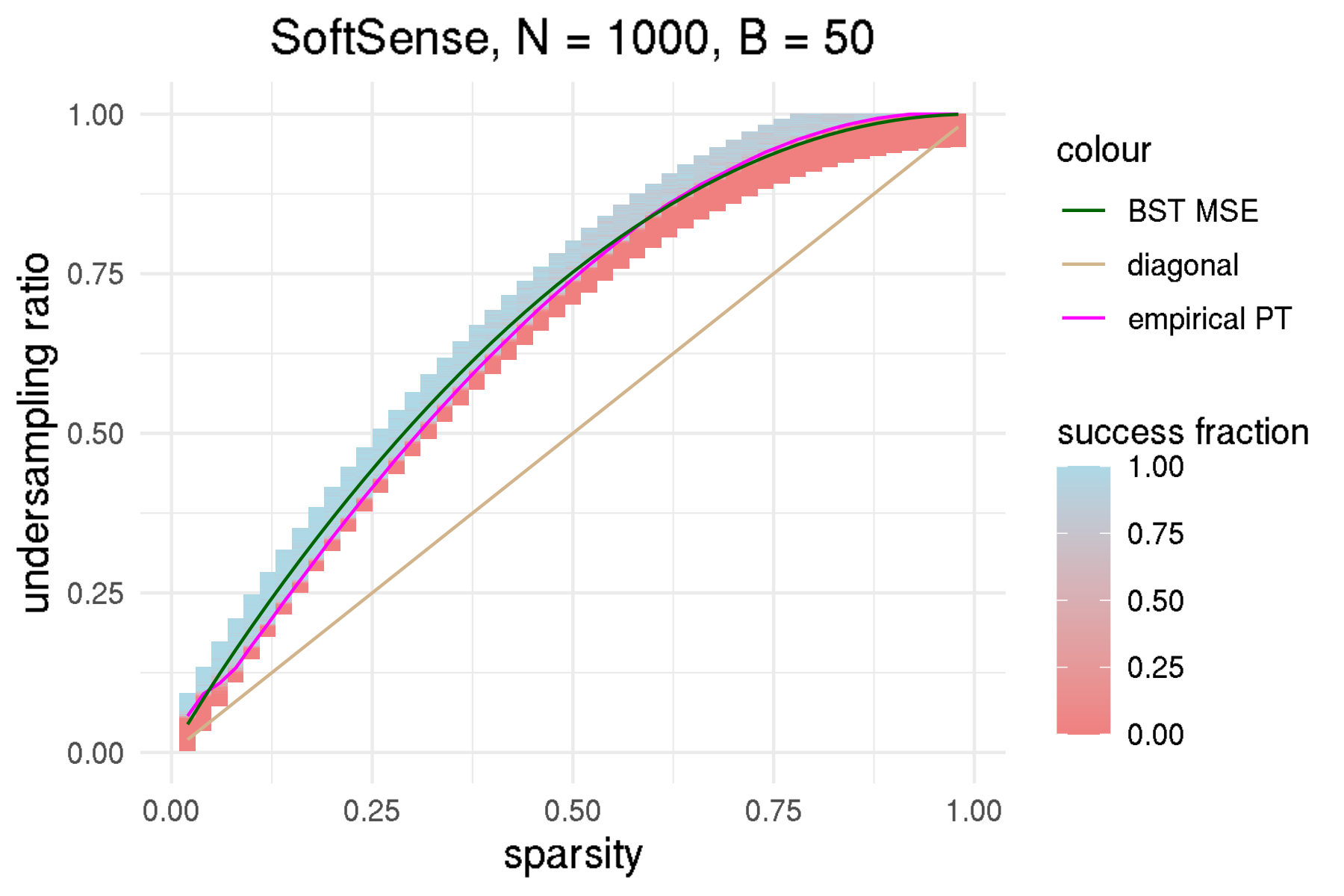}
  \end{subfigure}
  \caption{The nonzero entries are iid $N(0,1)$. The match between empirical phase transition and BlockSoft minimax risk definitely improves, but even at $N=1000$ we notice a little bit of numerical mismatch between the two. This seems to completely disappear when $N=5000$ as shown in Figure \ref{fig:softsense_jit_normal_B_20_50} in the main text.}
\end{figure}

% \begin{figure}[htbp]
%     \centering
%     \includegraphics[width=\linewidth]{figures/softsense/paper_softsense_jit_N_800_B_50.png}
%     \caption{Caption}
%     \label{fig:softsense_jit_normal_N_800_B_50}
% \end{figure}

% \vspace{1em}

% \begin{figure}[htbp]
%     \centering
%     \includegraphics[width=\linewidth]{figures/softsense/paper_softsense_jit_N_1000_B_50.png}
%     \caption{Caption}
%     \label{fig:softsense_jit_normal_N_1000_B_50}
% \end{figure}

% \vspace{1em}
% \input{sections/appendix/softsense_more_experiments}

\end{document}